%% file: main.tex
\documentclass[10pt]{article} 

\usepackage[accepted]{rlj}           

\usepackage{amssymb}            
\usepackage{mathtools}          
\usepackage{mathrsfs}           
\usepackage{graphicx}           
\usepackage{subcaption}         
\usepackage[space]{grffile}     
\usepackage{url}                
\usepackage{lipsum}             
\usepackage{microtype}
\usepackage{amsthm}
\usepackage[export]{adjustbox}
\usepackage{array}
\newcommand{\codeblue}[1]{\textcolor{textbluegray}{\textbf{\texttt{#1}}}}

\usepackage{booktabs}

\definecolor{myyellow}{HTML}{EBE450}
\definecolor{myblue}{HTML}{4C1155}

\include{math_commands}

\title{Unsupervised Behavioral Compression: \\
Learning Low-Dimensional Policy Manifolds through \\State-Occupancy Matching}

\setrunningtitle{Unsupervised Behavioral Compression}

\author{Andrea Fraschini\textsuperscript{$\clubsuit, \dagger$}, Davide Tenedini\textsuperscript{$\clubsuit, \dagger$}, Riccardo Zamboni\textsuperscript{$\clubsuit$}, Mirco Mutti\textsuperscript{$\spadesuit$}, Marcello Restelli\textsuperscript{$\clubsuit$}}

\coverPageAuthor{Andrea Fraschini\textsuperscript{$\dagger$}, Davide Tenedini\textsuperscript{$\dagger$}, Riccardo Zamboni, Mirco Mutti, Marcello Restelli}

\emails{andrea.fraschini@mail.polimi.it, \\\{davide.tenedini, riccardo.zamboni, marcello.restelli\}@polimi.it, mirco.m@technion.ac.il}

\affiliations{
$^{\clubsuit}$ \textbf{Politecnico di Milano}\\
$^{\spadesuit}$ \textbf{Technion}
\par 
$^\dagger$ Equal contribution}

\input{sections/_cover_page}

\begin{document}

\makeCover  
\maketitle  

\input{sections/0_abstract}

\input{sections/1_introduction}

\input{sections/2_preliminaries}

\input{sections/3_background}

\input{sections/4_methodology}
\raggedbottom

\input{sections/5_experiments}
\input{sections/6_conclusion}

\appendix


\subsubsection*{Acknowledgments}
This work was supported by the Italian Ministry of University and Research (MUR) under the National Recovery and Resilience Plan (NRRP), and by the European Union (EU) under the NextGenerationEU project (CUP D43C24001670008).


\bibliography{main}
\bibliographystyle{rlj}

\beginSupplementaryMaterials

\input{sections/A1_related_works}

\input{sections/A2_scores}

\input{sections/A3_importance_weights}

\input{sections/A4_training}

\input{sections/A5_experimental_settings}

\input{sections/A6_additional_experiments}

\end{document}

%% file: math_commands.tex
\usepackage{wrapfig}
\usepackage{amsmath,amsfonts,bm}

\def\Figgref#1{Fig.~\ref{#1}}

\def\quadfigref#1#2#3#4{figures \ref{#1}, \ref{#2}, \ref{#3} and \ref{#4}}

\def\Secref#1{Section~\ref{#1}}

\def\eqref#1{equation~\ref{#1}}

\def\Eqqref#1{Eq.~\ref{#1}}

\def\Apref#1{Appendix~\ref{#1}}

\def\Algref#1{Algorithm~\ref{#1}}

\def\1{\bm{1}}

\def\vmu{{\bm{\mu}}}
\def\vtheta{{\bm{\theta}}}
\def\vphi{{\bm{\phi}}}
\def\vxi{{\bm{\xi}}}
\def\vzeta{{\bm{\zeta}}}
\def\vsigma{{\bm{\sigma}}}

\def\vz{{\bm{z}}}

\DeclareMathAlphabet{\mathsfit}{\encodingdefault}{\sfdefault}{m}{sl}
\SetMathAlphabet{\mathsfit}{bold}{\encodingdefault}{\sfdefault}{bx}{n}

\def\gA{{\mathcal{A}}}

\def\gD{{\mathcal{D}}}

\def\gL{{\mathcal{L}}}
\def\gM{{\mathcal{M}}}

\def\gS{{\mathcal{S}}}

\def\gZ{{\mathcal{Z}}}

\def\sM{{\mathbb{M}}}

\def\sP{{\mathbb{P}}}

\def\sR{{\mathbb{R}}}

\newcommand{\E}{\mathbb{E}}

\newcommand{\KL}{D_{\mathrm{KL}}}

\DeclareMathOperator*{\argmax}{arg\,max}
\DeclareMathOperator*{\argmin}{arg\,min}

\definecolor{softbluegray}{HTML}{F2F6F9}
\definecolor{textbluegray}{HTML}{629999}

\NewDocumentCommand{\stackcomparison}{O{0.23} O{-1mm} O{-6mm} m m m }{%
    \begin{subfigure}[b]{#1\textwidth}
        \centering
        \includegraphics[width=\textwidth]{#4} \\
        \vspace{#2}
        \includegraphics[width=\textwidth]{#5}
        \vspace{#3}
        \caption{#6}
    \end{subfigure}
}

%% file: sections/_cover_page.tex
\contribution{
    This paper introduces an information-theoretic uniqueness score derived from the entropy decomposition of the mixture state occupancy to curate unsupervised policy datasets for manifold learning.
    }
    {
    Prior work on Latent Behavior Compression~\citep{tenedini2025from} curates datasets using novelty search based on $k$-nearest neighbors in the Euclidean space of immediate policy actions corresponding to randomly sampled states, which, as demonstrated by the experiments presented in~\Secref{sec:exp_curation}, often fails to capture complex behavioral uniqueness.
    }

\contribution{
    This paper proposes a fully differentiable Behavioral Compression objective that directly minimizes the divergence between true and reconstructed mixture occupancy distributions using an importance sampling estimator.
    }
    {
    Prior work~\citep{tenedini2025from} employed a behavioral reconstruction loss based on the divergence of the immediate action distribution relative to randomly sampled states. 
    However, such a measure has been shown to be a rather ultra-conservative proxy for behavioral similarity~\citep[Prop. E.1,][]{metelli2018policy}.
    }

\contribution{
    Through an extensive empirical campaign, we demonstrate that our methodology preserves unique, high-performing behaviors during dataset curation, yields a more behaviorally rich latent manifold, and successfully enables latent optimization in complex environments.
    }
    {
    Prior work~\citep{tenedini2025from} lacked quantitative analysis on the effect of dataset curation, presented in~\Secref{sec:exp_curation}, and proposed methods with action-based proxies that collapsed in more complex environments (e.g., Hopper).
    }

\keywords{reinforcement learning, unsupervised reinforcement learning, unsupervised representation learning} 

\summary{
Deep Reinforcement Learning (DRL) is widely recognized as sample-inefficient, a limitation attributable in part to the high dimensionality and substantial functional redundancy inherent to the policy parameter space. A recent framework, which we refer to as Action-based Policy Compression (APC), mitigates this issue by compressing the parameter space $\Theta$ into a low-dimensional latent manifold $\gZ$ using a learned generative mapping $g:\gZ \to \Theta$. However, its performance is severely constrained by relying on immediate action-matching as a reconstruction loss, a myopic proxy for behavioral similarity that suffers from compounding errors across sequential decisions.

To overcome this bottleneck, we introduce Occupancy-based Policy Compression (OPC), which enhances APC by shifting behavior representation from immediate action-matching to long-horizon state-space coverage. This modification forces the generative model to organize the latent space around true functional similarity, promoting a latent representation that generalizes over a broad spectrum of behaviors while retaining most of the original parameter space's expressivity. Finally, we empirically validate the advantages of our contributions across multiple continuous control benchmarks.
}



%% file: sections/0_abstract.tex
\begin{abstract}
Deep Reinforcement Learning (DRL) is widely recognized as sample-inefficient, a limitation attributable in part to the high dimensionality and substantial functional redundancy inherent to the policy parameter space. A recent framework, which we refer to as Action-based Policy Compression (APC), mitigates this issue by compressing the parameter space $\Theta$ into a low-dimensional latent manifold $\gZ$ using a learned generative mapping $g:\gZ \to \Theta$. However, its performance is severely constrained by relying on immediate action-matching as a reconstruction loss, a myopic proxy for behavioral similarity that suffers from compounding errors across sequential decisions. To overcome this bottleneck, we introduce Occupancy-based Policy Compression (OPC), which enhances APC by shifting behavior representation from immediate action-matching to long-horizon state-space coverage. Specifically, we propose two principal improvements: (1) we curate the dataset generation with an information-theoretic uniqueness metric that delivers a diverse population of policies; and (2) we propose a fully differentiable compression objective that directly minimizes the divergence between the true and reconstructed mixture occupancy distributions. These modifications force the generative model to organize the latent space around true functional similarity, promoting a latent representation that generalizes over a broad spectrum of behaviors while retaining most of the original parameter space's expressivity. Finally, we empirically\footnote{The source code is available in the supplementary material and at \href{https://github.com/DavideTenediniPoliMi/Unsupervised-Behavioral-Compression}{this link}.} validate the advantages of our contributions across multiple continuous control benchmarks.
\end{abstract}

%% file: sections/1_introduction.tex
\section{Introduction}
\label{sec:intro}
The success of Deep Reinforcement Learning (DRL) in complex continuous control relies heavily on the highly expressive nature of deep neural networks~\citep{lillicrap2015continuous, schulman2017proximal}. However, optimizing these high-dimensional policies \emph{tabula rasa} is notoriously sample-inefficient~\citep{duan2016benchmarking, agarwal2022reincarnating}. This inefficiency stems in part from a fundamental functional redundancy: Due to the complex symmetries of neural architectures, vast and disparate regions of the parameter space often collapse into identical agent behaviors. Searching this massive, unstructured space blindly ignores the underlying functional geometry of the environment. This limitation becomes a severe bottleneck in multi-task or Unsupervised Reinforcement Learning (URL) settings~\citep{laskin2021urlb}, where an agent must train a model that rapidly adapts to a variety of previously unknown downstream tasks rather than learning from scratch.

To address this redundancy in the policy parameter space, prior work has proposed various approaches to compress the policy space onto lower-dimensional latent manifolds~\citep[e.g.,][]{eysenbach2018diversity, mutti2022reward, faccio2023goal, tenedini2025from, li2026learning}. In this stream,
\codeblue{Action-based Policy Compression}~\citep[APC,][]{tenedini2025from} explicitly shifts the learning paradigm from parameter optimization to behavior optimization. APC leverages a generative model to compress a highly redundant policy parameter space into a low-dimensional, behaviorally structured latent manifold. By curating a diverse dataset of random policies and training an autoencoder to reconstruct them, agents can subsequently bypass the raw parameter space entirely. Instead, they perform task-specific fine-tuning by directly navigating the compact latent space.
By following this recipe, APC compresses policy architectures with several thousand parameters into tiny two- or three-dimensional latent spaces, while retaining most of the functional expressivity to address a variety of tasks in classical continuous control domains, such as Mountain Car and Reacher.

While conceptually and empirically promising, the practical adaptability of APC beyond these domains is severely constrained by the technical solutions adopted to train the latent space. In particular, to organize this latent space functionally rather than geometrically, APC relies on an action-matching objective, training the autoencoder to minimize the discrepancy between the immediate action outputs of the original and reconstructed policies. However, measuring behavioral similarity through immediate action divergence is a mathematically loose and myopic proxy for long-horizon control~\citep{metelli2018policy}. Because errors compound across sequential decisions, policies that appear functionally identical at the single-action level can lead to drastically different trajectories and global state visitations. Furthermore, \citet{tenedini2025from} curate the initial pre-training dataset of policies using novelty search in the action space, which frequently discards rare, high-performing behaviors, severely impoverishing the topological diversity of the resulting manifold.

To overcome these limitations, we introduce \codeblue{Occupancy-based Policy Compression} (OPC), representing a paradigm shift in latent manifold learning for DRL. We abandon immediate action-matching entirely in favor of global state-space coverage. Specifically, we introduce two principal algorithmic enhancements. First, we curate the pre-training dataset using an information-theoretic uniqueness metric that actively selects policies based on their contribution to the global state-visitation entropy, retaining a much broader spectrum of rare behaviors. Second, we propose a fully differentiable compression objective that explicitly matches the true mixture occupancy distribution of the policy population. By evaluating policies based on the exact states they visit over the full horizon, OPC completely bypasses the compounding errors of action-matching, forcing the generative model to organize the latent space around true, long-horizon functional similarity.

Then, to support the design choices behind OPC, we address the following:
\vspace{-5pt}
\begin{tcolorbox}[colback=softbluegray, colframe=softbluegray,  boxrule=0.5pt, arc=4pt, width=\linewidth]
\textbf{Research Questions:}
\begin{itemize}[leftmargin=5pt]
    \item[] (\textbf{Q1}) Does curating policies based on their contribution to state-space coverage preserve a wider spectrum of rare, high-performing behaviors compared to action-space novelty search?
    \item[] (\textbf{Q2}) Does optimizing a compression objective based on state-space visitation patterns yield a more behaviorally meaningful latent space than action-based behavioral reconstruction?
    \item[] (\textbf{Q3}) Does navigating an occupancy-matched latent space improve sample efficiency and task performance compared to standard baselines and action-matched manifolds?
\end{itemize}
\end{tcolorbox}

\textbf{Content Outline and Contributions.}~~In \Secref{sec:prel}, we establish the necessary notation. In \Secref{sec:background}, we detail the existing APC framework and expose the limitations of action-matching. In \Secref{sec:method}, we introduce our OPC framework. Finally, in \Secref{sec:exp}, we address our Research Questions with extensive empirical evidence, demonstrating that OPC retains richer behavioral modes, constructs superior latent topologies, and enables fast few-shot adaptation to various downstream tasks.

%% file: sections/2_preliminaries.tex
\section{Preliminaries}
\label{sec:prel}

\textbf{Notation.}~~For a set $\gA$ of finite size $|\gA|$, we denote $\Delta(\gA)$ its probability simplex. 
For two distributions $p,q$, the Kullback-Leibler (KL) divergence is $\KL(p\parallel q)$ and the differential entropy of $p$ is $H(p)$.

\textbf{Interaction Protocol.}~~We model the interactions between the agent and the environment as a (finite-horizon) \textcolor{textbluegray}{\textbf{Controlled Markov Process}}  (CMP). A CMP is a tuple $\gM \coloneq (\gS, \gA, \sP, \mu, T)$, where $\gS$ is the state space and $\gA$ is the action space. An episode begins with the initial state $s_0\sim\mu\in\Delta(\gS)$, where $\mu$ is the initial state distribution. At each timestep $t=0,\ldots,T$, where $T<\infty$ is the episode horizon, the agent observes the current state $s_t$, takes action $a_t \in \gA$, such that the model transitions to the next state $s_{t+1} \sim \sP(\cdot\mid s_t,a_t)$ according to the transition model $\sP:\gS\times\gA\to\Delta(\gS)$. The agent acts according to a \emph{stochastic policy} $\pi:\gS\to\Delta(\gA)$ such that $\pi(a\mid s)$ denotes the conditional probability of taking action $a$ in state $s$. This interaction induces a trajectory $\tau \coloneq (s_0, a_0, s_1, a_1, \ldots, s_T)$.

\textbf{Occupancy Measure.}~~The deployment of policy $\pi$ in $\gM$ induces a marginal distribution over the state space, known as the (state) \emph{occupancy measure} $d_\pi\in\Delta(\gS)$, which represents the average visitation probability of state $s$ over the episode length: $d_{\pi}(s)=\frac{1}{T+1}\sum_{t=0}^T \Pr(s_t=s\mid\pi)$. If we consider the deployment of a population of $M$ distinct policies $\Psi=\{\pi_1,\ldots,\pi_M\}$, we define the \codeblue{mixture occupancy measure} $d_\Psi\in\Delta(\gS)$ as the (uniform) mixture of individual occupancies: $d_{\Psi}(s)=\frac{1}{M}\sum_{i=1}^M d_{\pi_i}(s), \quad \forall s\in\gS$.

We consider parametric policies $\pi_\vtheta$ represented by neural networks parametrized by $P$-dimensional weight vectors $\vtheta\in\Theta\subseteq\sR^P$, where $\Theta$ represents the policy parameter space. We denote the collection of representable policies as the \emph{Policy Space} $\Pi_\Theta$. As shorthand, we use the parameter vector $\vtheta$ in place of its induced policy $\pi_\vtheta$ and the parameter space $\Theta$ in place of the policy space $\Pi_\Theta$.

\textbf{Non-Parametric Estimation.}~~In continuous spaces, a direct computation of the $\KL(f \parallel f')$ between distributions $f, f'$ is often intractable. Other approaches, such as \emph{particle-based estimators}, are used to approximate these quantities using the finite set of states of collected trajectories. Specifically,~\citet{AJGL2011KnnEstimator} propose an \codeblue{importance sampling $k$-NN estimator}:
\begin{equation}
    \label{eq:knn_estimator}
    \widehat{D}_{KL}\!\left(f \parallel f'\right)
    = \frac{1}{N} \sum_{i=1}^{N} \ln \frac{k/N}{\sum_{j \in \mathcal{N}_i^{k}} w_j},\quad
     w_j = \frac{f'(x_j) / f(x_j)}{\sum_{n=1}^{N} f'(x_n) / f(x_n)},
\end{equation}
where $N$ is the number of particles, $\mathcal{N}_i^{k}$ is the set of indices of the $k$-nearest neighbors of $x_i$, and $w_j$ are the self-normalized importance weights of samples $x_j$. 

%% file: sections/3_background.tex
\section{Background: Action-based Policy Compression}
\label{sec:background}

In Reinforcement Learning (RL), an agent interacts with a CMP equipped with a given reward signal $R:\gS\times\gA\to\sR$, augmenting the CMP into a Markov Decision Process~\citep[MDP,][]{puterman2014markov} $\gM_{R}:=\gM\cup R$. The agent's goal is typically to find a high-dimensional parameter vector $\vtheta^*=\argmax_{\vtheta\in\Theta}J^R(\vtheta)$ that maximizes the expected cumulative sum of rewards $J^R(\vtheta)$  obtained by playing the policy $\pi_\vtheta$. This process is notoriously sample-inefficient, partially due to a functional \emph{redundancy}: vast, distinct regions of the parameter space $\Theta$ collapse into identical agent behaviors. As first noted by~\citet{tenedini2025from}, this inefficiency is severely exacerbated in \codeblue{Unsupervised Reinforcement Learning}~\citep[URL,][]{laskin2021urlb}: the agent learns a task-agnostic representation $\sM$ via \emph{unsupervised pre-training} on a reward-free CMP, then leverages it during \emph{supervised fine-tuning} to rapidly optimize the newly revealed reward $R$.

\textbf{The APC Framework.}~~This redundancy in URL was first addressed directly via~\codeblue{Action-based Policy Compression}~\citep[APC,][]{tenedini2025from}: the high-dimensional policy parameter space $\Theta\subseteq\sR^P$ is compressed into a compact, behaviorally structured latent space $\gZ\subseteq\sR^k$, with $k\ll P$. By shifting the downstream search space from parameters to behaviors, APC enables efficient task adaptation. The framework operates in three sequential stages:

\emph{1. Policy Dataset Curation:}~~To model the behavior manifold, APC generates a large pool of policies via random weight initialization. This na\"ive approach is computationally light, but prone to generating redundant, low-energy policies.  To filter this pool into a diverse training dataset, APC employs a $k_n$-nearest-neighbor novelty search in the \emph{action space}, selecting policies that produce diverse immediate action distributions across a set of  collected states. Formally, they threshold the dataset to keep only the top percentile of policies according to the following score:
\begin{equation}
    \label{eq:apc_score}
    \rho^{\text{APC}}(\pi_\vtheta)=\frac{1}{k_{n}}\sum_{i\in\mathcal{N}^{k_n}}\KL(\pi_\vtheta\parallel \pi_{\vtheta_i}).
\end{equation}
\emph{2. Latent Behavior Compression:}~~APC defines the objective of Policy Compression as finding a generative mapping $g^*:\gZ\to\Theta$, such that:
\begin{equation}
    \label{eq:apc_latent}
    \forall\vtheta\in\Theta,\quad\exists \vz \in\gZ:\quad g^\star = \argmin_g\KL\left(d_{\pi_\vtheta}^{sa}\parallel d_{\pi_{\vz}}^{sa}\right),
\end{equation}
where $d_{\pi}^{sa}=\pi(a\mid s)\cdot d_{\pi}^s$ is the (state--action) occupancy measure. To achieve such an objective, APC employs an Autoencoder~\citep[AE,][]{hinton2006reducing}, consisting of an encoder $f_\vxi:\Theta\to\gZ$ and a decoder $g_\vzeta:\gZ\to\Theta$, parametrized by $\vxi$ and $\vzeta$, respectively, trained to minimize the divergence between immediate actions of the original and reconstructed policy, defined as $\gL_{B}(\vxi, \vzeta)=\E_{\vtheta\sim\gD_{\Theta}}\left[\KL\left(\pi_\vtheta\parallel \pi_{(g_\vzeta\circ f_\vxi)(\vtheta)}\right)\right]$. This serves as a computationally efficient proxy to \Eqqref{eq:apc_latent}.

\emph{3. Latent Behavior Optimization:}~~Once the decoder $g_\vzeta$ is trained and frozen, it serves as the task-agnostic representation $\sM$ for fine-tuning. For a downstream task $R$, the agent searches for an optimal code $\vz^*=\argmax_{\vz\in\gZ}J^R(\vtheta=g_\vzeta(\vz))$. Because $\gZ$ is low-dimensional, APC employs \codeblue{Policy Gradient with Parameter-based Exploration} \citep[PGPE,][]{sehnke2008policy}. PGPE treats the decoder as a black box, avoiding costly Jacobian computations by sampling latent codes $\vz$ from a hyper-policy $\nu_\vphi$ (e.g., a Gaussian distribution parameterized by $\vphi = (\vmu, \vsigma)$). The hyperparameters are updated via a Monte Carlo gradient estimator over $N$ trajectories:
\begin{equation}
\label{eq:pgpe}
\hat\nabla_\vphi J^R(\vphi)=\frac {1}{N}\sum_{i=1}^N\nabla_\vphi\log\nu_\vphi(\vz_i)R(\tau_i).
\end{equation}
\textbf{Limitations: The Action-matching Bottleneck.}~~While APC successfully reduces the dimensionality of the policy space, its reliance on immediate action-matching poses a fundamental limitation for long-horizon continuous control. Theoretically, APC justifies this proxy by relying on the fact that bounding immediate action divergence also bounds the true discrepancy between state occupancies.~\citet[Prop. E.1,][]{metelli2018policy} provide an upper bound to this quantity for $H$-horizon settings, namely $H\sup_{s\in\gS}\KL(\pi_\vtheta(\cdot\mid s)\parallel \pi_{\vtheta'}(\cdot\mid s))$.
In practice, due to compounding errors across sequential decisions, policies with near-identical action distributions can yield drastically different state-occupancy measures. This myopic behavioral representation in both the dataset curation and behavior compression phases hinders APC's ability to retain rare, but highly structured behaviors, severely impoverishing the topological diversity of the resulting latent manifold. Overcoming these limitations requires a paradigm shift: designing and optimizing behaviors based on long-horizon state-space coverage, rather than immediate actions.

%% file: sections/4_methodology.tex
\section{Methodology: Occupancy-based Policy Compression}
\label{sec:method}
To overcome the inherent limitation of action-matching proxies, we introduce \codeblue{Occupancy-based Policy Compression} (OPC). While retaining APC's three-stage pipeline, we introduce fundamental algorithm shifts to the dataset curation, manifold learning, and fine-tuning phases:

\emph{1. Policy Dataset Curation:}~~Like APC, we initially generate a large candidate pool of policies $\Psi$ via random weight initialization. However, rather than using a $k$-NN search in the action space, we propose an information-theoretic selection criterion. We aim to identify policies that maximize the coverage of the feasible state space. To do so, we score each candidate policy based on its contribution to the differential entropy of the mixture occupancy distribution $H(d_\Psi)=\frac 1 M\sum_{i=1}^M \rho^{\text{OPC}}(\vtheta_i)$, where each component is defined as:
\begin{equation}
    \label{eq:opc_score}
    \rho^{\text{OPC}}(\vtheta_i)=\left[ H(d_{\vtheta_i}) + \KL(d_{\vtheta_i} \parallel d_\Psi) \right], \quad \forall \vtheta_i\in\Psi.
\end{equation}
Intuitively, this metric selects policies that exhibit both high individual exploration (high entropy $H(d_{\vtheta_i})$) and are statistically diverse from the population average (high divergence from the mixture $d_\Psi$). To make this computationally tractable for large populations, we approximate each occupancy distribution $d_{\vtheta_i}$ with a Gaussian Mixture Model (GMM) fitted on trajectories collected in a trajectory dataset $\gD_\tau$. A complete derivation and tractable estimators of $\rho$ can be found in \Apref{app:derivation}. The final training dataset $\gD_\Theta$ is curated by selecting the top-scoring policies from $\Psi$ according to $\rho$.

\emph{2. Latent Behavior Compression:}~~In the compression phase, OPC redefines the fundamental theoretical objective of Policy Compression. The original APC framework aimed for strict, 1-to-1 state-action occupancy matching (\Eqqref{eq:apc_latent}). However, forcing a generative model to perfectly reproduce individual policies in isolation is overly restrictive and discourages the discovery of a smooth, continuous behavioral topology. Instead, we argue that the true goal of manifold learning should be to preserve the \codeblue{global state-space coverage} of the dataset as a whole. Therefore, OPC shifts the objective from individual matching to \codeblue{mixture-occupancy matching}. We seek a generative mapping $g^\star$ and a corresponding population of latent codes $Z \in\gZ^{|\Psi|}$ that minimizes the divergence between the mixture occupancy of the original population $\Psi$ and of the reconstructed population $g(Z) = \{g(\vz)\mid\vz\in Z\}$:
\begin{equation}
    \label{eq:obj_latent}
   \exists \;\; Z \in\gZ^{|\Psi|}:\quad g^\star = \argmin_g \KL(d_{\Psi}\parallel d_{g(Z)}).
\end{equation}
By relaxing the strict 1-to-1 reconstruction constraint, this objective allows the generative model to organize the latent space purely around global, task-agnostic behaviors. Furthermore, unlike APC, which falls back on an action-matching proxy to approximate its objective, we can optimize \Eqqref{eq:obj_latent} directly. The autoencoder minimizes the empirical loss $\mathcal{L}_{B}(\vxi, \vzeta) = \KL(d_{\gD_\Theta} \parallel d_{\gD_{\widehat\Theta}})$, where $\gD_{\widehat\Theta}=(g_\vzeta\circ f_\vxi)(\gD_\Theta)$. We achieve this end-to-end by estimating the divergence using the differentiable non-parametric importance sampling $k$-NN estimator defined in \Eqqref{eq:knn_estimator}, evaluated over the trajectory dataset $\gD_\tau$. The core of this estimator relies on the unnormalized importance weight $\overline w_j$ for a given trajectory $\tau_j = (s^j_0, a^j_0, \dots, s^j_T, a^j_T)$, representing the density ratio between the reconstructed and original mixtures, computed as in~\citet{Mutti2021Mepol}:
\begin{equation*} 
 \overline w_j =\frac{\sum_{\hat\vtheta \in \gD_{\widehat\Theta}} p(\tau_j \mid \hat\vtheta)}{\sum_{\vtheta \in \gD_{\Theta}} p(\tau_j \mid \vtheta)} = \frac{\sum_{\hat{\vtheta} \in \mathcal{D}_{\widehat{\Theta}}} \prod_{t=0}^{T} \pi_{\hat{\vtheta}}(a^j_t \mid s^j_t)}{\sum_{\vtheta \in \mathcal{D}_{\Theta}} \prod_{t=0}^{T} \pi_{\vtheta}(a^j_t \mid s^j_t)},
\end{equation*}
where $p(\tau\mid\vtheta)=\mu(s_0)\prod_{t=0}^T \sP(s_{t+1}\mid s_t, a_t)\ \pi_\vtheta(a_t\mid s_t)$ is the probability density of a trajectory and $\hat{\vtheta}$ is the vector of reconstructed parameters. Crucially, because the environment dynamics $\sP$ and initial state distribution $\mu$ are identical for both the original and reconstructed policies, they cancel out entirely. This yields a purely policy-explicit formulation, rendering the loss function fully differentiable with respect to the Autoencoder parameters $\vxi$ and $\vzeta$. The final weights are obtained via self-normalization, $w_j = {\overline{w}_j}/{\sum_{i=1}^N \overline{w}_i}$, and by assigning to each particle $s_i^j$ the weight $w_j$ of the trajectory $\tau_j$ that produced it. A numerically stable formulation of the importance weights can be found in~\Apref{app:IW_computation}. To manage the computational complexity of the $k$-NN estimator over large populations, we adopt a stochastic approach, computing the loss and updating the parameters over mini-batches of policies $\psi \subset \gD_\Theta$ and their corresponding trajectories. The detailed algorithm for training the autoencoder is provided in~\Apref{app:training_pipeline}.

\emph{3. Latent Behavior Optimization:}~~Similar to APC, we employ PGPE to optimize the latent code $\vz$ directly, avoiding analytical computation of the Jacobian of the decoder. To fully exploit the low dimensionality and topologically dense structure of the OPC latent space, we introduce a \codeblue{Latent Warm Start} phase prior to running PGPE. We first sample a small set of latent codes. We then evaluate the reconstructed policies and use the best-performing latent code as a starting point for standard PGPE (\Eqqref{eq:pgpe}). This coarse-to-fine strategy prevents PGPE from converging to poor local optima and accelerates fine-tuning by starting the gradient updates in a high-reward basin of attraction.

In the following section, we empirically validate how these structural enhancements translate to behaviorally richer learned latent manifolds.

%% file: sections/5_experiments.tex
\section{Experiments}
\label{sec:exp}

In this section, we empirically evaluate the proposed framework. We first outline the experimental setup and domains, then systematically address the three research questions introduced in~\Secref{sec:intro}.

\textbf{Experimental Settings.}~~We select continuous control domains that are challenging yet interpretable to clearly illustrate the features of the learned behavioral manifold. The first is \codeblue{Mountain Car Continuous}~\citep[MC,][]{Moore90efficient}. To evaluate the quality of the latent space, we define four downstream tasks: \codeblue{standard} and \codeblue{left} place the goal state on the right and left hill, respectively; \codeblue{speed} and \codeblue{height} incentivize the car to maintain high velocity and elevation ($z$-coordinate), respectively, without terminating the episode. We also consider two MuJoCo environments~\citep[][]{todorov2012mujoco}. In \codeblue{Reacher} (RC), a two-jointed robotic arm moving in a 2D plane, we remove target observations and define four tasks based on fingertip kinematics: \codeblue{speed} promotes high linear velocity; \codeblue{clockwise} and \codeblue{c-clockwise} promote sustained rotation; and \codeblue{radial} promotes fast arm extension and retraction. Finally, in \codeblue{Hopper} (HP), we evaluate locomotion via four downstream tasks: \codeblue{forward}, \codeblue{backward}, and \codeblue{standstill} reward the agent for positive, negative, and zero velocity along the $x$-axis, respectively, while \codeblue{standard} corresponds to the original Gymnasium composite reward. For all experiments, we generate a dataset of $50,000$ policies and retain the top 5\% most informative ones. Policies are fully-connected neural networks with $\approx 10^3$ parameters. Full experimental details are provided in~\Apref{app:experimental_settings}.

\begin{figure}[t]
    \vspace{-0.4cm}
    \centering 
    \includegraphics[width=0.4\textwidth, height=0.5cm]{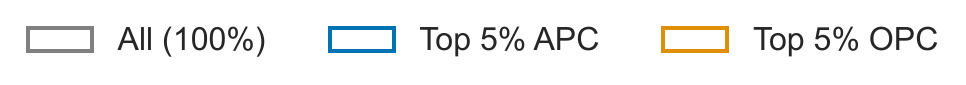}
    
    \begin{subfigure}[b]{0.24\textwidth}
        \includegraphics[width=\textwidth]{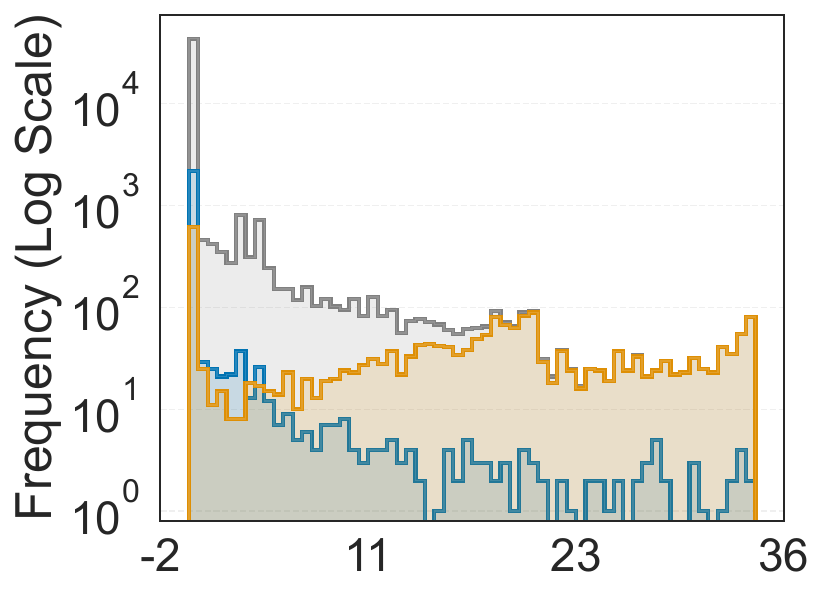}
        \vspace{-0.6cm} \caption{\codeblue{speed}}
        \label{fig:main_curation_a}
    \end{subfigure}
    \hfill 
    \begin{subfigure}[b]{0.24\textwidth}
        \includegraphics[width=\textwidth]{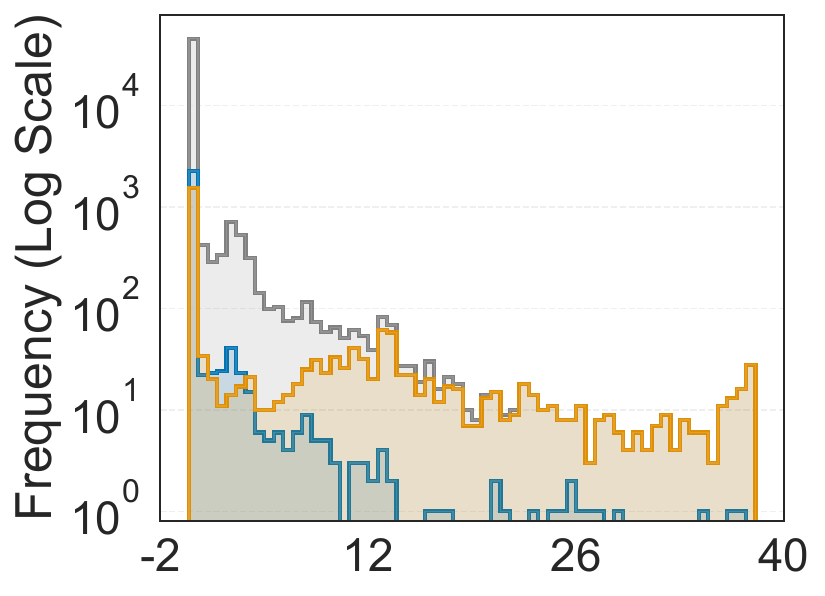}
        \vspace{-0.6cm}\caption{\codeblue{clockwise}}
        \label{fig:main_curation_b}
    \end{subfigure}
    \hfill
    \begin{subfigure}[b]{0.24\textwidth}
        \includegraphics[width=\textwidth]{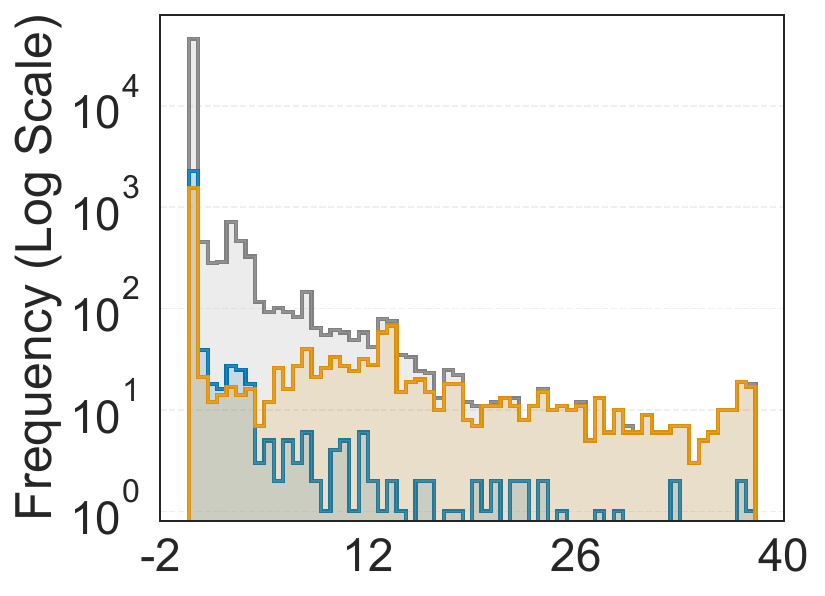}
        \vspace{-0.6cm}\caption{\codeblue{c-clockwise}}
        \label{fig:main_curation_c}
    \end{subfigure}
    \hfill
    \begin{subfigure}[b]{0.24\textwidth}
        \includegraphics[width=\textwidth]{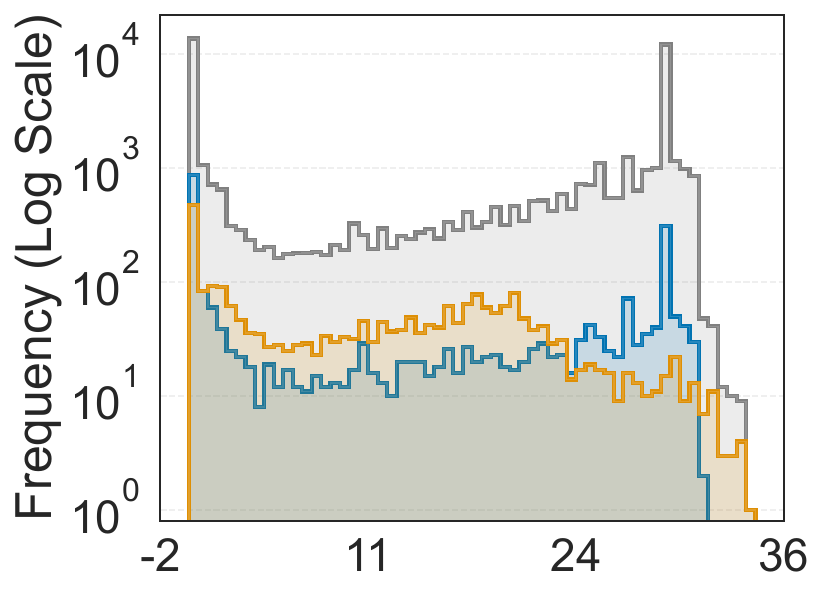}
        \vspace{-0.6cm}\caption{\codeblue{radial}}
        \label{fig:main_curation_d}
    \end{subfigure}

    \begin{subfigure}[b]{0.24\textwidth}
        \includegraphics[width=\textwidth]{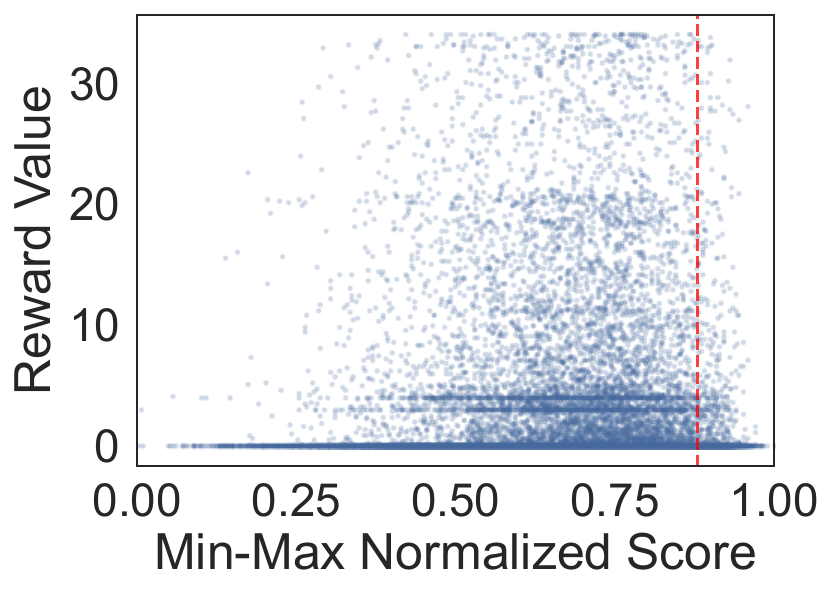}
        \vspace{-0.6cm} \caption{\codeblue{speed}, APC}
        \label{fig:main_curation_e}
    \end{subfigure}
    \hfill 
    \begin{subfigure}[b]{0.24\textwidth}
        \includegraphics[width=\textwidth]{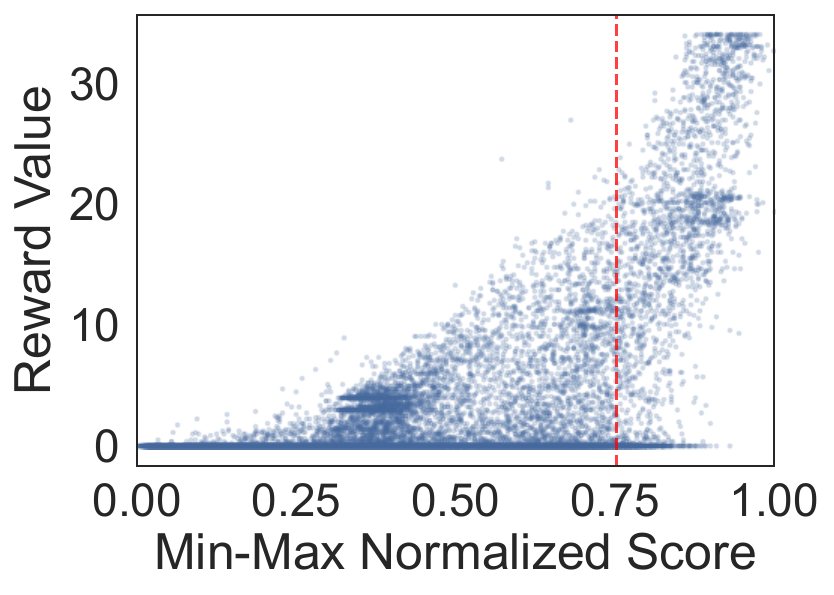}
        \vspace{-0.6cm}\caption{\codeblue{speed}, OPC}
        \label{fig:main_curation_f}
    \end{subfigure}
    \hfill
    \begin{subfigure}[b]{0.24\textwidth}
        \includegraphics[width=\textwidth]{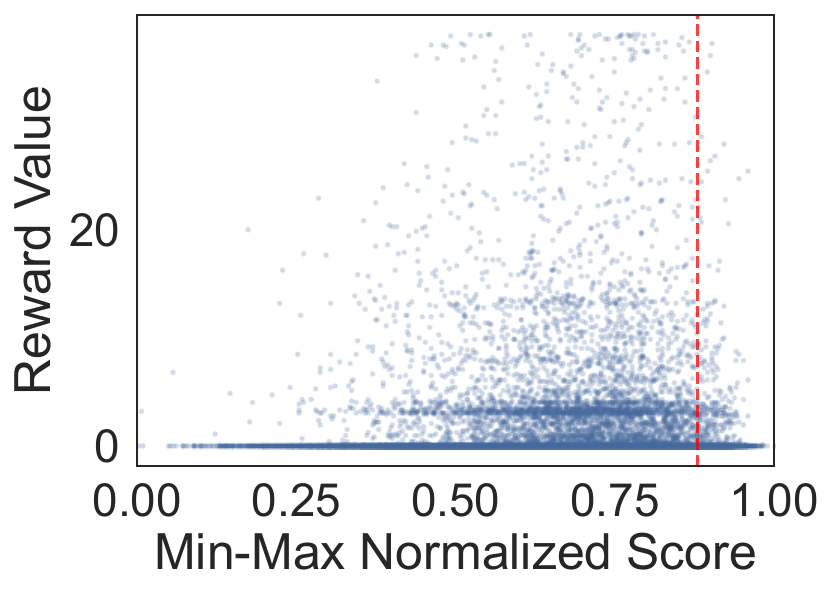}
        \vspace{-0.6cm}\caption{\codeblue{clockwise}, APC}
        \label{fig:main_curation_g}
    \end{subfigure}
    \hfill
    \begin{subfigure}[b]{0.24\textwidth}
        \includegraphics[width=\textwidth]{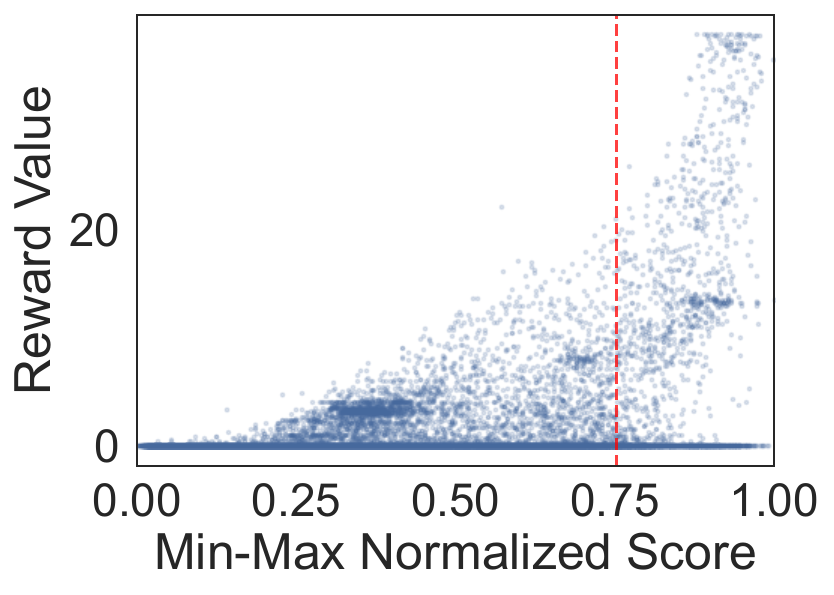}
        \vspace{-0.6cm}\caption{\codeblue{clockwise}, OPC}
        \label{fig:main_curation_h}
    \end{subfigure}

    \begin{minipage}{\textwidth} 
        \centering
        \caption{
        Comparison of APC and OPC thresholding in the RC environment: reward distribution in the curated datasets (top row, frequency is on a log scale) and score distribution comparison (bottom row, scores are min-max normalized). The vertical dashed line marks the 5\% cut-off.}
        \label{fig:main_curation}
    \end{minipage}
\end{figure}

\noindent\begin{minipage}[t]{0.45\linewidth}
\subsection{Policy Dataset Curation}
\label{sec:exp_curation}
\end{minipage}%
\hspace{-35pt}%
\begin{minipage}[t]{0.65\linewidth}
\vspace{-9pt}
\begin{tcolorbox}[colback=softbluegray, colframe=softbluegray, boxrule=0.2pt, arc=4pt, width=\linewidth]
\textbf{Q1:} On the role of curating policies for improved datasets.
\end{tcolorbox}
\end{minipage}

To answer \textbf{Q1}, we analyze the score distributions assigned to the policies and the reward distributions of the resulting datasets obtained during the curation phase. We compare the reward distribution of the original random dataset ($100\%$) against a 5\% thresholding according to both scores $\rho^{\text{APC}}$ (\Eqqref{eq:apc_score}) and  $\rho^{\text{OPC}}$ (\Eqqref{eq:opc_score}). We also map the min--max normalized scores against the rewards across all tasks and environments. The core results for the RC environment are presented in~\Figgref{fig:main_curation}, while the extended experimental results for all other domains are detailed in~\Apref{app:add_curation}.

As illustrated in~\Figgref{fig:main_curation_a},\ref{fig:main_curation_b},\ref{fig:main_curation_c},\ref{fig:main_curation_d}, the random initialization process naturally yields a heavy-tailed distribution where high-reward behaviors are exceedingly rare; this phenomenon is exacerbated in complex control tasks such as Hopper. The APC novelty search, relying strictly on immediate action differences, frequently discards these rare behavioral modes. In contrast, the OPC curation successfully isolates and preserves these critical outlier policies, maintaining a significantly heavier tail in the high-reward regions across the evaluated tasks. This effect is explained by the score distributions: while scores $\rho^{\text{APC}}$ (\Figgref{fig:main_curation_e},\ref{fig:main_curation_g}) are generally uncorrelated with the actual task rewards or global behavioral uniqueness, scores $\rho^{\text{OPC}}$ (\Figgref{fig:main_curation_f},\ref{fig:main_curation_h}) are highly structured, ensuring that the most functionally unique policies consistently fall within the top percentile of the scoring distribution.

\begin{figure}[t]
\vspace{-0.4cm}
    \centering
    \newcommand{\imgwidth}{0.19\textwidth}
    \setlength{\tabcolsep}{8pt}
    \renewcommand{\arraystretch}{1.4}
    \begin{tabular}{m{5mm} c c c c}
        & \multicolumn{2}{c}{APC score} & \multicolumn{2}{c}{OPC score} \\
        \cmidrule(lr){2-3} \cmidrule(lr){4-5}
        & APC loss & OPC loss & APC loss & OPC loss \\[-.6ex]
        
        \rotatebox{90}{\hspace{0pt}\codeblue{Standard}} &
        \includegraphics[trim=0 0 0 80, clip, width=\imgwidth, valign=m]{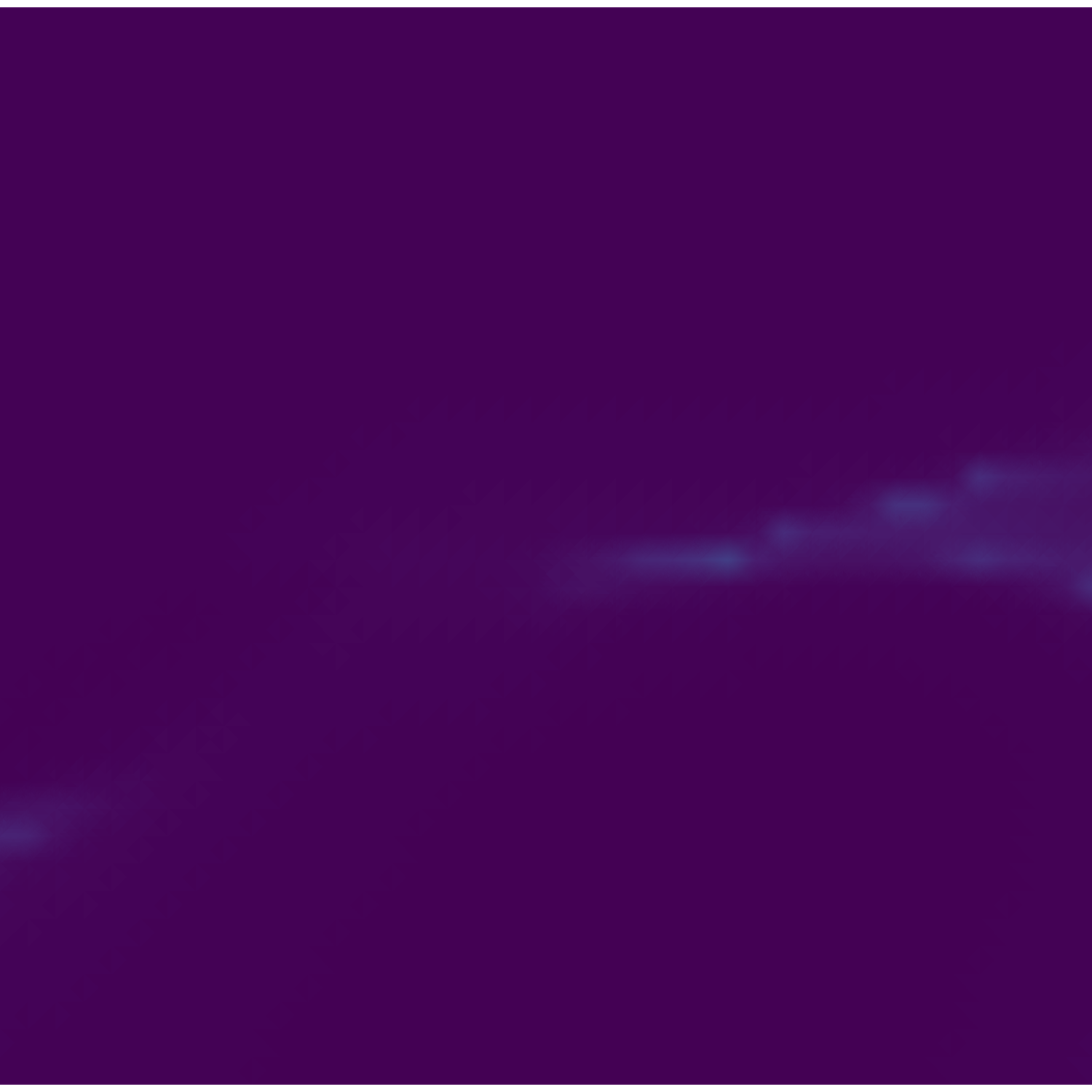} &
        \includegraphics[trim=0 0 0 85, clip, width=\imgwidth, valign=m]{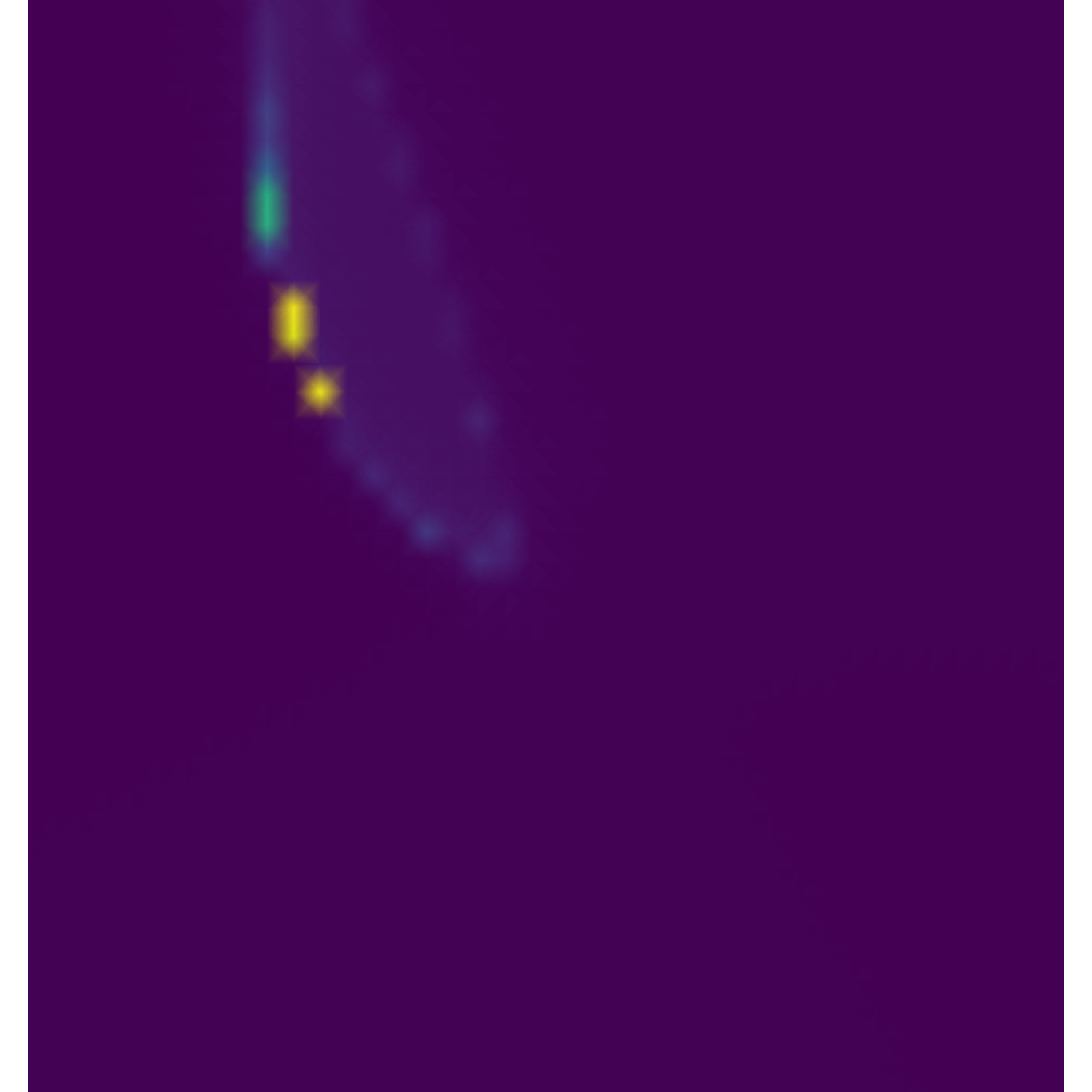} &
        \includegraphics[trim=0 0 0 85, clip, width=\imgwidth, valign=m]{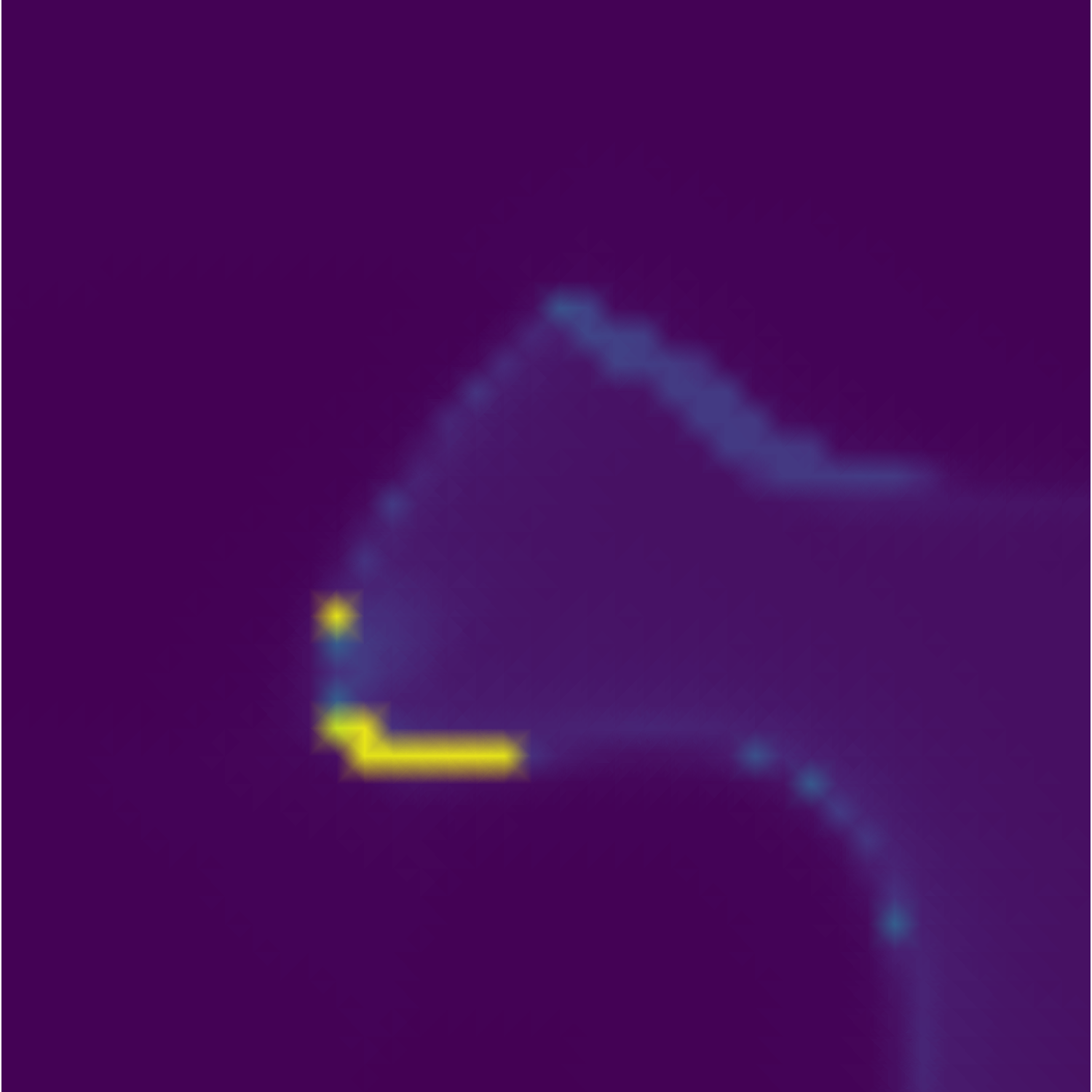} &
        \includegraphics[trim=0 0 0 80, clip, width=\imgwidth, valign=m]{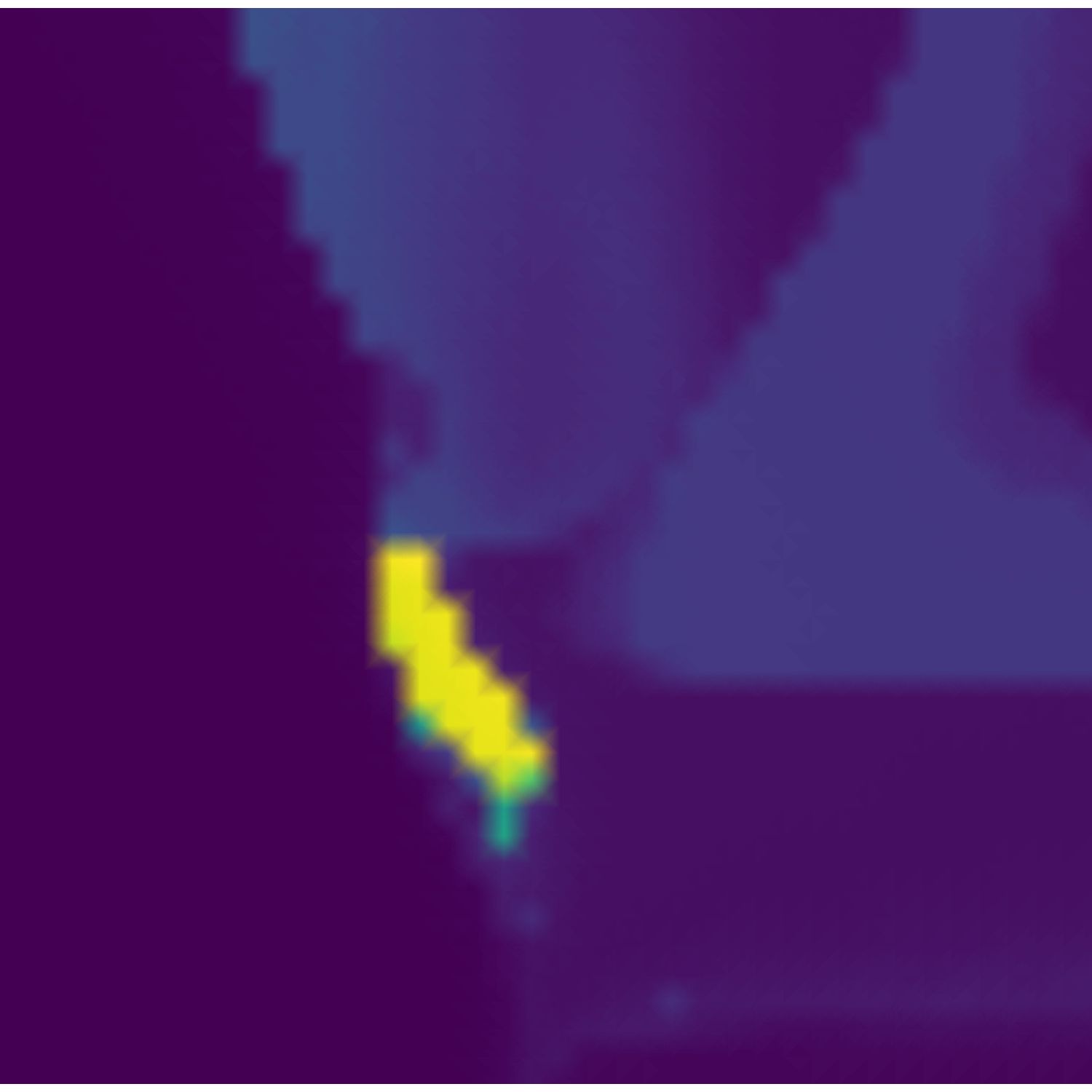} \\[1ex]
        
        \rotatebox{90}{\hspace{0pt}\codeblue{Forward}} &
        \includegraphics[trim=0 75 0 75, clip, width=\imgwidth, valign=m]{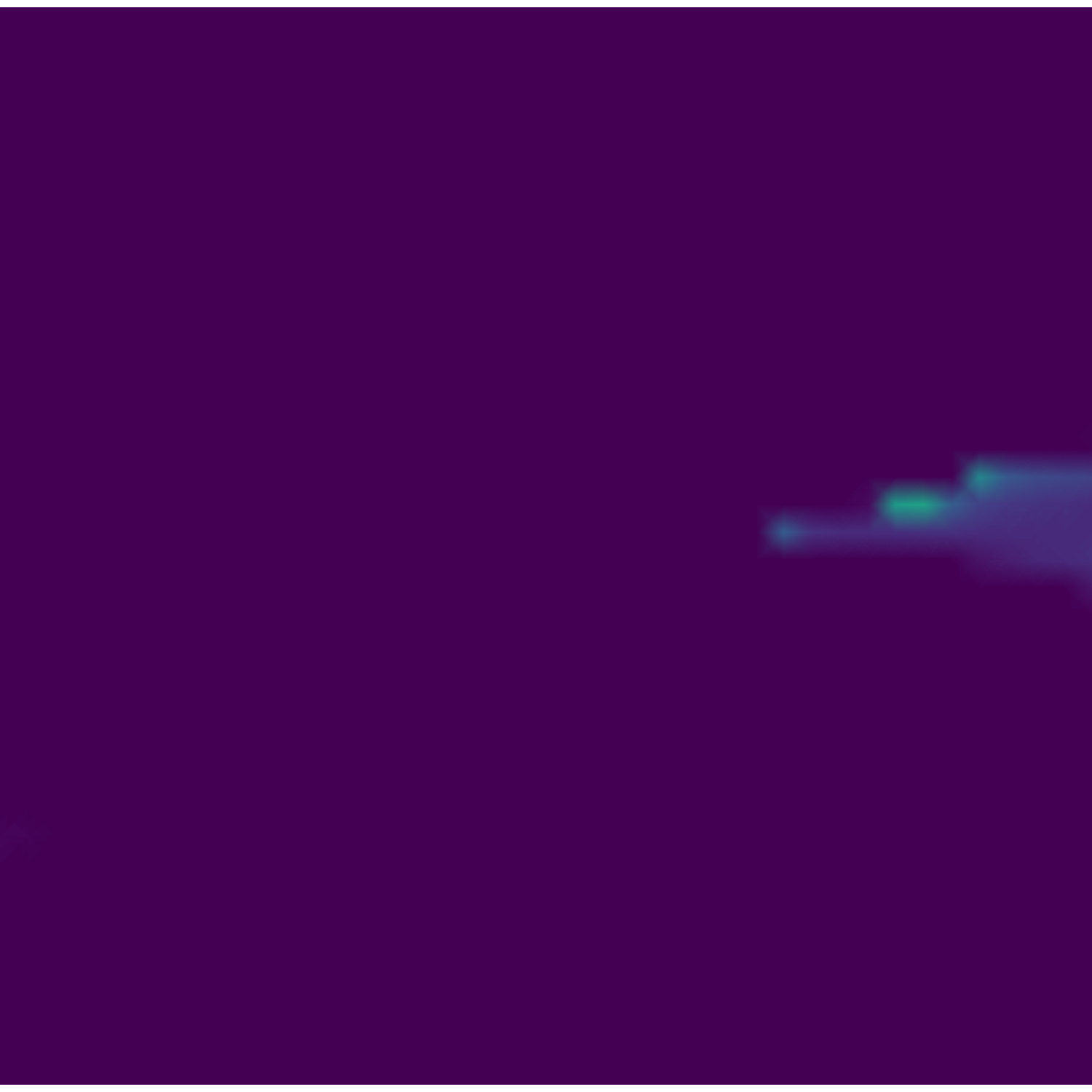} &
        \includegraphics[trim=0 80 0 75, clip, width=\imgwidth, valign=m]{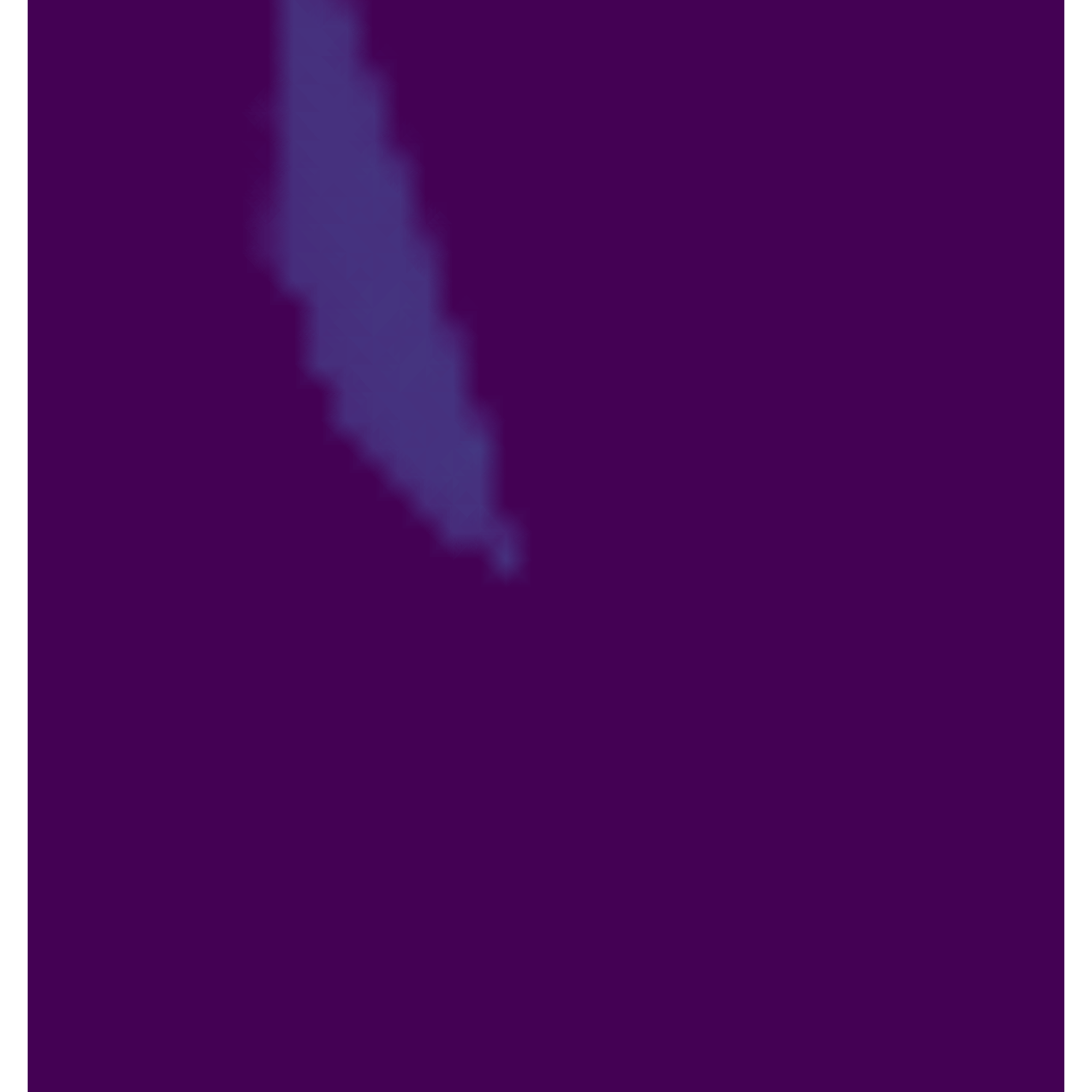} &
        \includegraphics[trim=0 80 0 75, clip, width=\imgwidth, valign=m]{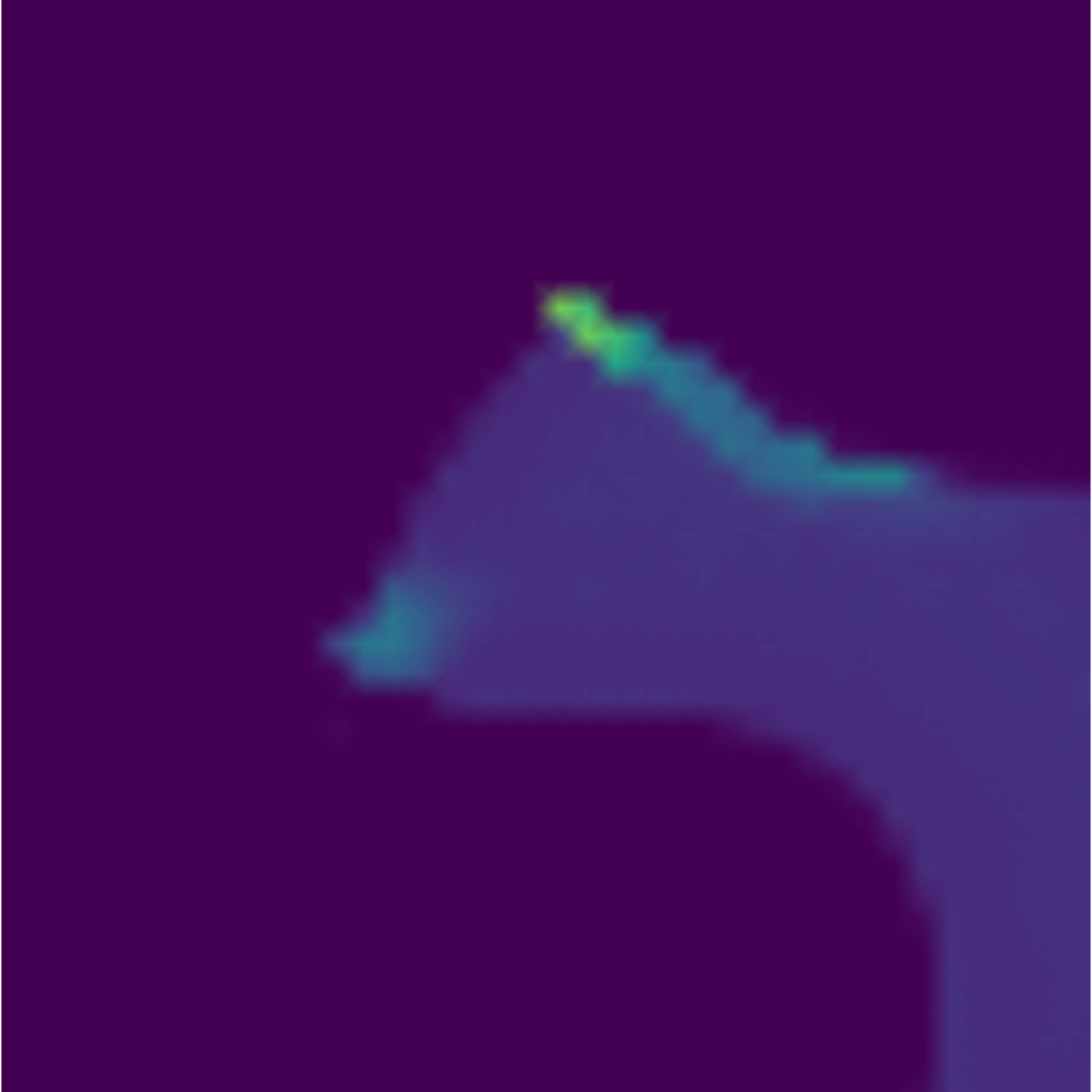} &
        \includegraphics[trim=0 75 0 75, clip, width=\imgwidth, valign=m]{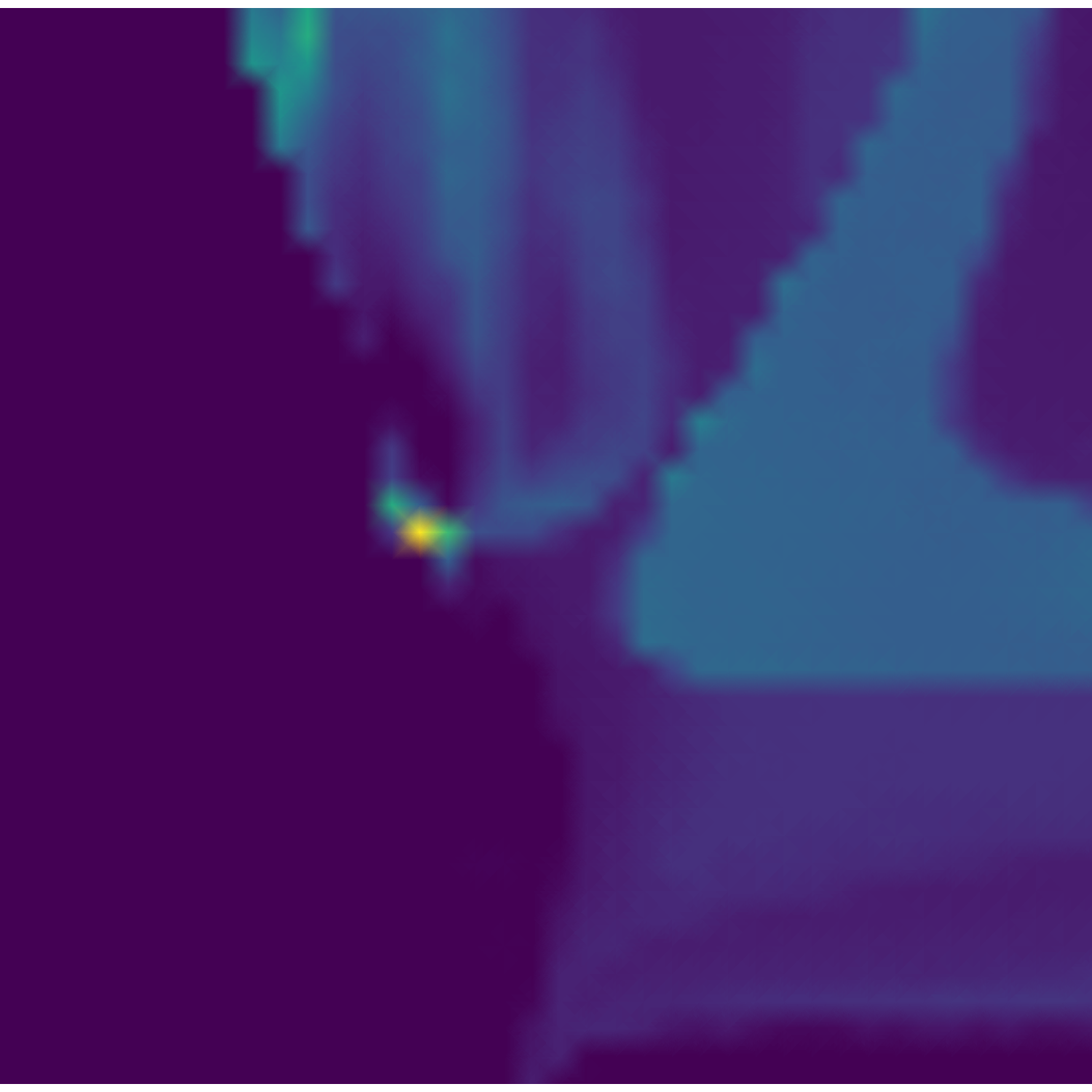} \\
    \end{tabular}
    
    \caption{Ablation study comparing Latent Space topologies in the HP environment. Rows represent reward functions, while columns represent different score/loss combinations. \textcolor{myyellow}{\textbf{Lighter}} and \textcolor{myblue}{\textbf{darker}} colors indicate higher and lower returns of the decoded policy.}
    \label{fig:ablation_study}
\end{figure}

\noindent\begin{minipage}[t]{0.45\linewidth}
\subsection{Latent Behavior Compression}
\label{sec:exp_compression}
\end{minipage}%
\hspace{-20pt}%
\begin{minipage}[t]{0.60\linewidth}
\vspace{-9pt}
\begin{tcolorbox}[colback=softbluegray, colframe=softbluegray, boxrule=0.2pt, arc=4pt, width=\linewidth]
\textbf{Q2:} On the role of OPC in the quality of  latent spaces.
\end{tcolorbox}
\end{minipage}

To answer \textbf{Q2}, we perform a qualitative visual inspection of the latent spaces generated by both APC and OPC pipelines. We conduct an ablation study to assess the effects of the occupancy-based components (uniqueness scores and the compression objective). We focus on the HP environment, as it offers a more complex locomotion task. We fix the AE bottleneck size to 2 to provide clear visualizations. To achieve better performances in \Secref{sec:exp_optimization} we increase the bottleneck sizes to match the complexity of the HP environment. The full ablation study is reported in~\Apref{app:add_compression}.

In \Figgref{fig:ablation_study}, we show 2D latent embeddings colored by task performance. We sample a uniform grid in the latent space and evaluate the reconstructed policies. The fully action-based baseline (\textbf{APC score \& APC loss}) fails to recover a meaningful topological structure, resulting in an unstructured space dominated by low-performing behaviors. Evaluating the components in isolation reveals their individual limitations. Introducing only the occupancy-based objective (\textbf{APC score \& OPC loss}) yields a smoother geometric gradient, but the lack of diverse high-performing policies in the APC-curated dataset restricts the presence of high-reward regions. Conversely, using the occupancy-based metric for curation with the action-based objective (\textbf{OPC score \& APC loss}) successfully introduces high-reward policies into the latent space, but the action-matching loss scatters them erratically, failing to form cohesive clusters. The full occupancy-based pipeline (\textbf{OPC score \& OPC loss}) effectively synergizes both components. The latent space is organized into a continuous, traversable geometry, with high-reward policies densely clustered in cohesive, structured regions.

\begin{figure}[p]
    \centering 
    \vspace{-0.5cm}
\noindent
\includegraphics[width=0.95\textwidth, height=0.4cm]{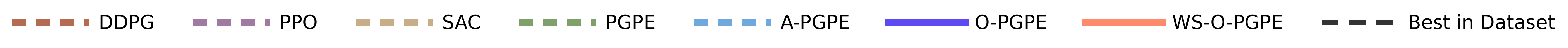}

    \begin{subfigure}[b]{0.48\textwidth}
        \includegraphics[width=\textwidth]{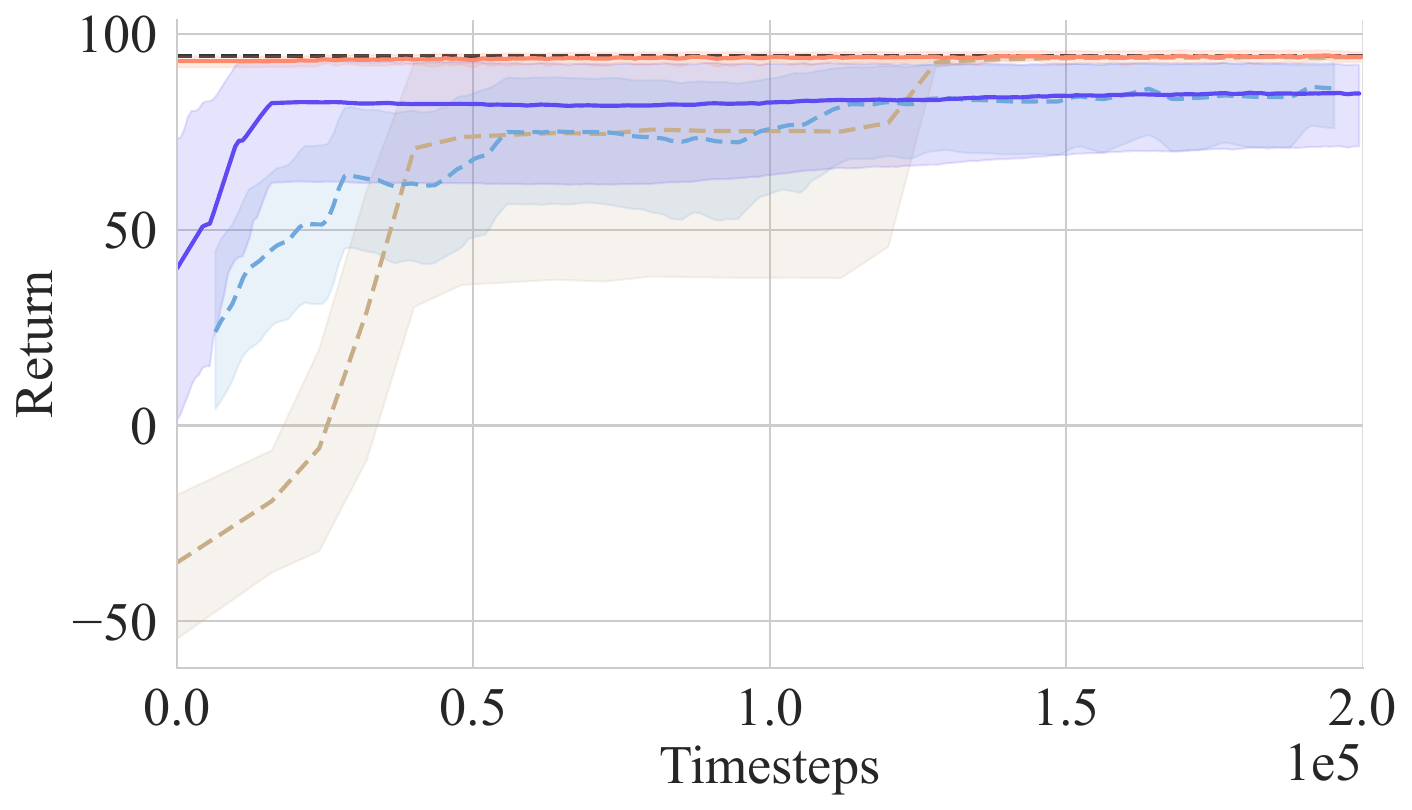}
        \vspace{-0.6cm} \caption{MC, \codeblue{standard}}
        \label{fig:pgpe_a}
    \end{subfigure}
    \hfill
    \begin{subfigure}[b]{0.48\textwidth}
        \includegraphics[width=\textwidth]{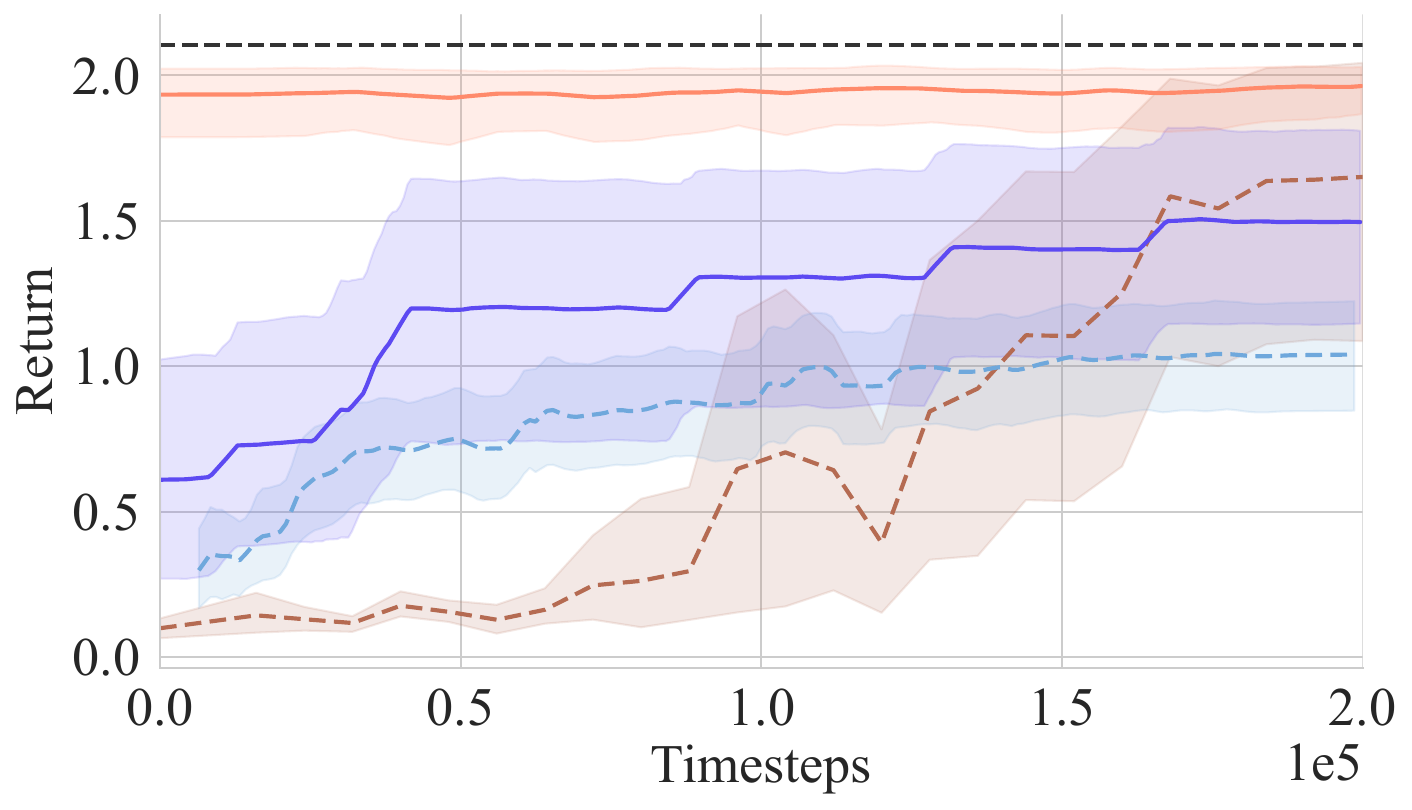}
        \vspace{-0.6cm}\caption{MC, \codeblue{speed}}
        \label{fig:pgpe_c}
    \end{subfigure}
    
    \begin{subfigure}[b]{0.48\textwidth}
        \includegraphics[width=\textwidth]{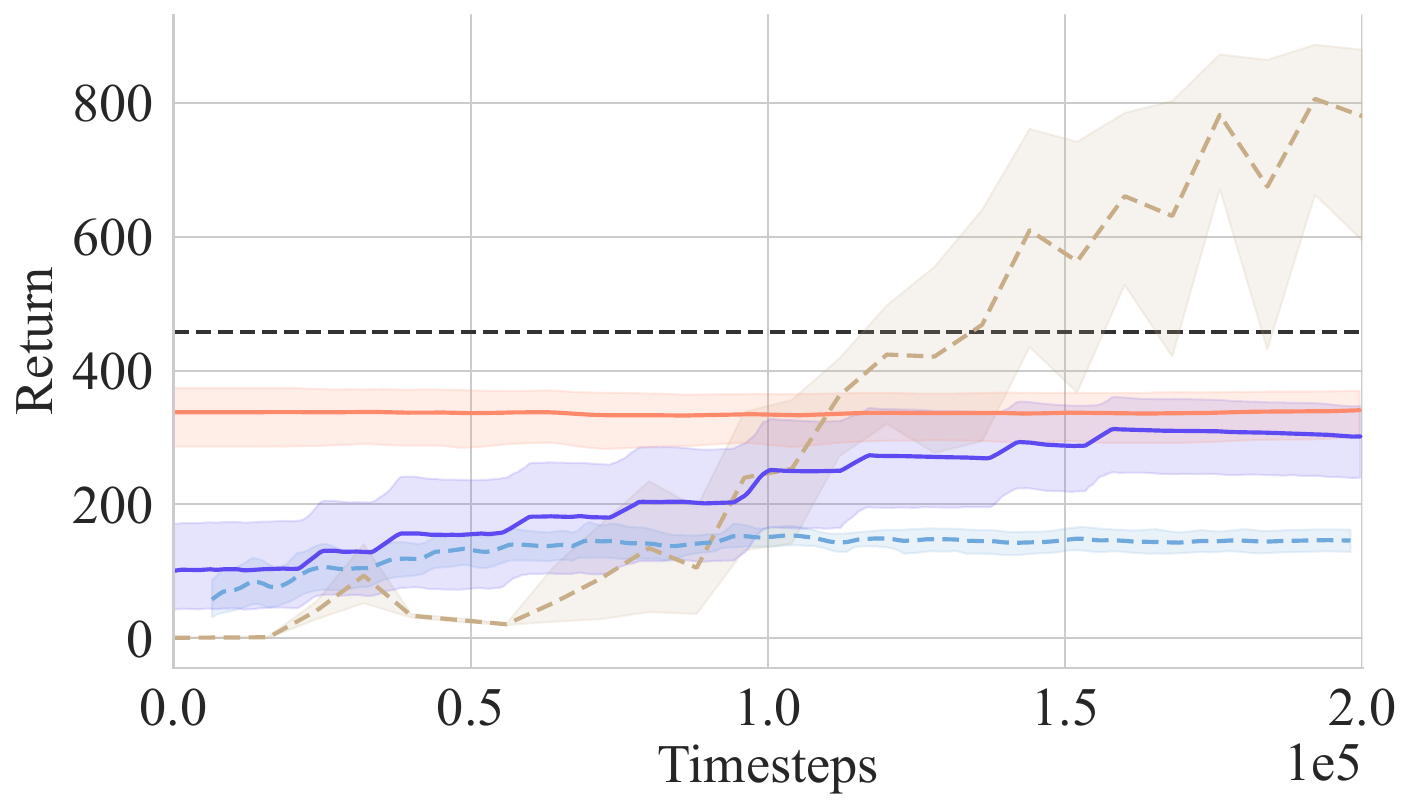}
        \vspace{-0.6cm}\caption{MC, \codeblue{height}}
        \label{fig:pgpe_d}
    \end{subfigure}
    \hfill
    \begin{subfigure}[b]{0.48\textwidth}
        \includegraphics[width=\textwidth]{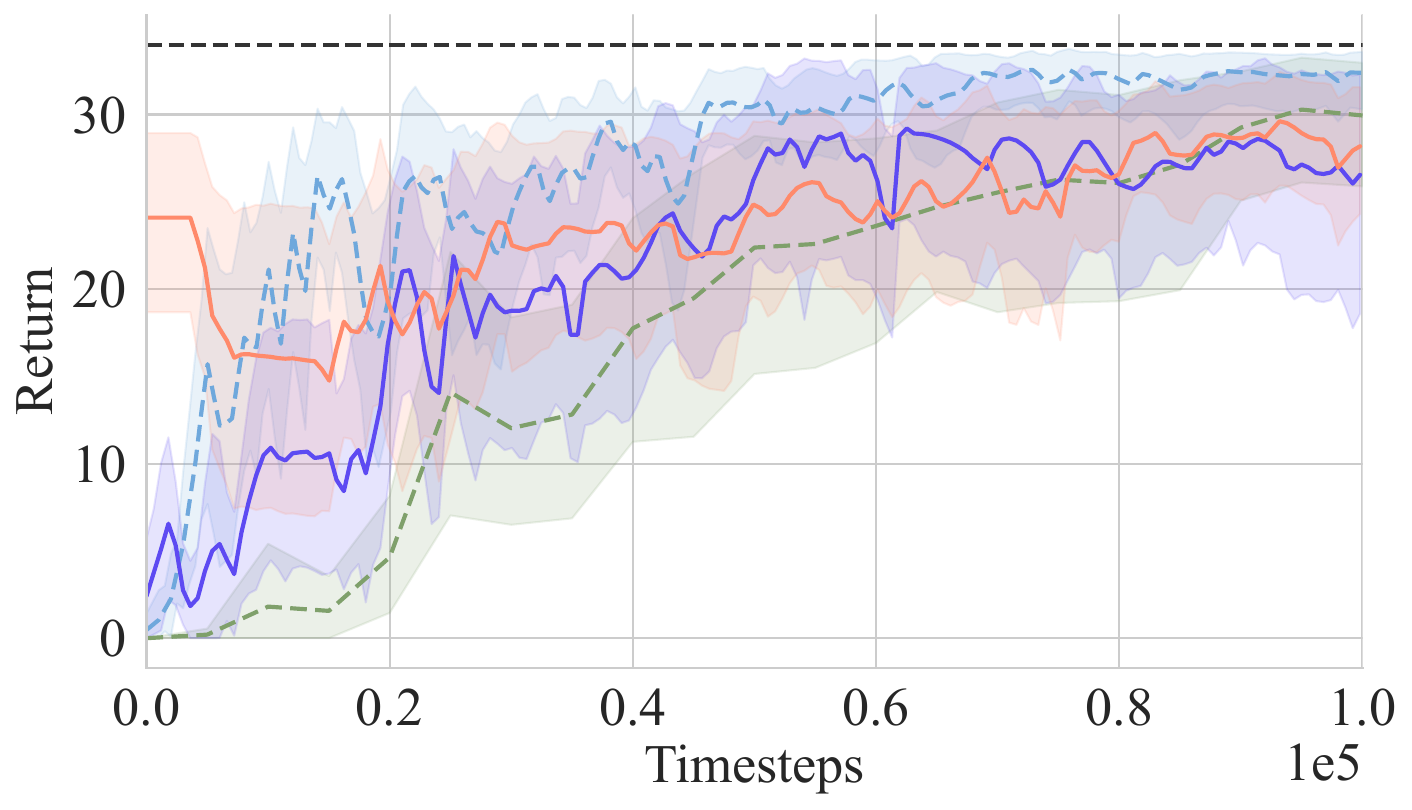}
        \vspace{-0.6cm} \caption{RC, \codeblue{speed}}
        \label{fig:pgpe_e}
    \end{subfigure}

    \begin{subfigure}[b]{0.48\textwidth}
        \includegraphics[width=\textwidth]{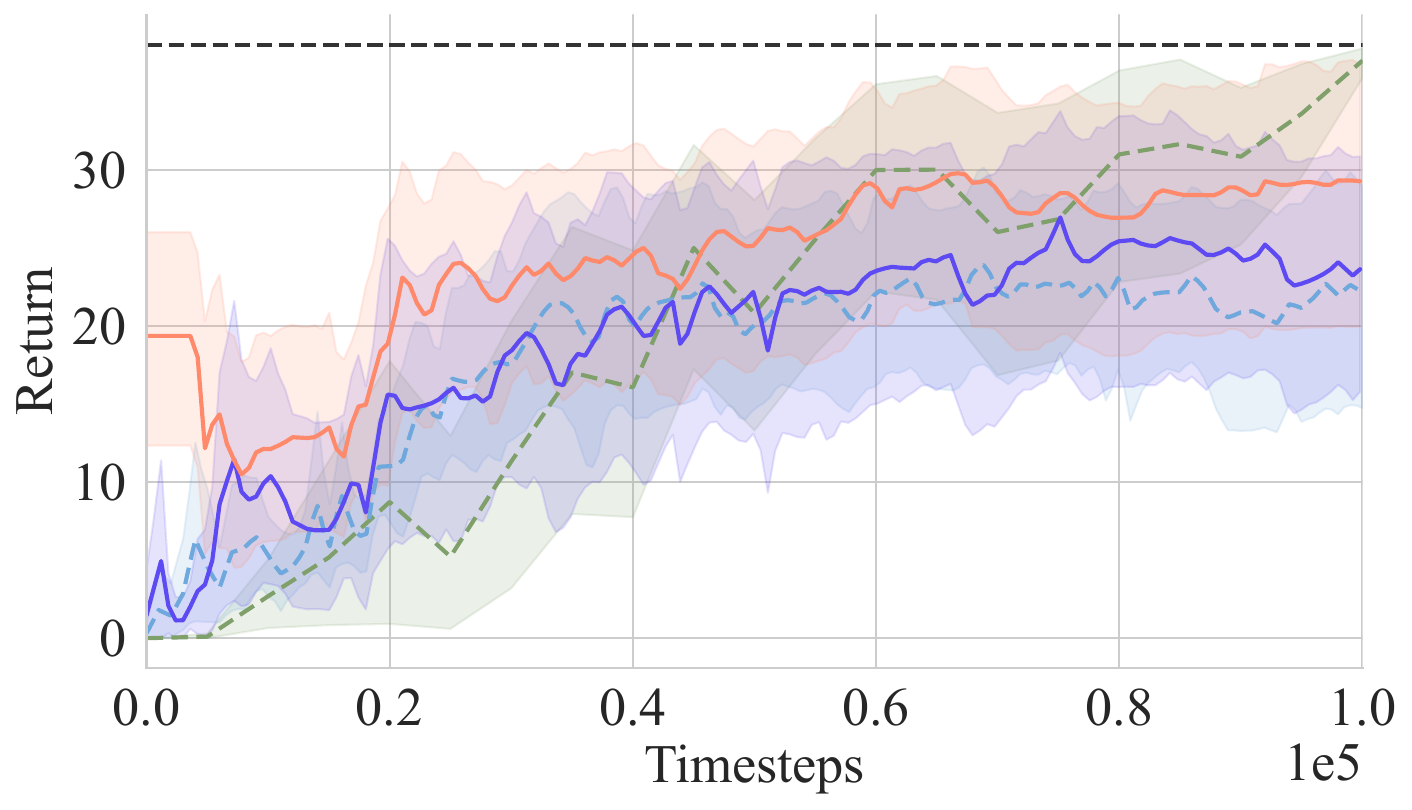}
        \vspace{-0.6cm}\caption{RC, \codeblue{clockwise}}
        \label{fig:pgpe_f}
    \end{subfigure}
    \hfill
    \begin{subfigure}[b]{0.48\textwidth}
        \includegraphics[width=\textwidth]{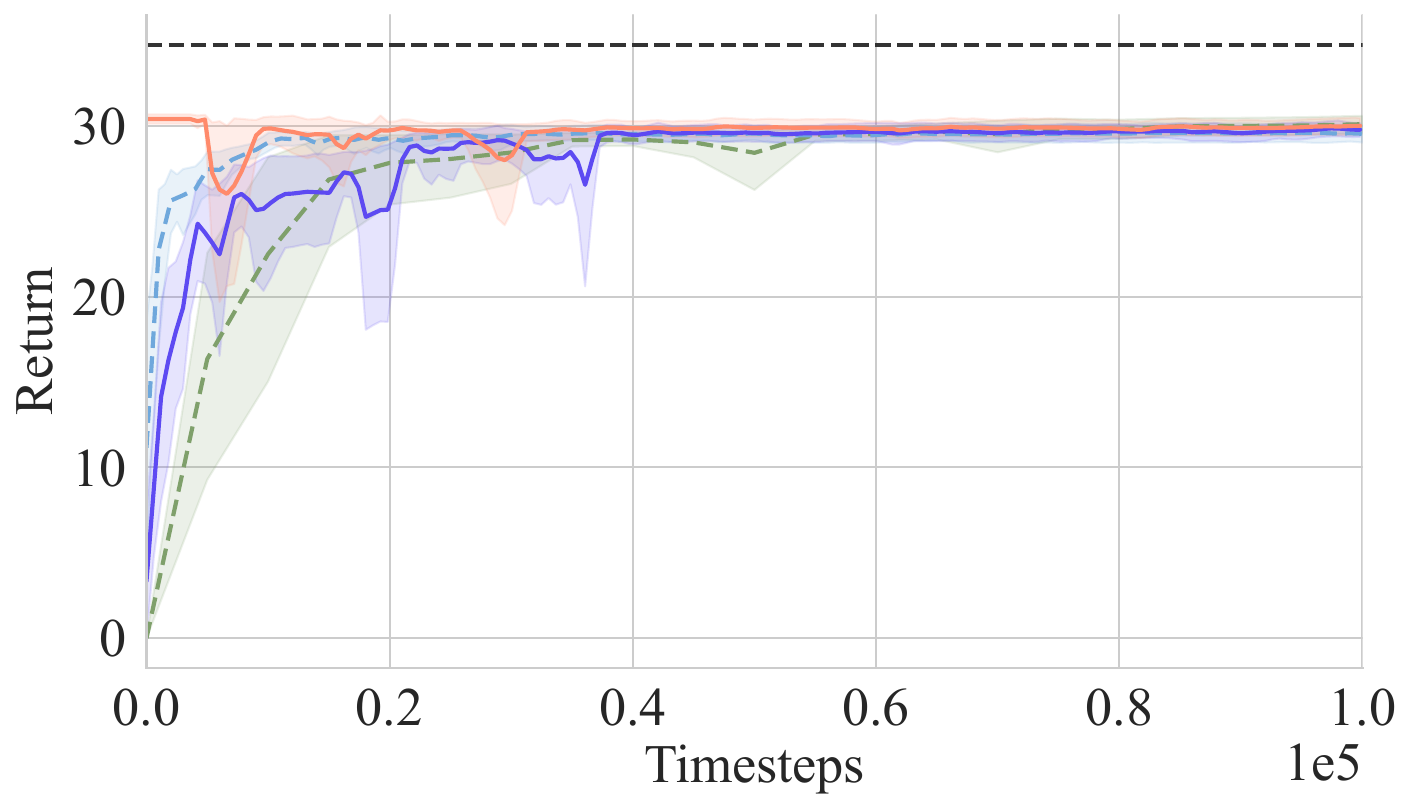}
        \vspace{-0.6cm}\caption{RC, \codeblue{radial}}
        \label{fig:pgpe_h}
    \end{subfigure}

    \begin{subfigure}[b]{0.48\textwidth}
        \includegraphics[width=\textwidth]{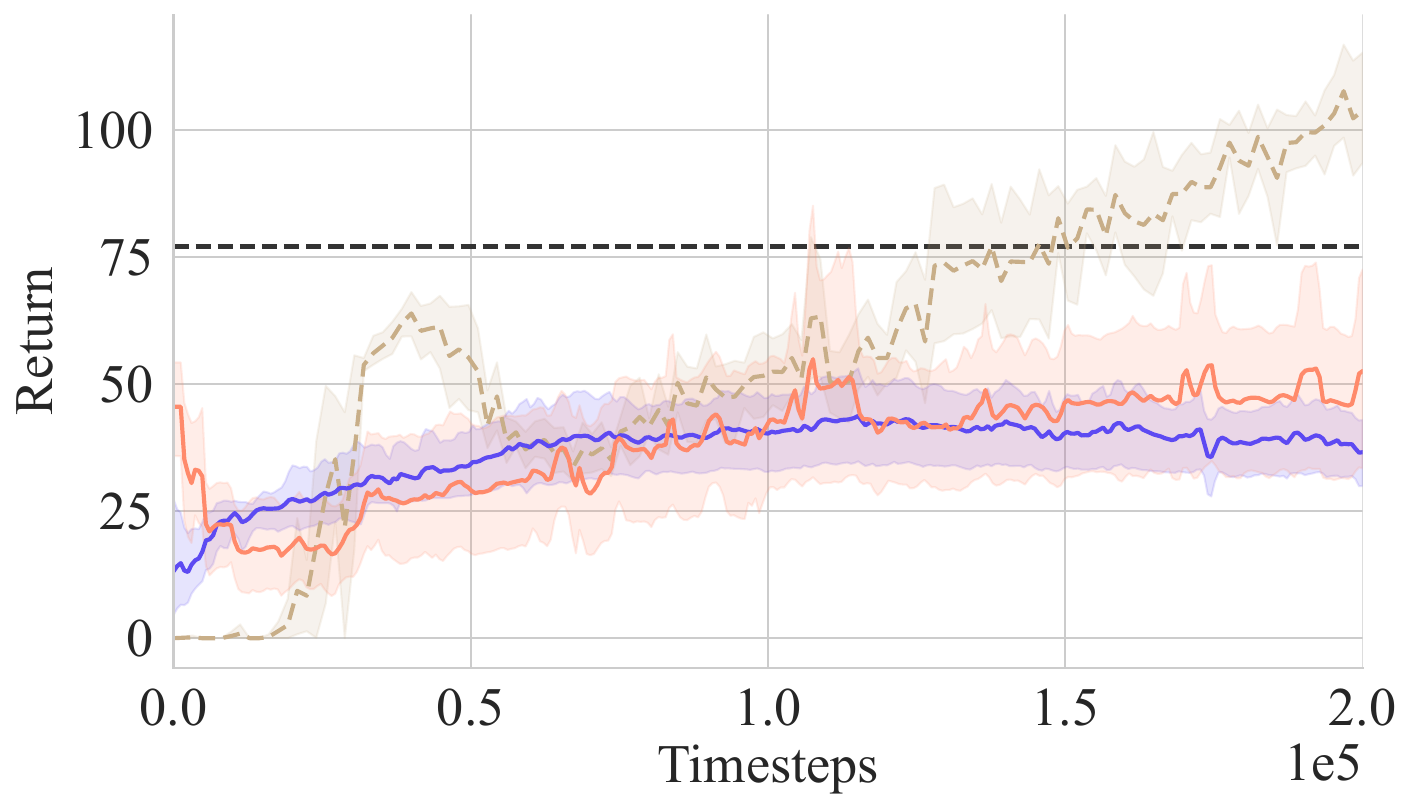}
        \vspace{-0.6cm}\caption{HP, \codeblue{forward}}
        \label{fig:pgpe_i}
        \vspace{-0.1cm}
    \end{subfigure}
    \hfill 
    \begin{subfigure}[b]{0.48\textwidth}
        \includegraphics[width=\textwidth]{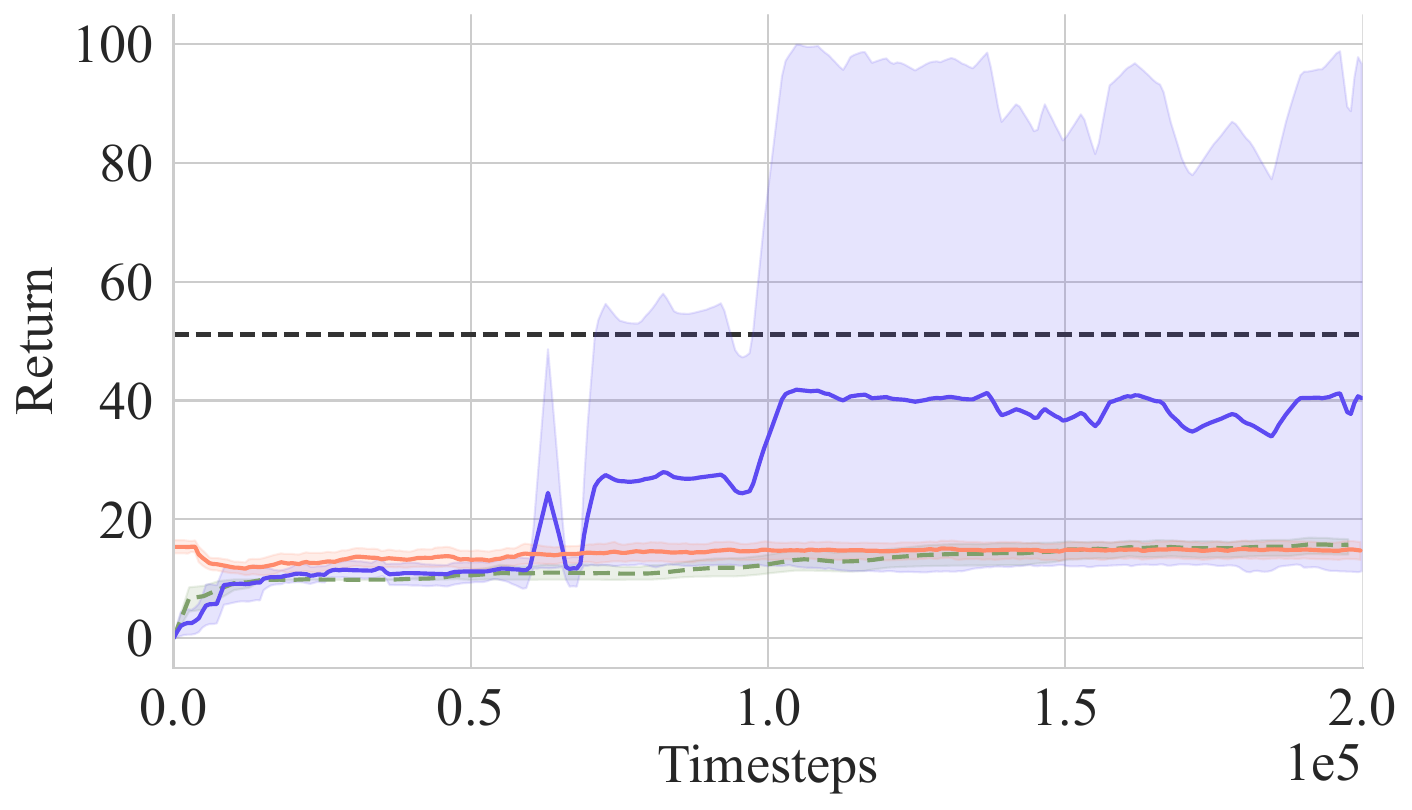}
        \vspace{-0.6cm}\caption{HP, \codeblue{backward}}
        \label{fig:pgpe_j}
        \vspace{-0.1cm}
    \end{subfigure}

    \begin{subfigure}[b]{0.48\textwidth}
        \includegraphics[width=\textwidth]{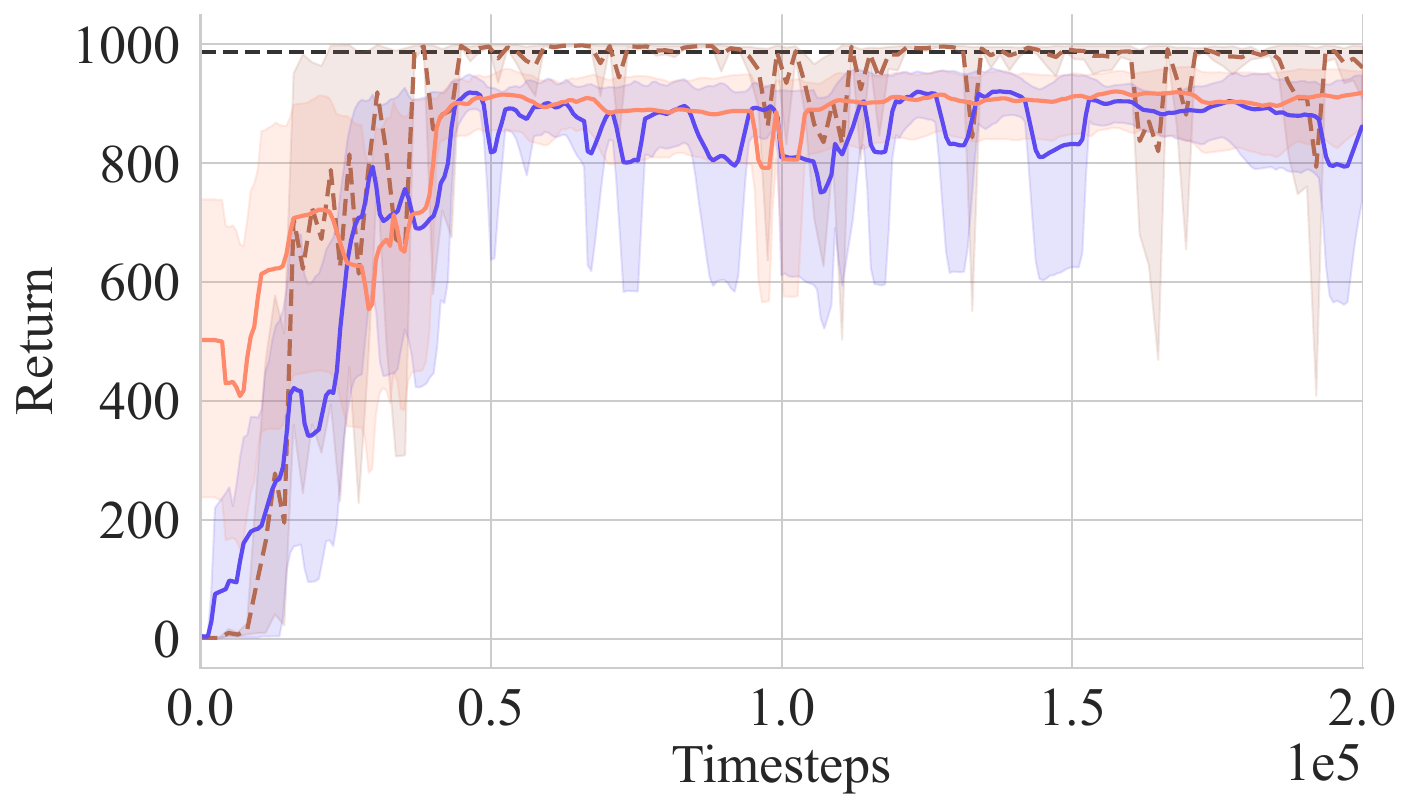}
        \vspace{-0.6cm}\caption{HP, \codeblue{standstill}}
        \label{fig:pgpe_k}
        \vspace{-0.1cm}
    \end{subfigure}
    \hfill
    \begin{subfigure}[b]{0.48\textwidth}
        \includegraphics[width=\textwidth]{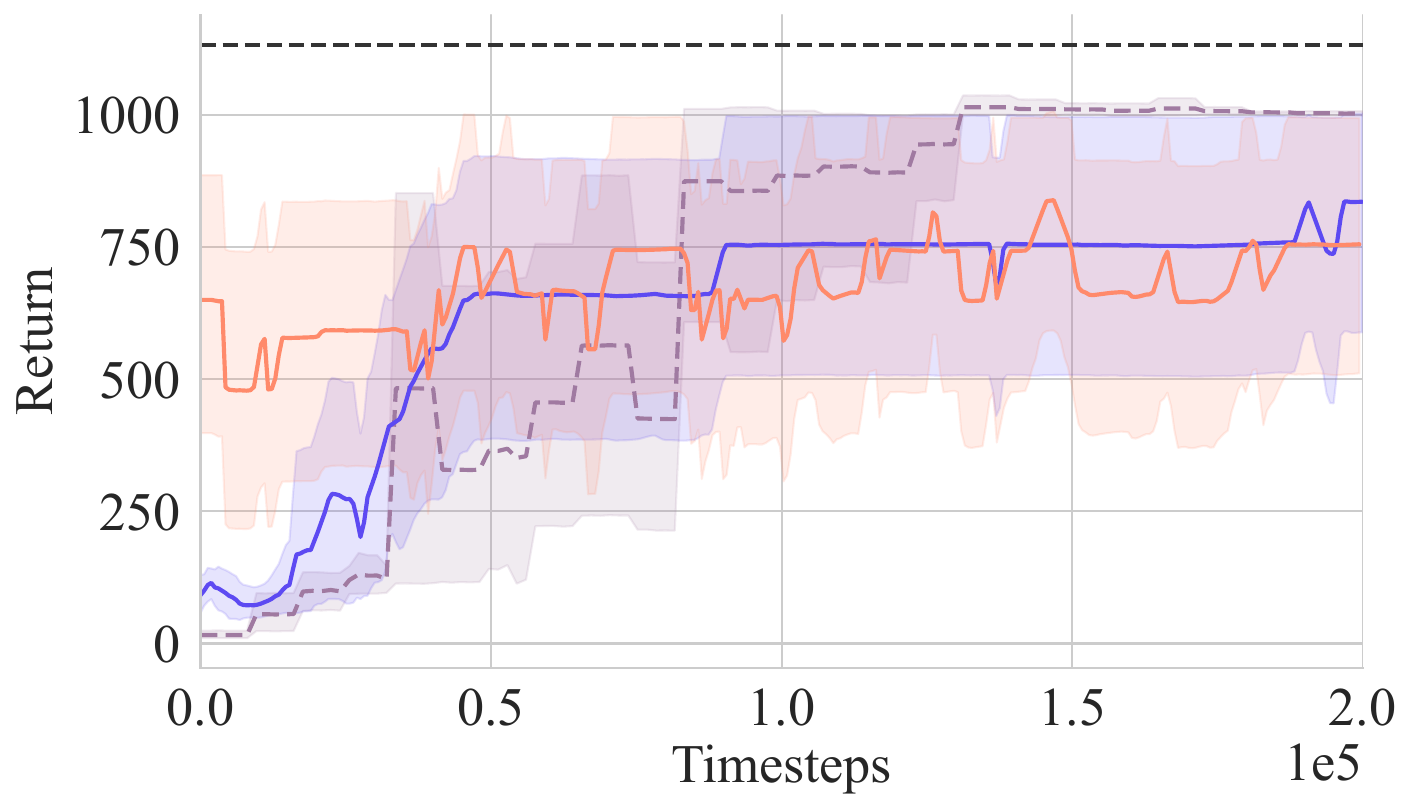}
        \vspace{-0.6cm}\caption{HP, \codeblue{standard}}
        \label{fig:pgpe_l}
        \vspace{-0.1cm}
    \end{subfigure}
    \centering
    \caption{
        Performance comparison in MC, RC, and HP  across different tasks. We report the average and 95\% confidence interval over 10 runs of the best-performing algorithms for each task.
    }
    \label{fig:pgpe}
\end{figure}

\noindent\begin{minipage}[t]{0.45\linewidth}
\subsection{Latent Behavior Optimization}
\label{sec:exp_optimization}
\end{minipage}%
\hspace{-20pt}%
\begin{minipage}[t]{0.60\linewidth}
\vspace{-9pt}
\begin{tcolorbox}[colback=softbluegray, colframe=softbluegray, boxrule=0.2pt, arc=4pt, width=\linewidth]
\textbf{Q3:} On the role of OPC for latent optimization.
\end{tcolorbox}
\end{minipage}

To answer \textbf{Q3}, we evaluate the downstream optimization performance. Following the experimental setting in~\citet{tenedini2025from}, we compare the performance of PGPE~\citep{sehnke2008policy} operating in the parameter space, in the latent space derived from the APC pipeline (A-PGPE), and in the latent space derived from our OPC pipeline (O-PGPE) with various baselines, including DDPG~\citep{lillicrap2015continuous}, TD3~\citep{fujimoto2018addressing}, and SAC~\citep{haarnoja2018soft}. We perform the optimization over all tasks and environments, and across several latent dimensions. The results can be observed in \Figgref{fig:pgpe}. For each task, we plot only the best curve for each family of algorithms: A-PGPE, O-PGPE, WS-O-PGPE, and the baselines. We also show the best-performing policy in the training dataset. The complete collection of training curves for all algorithms can be found in~\Apref{app:add_optimization}.

As shown in \Figgref{fig:pgpe}, optimizing over the occupancy-matched latent space (O-PGPE) yields sample efficiency and final returns comparable to those of standard DRL baselines and action-matched manifolds (A-PGPE). Incorporating the Latent Warm Start (WS-O-PGPE) often accelerates the initial learning phase by initializing the agent in a high-reward region. For example, in MC, using Warm Start seemingly transforms the problem into a zero-shot optimization, superseding all other methods (\Figgref{fig:pgpe_a},\ref{fig:pgpe_c}), except for the task \codeblue{height} (\Figgref{fig:pgpe_d}).

The main advantage of our framework, however, emerges in complex environments, such as HP. In this domain, the action-matching proxy of APC fails to learn a behaviorally meaningful latent representation, leading to collapsed fine-tuning performance. Conversely, OPC successfully compresses high-performing behaviors in the latent manifold, which it later recovers to reliably combat standard baselines. Notably, in the complex locomotion task \codeblue{backward} (\Figgref{fig:pgpe_j}), OPC demonstrates an impressive capacity to generalize, discovering latent behaviors that achieve returns surpassing even the best-performing policies present in the training dataset. This task also highlights a trade-off of Latent Warm Start: coarse initialization can occasionally trap the optimization in local optima.

%% file: sections/6_conclusion.tex
\addvspace{1.5em} 
\section{Conclusion}
\label{sec:conclusion}

In this paper, we presented an enhanced framework for Latent Behavior Compression that fundamentally shifts the characterization of policy behaviors from immediate-action proxies to global state-visitation distributions. By combining an information-theoretic uniqueness metric for dataset curation with a differentiable mixture-occupancy matching objective, our pipeline successfully captures complex, long-horizon dynamics. Empirically, we demonstrated that it preserves critical outlier behaviors, constructs a behaviorally meaningful latent manifold, and enables highly sample-efficient adaptation. Crucially, optimizing within this occupancy-matched space enables agents to solve complex continuous control tasks that previous action-based compression methods fail to solve.

\textbf{Limitations and Future Directions.}~~While our framework establishes a robust foundation for state-occupancy manifold learning, its maximum asymptotic performance is ultimately limited by the behavioral diversity of the pre-training dataset. Future research could explore replacing random policy initialization with active, task-agnostic exploration frameworks, such as maximum-entropy or intrinsic-motivation algorithms~\citep[e.g.,][]{eysenbach2018diversity, Mutti2021Mepol}. Furthermore, both the non-parametric $k$-NN estimator and the zeroth-order PGPE optimizer can exhibit high variance and fail to exploit the differentiable latent manifold. Consequently, employing variance-reduction techniques or exploring alternative latent optimization strategies beyond PGPE, such as Bayesian Optimization or gradient-aware methods like Latent-PPO~\citep{han_lg-h-ppo_2026}, could unlock even greater downstream adaptation capabilities within these highly compressed behavioral spaces.


%% file: sections/A1_related_works.tex
\section{Extended Related Work}
\label{app:related_works}

Our framework sits at the intersection of unsupervised behavior discovery, weight-space learning, and occupancy matching, thereby unifying these domains to learn a functionally grounded, low-dimensional policy manifold.

\textbf{Unsupervised RL and Behavior Discovery.}~~A key challenge in Deep Reinforcement Learning is the dependency on extrinsic rewards, which limits an agent's ability to generalize. To address this, Unsupervised RL (URL) methods seek to learn useful behaviors before a specific goal is defined. Information-theoretic frameworks like DIAYN~\citep{eysenbach2018diversity}, as well as other skill-discovery methods~\citep{gregor2017skill, achiam2018variational, hansen2019fast}, aim to discover diverse skills by maximizing the mutual information between states and latent variables. Similarly, MEPOL~\citep{Mutti2021Mepol} utilizes a maximum-entropy objective to drive an agent toward maximal state-space coverage, while more recently, TRPE~\citep{zamboni2025towards} utilizes a similar approach to enforce behavioral diversity in multi-agent systems. While these methods successfully identify robust task-agnostic initializations, they typically revert to operating in the original high-dimensional parameter space once supervised fine-tuning begins. Our framework, instead, identifies a continuous global behavioral manifold that inherently facilitates and accelerates this downstream adaptation process.

\textbf{Policy Compression and Weight Space Learning.}~~The goal of learning a latent representation of neural network parameters connects our approach to the field of Weight Space Learning (WSL). In RL, recent generative models have explored task-specific parameter generation conditioned on goals~\citep{faccio2023goal} or performance checkpoints~\citep{peebles2022learning}. In supervised learning, \emph{hyper-representations}~\citep{schurholt2021self, schurholt2022hyper, schurholt2024towards} aim to build task-agnostic embeddings from a ``zoo'' of trained models. However, standard WSL methods and traditional policy compression techniques~\citep{rusu_policy_2016} face a critical hurdle: the vast number of parameter-space symmetries (e.g., neuron permutations and scaling) inherent in neural architectures~\citep{kuurkova1994functionally}. To circumvent this, recent works in supervised learning supplement standard parameter reconstruction losses with behavioral output matching~\citep{meynent2025structure}. In Unsupervised Reinforcement Learning, other approaches tackle policy space simplification mathematically, such as reducing the cardinality of the policy space~\citep{mutti2022reward}. However, these strict constraints often lead to NP-hard optimization problems. Our approach sidesteps permutation symmetries entirely. Building on the Latent Behavior Compression framework~\citep{tenedini2025from}, we use a functional compression metric rather than parameter reconstruction, thereby allowing the generative model to naturally assimilate mathematically divergent yet functionally identical parameterizations. 

\textbf{Policy Manifolds and Quality Diversity.}~~The existence of low-dimensional manifolds embedded within policy parameter spaces has been hypothesized~\citep{rakicevic2021policy} and explored using Variational Autoencoders (VAEs) to analyze pre-trained expert embeddings~\citep{chang2019agent}. Similar architectures have been applied in Quality Diversity to improve the sample efficiency of diversity-based search~\citep{rakicevic2021policy} or to distill large policy archives into generative models~\citep{hegde2023generating}. Notably, these methods rely on \emph{parameter-reconstruction losses}, which fundamentally restrict their compression ratios (e.g., up to $19:1$ in \citet{hegde2023generating}). By optimizing a behavioral objective directly, our unsupervised pipeline achieves substantially greater compression while providing a topologically smooth space for downstream optimization.

\textbf{Occupancy Matching and Policy Optimization.}~~Comparing behaviors formally often involves analyzing state occupancy distributions, a concept recently utilized to learn steerable latent policy representations via expected state-action features~\citep{li2026learning}. Comparing behaviors is also the foundation of Imitation Learning (IL), where methods like GAIL~\citep{ho_generative_2016} and Wasserstein-GANs~\citep{goodfellow_generative_2014, arjovsky_wasserstein_2017, gulrajani_improved_2017, zhang_wasserstein_2020} minimize the divergence between an agent’s behavior and a target expert distribution. Our contribution scales these concepts from single-expert matching to population-level alignment by developing a mixture-occupancy matching objective. Once this manifold is learned, we optimize downstream tasks using Policy Gradient with Parameter-based Exploration (PGPE)~\citep{sehnke2008policy, ruckstiess2010exploring, miyamae2010natural, montenegro2024learning}. While zeroth-order methods like PGPE and first-order methods~\citep{peters2008reinforcement, lillicrap2015continuous, kakade2001natural, schulman2017proximal} traditionally struggle to scale in highly redundant parameter spaces, executing PGPE within our highly compressed, behaviorally organized latent space resolves these scalability bottlenecks, avoiding the complex decoder Jacobian calculations required by prior manifold optimization approaches~\citep{rakicevic2021policy}.

%% file: sections/A2_scores.tex
\section{Policy Dataset Curation: Derivations and Estimators}
\label{app:derivation}

This section provides the formal derivations and practical estimation details for the information-theoretic uniqueness score introduced in~\Secref{sec:method}.

\subsection{Derivation of the Uniqueness Score}
\label{app:entropy_derivation}

Let $M = |\Psi|$ denote the number of policies in the population. Let $d_i(s)$ for $i=1,\dots,M$ be the state occupancy distribution induced by the $i$-th policy $\pi_i \in \Psi$. We consider the uniform mixture over these occupancy distributions by assuming that each policy is selected with equal probability, $p(i) = \frac{1}{M}$.

This induces a joint distribution over states and policies given by $p(s,i) = p(s \mid i)p(i) = \frac{1}{M} d_i(s)$, from which the marginal state distribution, corresponding to the mixture occupancy $d_\Psi$, is obtained as $d_\Psi(s) = \sum_{i=1}^M p(s,i) = \frac{1}{M}\sum_{i=1}^M d_i(s)$.

Let $H(X)$ denote the differential entropy of a random variable $X$. Starting from the chain rule $H(X,Y) = H(X) + H(Y\mid X)$ and using the symmetry of joint differential entropy, we can write:
$$
\begin{aligned}
    H(X) + H(Y\mid X) &= H(Y) + H(X\mid Y) \\
    H(X) &= H(X\mid Y) + MI(X,Y), 
\end{aligned}
$$

where $MI(X,Y)$ denotes the mutual information. We now apply this to the random variables $S \sim d_\Psi(s)$ and $\pi \sim p(i)$. The conditional entropy becomes:
$$
\begin{aligned}
    H(S\mid \pi)
    &= -\sum_{i=1}^M \int_{\mathcal{S}} p(s,i)\log p(s\mid i)\,ds \\
    &= -\sum_{i=1}^M \int_{\mathcal{S}} \frac{1}{M} d_i(s)\log d_i(s)\,ds \\
    &= \frac{1}{M}\sum_{i=1}^M H(d_i).
\end{aligned}
$$

The mutual information can be expanded as:
$$
\begin{aligned}
    MI(S,\pi)
    &= \sum_{i=1}^M \int_{\mathcal{S}} p(s,i) \log\frac{p(s,i)}{d_\Psi(s)p(i)}\,ds \\
    &= \sum_{i=1}^M \int_{\mathcal{S}} \frac{1}{M} d_i(s) \log\frac{\frac{1}{M}d_i(s)}{d_\Psi(s)\cdot\frac{1}{M}}\,ds \\
    &= \frac{1}{M}\sum_{i=1}^M \int_{\mathcal{S}} d_i(s)\log\frac{d_i(s)}{d_\Psi(s)}\,ds \\
    &= \frac{1}{M}\sum_{i=1}^M \KL(d_i \parallel d_\Psi).
\end{aligned}
$$

Combining the two terms, we obtain the entropy decomposition of the mixture occupancy distribution:
$$
H(d_\Psi) = \frac{1}{M}\sum_{i=1}^M \left[ H(d_i) + \KL(d_i \parallel d_\Psi) \right].
$$

The term inside the summation represents the marginal contribution of the $i$-th policy to the total entropy of the mixture, which we utilize as our uniqueness score $\rho(\vtheta_i)$.

\subsection{Tractable Estimation via GMMs}
\label{app:gmm_estimation}

Because the analytical form of the occupancy $d_{\vtheta_i}$ is unknown, we estimate the scores using trajectory data. We first fit Gaussian Mixture Models (GMMs) to the temporally downsampled trajectories of each policy, yielding the continuous density estimators $\hat{d}_{\vtheta_i}(s)$ for individual policies and $\hat{d}_\Psi(s)$ for the global mixture. 

Using a set of $N$ particles $\mathcal{D}_{\vtheta_i} = \{s_j\}_{j=1}^{N}$ sampled from the individual density estimate $\hat{d}_{\vtheta_i}$, we apply Monte Carlo (MC) integration to approximate the values of the differential entropy and the KL divergence:
$$
\begin{aligned}
    \widehat{H}(d_{\vtheta_i}) &= -\frac{1}{N} \sum_{j=1}^N\log \hat{d}_{\vtheta_i}(s_j), \\
    \widehat{\KL}(d_{\vtheta_i} \parallel d_\Psi) &= \frac{1}{N} \sum_{j=1}^N \left[ \log \hat{d}_{\vtheta_i}(s_j) - \log \hat{d}_\Psi(s_j) \right].
\end{aligned}
$$

\subsection{KL Divergence Bound and the Saturation Effect}
\label{app:kl_bound}

When calculating $\widehat{\KL}$, the estimation of the mixture occupancy distribution $\hat{d}_\Psi$ must be handled carefully to avoid a \textit{saturation effect} that degrades the discriminative power of the scores as the population grows. This is due to a strict upper bound on the KL divergence between a single component and its mixture.

\textbf{Proposition 1.} \textit{Let $m(x) = \sum_{j=1}^M w_j p_j(x)$ be a mixture of $M$ probability density functions with weights $w_j \ge 0$ and $\sum w_j = 1$. For any component $i$, the KL divergence is bounded: $\KL(p_i \parallel m) \le -\log w_i$.}

\textit{Proof.} By the definition of the mixture, $m(x) = w_i p_i(x) + \sum_{j \neq i} w_j p_j(x)$. Because density functions and weights are non-negative, $m(x) \ge w_i p_i(x)$, which implies $\frac{p_i(x)}{m(x)} \le \frac{1}{w_i}$.
    
Since the logarithm is monotonically increasing, $\log \frac{p_i(x)}{m(x)} \le -\log w_i$. The KL divergence is the expectation of this log-ratio over $p_i$:
$$
\KL(p_i \parallel m) = \int p_i(x) \log \frac{p_i(x)}{m(x)} dx \le \int p_i(x) (-\log w_i) dx.
$$

Since $-\log w_i$ is a constant and $p_i(x)$ integrates to $1$, $\KL(p_i \parallel m) \le -\log w_i$. $\blacksquare$

In our uniform mixture, $w_i = 1/M$, yielding the bound $\KL(d_{\vtheta_i} \parallel d_\Psi) \le \log M$. If $\hat{d}_\Psi$ is estimated using exhaustive aggregation across a massive population $\Psi$, the KL term mathematically vanishes relative to the scale of the state space, failing to capture uniqueness. To preserve metric sensitivity, we restrict the calculation of the global reference $\hat{d}_\Psi$ to a representative subset of the population.

%% file: sections/A3_importance_weights.tex
\section{Numerically Stable Importance Weight Computation}
\label{app:IW_computation}

As introduced in~\Secref{sec:method}, the computation of the importance weights follows the density-ratio methodology of \citet{Mutti2021Mepol}, adapted here to accommodate mixtures of policies. We associate each particle $s^j_t$ with the importance weight $w_j$ calculated for the full trajectory $\tau_j$ that produced it.

Because the non-normalized weight $\overline{w}_j$ involves the product of numerous probability values lying in the interval $[0, 1]$, both the numerator and denominator can quickly vanish to zero over long horizons. To prevent arithmetic underflow and a loss of numerical precision, we transform the product of probabilities into a sum of log-probabilities and compute the weights entirely in log-space using the \textit{Log-Sum-Exp} (LSE) trick.

Let $\ell_{i,j}$ be the log-probability of trajectory $\tau_j$ under the $i$-th policy of the original dataset mixture $\pi_{\vtheta_i} \in \gD_\Theta$:
$$
\ell_{i,j} = \log p(\tau_j \mid \pi_{\vtheta_i}) = \sum_{t=0}^{T} \log \pi_{\vtheta_i}(a^j_t \mid s^j_t).
$$

Similarly, let $\ell'_{i,j}$ be the log-probability of the same trajectory under the $i$-th reconstructed target policy $\pi_{\hat{\vtheta}_i} \in \gD_{\hat{\Theta}}$. The non-normalized weight $\overline{w}_j$ is the ratio of these two sums of exponentials. In log-space, letting $\lambda_j = \log \overline{w}_j$, this is expressed as:
$$
\lambda_j = \log \left( \sum_{i} \exp(\ell'_{i,j}) \right) - \log \left( \sum_{i} \exp(\ell_{i,j}) \right).
$$

To prevent overflow or underflow during these summations, each term is computed using the LSE function:
$$
\text{LSE}(x_1, \dots, x_n) = c + \log \sum_{i=1}^{n} \exp(x_i - c), \quad \text{where } c = \max_{i} \{x_i\}.
$$

By extracting the maximum value $c$, we ensure that the largest exponentiated term inside the summation is strictly $\exp(0) = 1$, safely shifting the range of the calculation into a stable floating-point region. Thus, the non-normalized log-weight is robustly computed as:
$$
\lambda_j = \text{LSE}_{i \in \gD_{\hat{\Theta}}}(\ell'_{i,j}) - \text{LSE}_{i \in \gD_{\Theta}}(\ell_{i,j}).
$$

The final step is self-normalization, which ensures that the weights over the dataset sum to unity, $w_j = \overline{w}_j / \sum_k \overline{w}_k$. To maintain numerical stability until the very end of the computation, we perform this normalization in the log-domain as well:
$$
w_j = \exp \left( \lambda_j - \text{LSE}_k(\lambda_k) \right).
$$

This approach guarantees that we exponentiate only the final output, preserving high numerical precision throughout the density-ratio estimation process.

%% file: sections/A4_training.tex
\section{Autoencoder Training Pipeline}
\label{app:training_pipeline}

Optimizing the mixture-occupancy matching objective over the entire policy dataset $\mathcal{D}_\Theta$ at once is computationally prohibitive. To scale the training process, we employ a mini-batch stochastic gradient descent strategy over the policy population.

During each training step, we uniformly sample a subset $\psi$ of $P$ policies from the dataset, and fetch their corresponding pre-collected trajectories $\mathcal{D}_\psi\subset\mathcal{D}_\tau$. A major advantage of our non-parametric importance-sampling formulation is the ability to reuse the exact trajectory batch across multiple sequential parameter updates, thereby drastically increasing sample efficiency. Specifically, for each sampled batch $\psi$, we execute an inner loop consisting of $I$ consecutive gradient descent steps of the autoencoder's parameters $\{\vxi, \vzeta\}$.

Critically, because the decoder's weights (and consequently, the reconstructed policies $\hat{\psi}$) evolve after each step, the true density ratio shifts. Therefore, the importance weights are re-evaluated at every step using the numerically stable Log-Sum-Exp procedure described in Appendix~\ref{app:IW_computation}. This nested update scheme significantly accelerates convergence and improves sample efficiency. Training iterations are repeated until the population has been sufficiently sampled or the autoencoder has converged. The complete training procedure is summarized in~\Algref{alg:AE_training}.

\begin{algorithm}[tb]
\caption{Latent Behavior Compression}
\label{alg:AE_training}
\begin{algorithmic}[1]
    \STATE \textbf{Input:} Policy dataset $\mathcal{D}_\Theta$, trajectory dataset $\mathcal{D}_\tau$, inner iterations $I$, batch size $P$, nearest neighbors parameter $k$, learning rate $\alpha$.
    \STATE \textbf{Output:} Generative behavioral mapping $g_{\vzeta}$.
    \STATE Randomly initialize autoencoder parameters: encoder $\vxi$ and decoder $\vzeta$.
    \REPEAT
        \STATE Sample a mini-batch of $P$ policies $\psi = \{\vtheta_i\}_{i=1}^P \subset \mathcal{D}_\Theta$
        \STATE Retrieve corresponding trajectories $\mathcal{D}_\psi \subset \mathcal{D}_\tau$
        \FOR{$i=1$ \textbf{to} $I$}
            \STATE Reconstruct parameters: $\hat{\psi} = \{ \hat{\vtheta}_i = g_{\vzeta}(f_{\vxi}(\vtheta_i)) \mid \vtheta_i \in \psi \}$
            \STATE Compute mixture-matching loss: $\mathcal{L}_{B}(\vxi, \vzeta) = \KL\!\left(d_\psi \parallel d_{\hat{\psi}}\right)$ \COMMENT{via Eq.~\ref{eq:knn_estimator}}
            \STATE Update weights: $\{\vxi, \vzeta\} \leftarrow \{\vxi, \vzeta\} - \alpha \nabla_{\{\vxi, \vzeta\}} \mathcal{L}_{B}(\vxi, \vzeta)$
        \ENDFOR
    \UNTIL{convergence or full population coverage}
    \RETURN $g_{\vzeta}$
\end{algorithmic}
\end{algorithm}

\textbf{Practical Implementation of Policy Stochasticity.}~~While our theoretical framework models policies as strictly stochastic ($\pi \in \Delta(\mathcal{A})$) to accommodate density ratio estimation, evaluating highly exploratory, randomly initialized continuous policies can yield erratic, noisy actions due to untrained variance parameters. To ensure meaningful and stable state-space coverage during the Policy Dataset Curation phase, and during downstream PGPE optimization, we deploy the deterministic version of the policies (i.e., we execute only the mean of the action distribution). Consequently, during the autoencoder training phase, we evaluate the stochastic likelihood of these deterministically collected trajectories. We acknowledge that evaluating stochastic probabilities on deterministically generated data introduces bias into the importance-sampling estimator. However, this is a necessary and highly effective empirical trade-off: it strips catastrophic noise from the behavioral rollouts while utilizing the variance parameters as a continuous smoothing mechanism. This guarantees a strictly positive probability density during training, circumventing the vanishing denominators and intractable density ratios that would occur if purely deterministic Dirac-delta distributions were used.

%% file: sections/A5_experimental_settings.tex
\section{Experimental Settings}
\label{app:experimental_settings}

This section provides the details regarding the architecture, environment, and training procedure used to conduct the experiments presented in~\Secref{sec:exp}.

\subsection{Architecture and Training}

\textbf{Policy Networks.}~~To model the behaviors, we use fully-connected, feed-forward Multi-Layer Perceptrons (MLPs). The input layer has $|\mathcal{S}|$ neurons and is preceded by a normalization layer that standardizes the state features to have zero mean and unit variance. The policies consist of two hidden layers of 32 neurons, each followed by Exponential Linear Unit (ELU) activations. The final output layer consists of $|\mathcal{A}|$ neurons with a \verb|tanh| activation to squash the actions into the valid environmental bounds. The policies have roughly $10^3$ parameters: $1,218$ for MC, $1,412$ for RC, and $1,638$ for HP. The weights are sampled from independent uniform distributions $\vtheta_i\sim U(-2.5,2.5)^n$, where $n$ is the number of parameters.

\textbf{Autoencoder Architecture.}~~We utilize a fixed, symmetric autoencoder architecture to construct the behavioral manifold. First, the weights are standardized to have zero mean and unit variance. Then, the encoder projects the $n$-dimensional policy parameter vector through two fully connected hidden layers with $25$ and $10$ neurons, respectively, into a $k$-dimensional bottleneck. The decoder symmetrically maps the $k$-dimensional latent code back through layers of $10$ and $25$ neurons to reconstruct the $n$-dimensional weight vector. The encoder and decoder output layers are not activated. All other layers utilize ELU activation functions. Based on the environment, the bottleneck can have shape $k\in\{1,2,3,5,8\}$.

\textbf{Autoencoder Training.}~~During the Latent Behavior Compression phase, the autoencoder is trained on a curated dataset of $|\mathcal{D}_\Theta| = 50,000$ policies. The dataset is then reduced to the top 5\%. To compute the mixture-matching loss, we sample $20$ trajectories per policy, which we found to be an optimal trade-off between estimator variance and computational efficiency. For the importance-sampling estimator, the nearest-neighbors parameter is fixed at $k=30$ across all domains. We train with a small batch size of 5; while this introduces gradient noise, it crucially prevents the model from collapsing into a trivial mean-policy solution, thereby forcing the manifold to capture distinct, high-reward behavioral modes that would otherwise be smoothed out. The total number of training iterations is computed to ensure that each policy in the dataset is sampled at least once with probability greater than $0.99$. Furthermore, policies are sampled uniformly at random rather than sequentially to prevent catastrophic forgetting.

\subsection{Environments}
We evaluate our framework on three continuous control environments from the Gymnasium library. For each environment, the unsupervised pre-training phase is entirely reward-free, while the downstream optimization phase evaluates few-shot adaptation to four distinct reward functions.

\textbf{Mountain Car Continuous (MC).}~~In this environment, a car is stochastically placed in a sinusoidal valley and must build momentum to traverse the hills. The state space $\mathcal{S} \subseteq \mathbb{R}^2$ captures position $p \in [-1.2, 0.6]$ and velocity $v \in [-0.07, 0.07]$, while the action space $\mathcal{A} \subseteq \mathbb{R}^1$ represents the applied acceleration. We define four tasks: (1) \codeblue{standard}: a sparse $+100$ reward for reaching the right goal ($p \ge 0.45$), with a $-0.1 a^2$ control penalty; (2) \codeblue{left}: identical to \codeblue{standard}, but the goal is at the left peak ($p \le -1.1$); (3) \codeblue{height}: a dense reward $R_t = h^2$ for height $h \ge 0.2$; and (4) \codeblue{speed}: a dense reward $R_t = v^2$. Episodes terminate upon reaching the standard goal or after 999 steps.

\textbf{Reacher (RC).}~~RC features a two-jointed robotic arm moving in a 2D plane. To focus strictly on task-agnostic behavioral discovery, we remove all target-related information from the observations. The normalized state space $\mathcal{S} \subseteq \mathbb{R}^6$ captures the sines, cosines, and angular velocities of the two joints, and the action space $\mathcal{A} \subseteq \mathbb{R}^2$ controls the hinge torques. For the purpose of normalization, we consider the state bounded between the vectors $[-1, -1, -1, -1, -5, -5]$ and $[1, 1, 1, 1, 5, 5]$. We define four binary tasks evaluated on the fingertip's kinematics ($R_t = 1$ if met, else $0$): (1) \codeblue{speed}: linear velocity $> 6$; (2) \codeblue{clockwise}: tangential velocity $< -1$; (3) \codeblue{c-clockwise}: tangential velocity $> 1$; and (4) \codeblue{radial}: radial velocity $> 3$. Episodes terminate after 50 steps. 

\textbf{Hopper (HP).}~~HP is a complex planar monopod robot governed by a normalized $11$-dimensional state space capturing height, joint angles, and velocities, controlled via a $3$-dimensional action space governing the hip, knee, and ankle torques. For the purpose of normalization, we consider the state bounded between the vectors $[0.7,\allowbreak -0.2,\allowbreak -2.7,\allowbreak -2.7,\allowbreak -0.8,\allowbreak -5.0,\allowbreak -5.0,\allowbreak -5.0,\allowbreak -5.0,\allowbreak -5.0,\allowbreak -5.0]$ and $[1.5,\allowbreak 0.2,\allowbreak 0.0,\allowbreak 0.0,\allowbreak 0.8,\allowbreak 5.0,\allowbreak 5.0,\allowbreak 5.0,\allowbreak 5.0,\allowbreak 5.0,\allowbreak 5.0]$. We define four locomotion tasks: (1) \codeblue{forward}: $x$-axis velocity $> 1$; (2) \codeblue{backward}: $x$-axis velocity $< -1$; (3) \codeblue{standstill}: $x$-axis velocity within $[-0.05, 0.05]$; and (4) \codeblue{standard}: the default Gymnasium composite reward consisting of a $+1$ healthy survival bonus, a forward velocity term, and a control penalty. The first three tasks use binary rewards. Episodes terminate if the robot falls or after 1,000 steps.

\subsection{Hyperparameters}
\textbf{Baselines.}~~We employ the baseline algorithms (PPO, DDPG, and SAC) from the \textbf{StableBaselines3}~\citep{stable-baselines3} library, using the same hyperparameters as in~\citet{tenedini2025from} (for MC and RC), specifically:
\begin{itemize}
\item \textbf{Mountain Car:}
\begin{itemize}
\item \textbf{DDPG:} Exploration noise standard deviation of 0.75 and 0.65, and a replay buffer size of 50,000.
\item \textbf{SAC:} Generalized State Dependent Exploration (gSDE), entropy coefficient of 0.1, soft update coefficient of 0.01, train frequency of 32, and 32 gradient steps per rollout.
\item \textbf{PPO:} Generalized State Dependent Exploration (gSDE), learning rate of 0.0001, batch size of 256, and 4 optimization epochs per rollout.
\end{itemize}
\item \textbf{Reacher:} Hyperparameters are identical to those used in Mountain Car, with the exception of the DDPG exploration noise standard deviation, which is set to 0.5.
\end{itemize}
Where not specified, including for all the algorithms in the HP environment, the hyperparameters are set to the already-fine-tuned values from the RL Baselines3 Zoo.

\textbf{PGPE.}~~Latent Behavior Optimization is performed using Policy Gradients with Parameter-based Exploration (PGPE), optimized via Adam. We maintain a population size of $8$ across all environments and use an exponential scheduler for learning-rate decay. To prevent overshooting high-reward regions and becoming trapped in zero-gradient plateaus, we reduce the first moment decay to $\beta_1 = 0.1$. Hyperparameters are tailored to specific environment objectives: for the \textbf{Mountain Car} (MC) \codeblue{standard} and \codeblue{left} rewards, we set the center and standard deviation learning rates to $0.11$ and $0.1$, respectively (initial $\sigma=1.0$). The \codeblue{speed} task uses more conservative rates ($0.02$ center, $0.01$ std), while the \codeblue{height} task requires a center rate of $0.01$ and a reduced initial $\sigma$ of $0.1$. For \textbf{Reacher} (RC), we utilize a center learning rate of $0.1$ and a standard deviation learning rate of $0.05$, whereas Hopper (HP) is optimized with a center learning rate of $0.11$.

For the \textbf{Warm Start} variation of PGPE, we adopt the hyperparameters previously used for Reacher in the Hopper environment (center LR $= 0.1$, std LR $= 0.05$). In contrast, for Mountain Car, we employ a more conservative configuration: a center learning rate of $0.01$, an initial standard deviation of $0.1$, and a standard deviation learning rate of $0.05$. This choice is informed by the observation that initial random sampling often identifies high-reward regions early, necessitating only marginal refinement rather than extensive further optimization. For the warm-start initialization, we sample an initial set of policies with size proportional to the latent space dimensionality: $40$, $56$, and $80$ policies for the 3D, 5D, and 8D spaces, respectively.

\textbf{Reproducibility.}~~All experimental components are fixed with specific random seeds to ensure reproducibility and facilitate fair comparisons. The seeding strategy for the pipeline is as follows:
 \begin{itemize}
     \item \textbf{Dataset Generation and Curation:} A constant seed of $0$ is applied across all stages, including policy generation, particle collection, and score computation.
     \item \textbf{Latent Behavior Compression:} For the main optimization experiments, ten distinct latent spaces are trained using seeds $0, \dots, 9$. The ablation studies that compare latent spaces specifically utilize seeds $888$ and $999$.
     \item \textbf{Latent Behavior Optimization:} The PGPE optimization scripts are initialized with seed $0$, while the baselines are evaluated across ten independent runs each using seeds $0, \dots, 9$.
 \end{itemize}

%% file: sections/A6_additional_experiments.tex
\section{Additional Experiments}
\label{app:add_exp}
In this section, we present the full list of experiments presented in \Secref{sec:exp}, including those that could not fit in the main text.

\begin{figure}[t]

    \centering 
    \includegraphics[width=0.5\textwidth, height=0.6cm]{figures/experiments/scores/comparison_legend.pdf}

    \begin{subfigure}[b]{0.24\textwidth}
        \includegraphics[width=\textwidth]{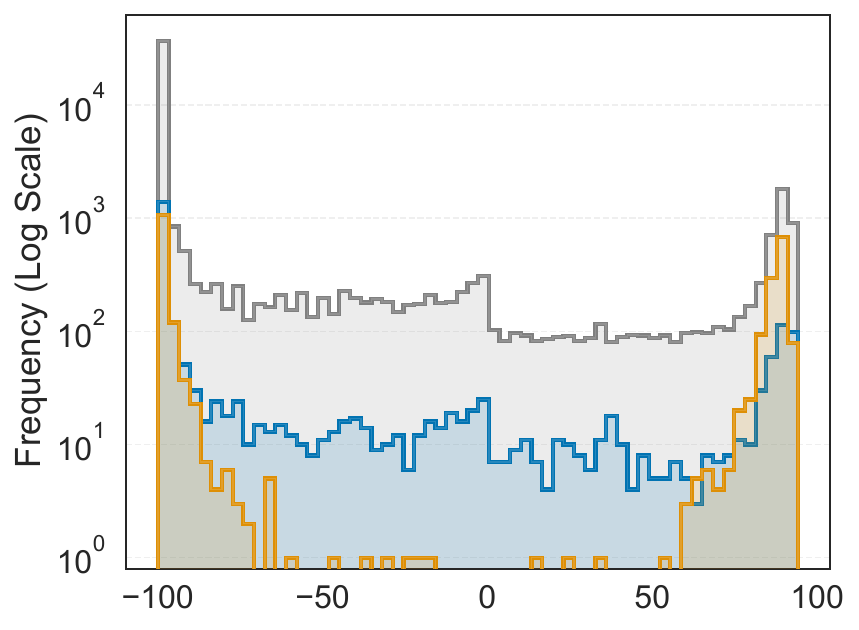}
        \vspace{-0.6cm} \caption{MC, \codeblue{standard}}
        
    \end{subfigure}
    \hfill 
    \begin{subfigure}[b]{0.24\textwidth}
        \includegraphics[width=\textwidth]{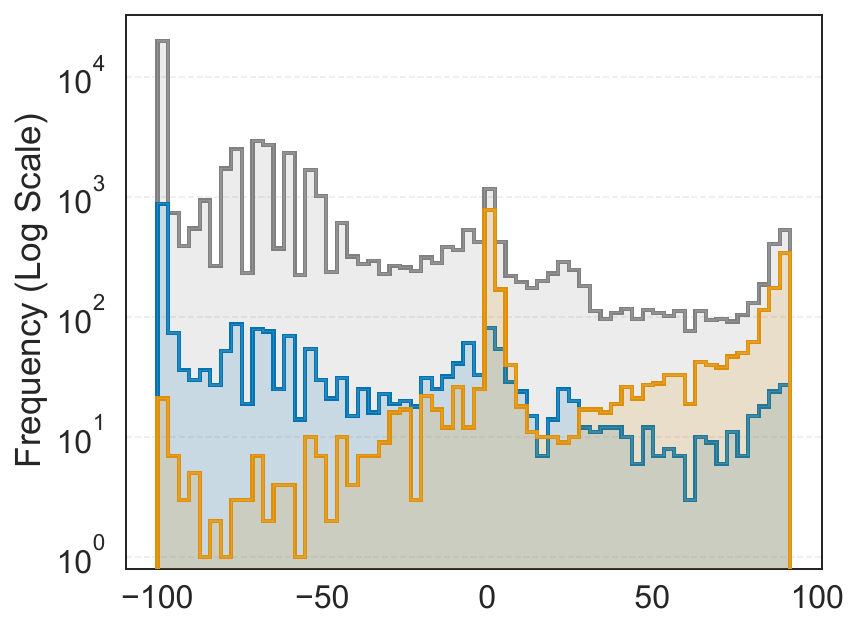}
        \vspace{-0.6cm}\caption{MC, \codeblue{left}}
        
    \end{subfigure}
    \hfill
    \begin{subfigure}[b]{0.24\textwidth}
        \includegraphics[width=\textwidth]{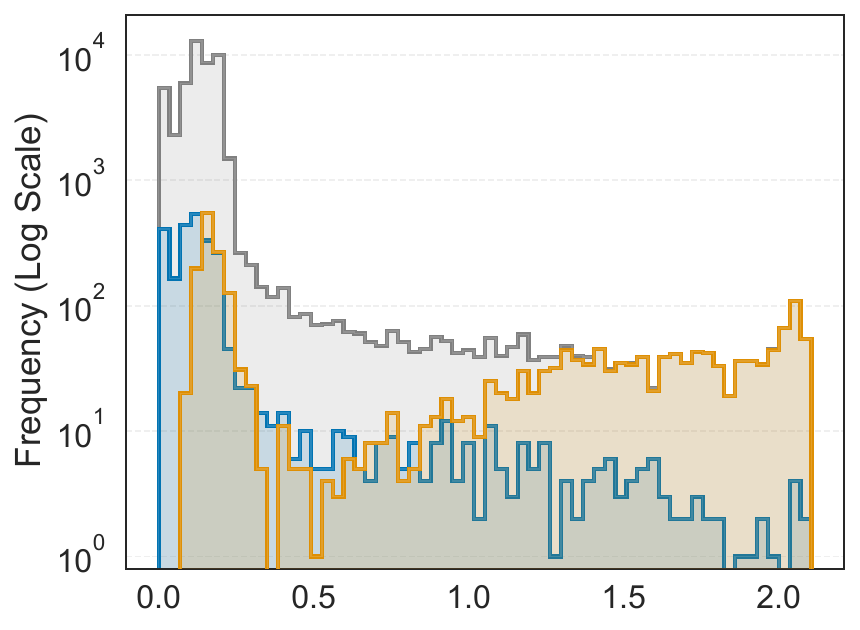}
        \vspace{-0.6cm}\caption{MC, \codeblue{speed}}
        
    \end{subfigure}
    \hfill
    \begin{subfigure}[b]{0.24\textwidth}
        \includegraphics[width=\textwidth]{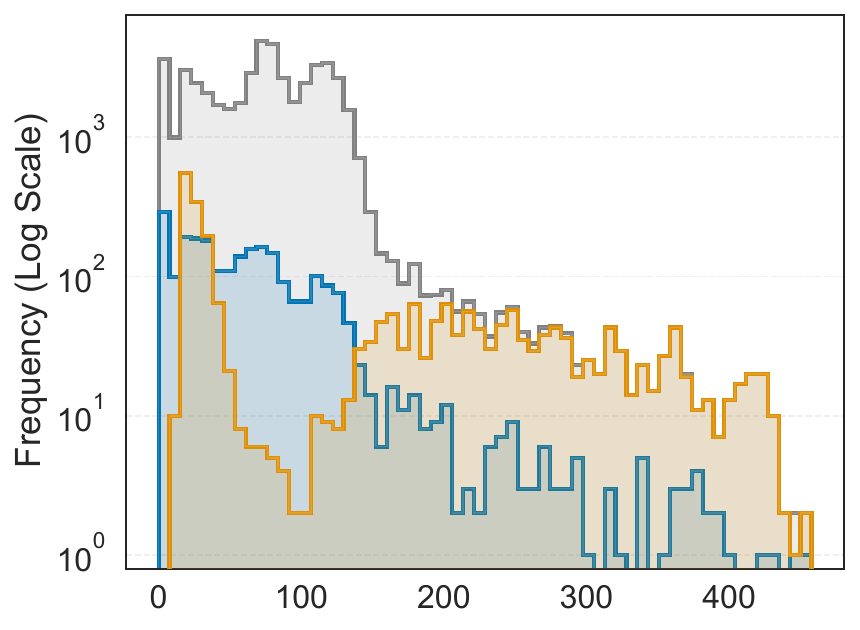}
        \vspace{-0.6cm}\caption{MC, \codeblue{height}}
        
    \end{subfigure}

    \begin{subfigure}[b]{0.24\textwidth}
        \includegraphics[width=\textwidth]{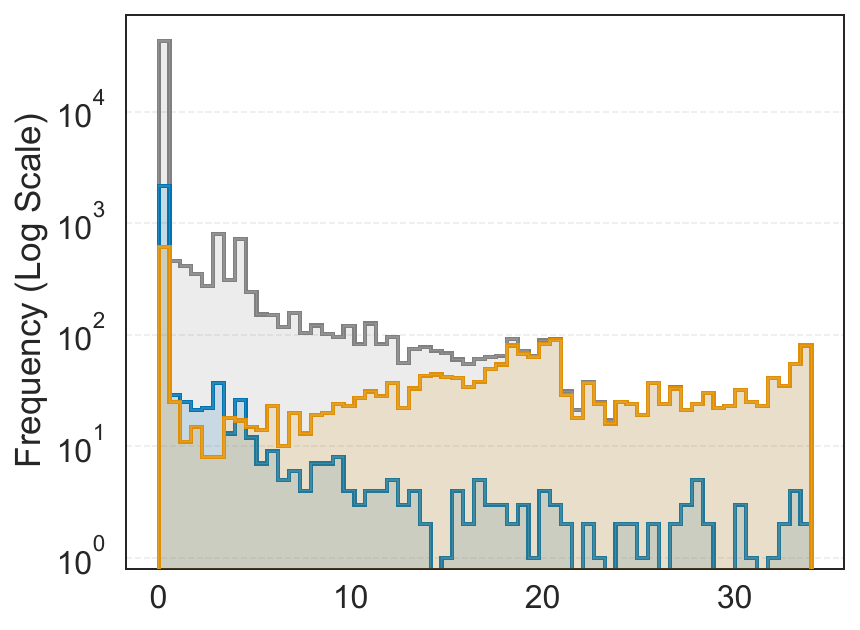}
        \vspace{-0.6cm} \caption{RC, \codeblue{speed}}
        
    \end{subfigure}
    \hfill 
    \begin{subfigure}[b]{0.24\textwidth}
        \includegraphics[width=\textwidth]{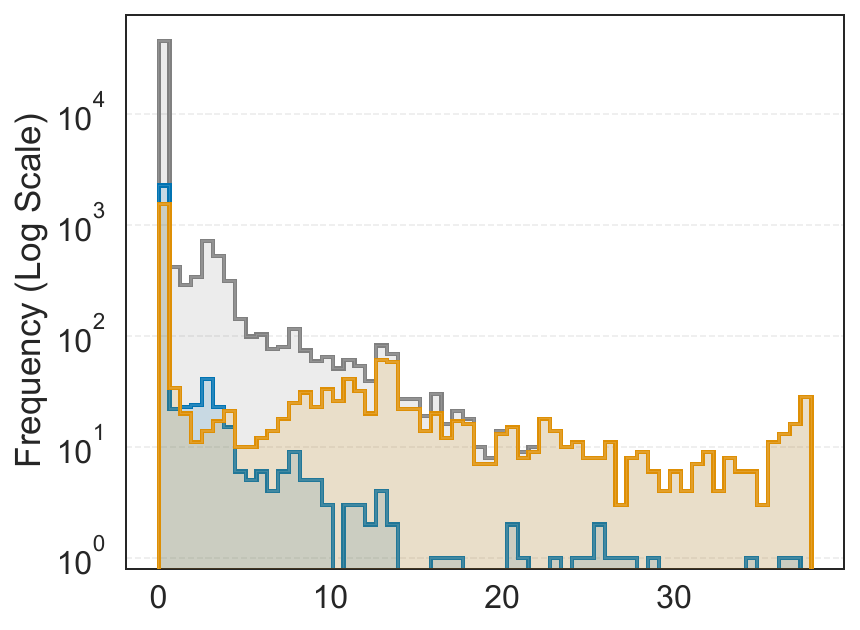}
        \vspace{-0.6cm}\caption{RC, \codeblue{clockwise}}
        
    \end{subfigure}
    \hfill
    \begin{subfigure}[b]{0.24\textwidth}
        \includegraphics[width=\textwidth]{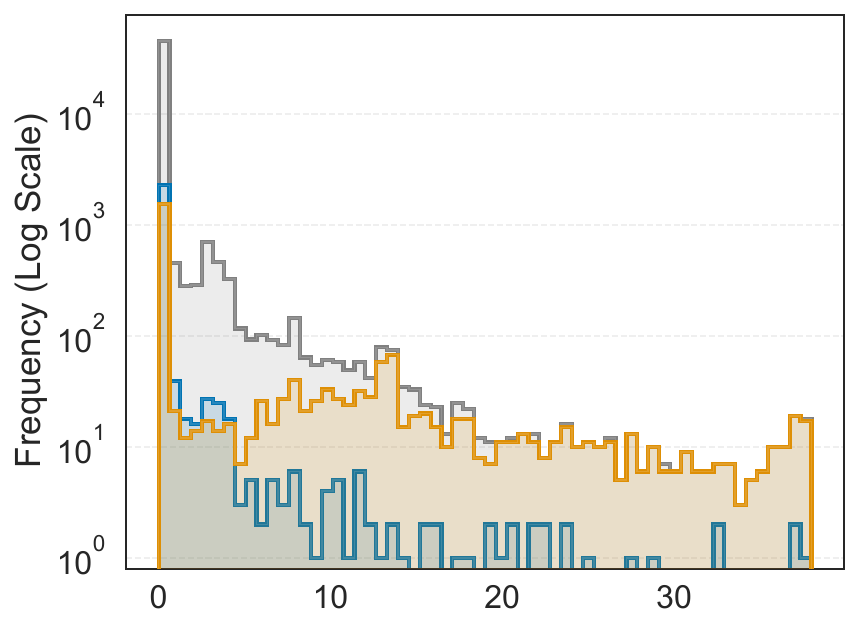}
        \vspace{-0.6cm}\caption{RC, \codeblue{c-clockwise}}
                
    \end{subfigure}
    \hfill
    \begin{subfigure}[b]{0.24\textwidth}
        \includegraphics[width=\textwidth]{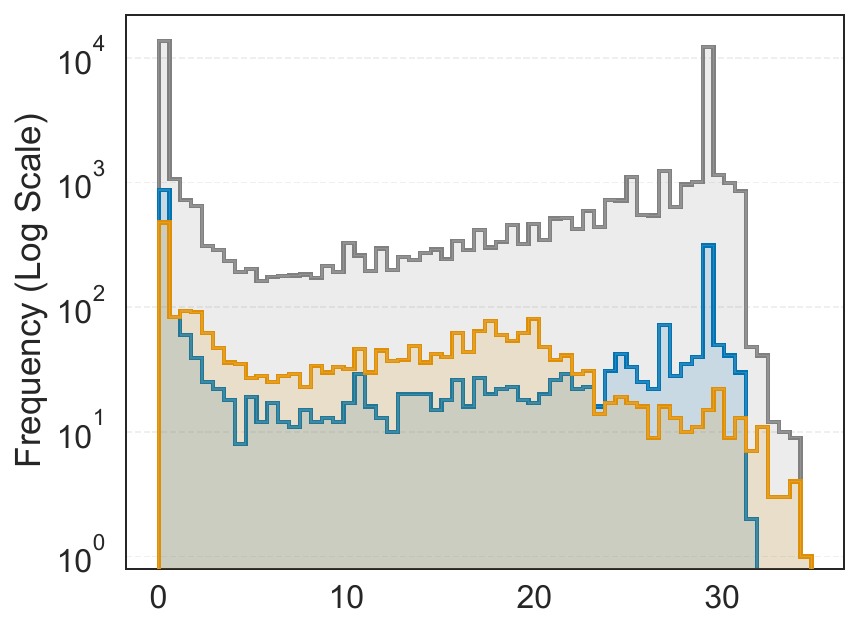}
        \vspace{-0.6cm}\caption{RC, \codeblue{radial}}
        \label{fig:full_distribution_grid_radial}
    \end{subfigure}

    \begin{subfigure}[b]{0.24\textwidth}
        \includegraphics[width=\textwidth]{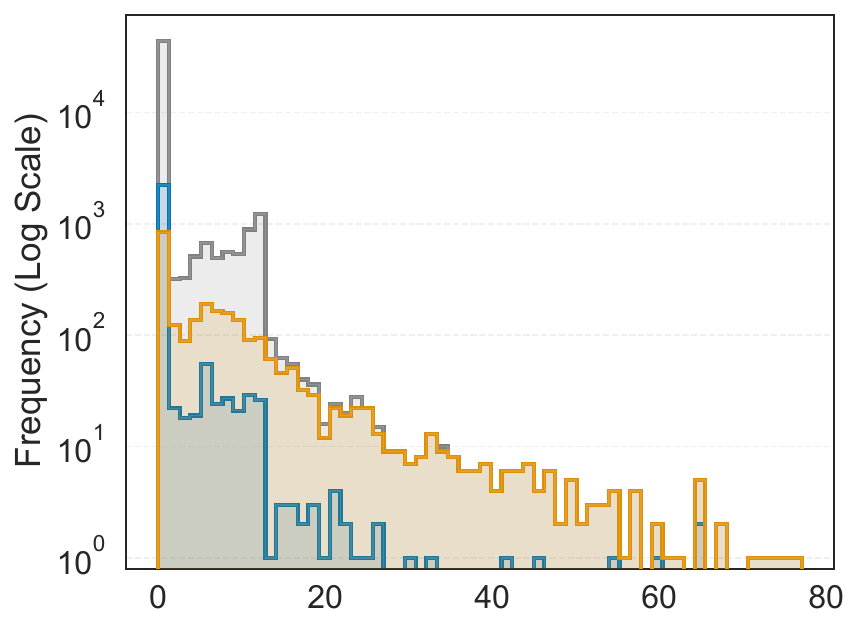}
        \vspace{-0.6cm}\caption{HP, \codeblue{forward}}
        
        \vspace{-0.1cm}
    \end{subfigure}
    \hfill 
    \begin{subfigure}[b]{0.24\textwidth}
        \includegraphics[width=\textwidth]{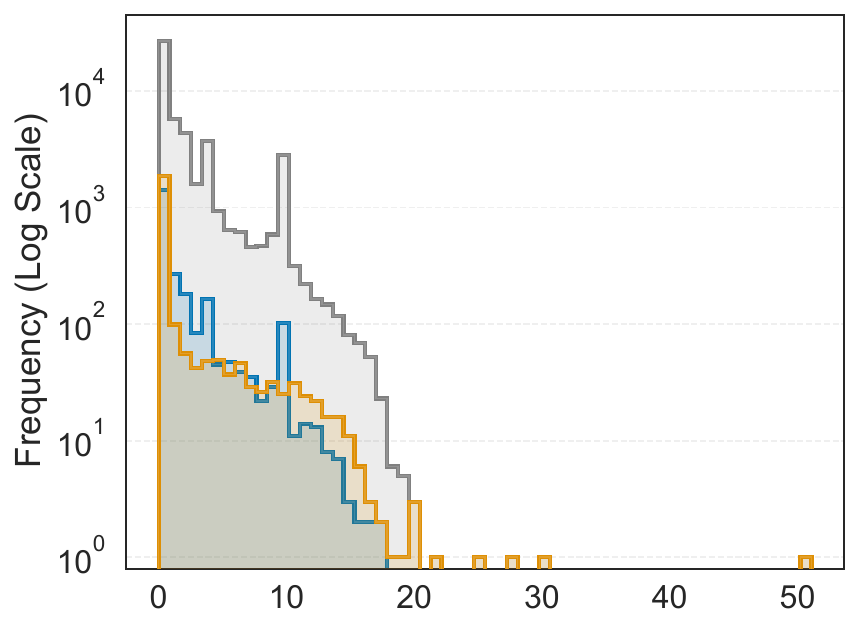}
        \vspace{-0.6cm}\caption{HP, \codeblue{backward}}
        
        \vspace{-0.1cm}
    \end{subfigure}
    \hfill
    \begin{subfigure}[b]{0.24\textwidth}
        \includegraphics[width=\textwidth]{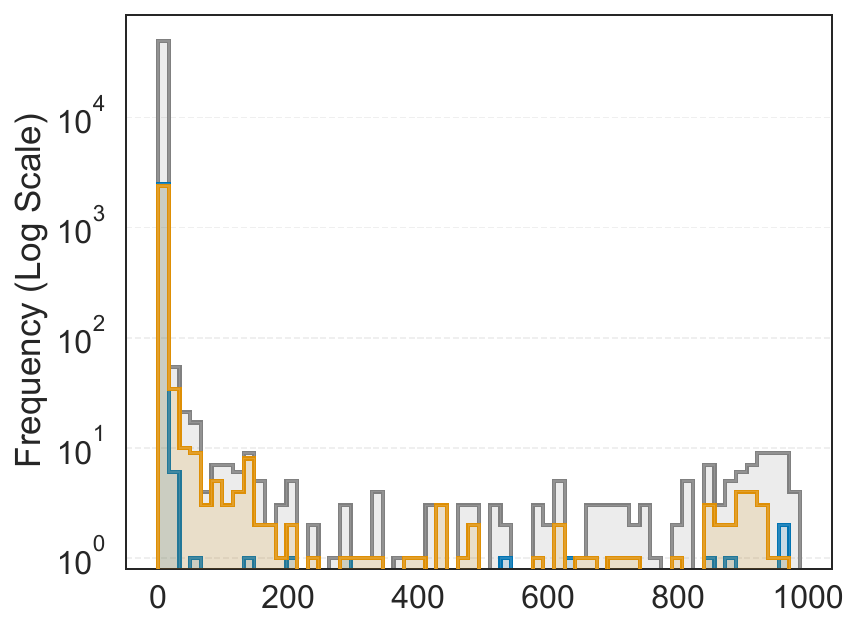}
        \vspace{-0.6cm}\caption{HP, \codeblue{standstill}}
        
        \vspace{-0.1cm}
    \end{subfigure}
    \hfill
    \begin{subfigure}[b]{0.24\textwidth}
        \includegraphics[width=\textwidth]{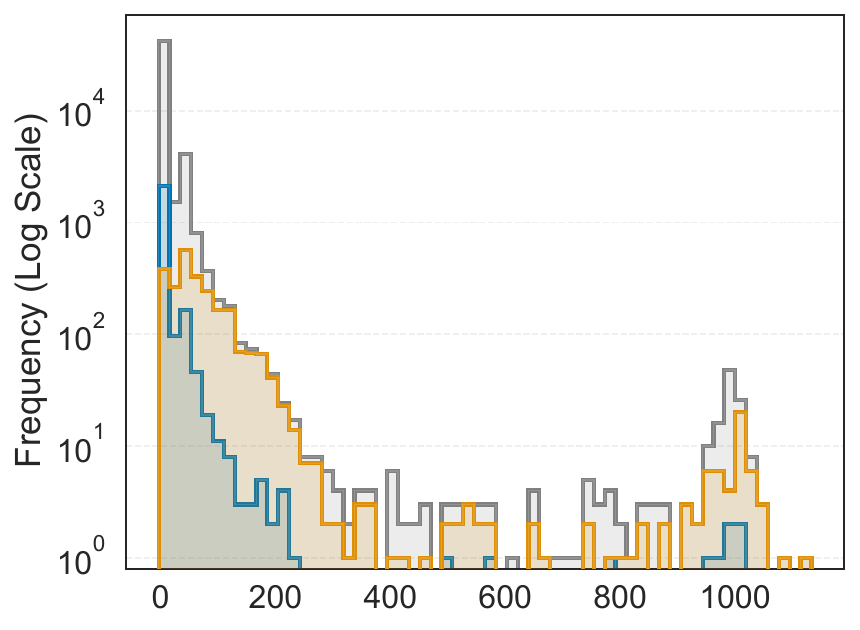}
        \vspace{-0.6cm}\caption{HP, \codeblue{standard}}
        
        \vspace{-0.1cm}
    \end{subfigure}
    \begin{minipage}{\textwidth} 
        \centering
        
        \caption{
        Reward distributions comparison in MC, RC, and HP. }
        \label{fig:full_distribution_grid}
    \end{minipage}
\end{figure}
\subsection{Policy Dataset Curation}
\label{app:add_curation}

\begin{figure}[tp]
    \centering

    \stackcomparison{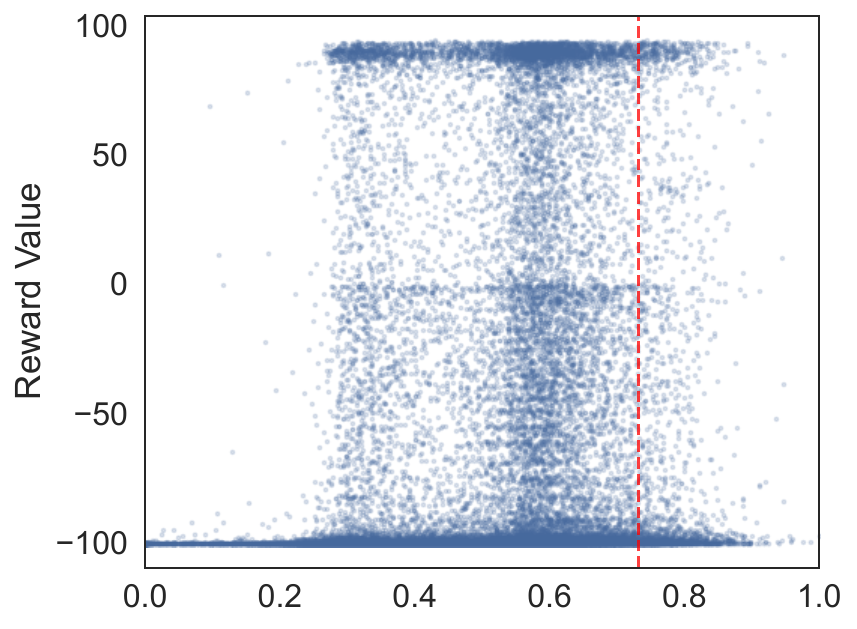}{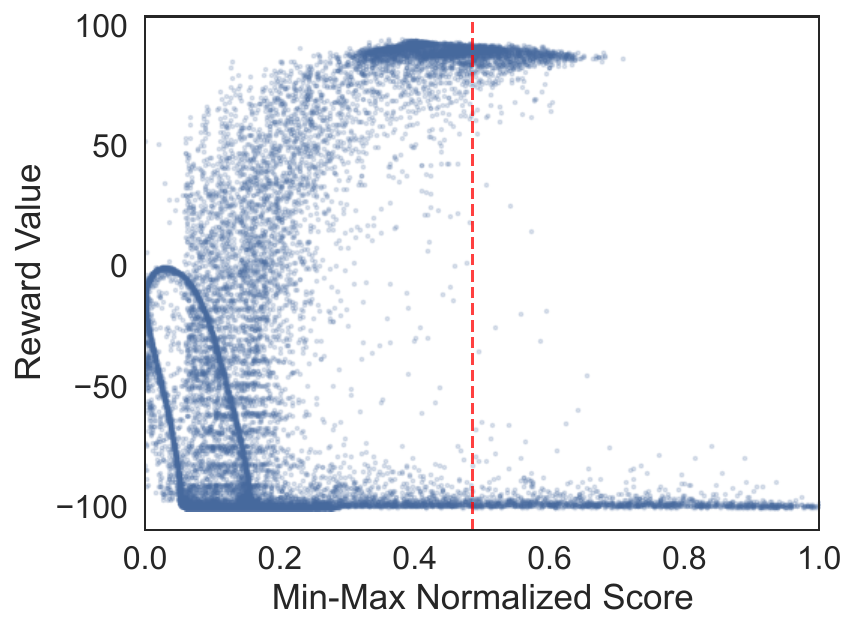}{MC, \codeblue{standard}} \hfill
    \stackcomparison{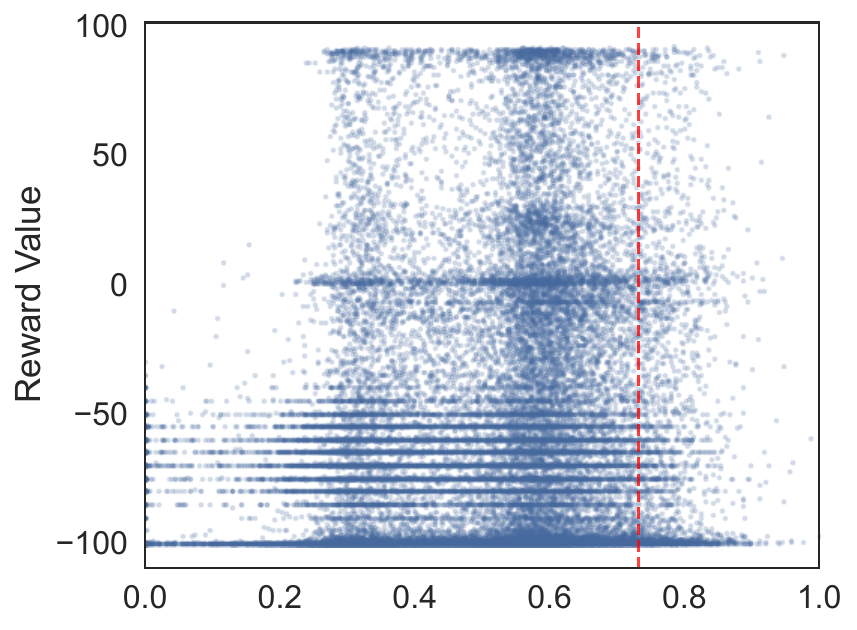}{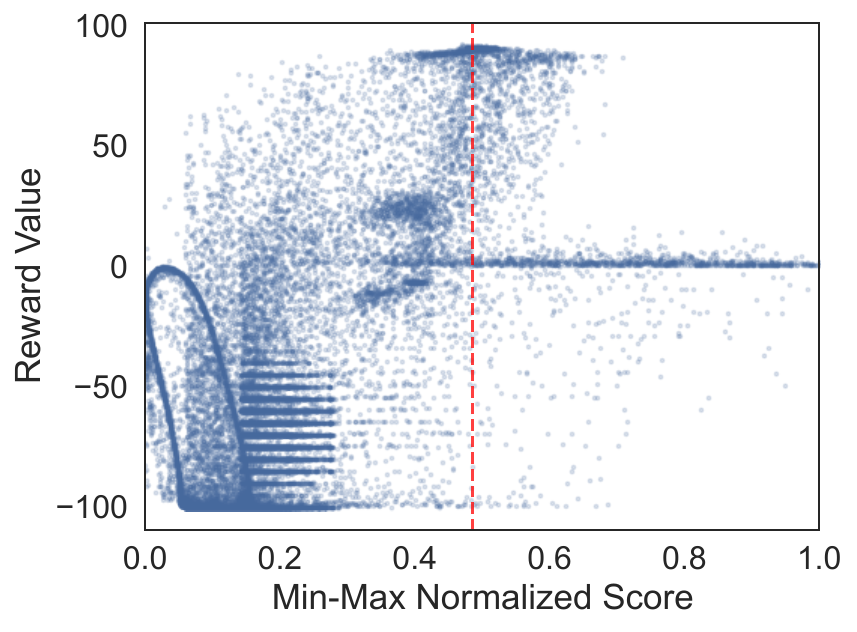}{MC, \codeblue{left}} \hfill
    \stackcomparison{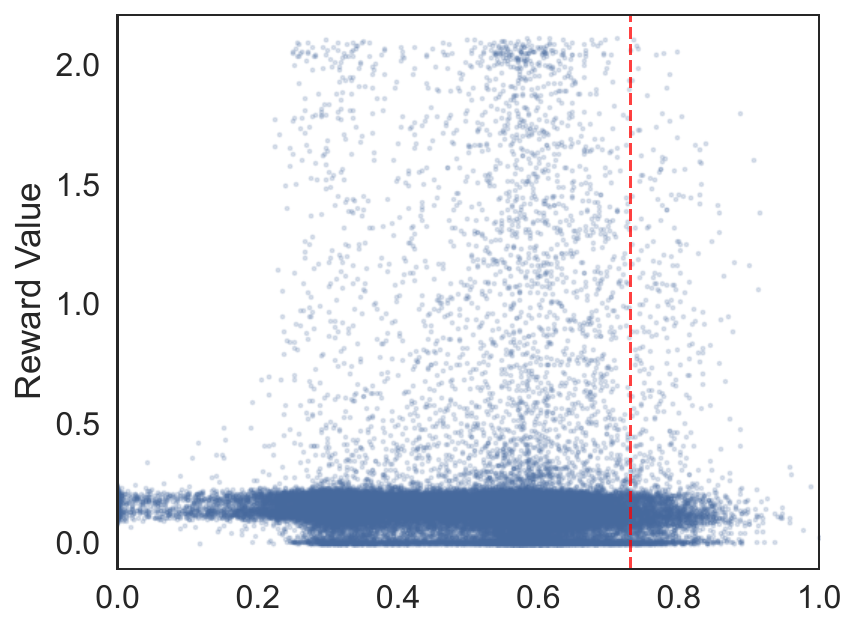}{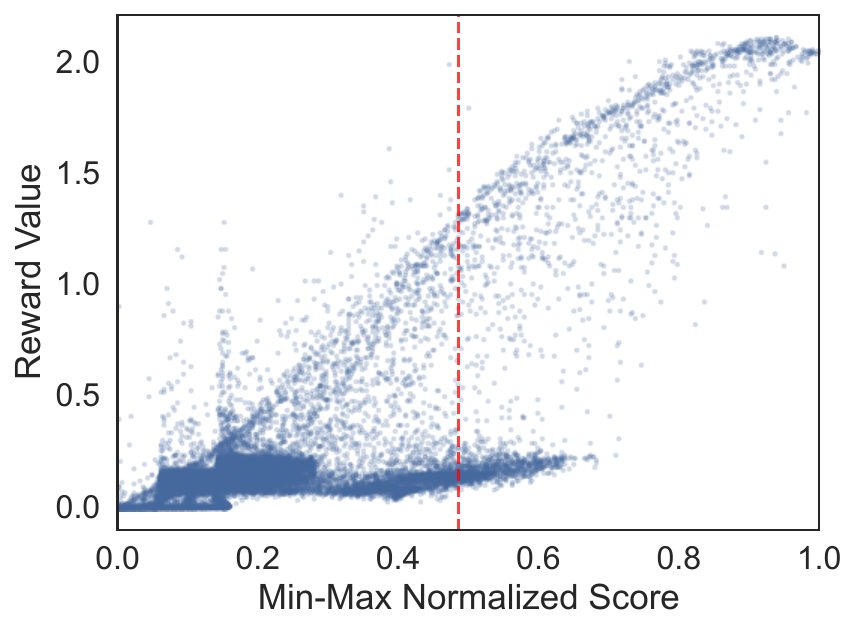}{MC, \codeblue{speed}} \hfill
    \stackcomparison{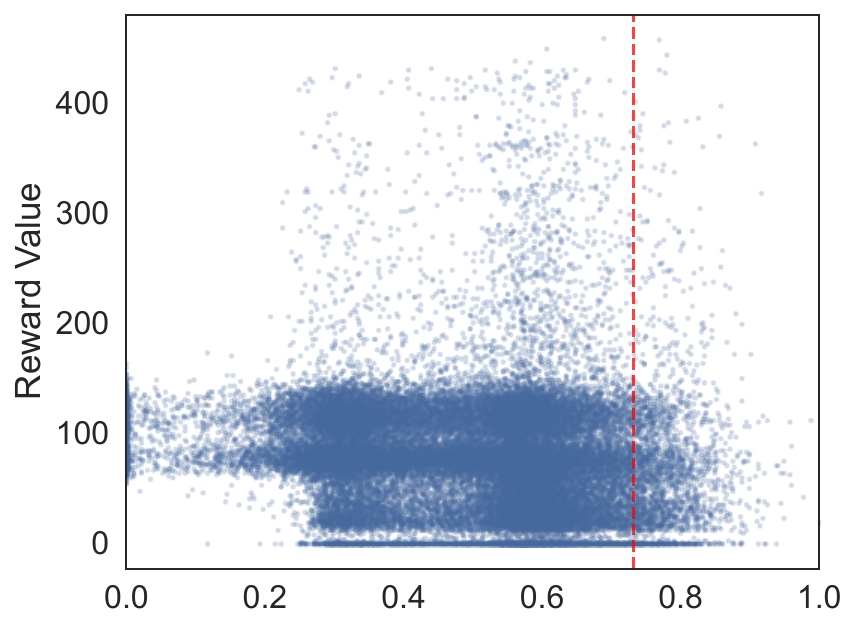}{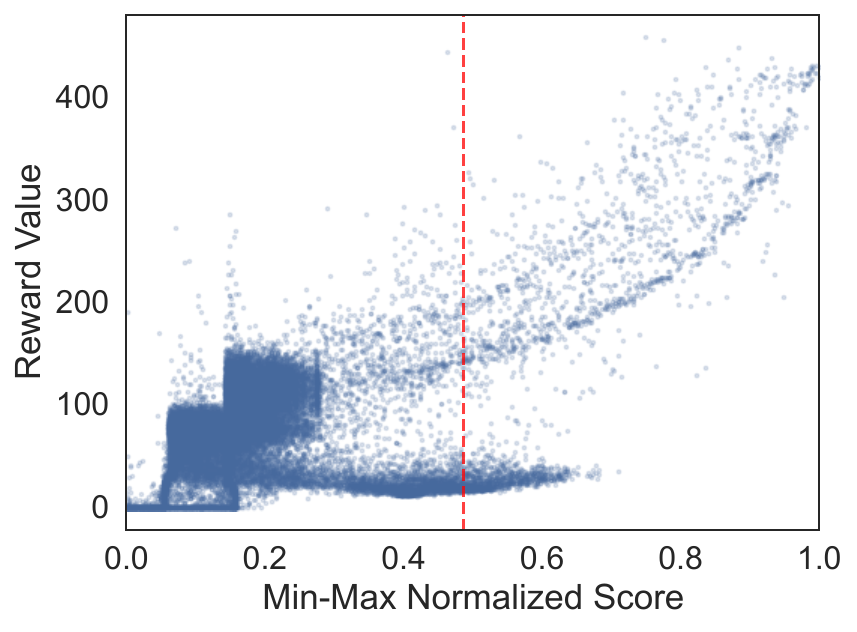}{MC, \codeblue{height}}

    \vspace{4mm}

    \stackcomparison{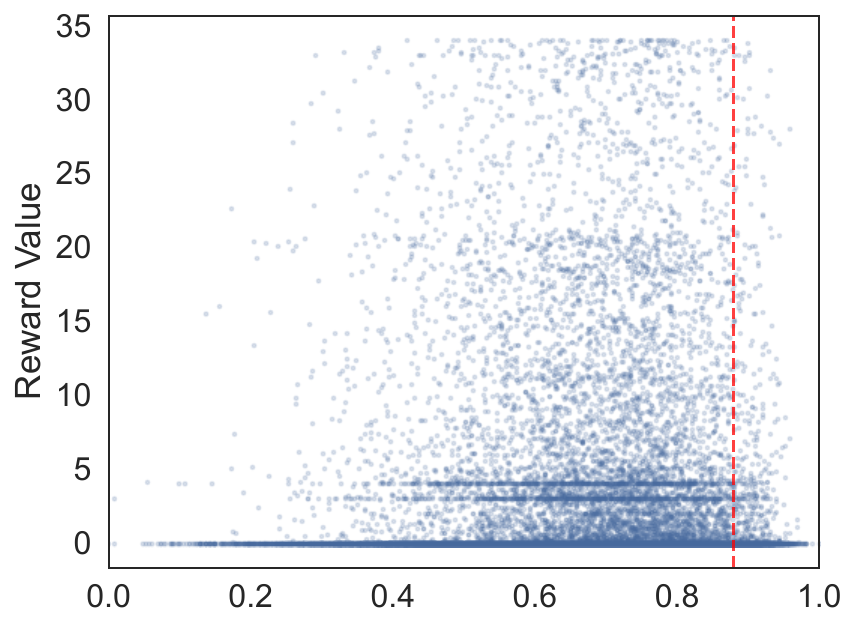}{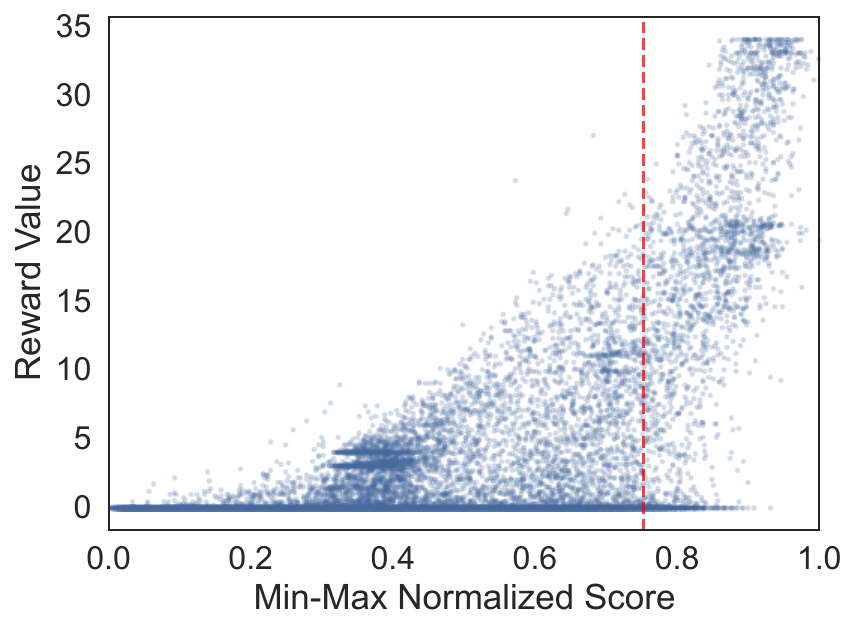}{RC, \codeblue{speed}} \hfill
    \stackcomparison{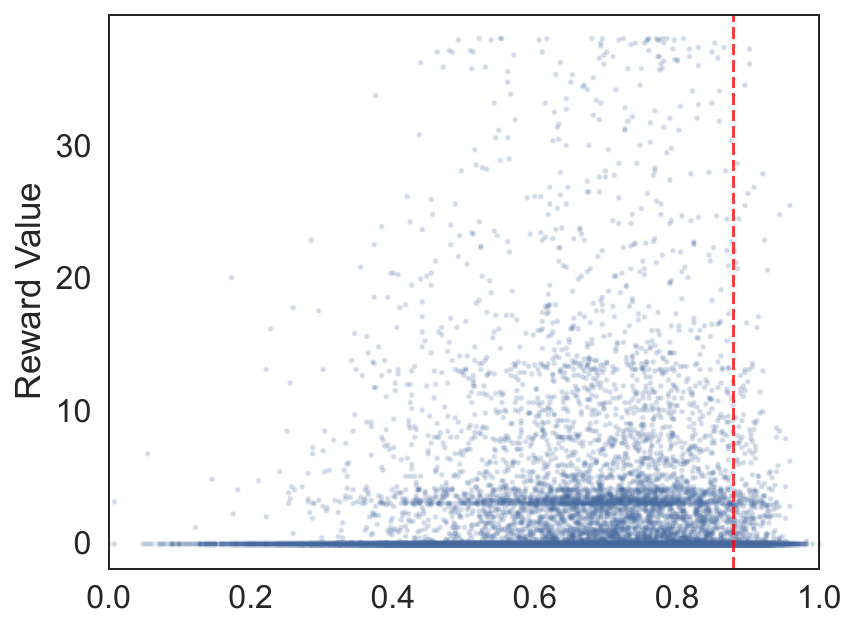}{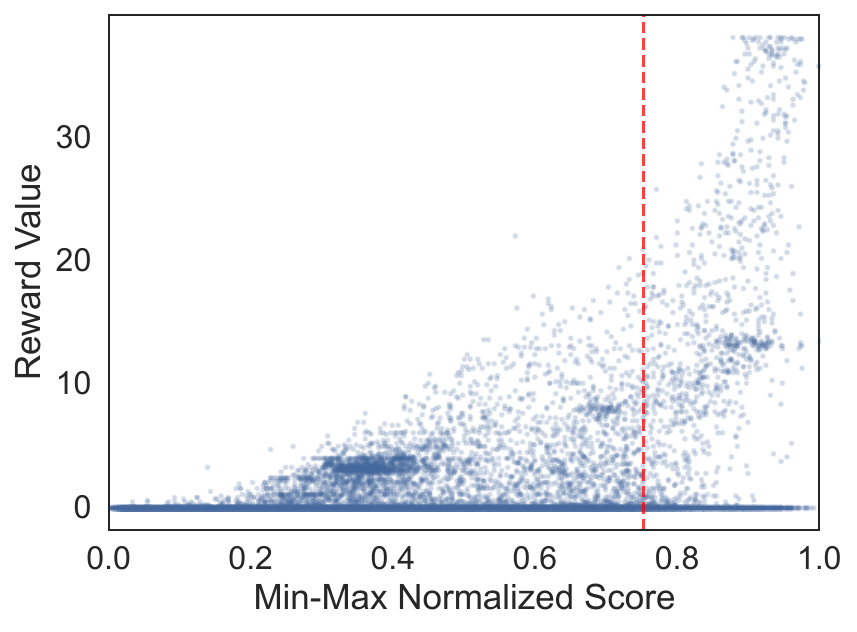}{RC, \codeblue{clockwise}} \hfill
    \stackcomparison{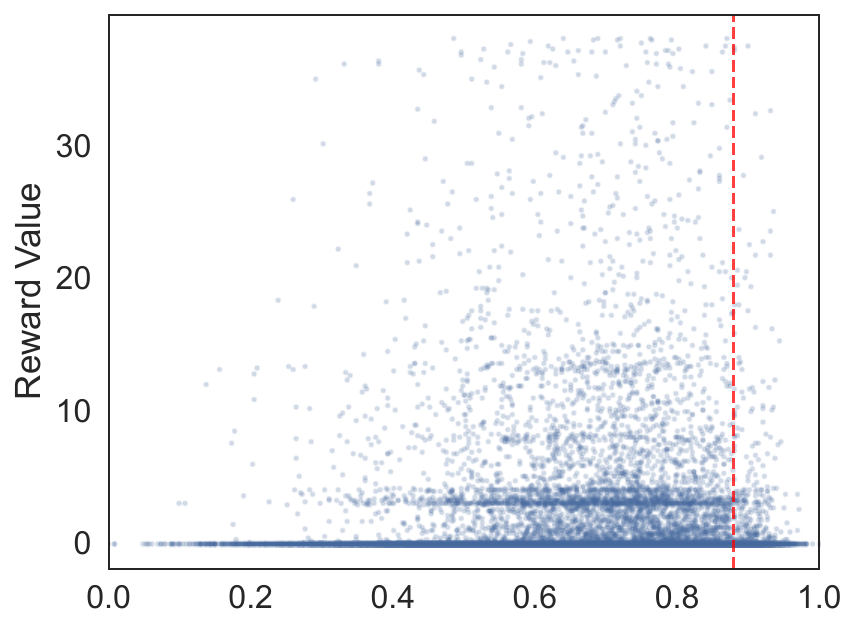}{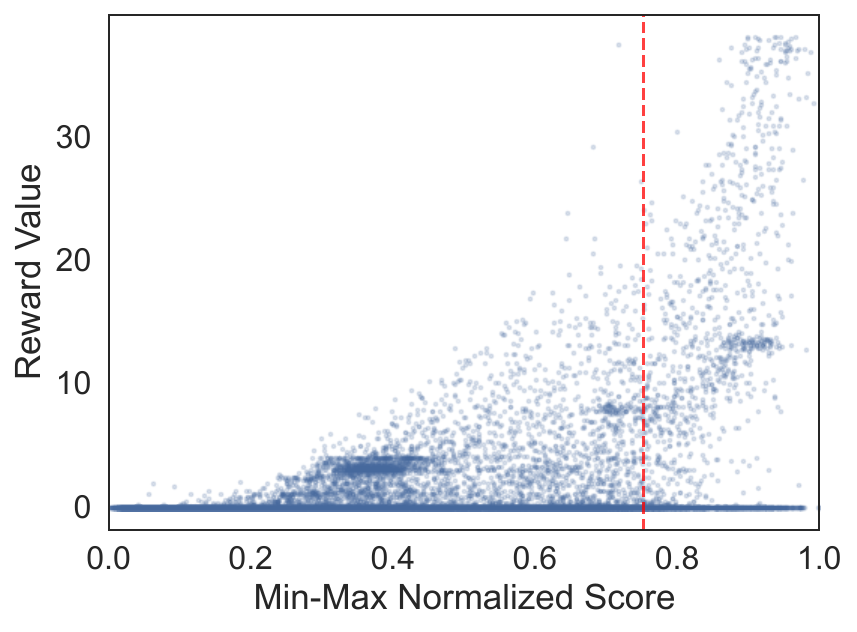}{RC, \codeblue{c-clockwise}} \hfill
    \stackcomparison{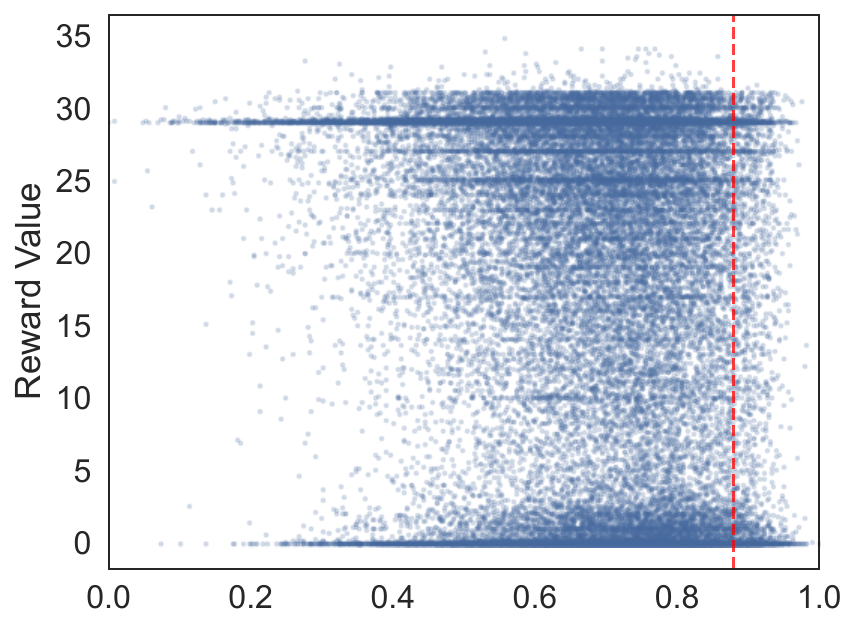}{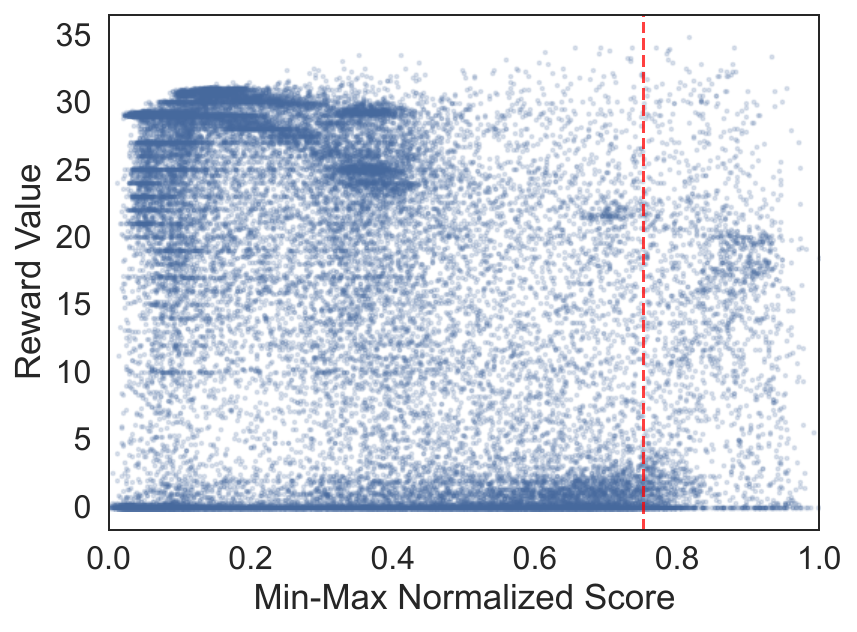}{RC, \codeblue{radial}}

    \vspace{4mm}

    \stackcomparison{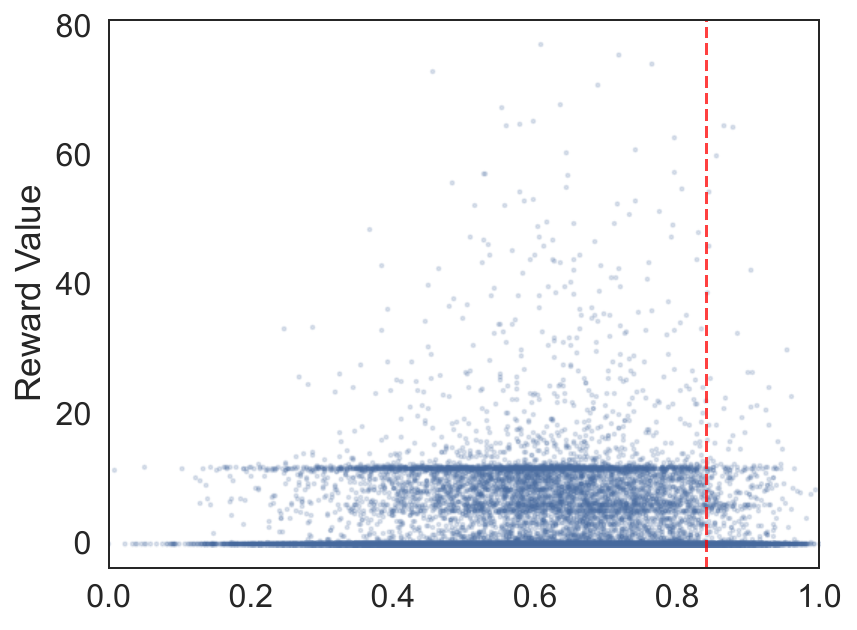}{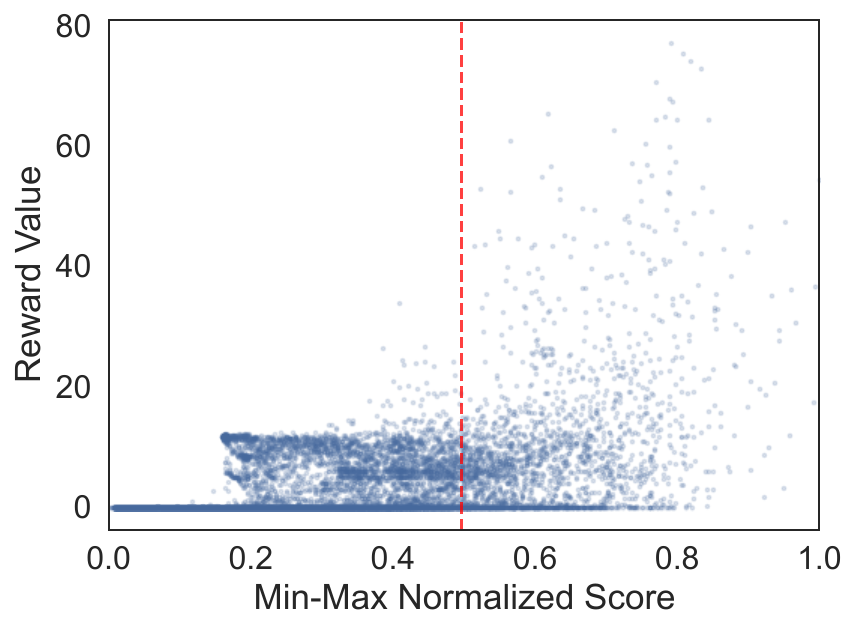}{HP, \codeblue{forward}} \hfill
    \stackcomparison{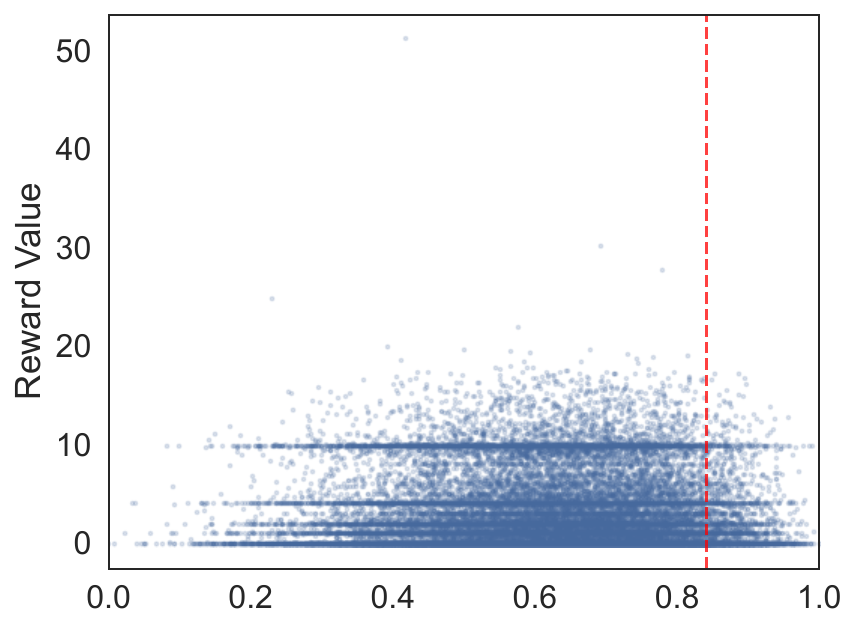}{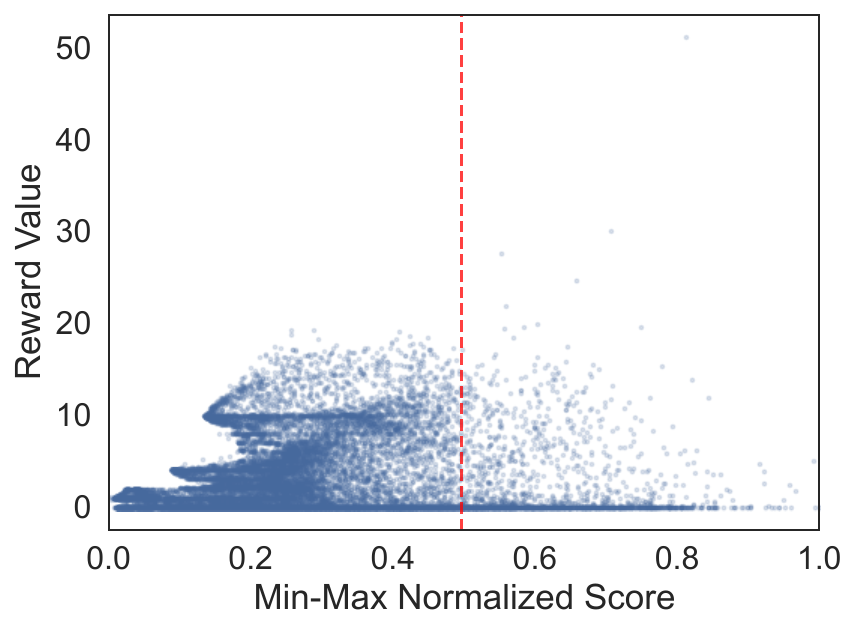}{HP, \codeblue{backward}} \hfill
    \stackcomparison{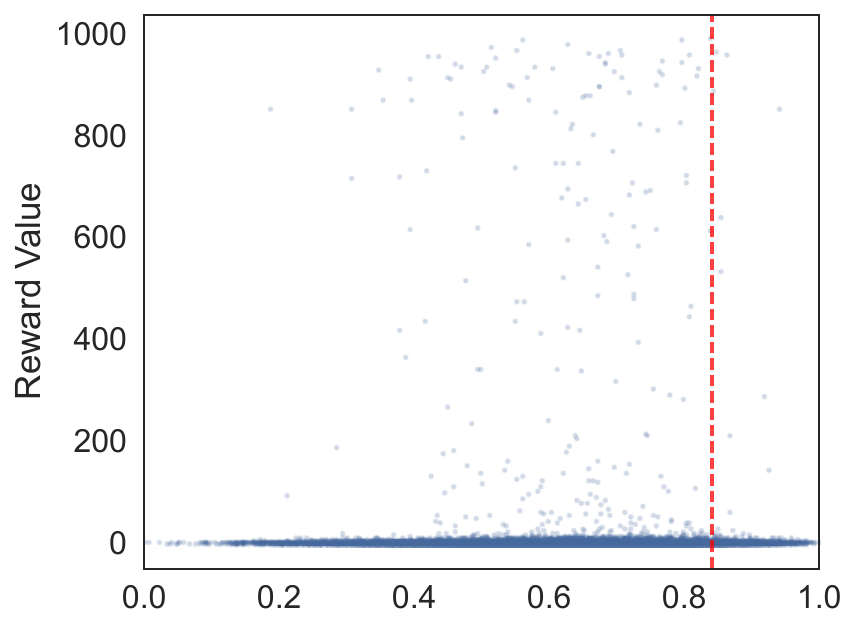}{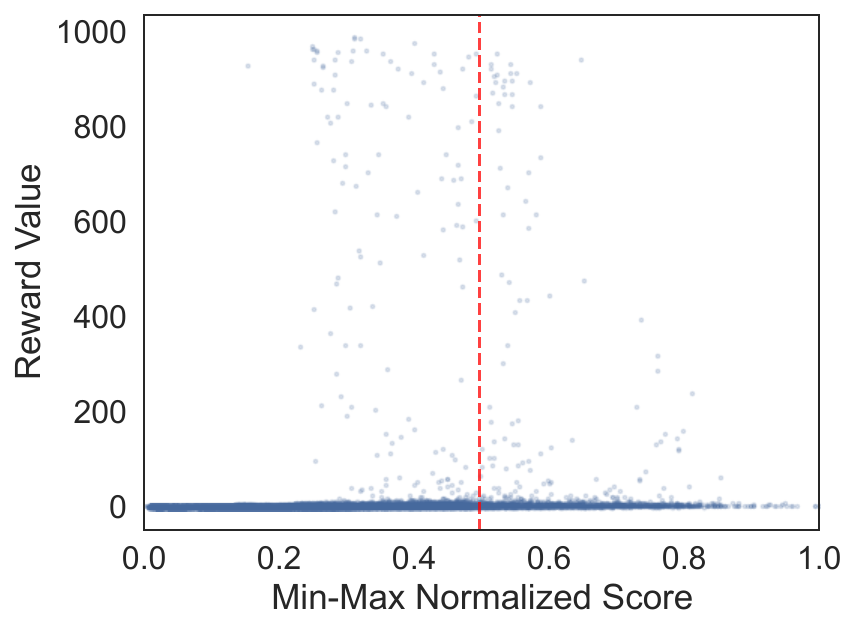}{HP, \codeblue{standstill}} \hfill
    \stackcomparison{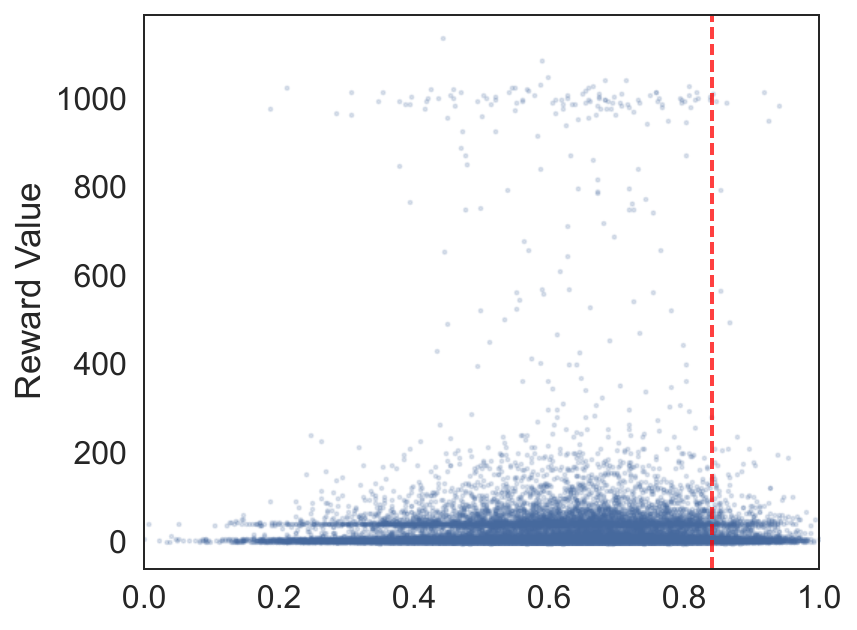}{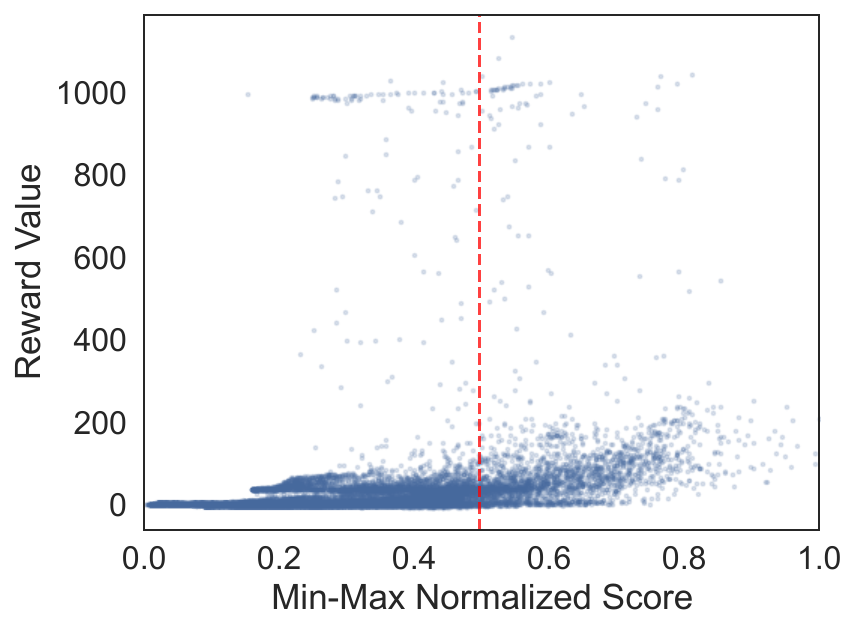}{HP, \codeblue{standard}}

    \caption{Correlation between APC (top row) and OPC (bottom row) scores and rewards in MC, RC, and HP. The vertical dashed line indicates the top 5\% selection threshold.}
    \label{fig:full_comparison_grid}
\end{figure}

In this experiment, we generate a dataset of $50,000$ policies for each environment (MC, RC, and HP). We then apply a top 5\% threshold based on scores generated by both the APC pipeline (obtained via novelty search in the Euclidean space of immediate actions) and the OPC pipeline (obtained via our information-theoretic uniqueness metric in the occupancy space). In \Figgref{fig:full_distribution_grid}, we visualize both dataset distributions compared to the complete (100\%) dataset. Consequently, in \Figgref{fig:full_comparison_grid}, we compare the distribution of scores that led to the previously mentioned thresholding.

As noted in \Secref{sec:exp_curation}, the scores generated by the APC pipeline tend to have less structure, consequently leading to the loss of rare, high-performing behaviors in the thresholded dataset. Conversely, the OPC scores generally favor unique behaviors. Interestingly, for tasks in which achieving a good performance is trivial, the scores successfully mitigate both the probability mass of low-performing and high-performing policies introduced by the random initialization (e.g., the \codeblue{radial} task for RC in \Figgref{fig:full_distribution_grid_radial}).

\subsection{Latent Behavior Compression}
\label{app:add_compression}

\begin{figure*}[t]
\vspace{-0.4cm}
    \centering
    \newcommand{\imgwidth}{0.20\textwidth}
    \setlength{\tabcolsep}{3pt}
    \begin{tabular}{m{5mm} c c c c}
        & \multicolumn{2}{c}{APC score} & \multicolumn{2}{c}{OPC score} \\
        \cmidrule(lr){2-3} \cmidrule(lr){4-5}
        & APC loss & OPC loss & APC loss & OPC loss \\[-.6ex]
        
        \rotatebox{90}{\hspace{0pt}\codeblue{Standard}} &
        \includegraphics[width=\imgwidth, valign=m]{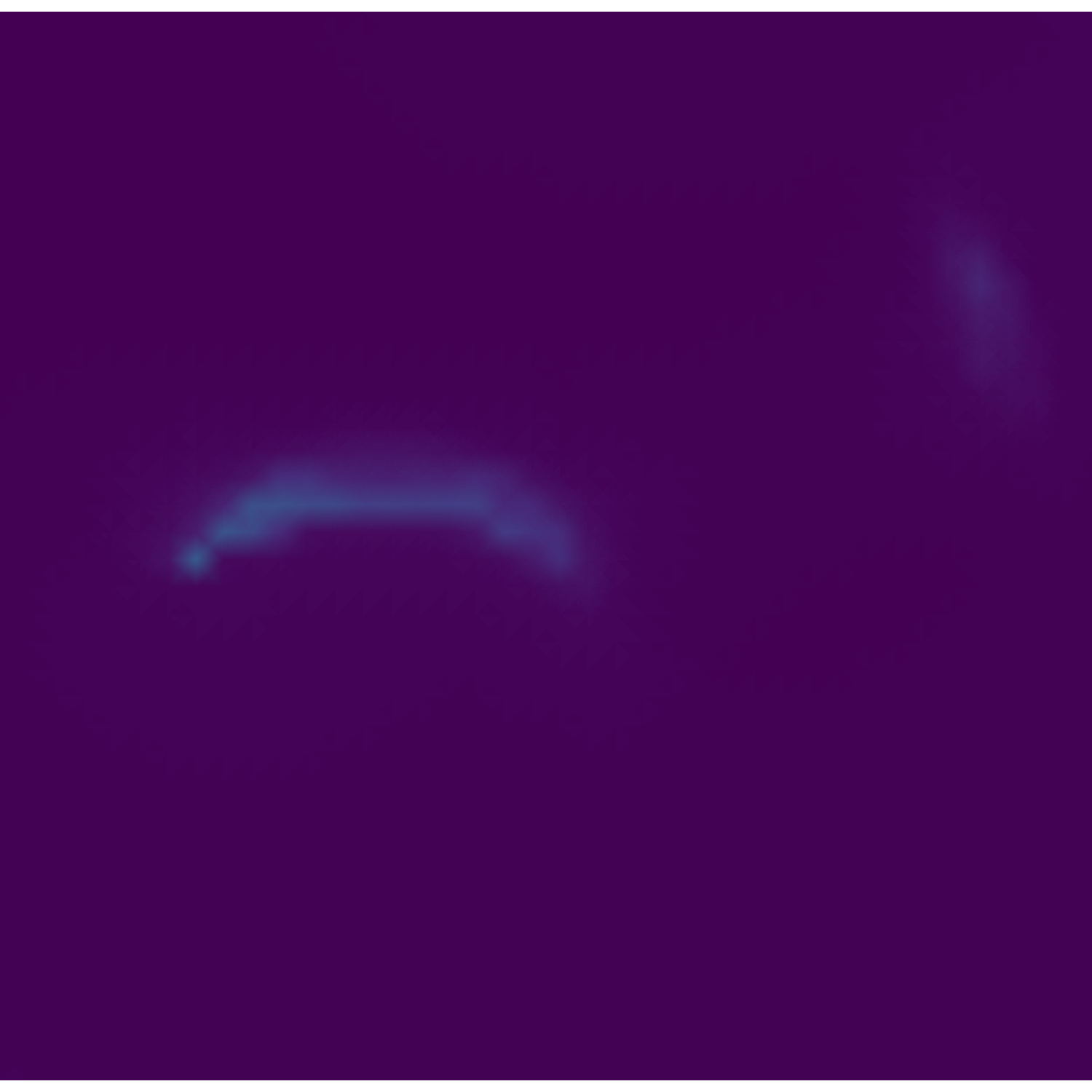} &
        \includegraphics[width=\imgwidth, valign=m]{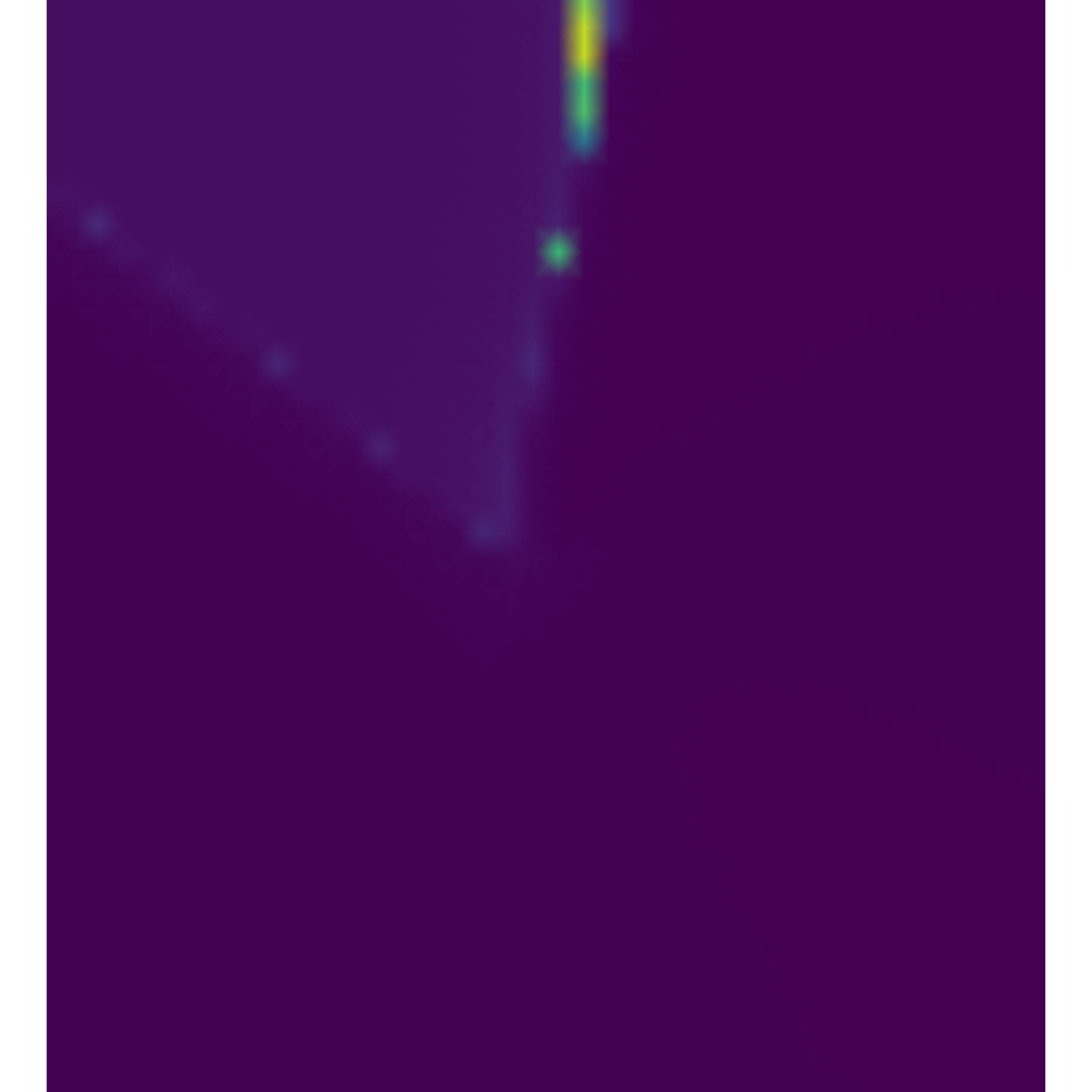} &
        \includegraphics[width=\imgwidth, valign=m]{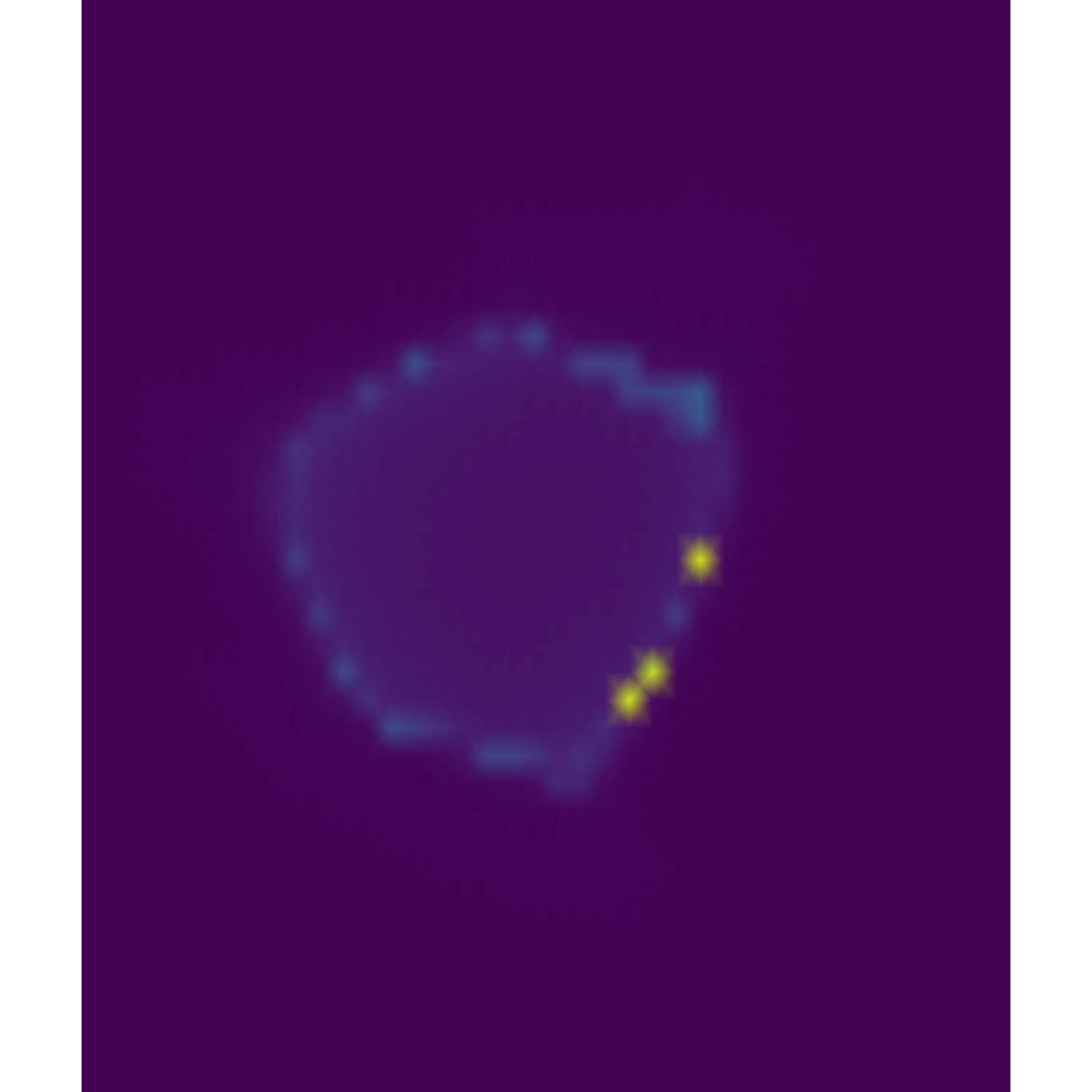} &
        \includegraphics[width=\imgwidth, valign=m]{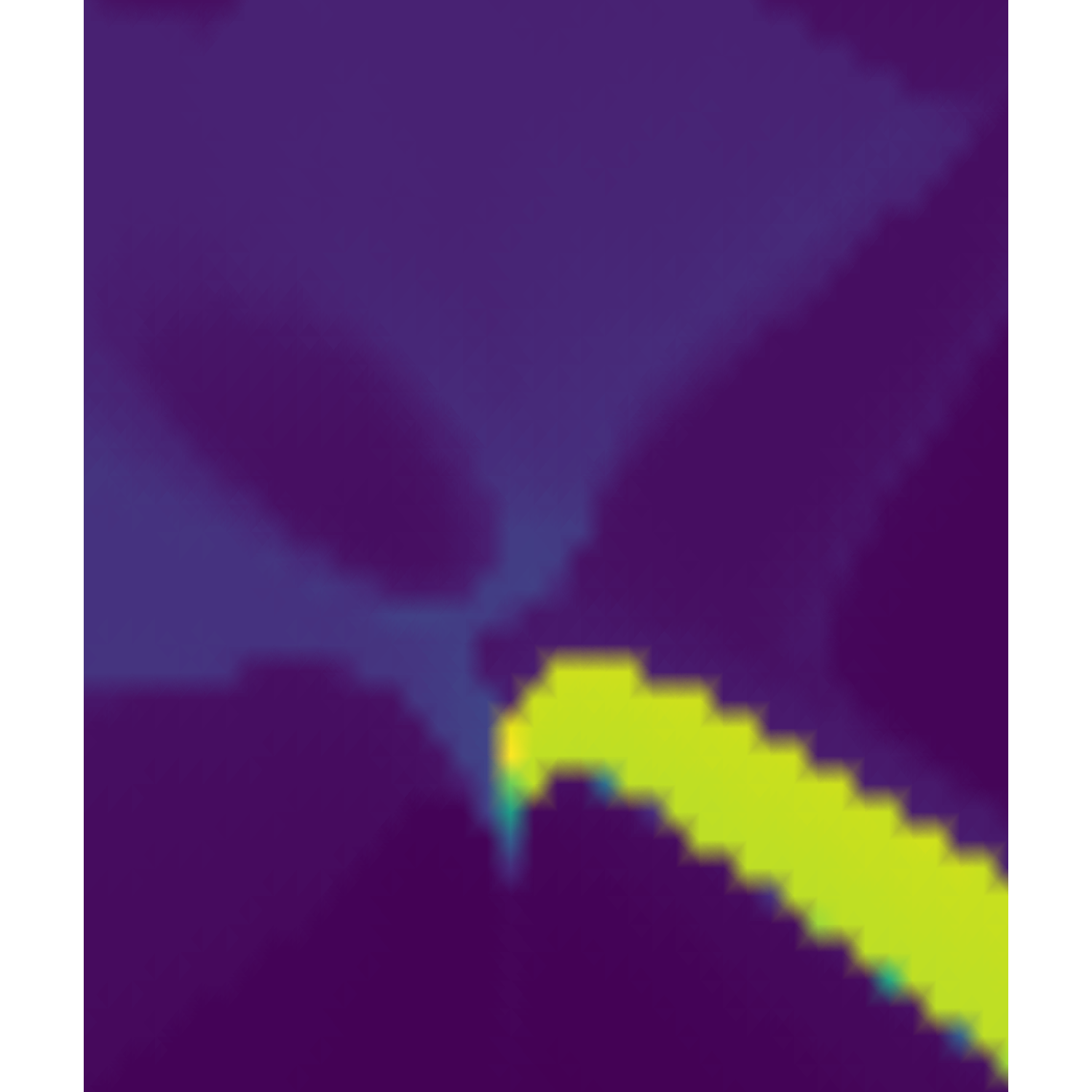} \\
        
        \rotatebox{90}{\hspace{0pt}\codeblue{Forward}} &
        \includegraphics[width=\imgwidth, valign=m]{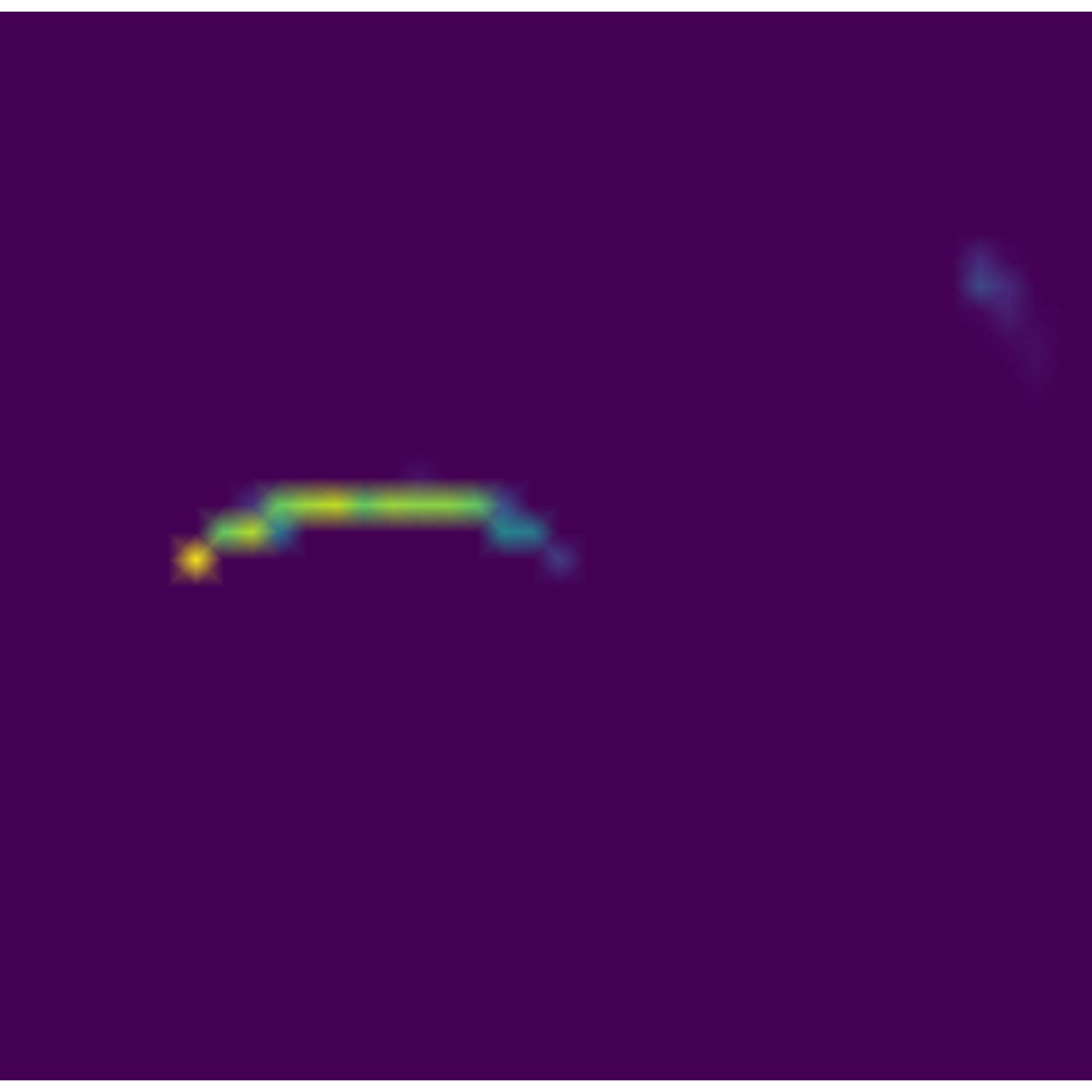} &
        \includegraphics[width=\imgwidth, valign=m]{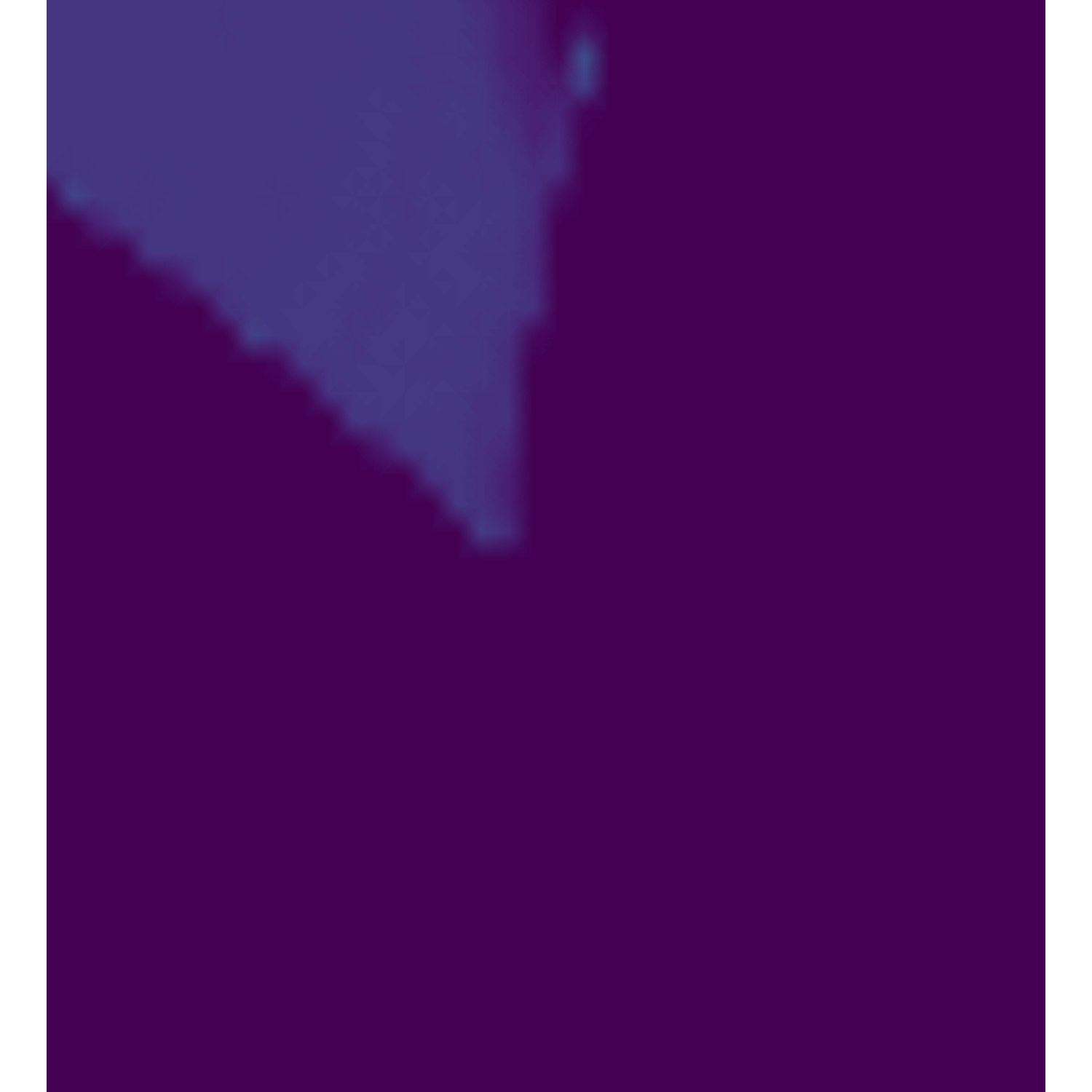} &
        \includegraphics[width=\imgwidth, valign=m]{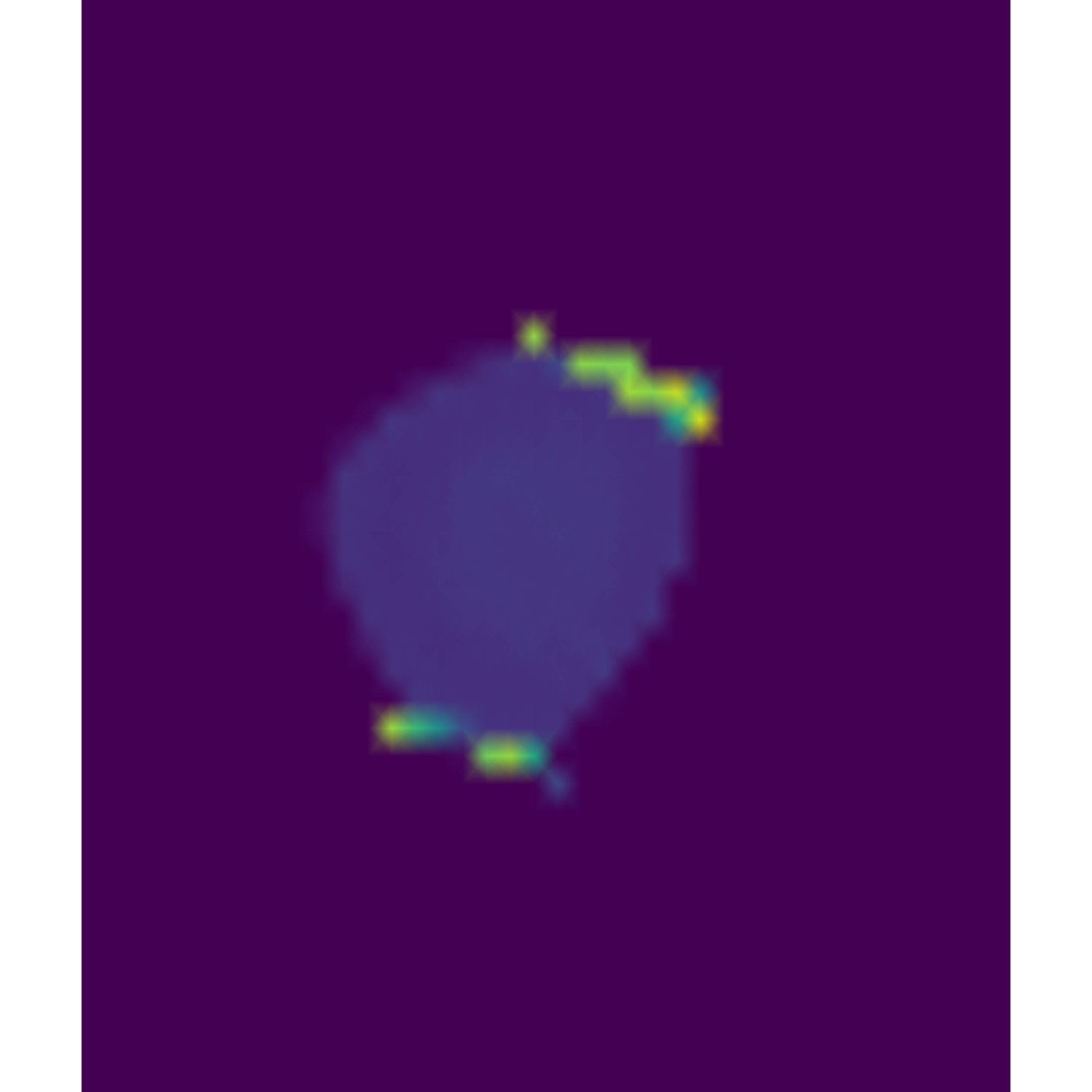} &
        \includegraphics[width=\imgwidth, valign=m]{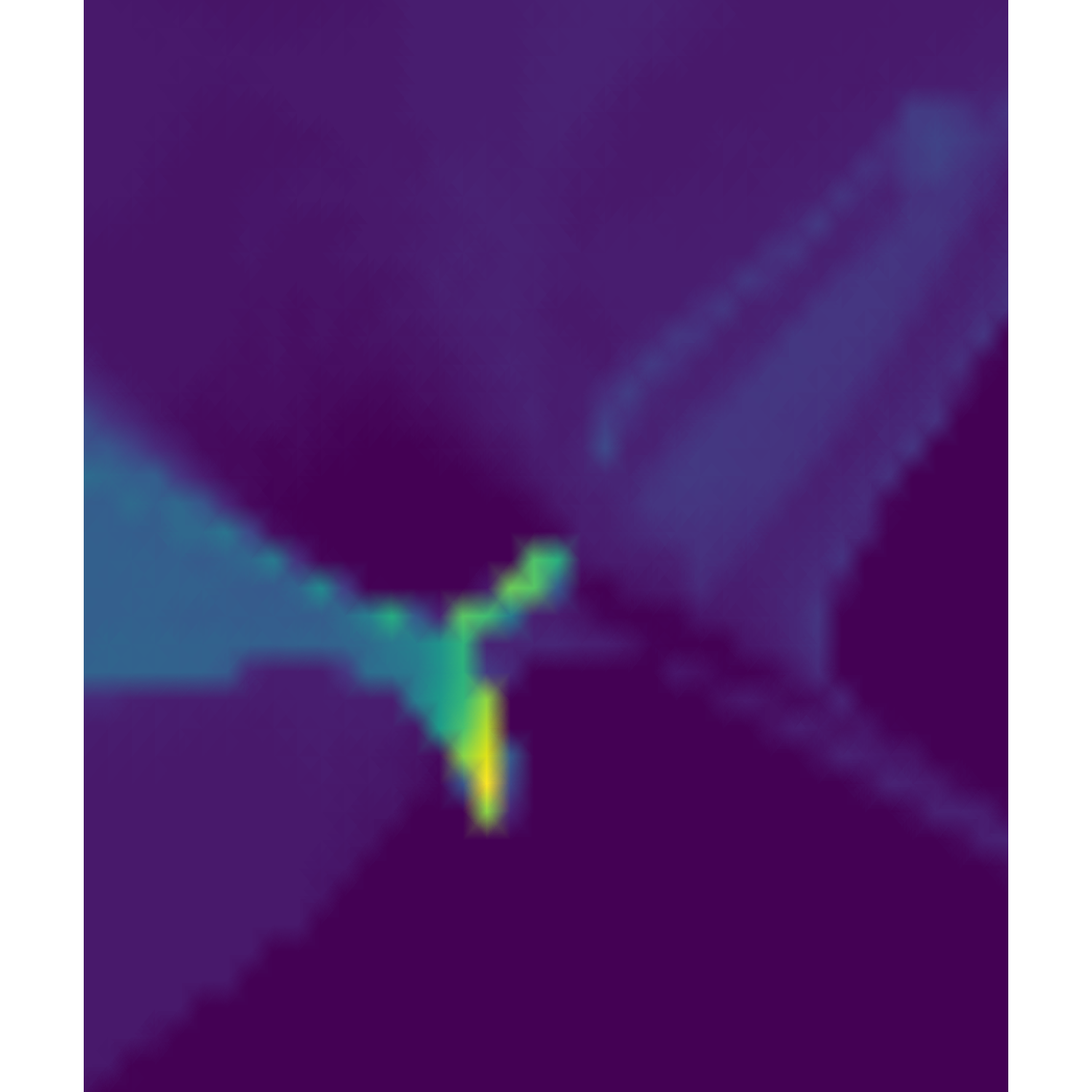} \\

        \rotatebox{90}{\hspace{0pt}\codeblue{Backward}} &
        \includegraphics[width=\imgwidth, valign=m]{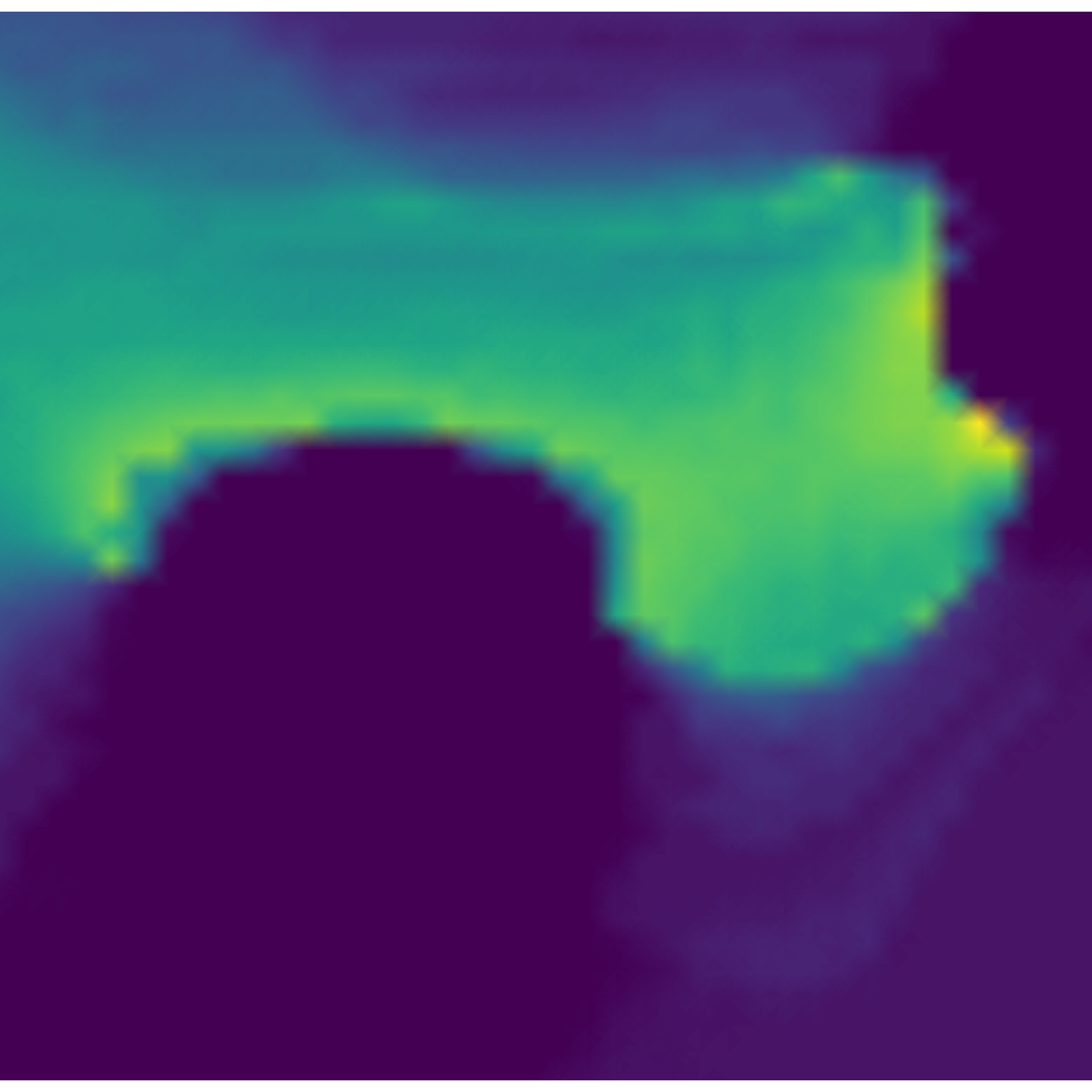} &
        \includegraphics[width=\imgwidth, valign=m]{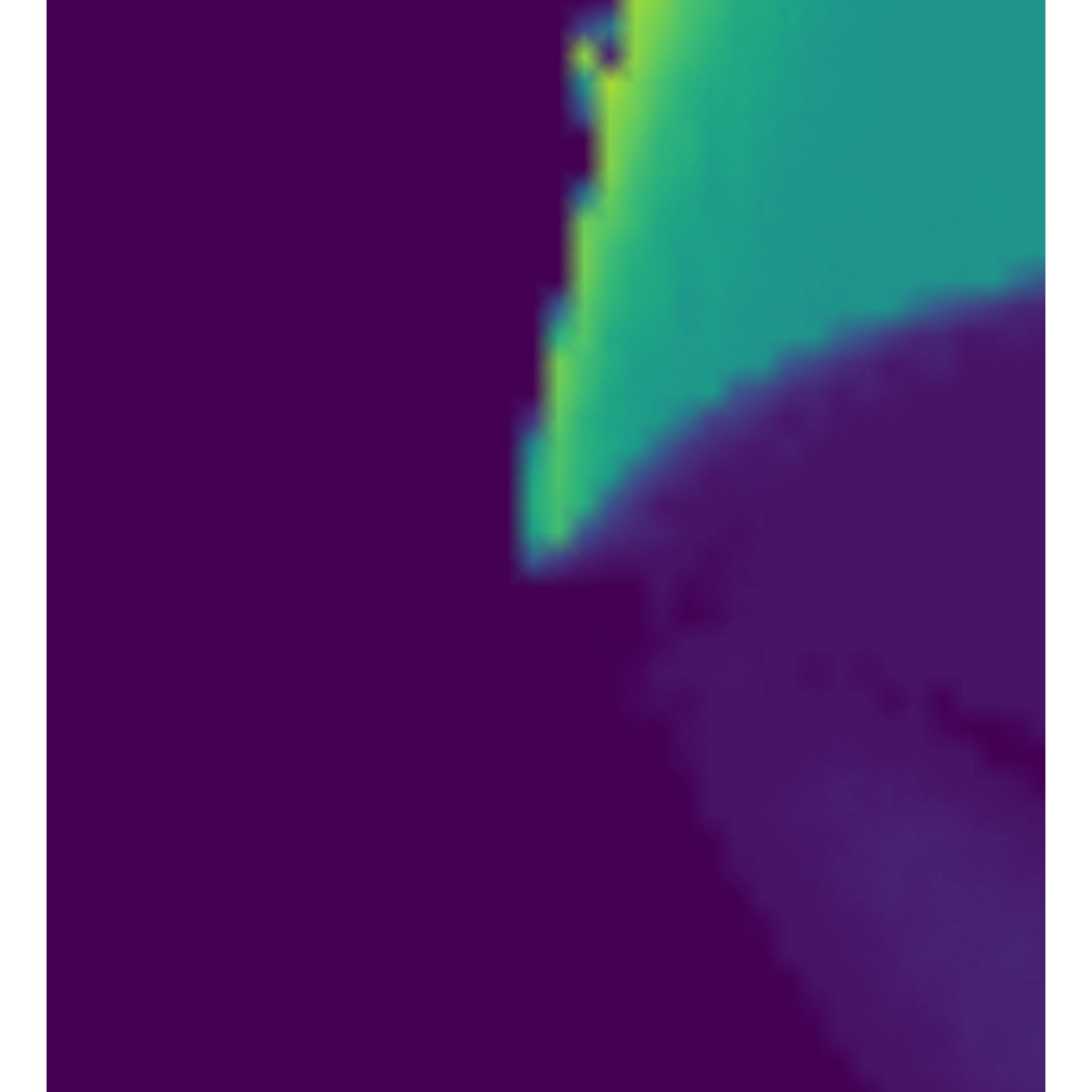} &
        \includegraphics[width=\imgwidth, valign=m]{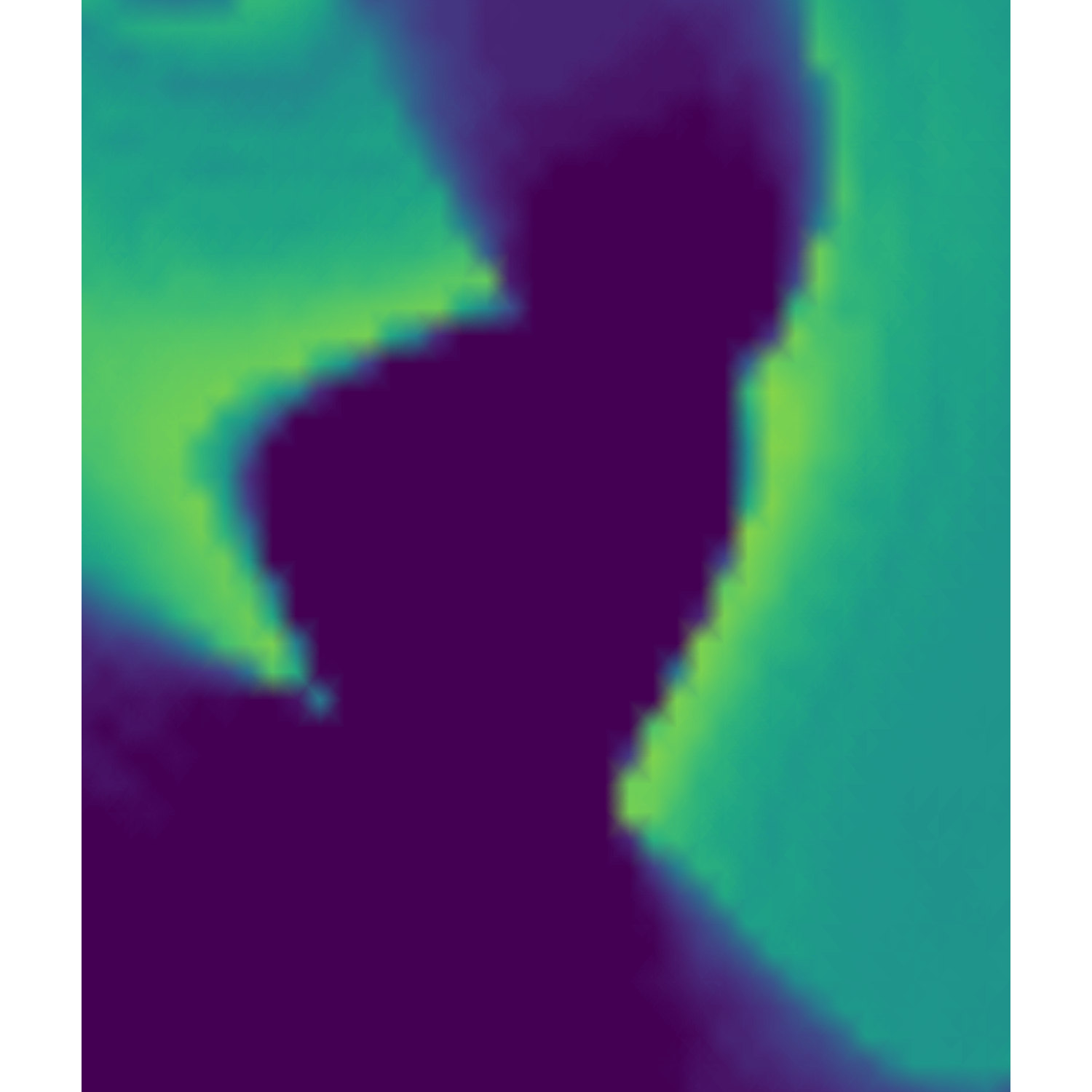} &
        \includegraphics[width=\imgwidth, valign=m]{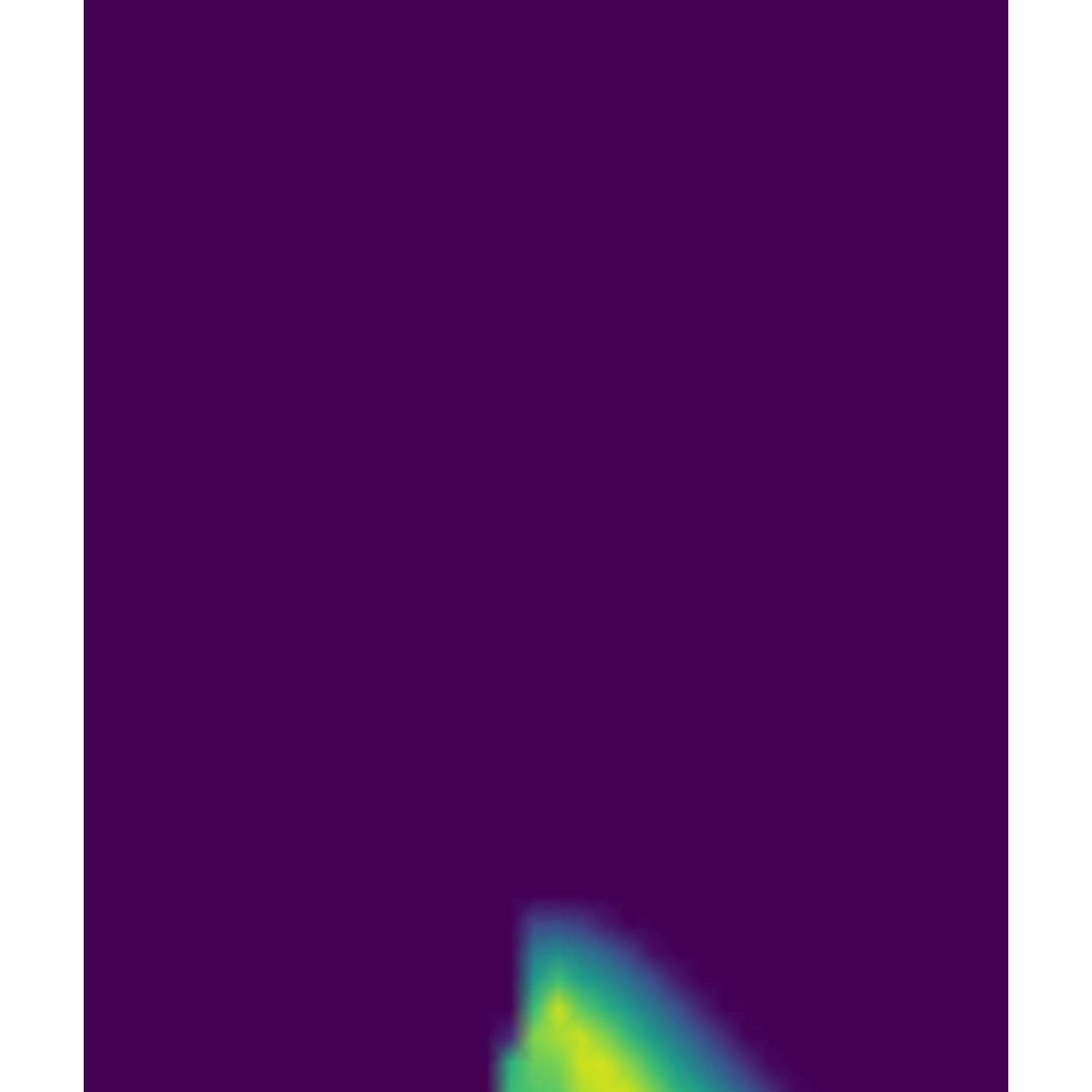} \\

        \rotatebox{90}{\hspace{0pt}\codeblue{Standstill}} &
        \includegraphics[width=\imgwidth, valign=m]{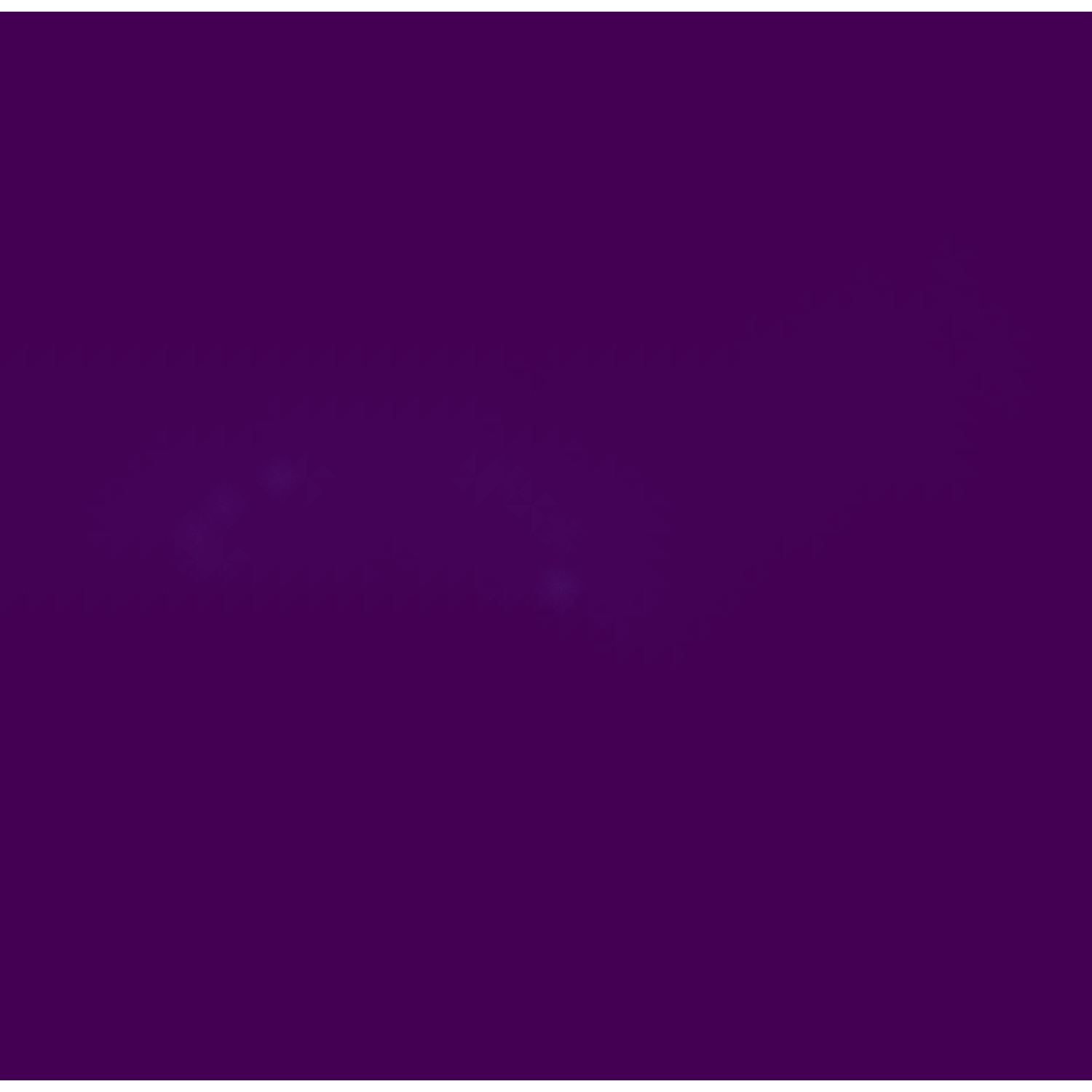} &
        \includegraphics[width=\imgwidth, valign=m]{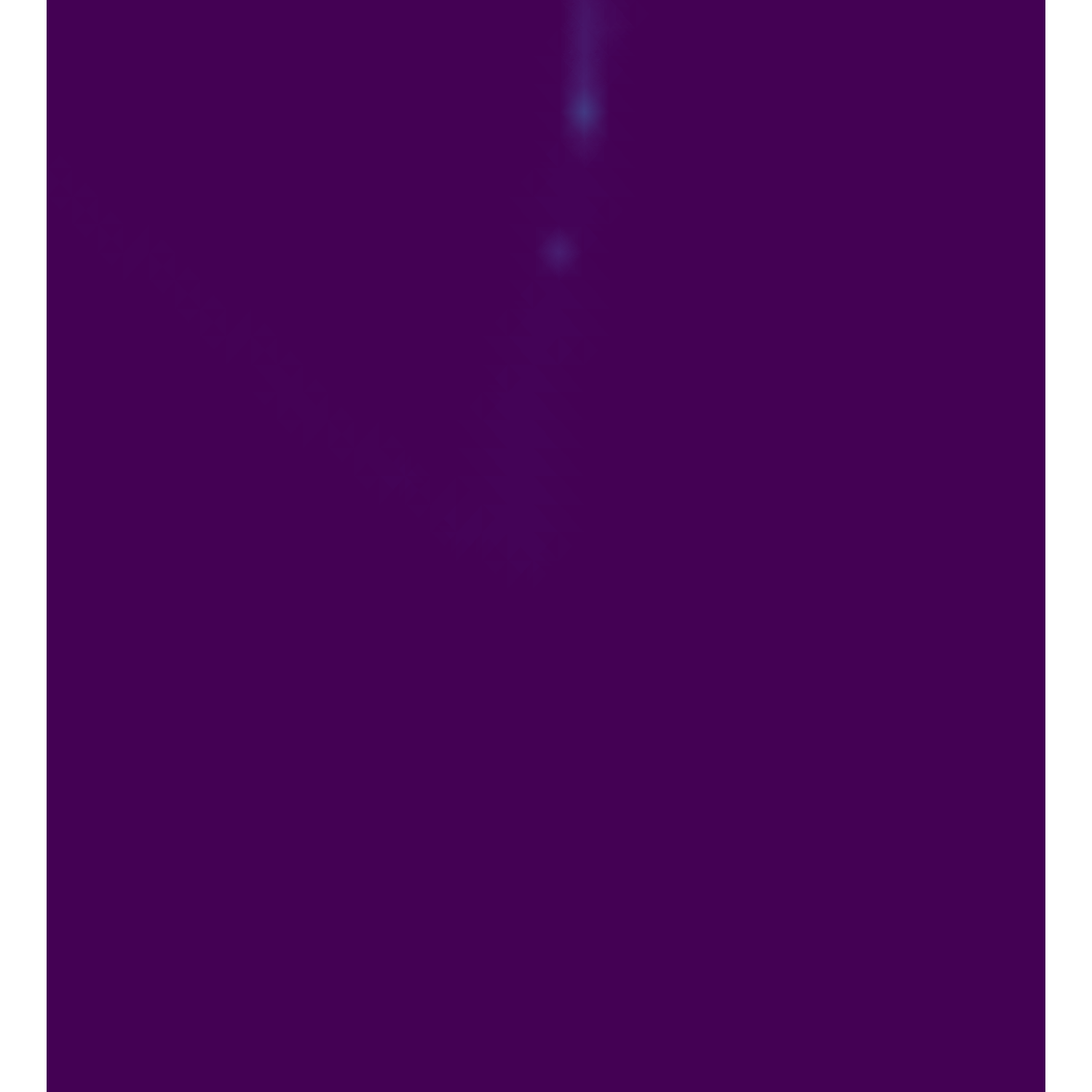} &
        \includegraphics[width=\imgwidth, valign=m]{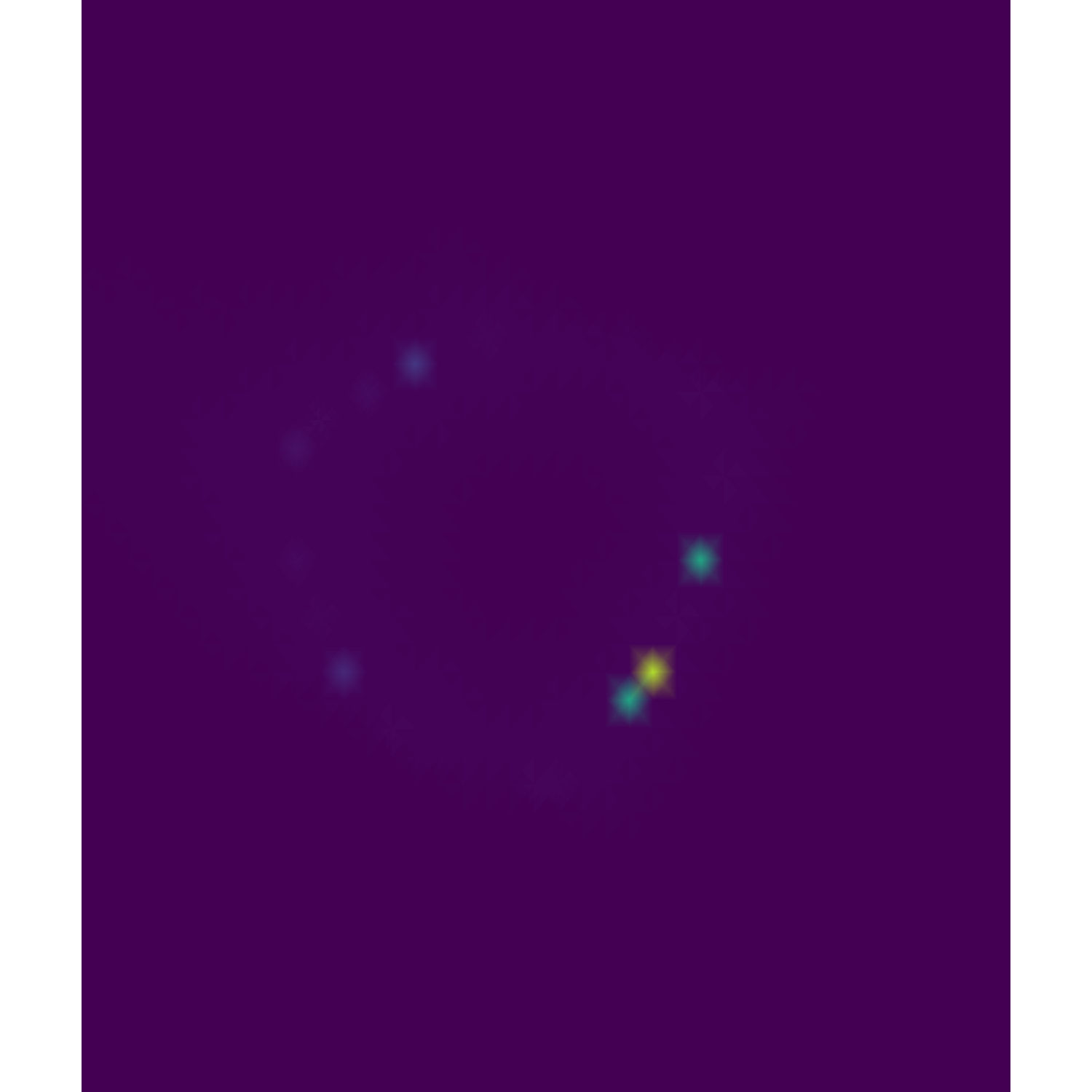} &
        \includegraphics[width=\imgwidth, valign=m]{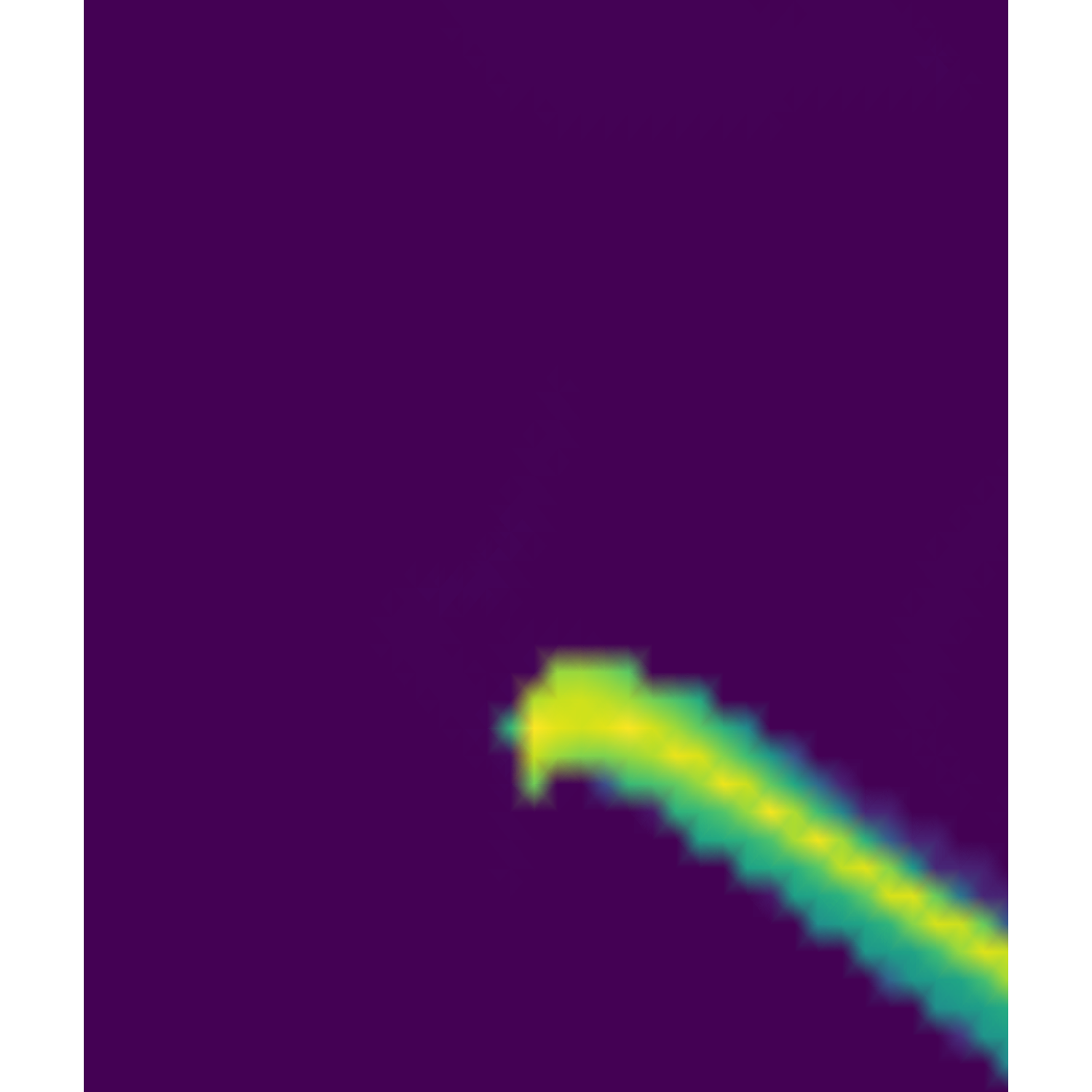} \\
    \end{tabular}
    \caption{HP, seed 888, 2D}
    \label{fig:full_latent_comparison_888_2d}
\end{figure*}

In this ablation study, we first generate a dataset of $50,000$ policies for the HP environment. We then curate it to 5\% using both the scores from the OPC and APC pipelines. Finally, we train two autoencoders: one with the action-matching objective from the APC pipeline and one with the occupancy-matching objective from the OPC pipeline. In all, we obtain four latent spaces: one obtained by running the full APC pipeline; one obtained by running the full OPC pipeline; one obtained by optimizing the OPC objective on the APC curated dataset; and, finally, one obtained by optimizing the APC objective on the OPC curated dataset. We repeat this process with two different latent dimensions: 2D and 3D. We compare the obtained topologies in \quadfigref{fig:full_latent_comparison_888_2d}{fig:full_latent_comparison_999_2d}{fig:full_latent_comparison_888_3d}{fig:full_latent_comparison_999_3d}.

\begin{figure*}[t]
\vspace{-0.4cm}
    \centering
    \newcommand{\imgwidth}{0.20\textwidth}
    \setlength{\tabcolsep}{3pt}
    \begin{tabular}{m{5mm} c c c c}
        & \multicolumn{2}{c}{APC score} & \multicolumn{2}{c}{OPC score} \\
        \cmidrule(lr){2-3} \cmidrule(lr){4-5}
        & APC loss & OPC loss & APC loss & OPC loss \\[-.6ex]
        
        \rotatebox{90}{\hspace{0pt}\codeblue{Standard}} &
        \includegraphics[width=\imgwidth, valign=m]{figures/experiments/mega_compare/999/2D/standard_A_scores_A_loss.pdf} &
        \includegraphics[width=\imgwidth, valign=m]{figures/experiments/mega_compare/999/2D/standard_A_scores_O_loss.pdf} &
        \includegraphics[width=\imgwidth, valign=m]{figures/experiments/mega_compare/999/2D/standard_O_scores_A_loss.pdf} &
        \includegraphics[width=\imgwidth, valign=m]{figures/experiments/mega_compare/999/2D/standard_O_scores_O_loss.pdf} \\
        
        \rotatebox{90}{\hspace{0pt}\codeblue{Forward}} &
        \includegraphics[width=\imgwidth, valign=m]{figures/experiments/mega_compare/999/2D/forward_1_A_scores_A_loss.pdf} &
        \includegraphics[width=\imgwidth, valign=m]{figures/experiments/mega_compare/999/2D/forward_1_A_scores_O_loss.pdf} &
        \includegraphics[width=\imgwidth, valign=m]{figures/experiments/mega_compare/999/2D/forward_1_O_scores_A_loss.pdf} &
        \includegraphics[width=\imgwidth, valign=m]{figures/experiments/mega_compare/999/2D/forward_1_O_scores_O_loss.pdf} \\

        \rotatebox{90}{\hspace{0pt}\codeblue{Backward}} &
        \includegraphics[width=\imgwidth, valign=m]{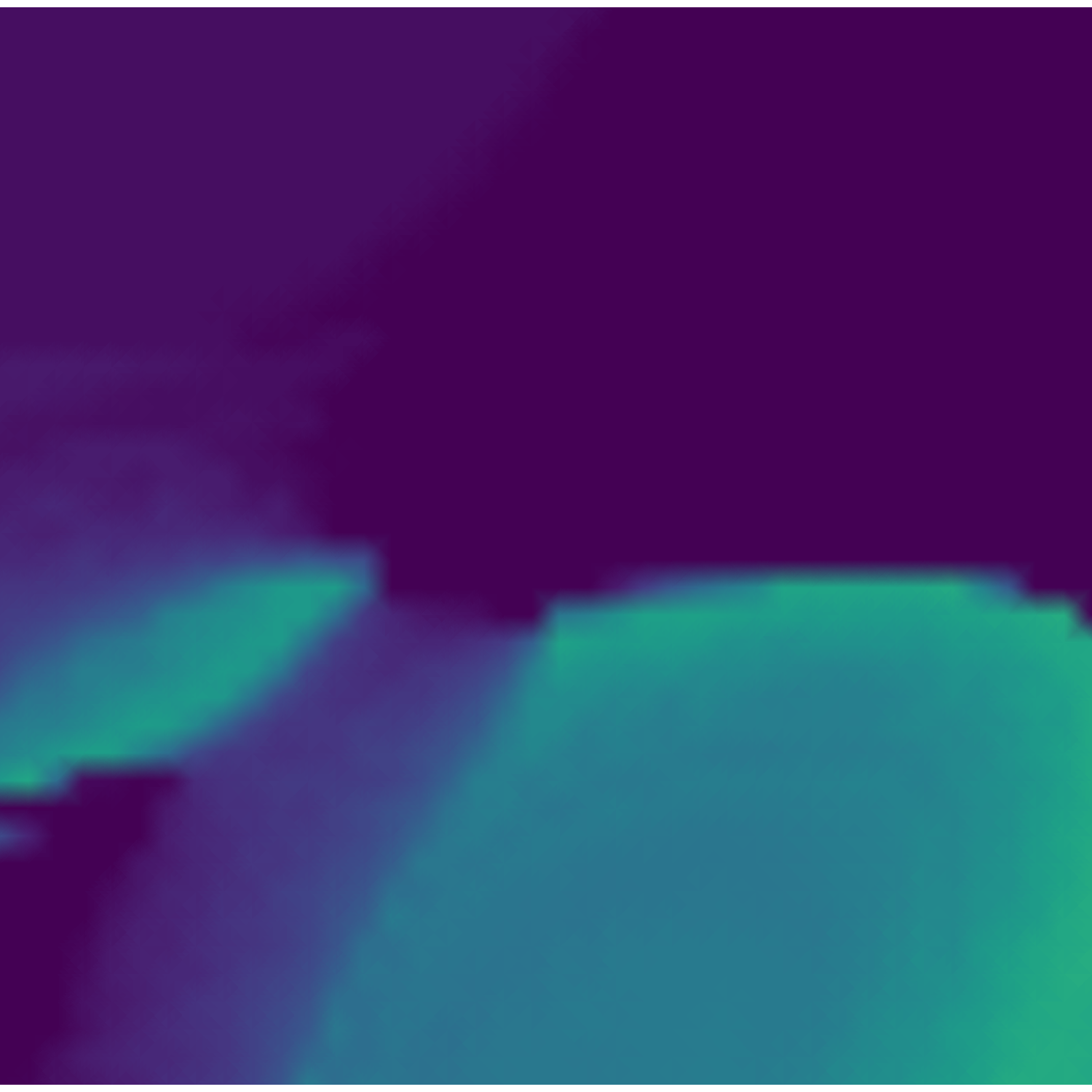} &
        \includegraphics[width=\imgwidth, valign=m]{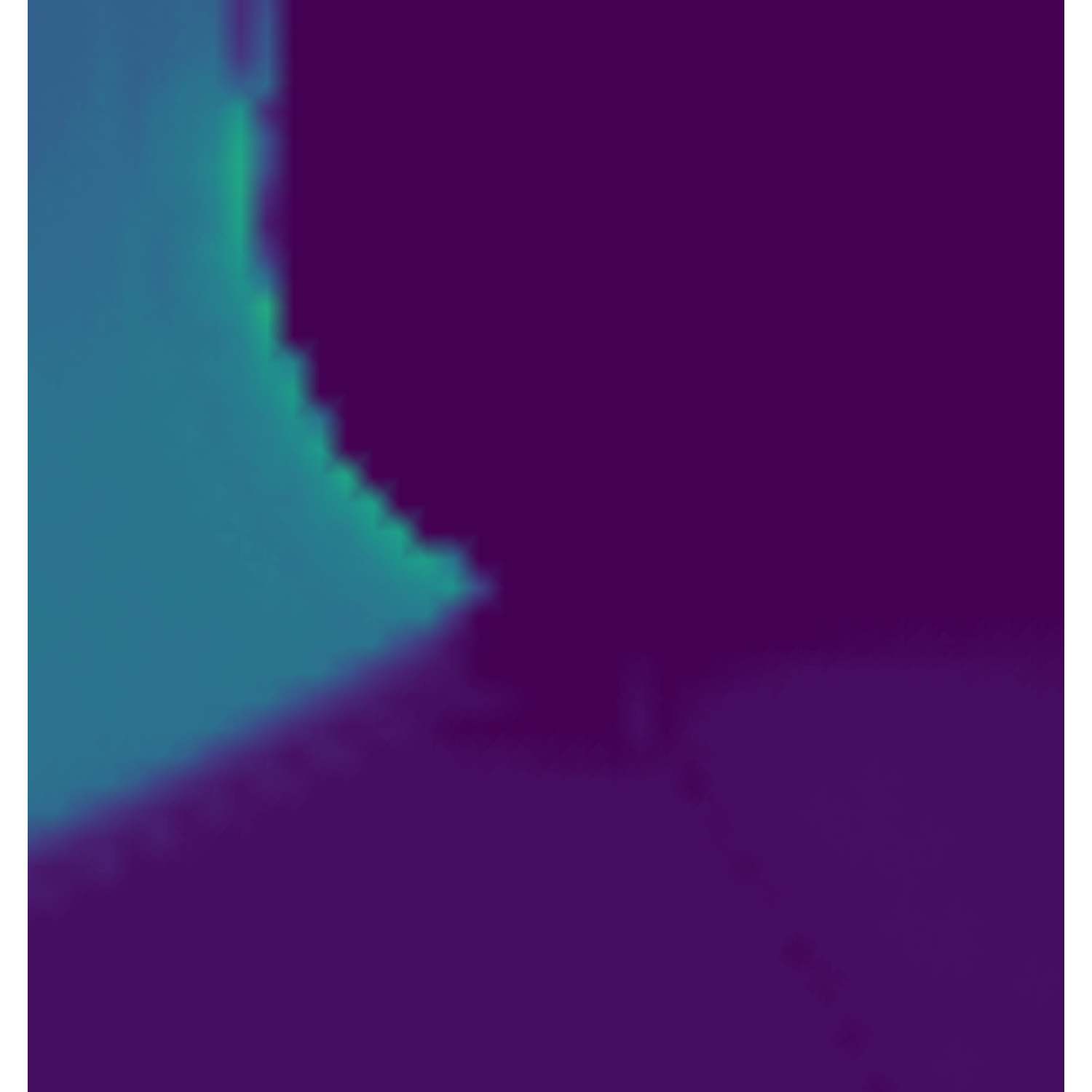} &
        \includegraphics[width=\imgwidth, valign=m]{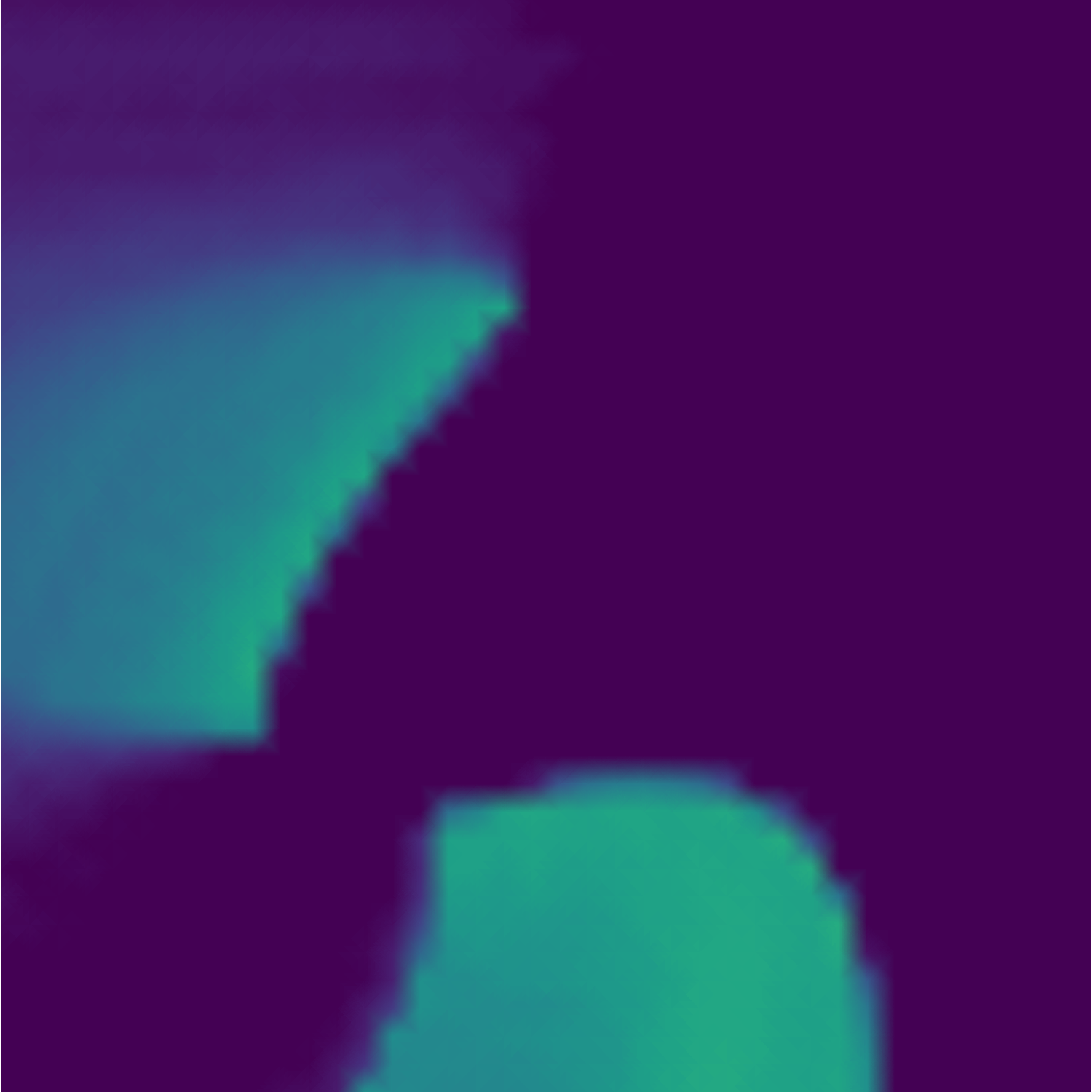} &
        \includegraphics[width=\imgwidth, valign=m]{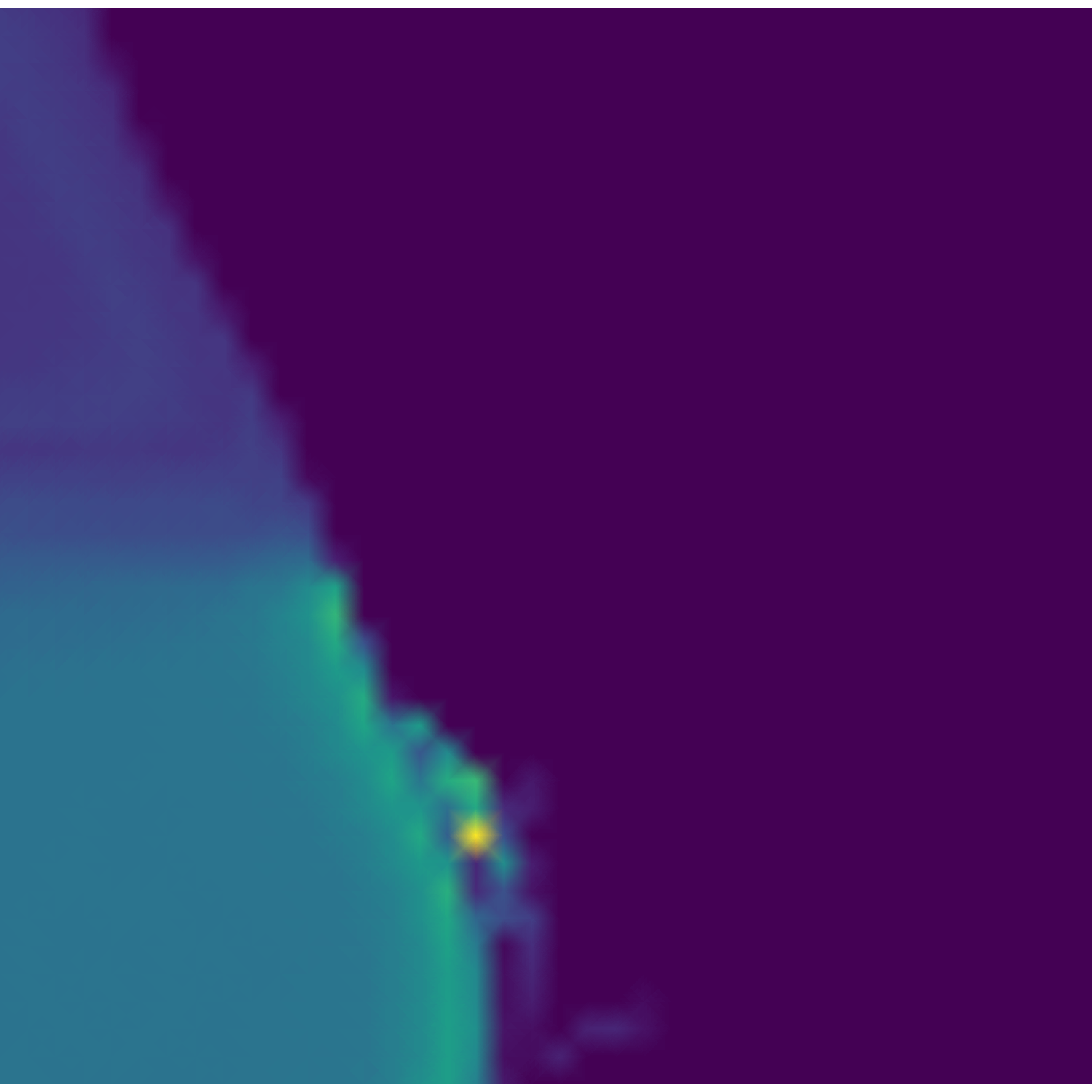} \\

        \rotatebox{90}{\hspace{0pt}\codeblue{Standstill}} &
        \includegraphics[width=\imgwidth, valign=m]{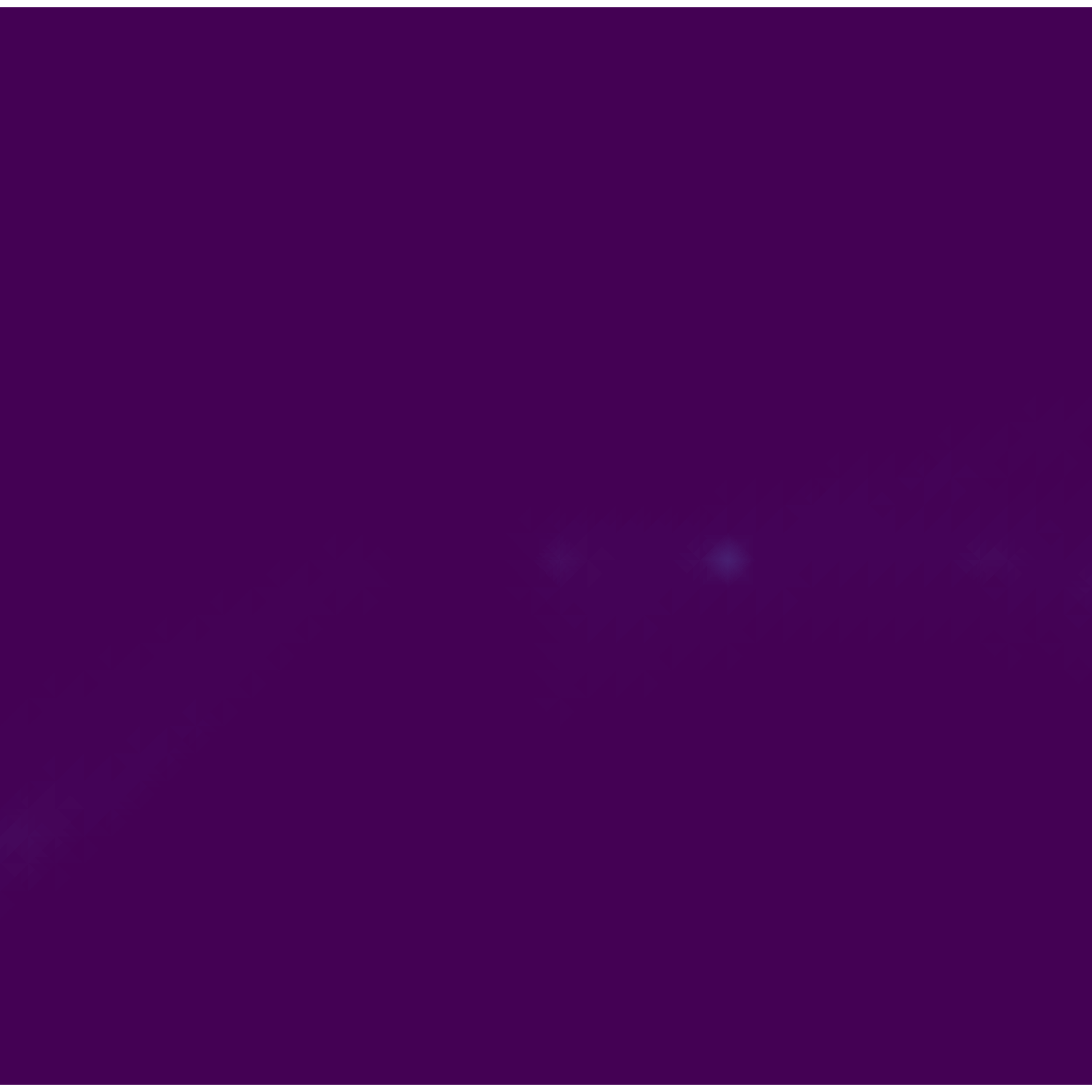} &
        \includegraphics[width=\imgwidth, valign=m]{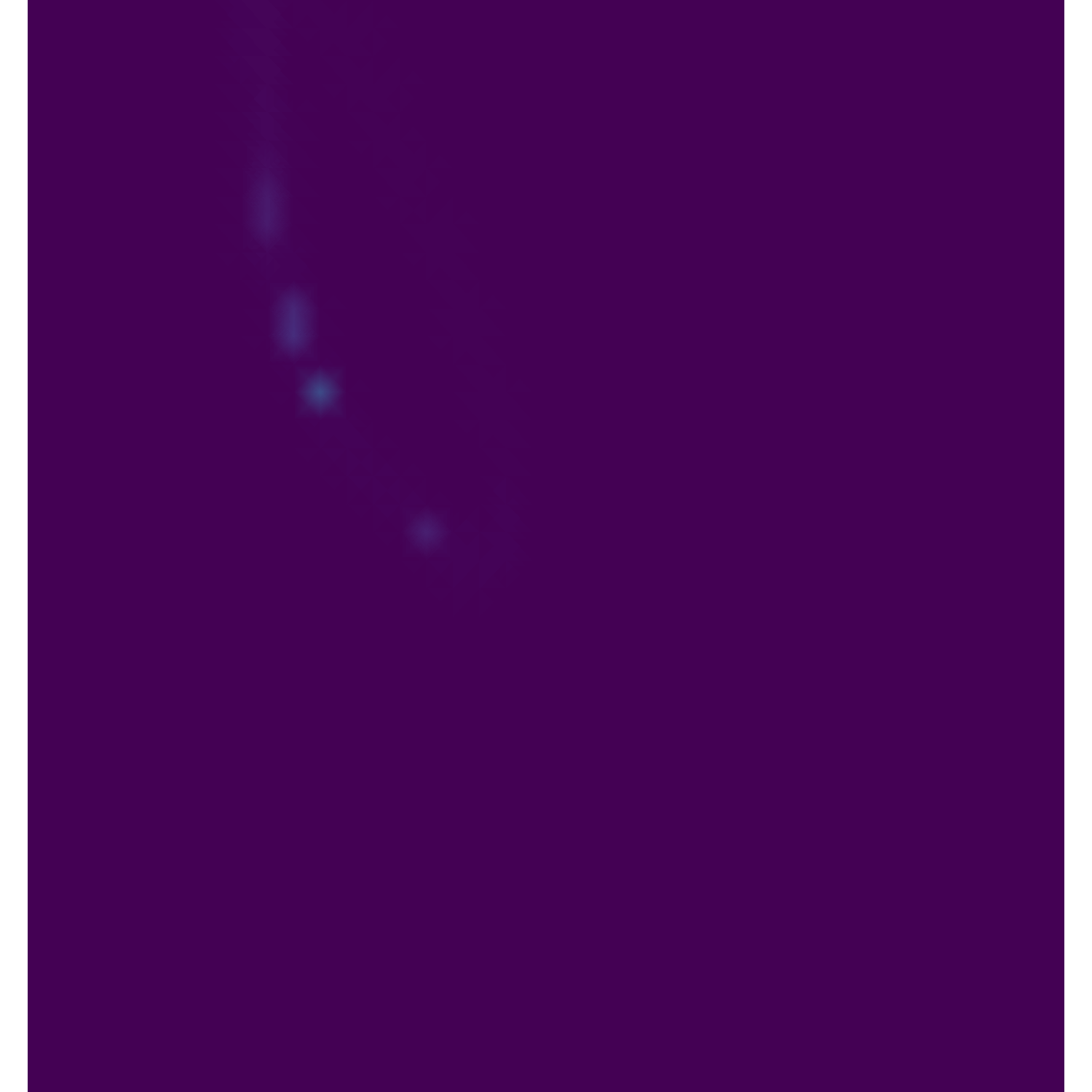} &
        \includegraphics[width=\imgwidth, valign=m]{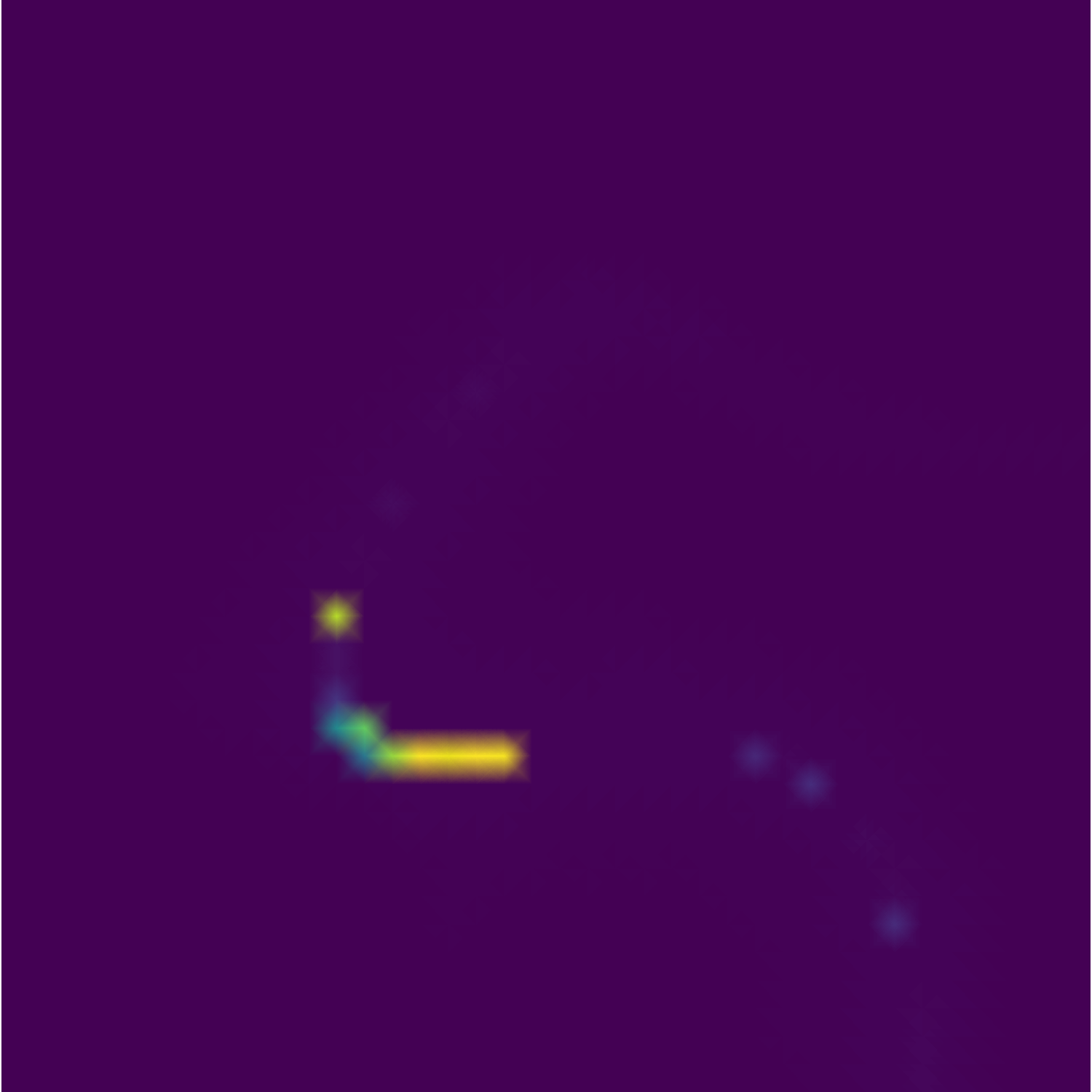} &
        \includegraphics[width=\imgwidth, valign=m]{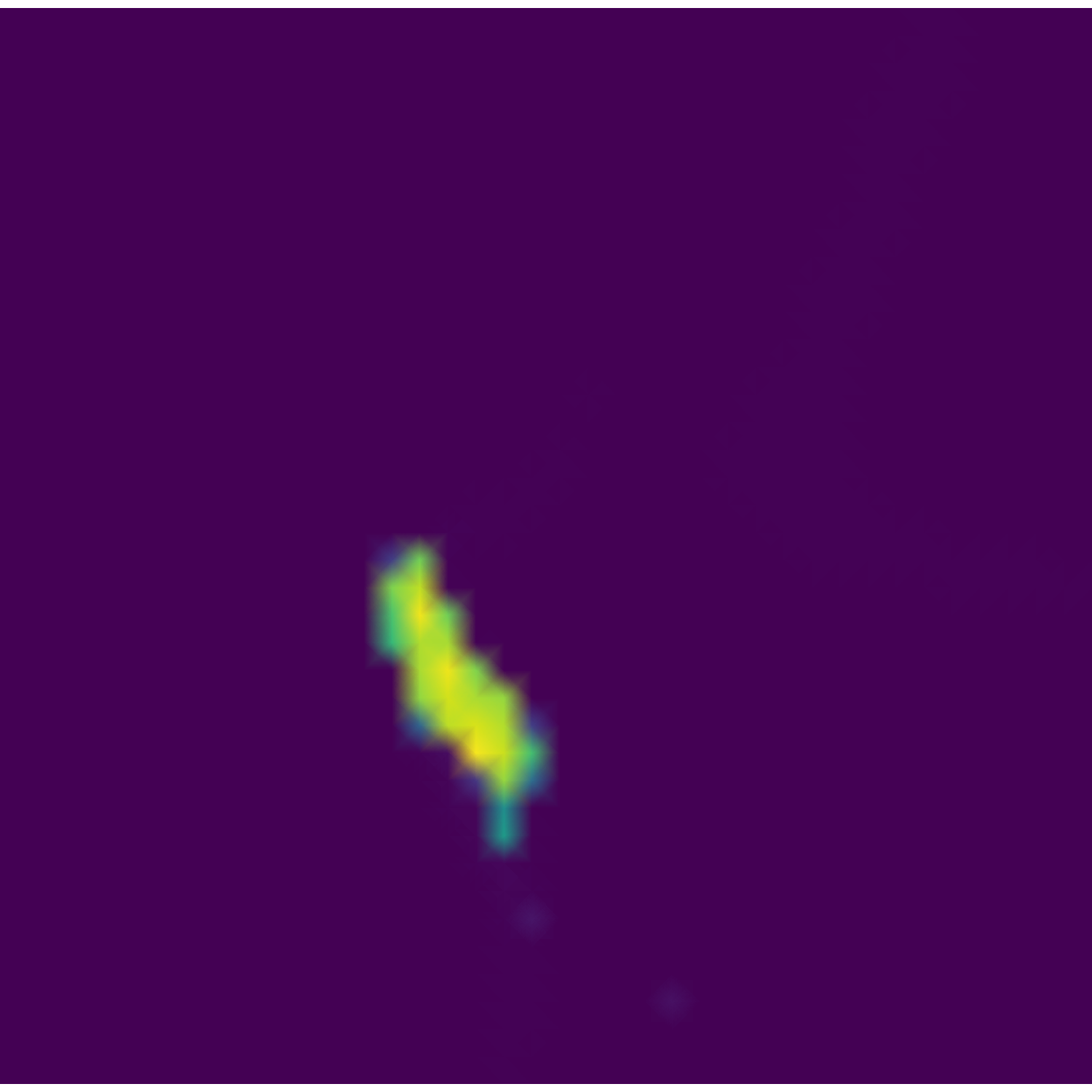} \\
    \end{tabular}
    \caption{HP, seed 999, 2D}
    \label{fig:full_latent_comparison_999_2d}
\end{figure*}

\begin{figure*}[t]
\vspace{-0.4cm}
    \centering
    \newcommand{\imgwidth}{0.22\textwidth}
    \setlength{\tabcolsep}{3pt}
    \begin{tabular}{m{5mm} c c c c}
        & \multicolumn{2}{c}{APC score} & \multicolumn{2}{c}{OPC score} \\
        \cmidrule(lr){2-3} \cmidrule(lr){4-5}
        & APC loss & OPC loss & APC loss & OPC loss \\[-.6ex]
        
        \rotatebox{90}{\hspace{0pt}\codeblue{Standard}} &
        \includegraphics[trim=0 0 0 80, clip, width=\imgwidth, valign=m]{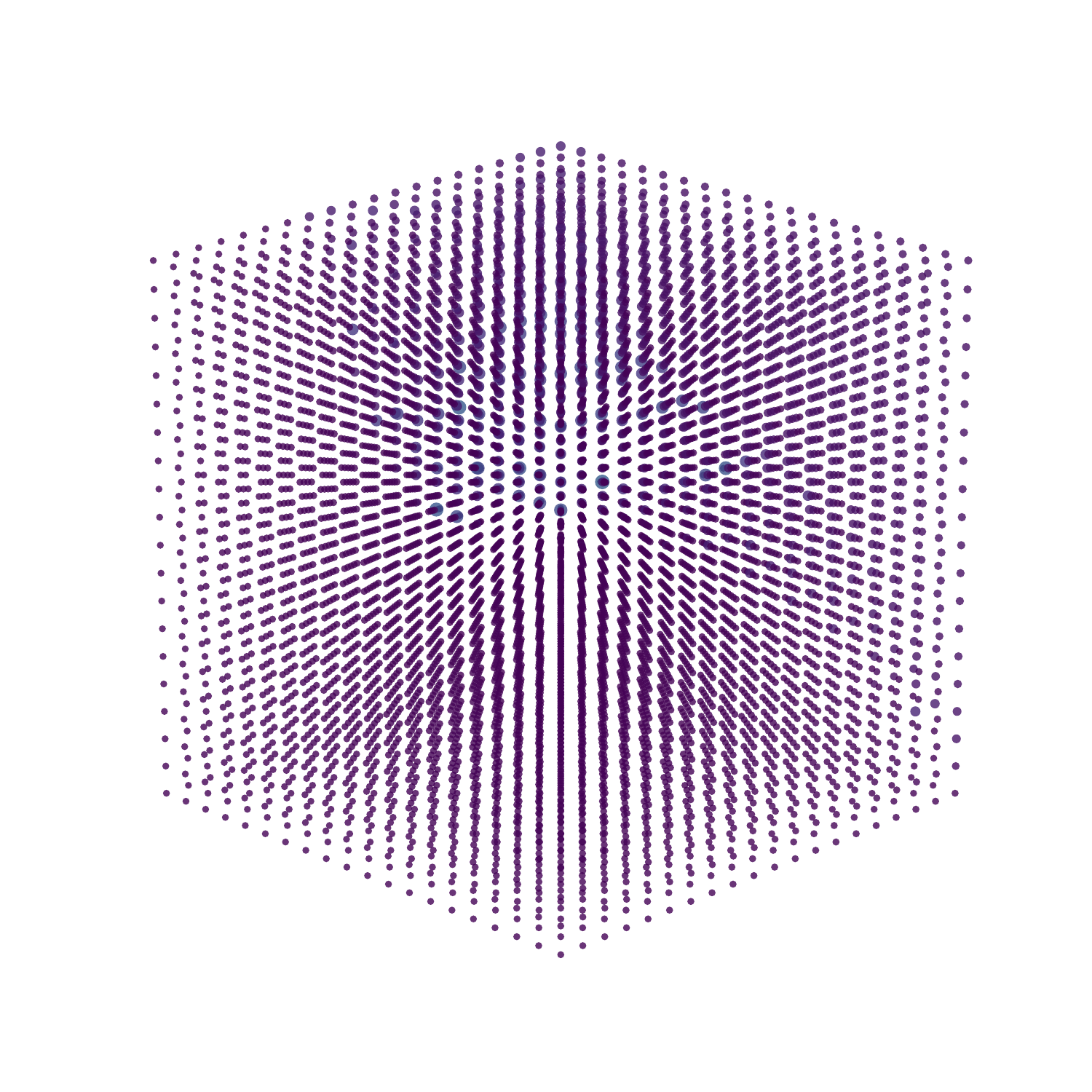} &
        \includegraphics[trim=0 0 0 80, clip, width=\imgwidth, valign=m]{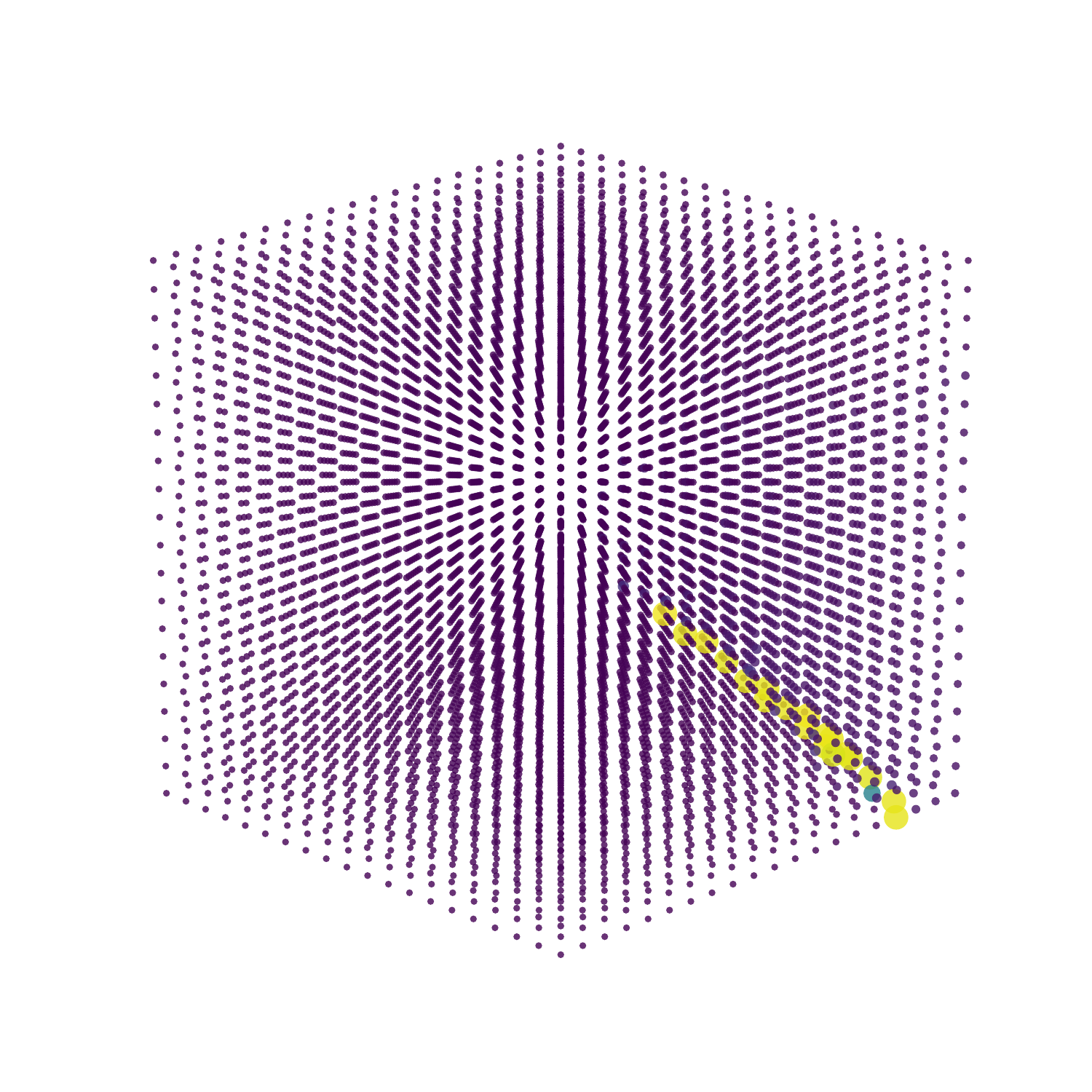} &
        \includegraphics[trim=0 0 0 80, clip, width=\imgwidth, valign=m]{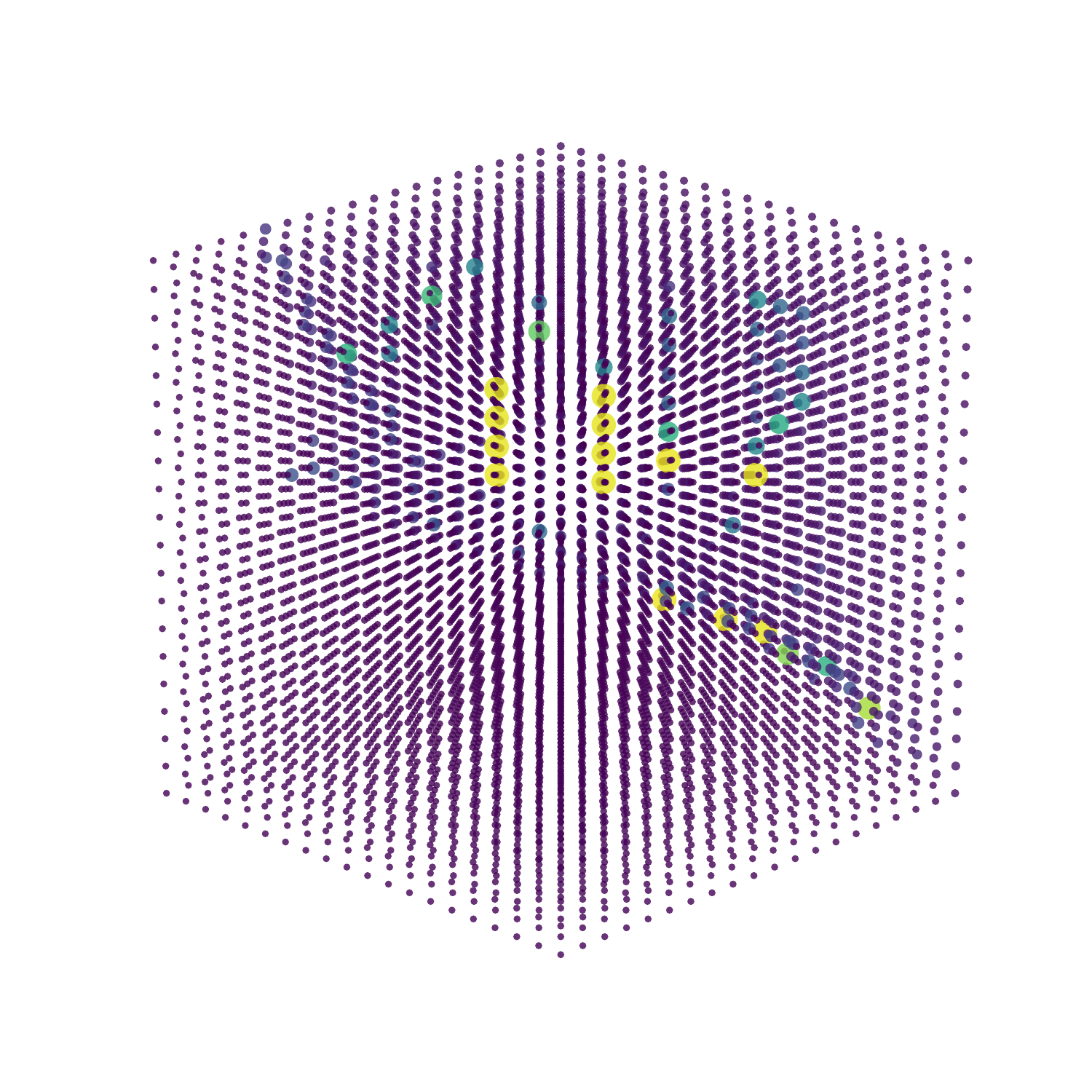} &
        \includegraphics[trim=0 0 0 80, clip, width=\imgwidth, valign=m]{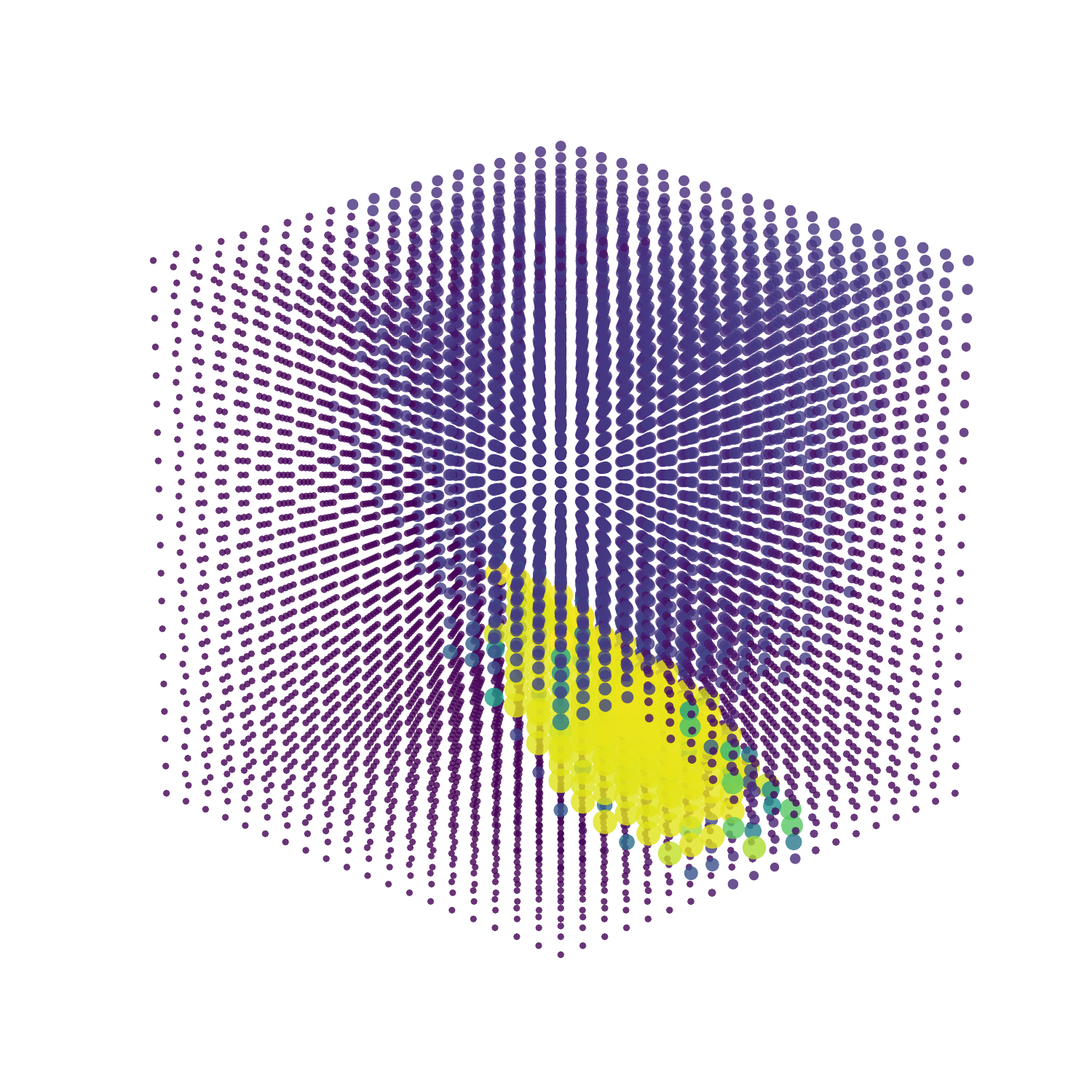} \\[-2.3ex]
        
        \rotatebox{90}{\hspace{0pt}\codeblue{Forward}} &
        \includegraphics[trim=0 0 0 80, clip, width=\imgwidth, valign=m]{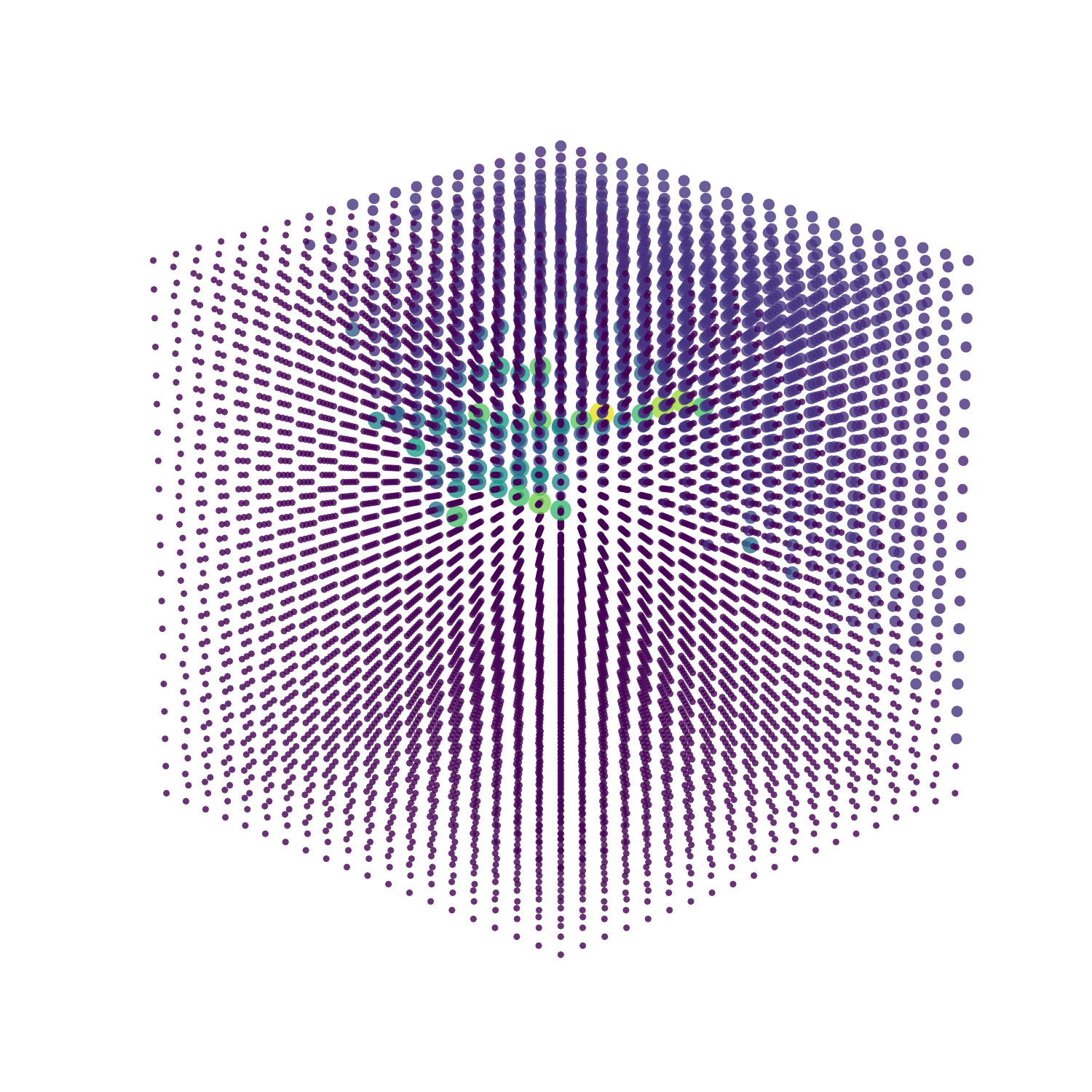} &
        \includegraphics[trim=0 0 0 80, clip, width=\imgwidth, valign=m]{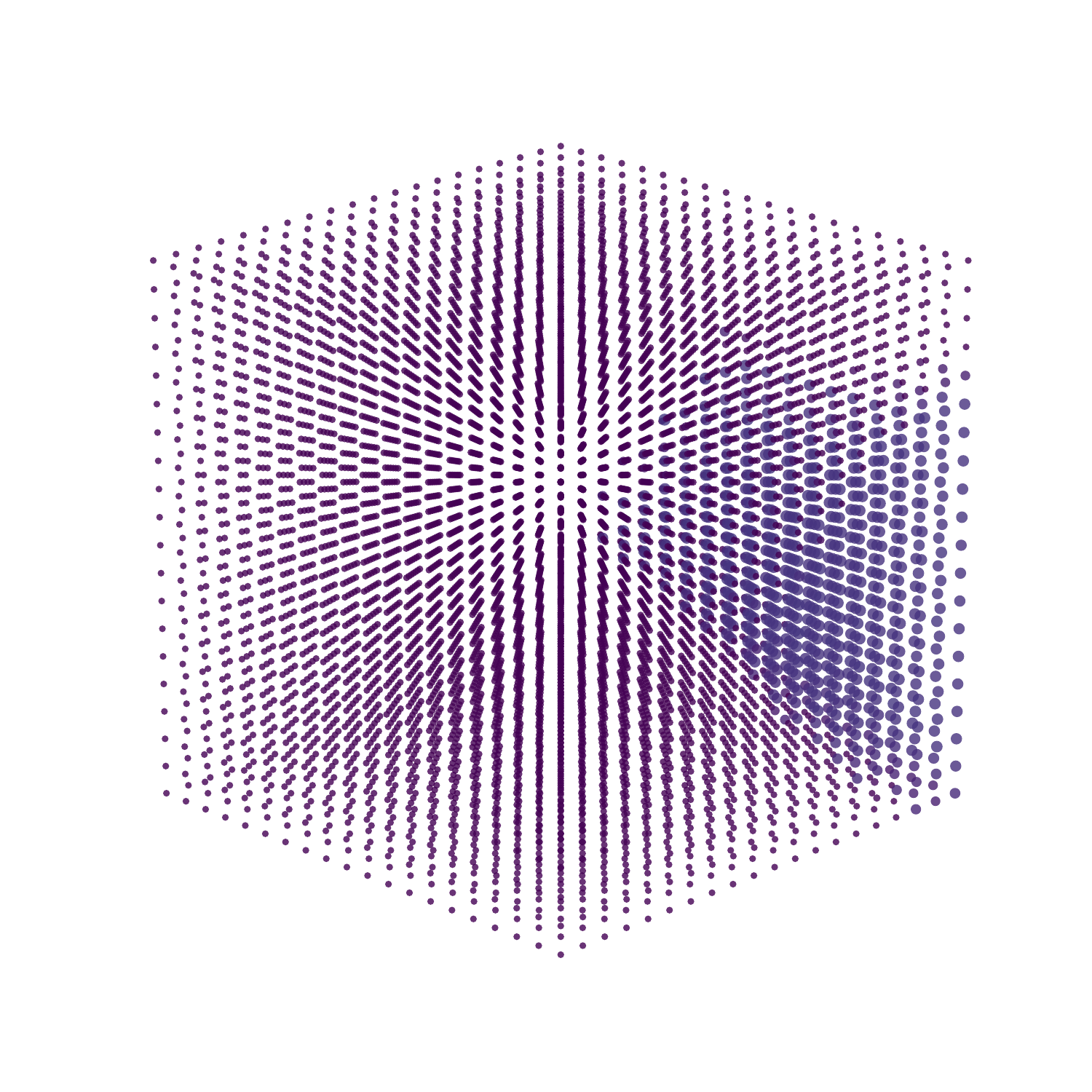} &
        \includegraphics[trim=0 0 0 80, clip, width=\imgwidth, valign=m]{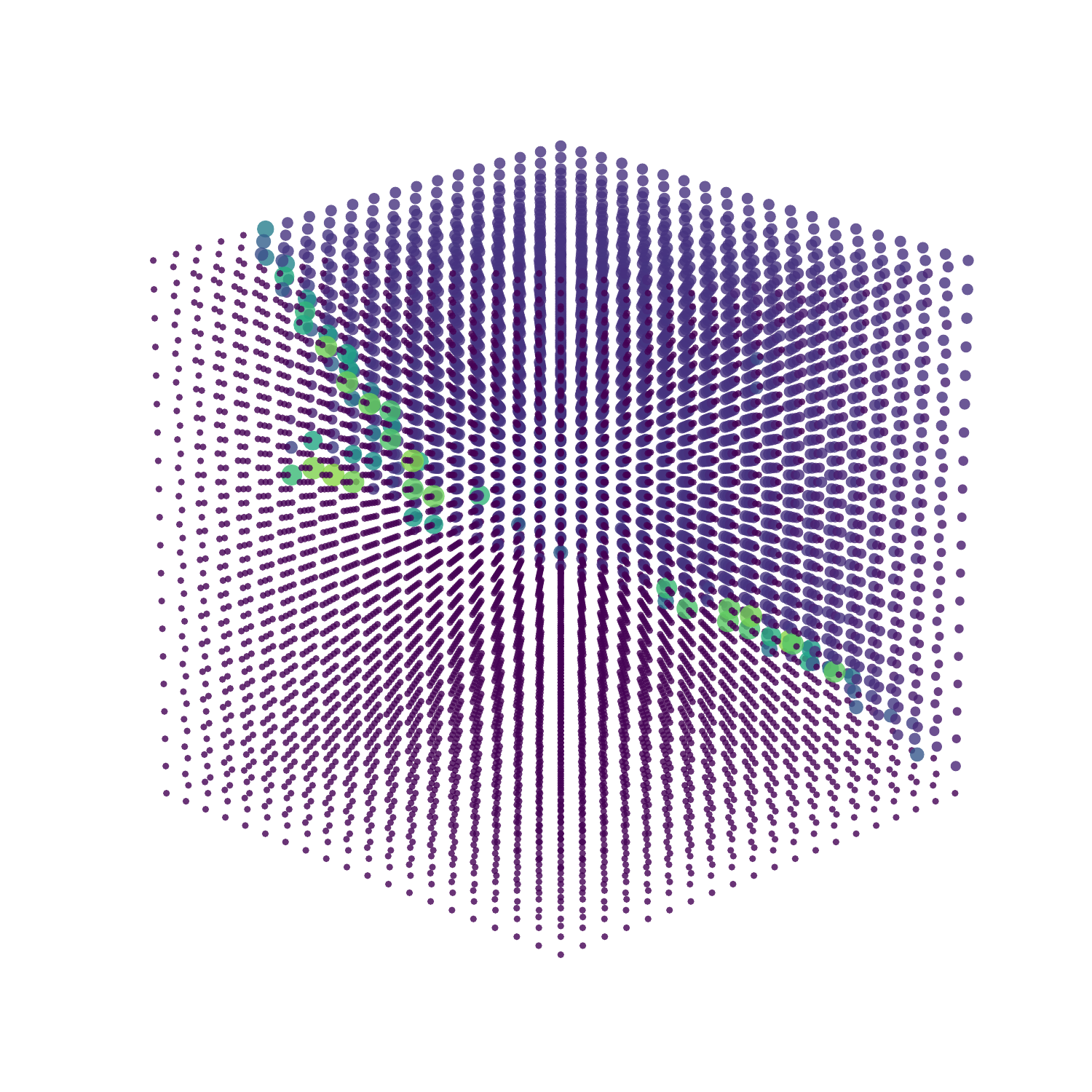} &
        \includegraphics[trim=0 0 0 80, clip, width=\imgwidth, valign=m]{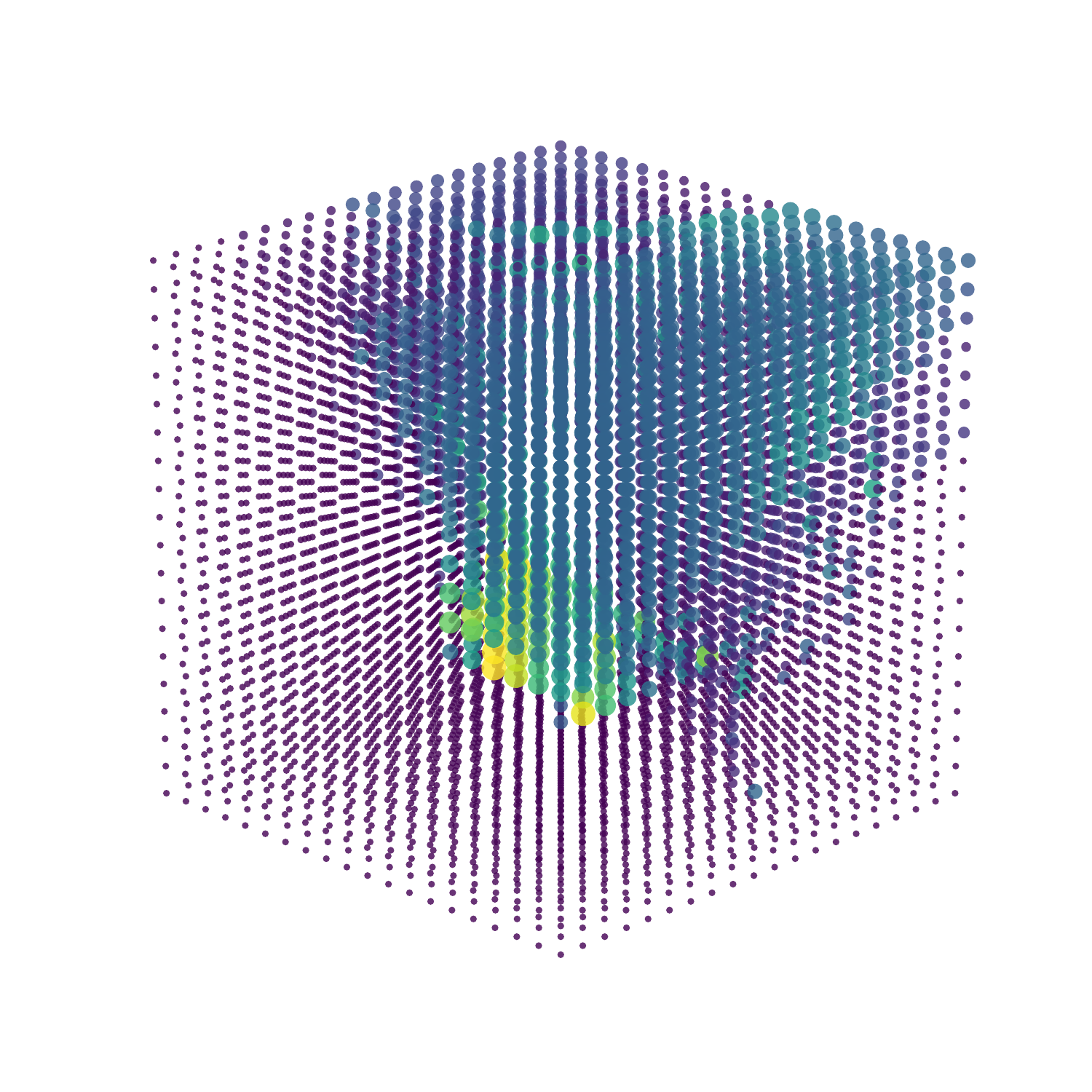} \\[-2.3ex]

        \rotatebox{90}{\hspace{0pt}\codeblue{Backward}} &
        \includegraphics[trim=0 0 0 80, clip, width=\imgwidth, valign=m]{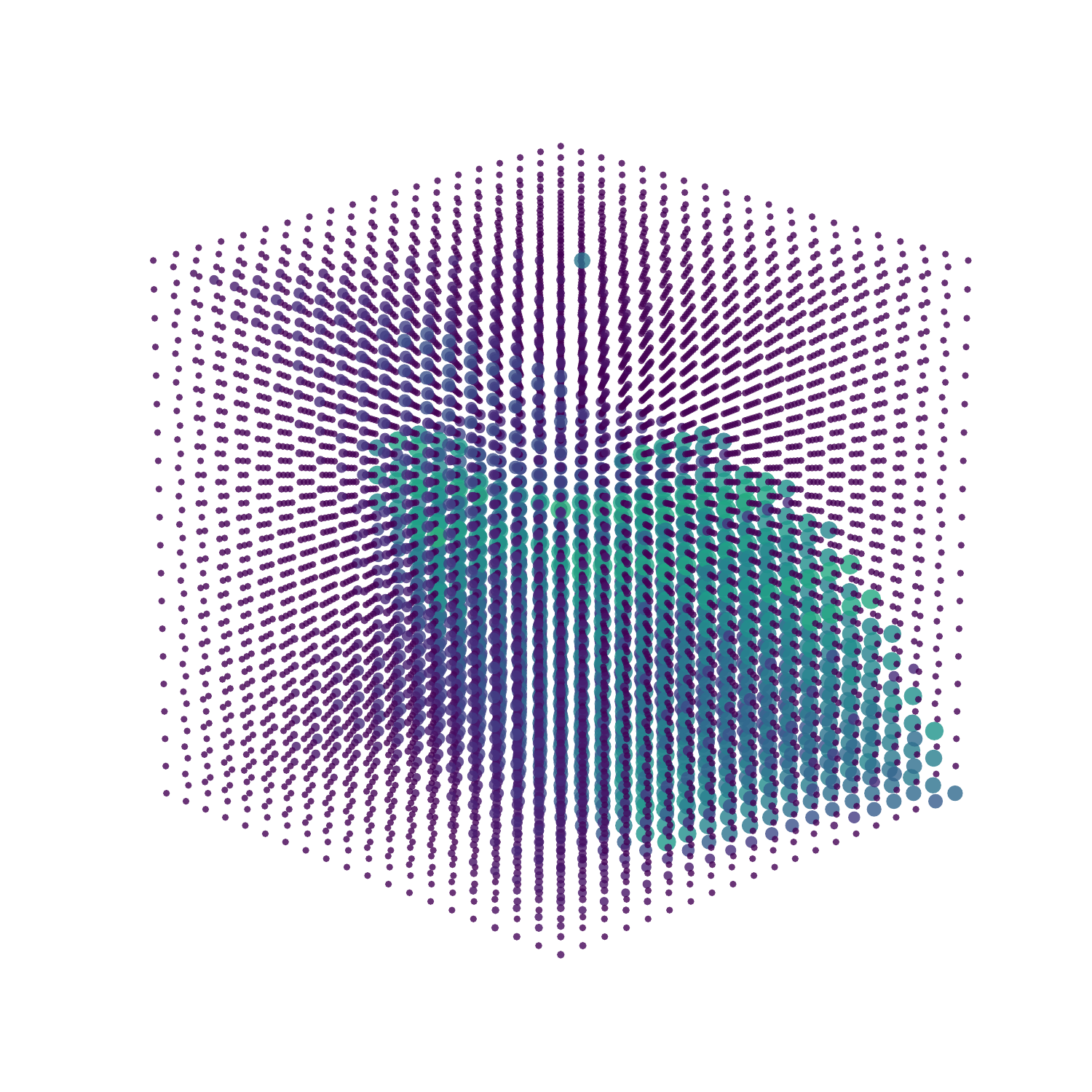} &
        \includegraphics[trim=0 0 0 80, clip, width=\imgwidth, valign=m]{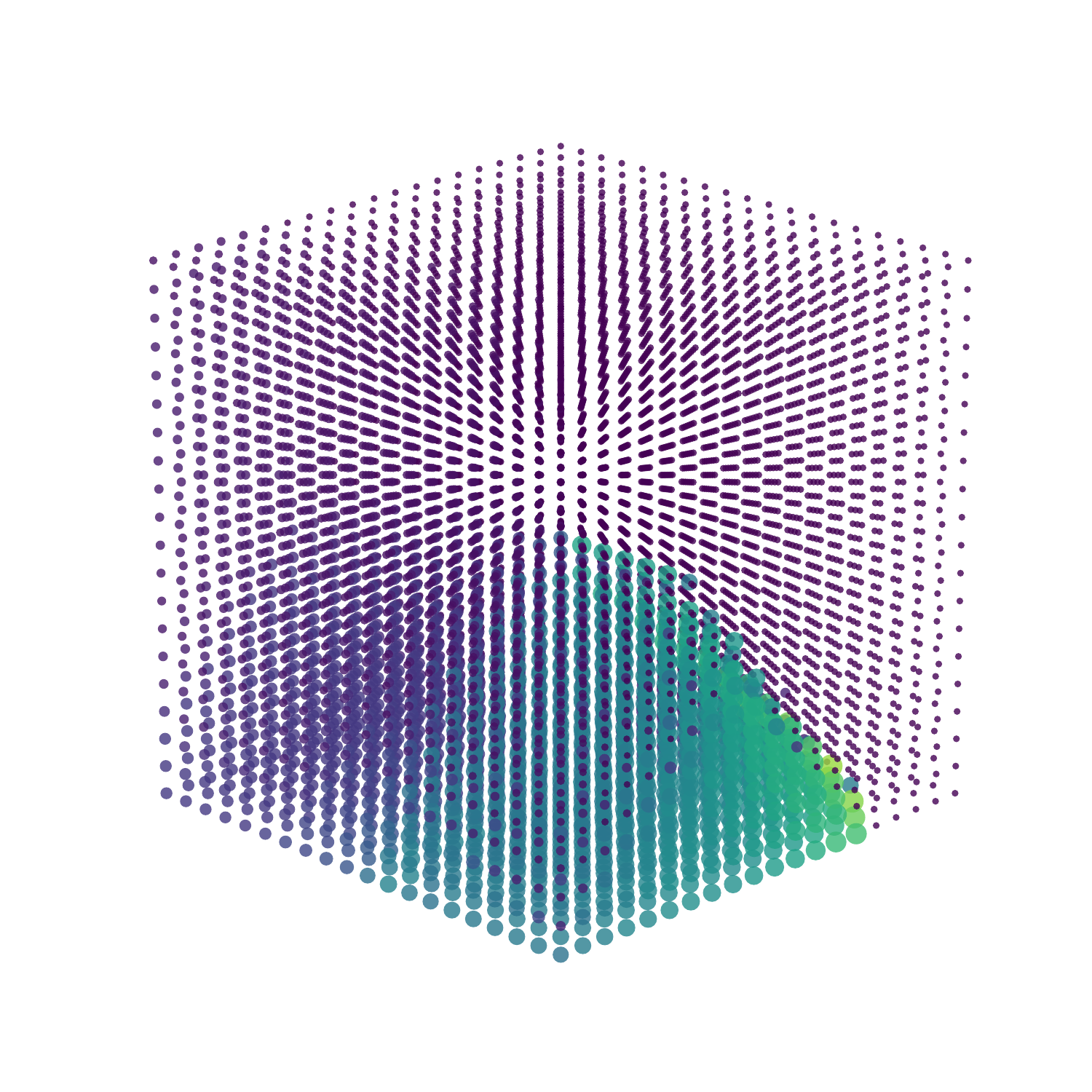} &
        \includegraphics[trim=0 0 0 80, clip, width=\imgwidth, valign=m]{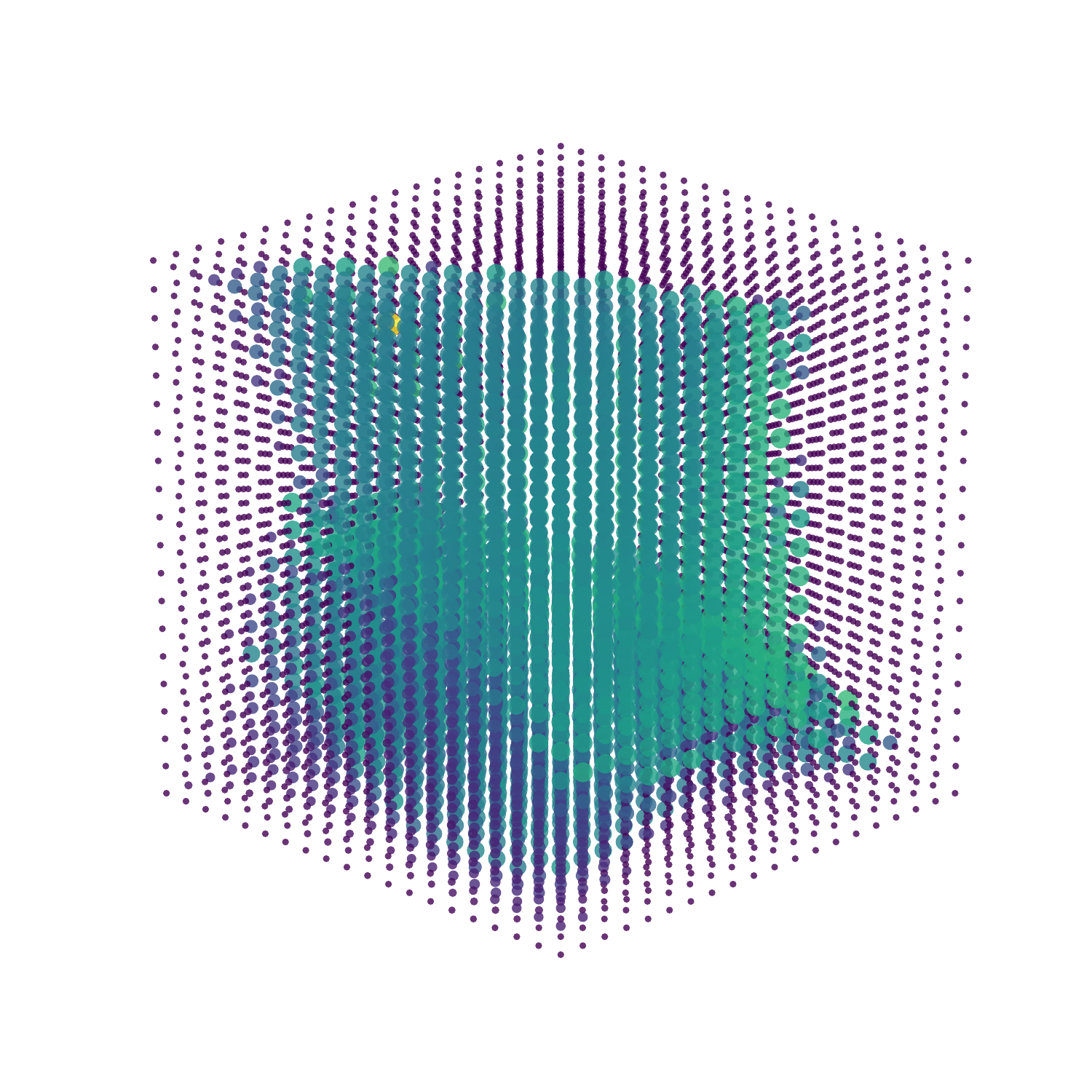} &
        \includegraphics[trim=0 0 0 80, clip, width=\imgwidth, valign=m]{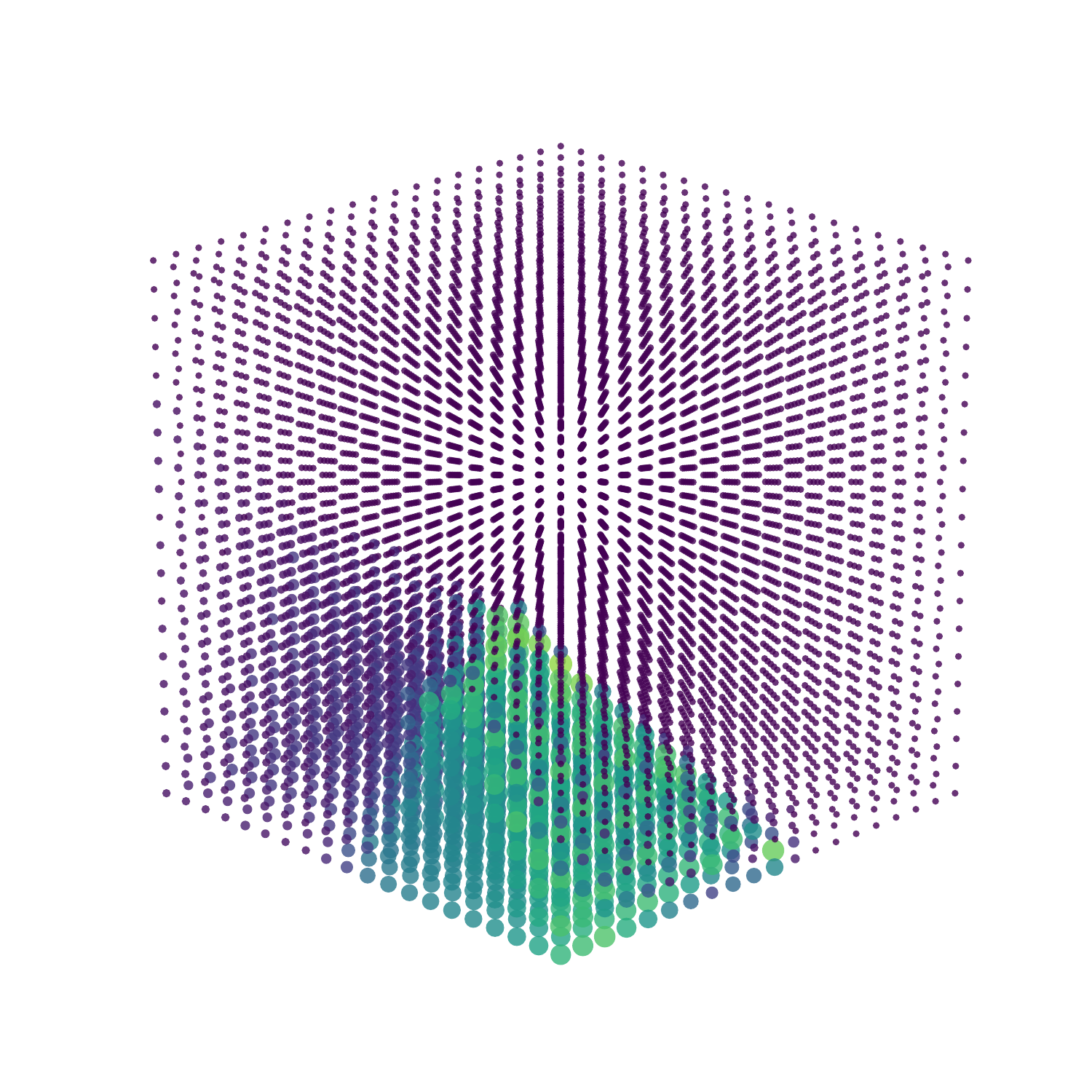} \\[-2.3ex]

        \rotatebox{90}{\hspace{0pt}\codeblue{Standstill}} &
        \includegraphics[trim=0 75 0 75, clip, width=\imgwidth, valign=m]{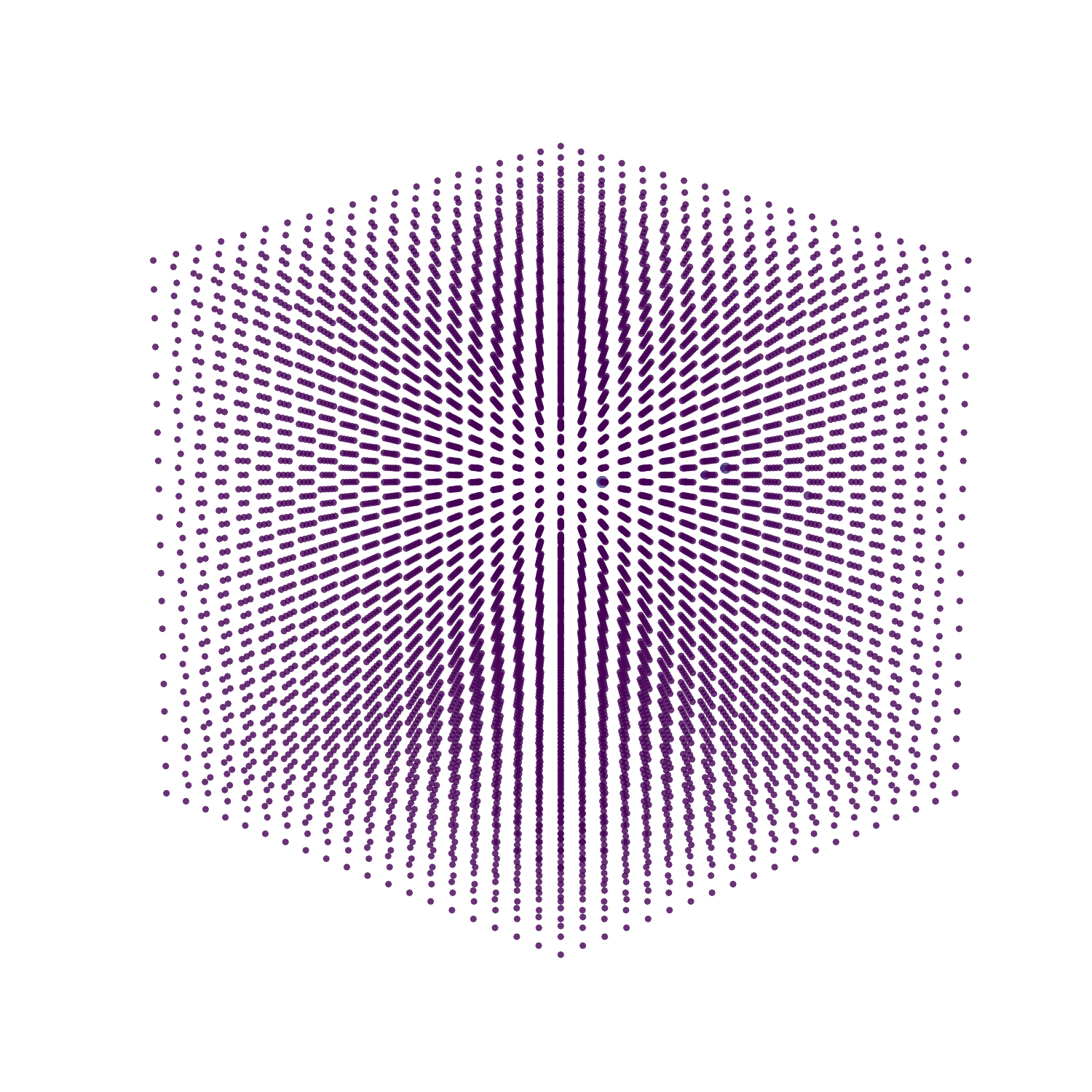} &
        \includegraphics[trim=0 75 0 75, clip, width=\imgwidth, valign=m]{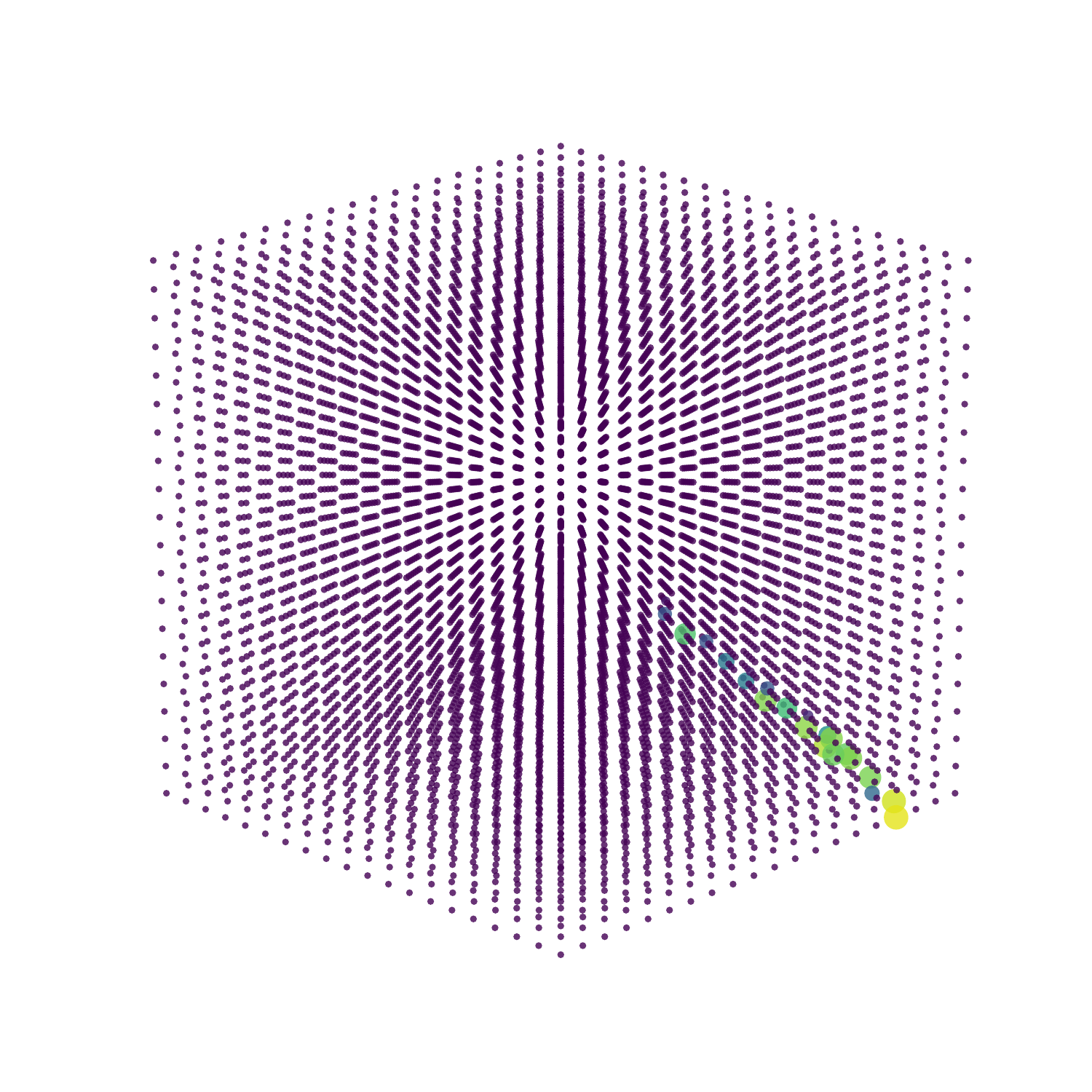} &
        \includegraphics[trim=0 75 0 75, clip, width=\imgwidth, valign=m]{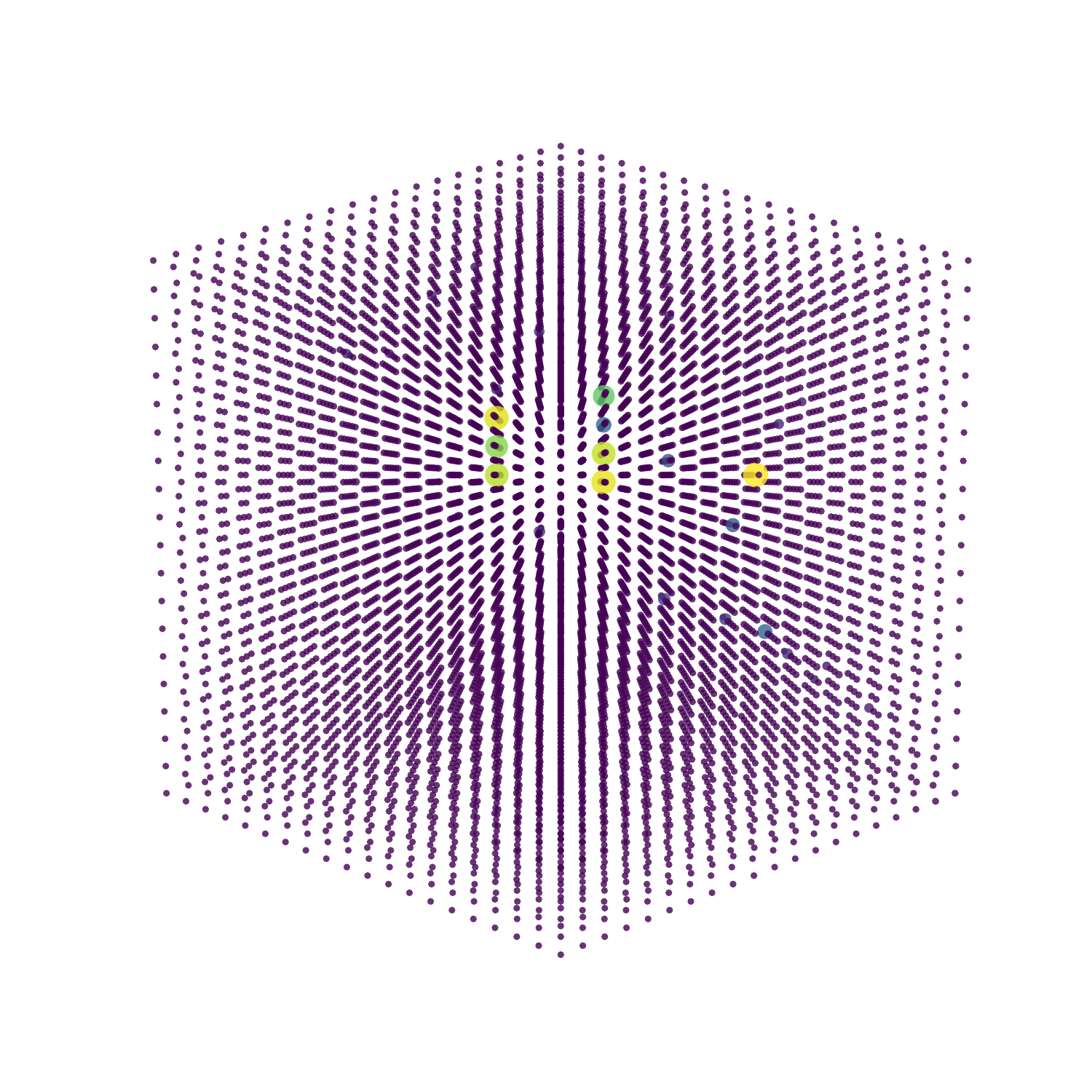} &
        \includegraphics[trim=0 75 0 75, clip, width=\imgwidth, valign=m]{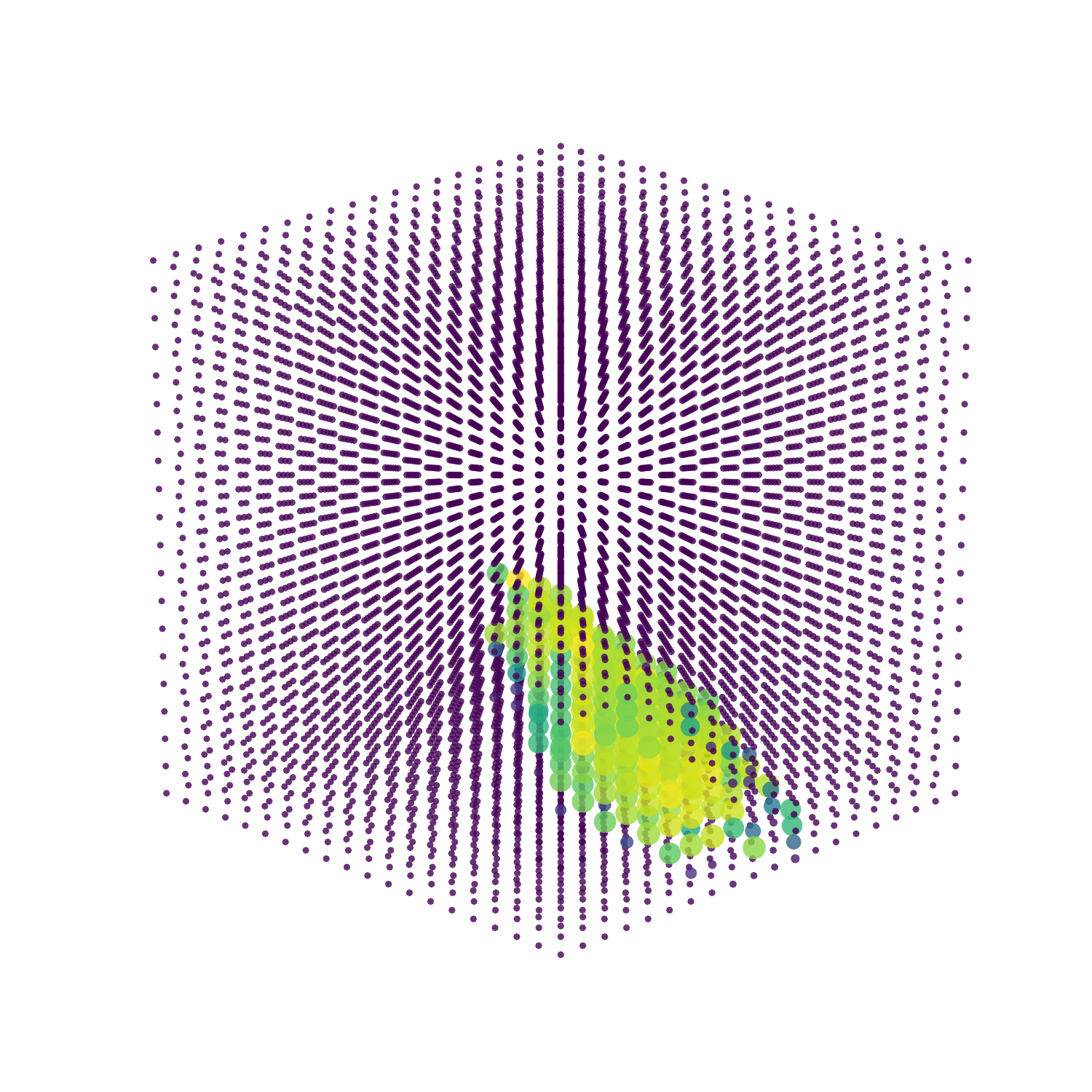} \\
    \end{tabular}
    \caption{HP, seed 888, 3D}
    \label{fig:full_latent_comparison_888_3d}
\end{figure*}

\begin{figure*}[t]
\vspace{-0.4cm}
    \centering
    \newcommand{\imgwidth}{0.22\textwidth}
    \setlength{\tabcolsep}{3pt}
    \begin{tabular}{m{5mm} c c c c}
        & \multicolumn{2}{c}{APC score} & \multicolumn{2}{c}{OPC score} \\
        \cmidrule(lr){2-3} \cmidrule(lr){4-5}
        & APC loss & OPC loss & APC loss & OPC loss \\[-.6ex]
        
        \rotatebox{90}{\hspace{0pt}\codeblue{Standard}} &
        \includegraphics[trim=0 0 0 80, clip, width=\imgwidth, valign=m]{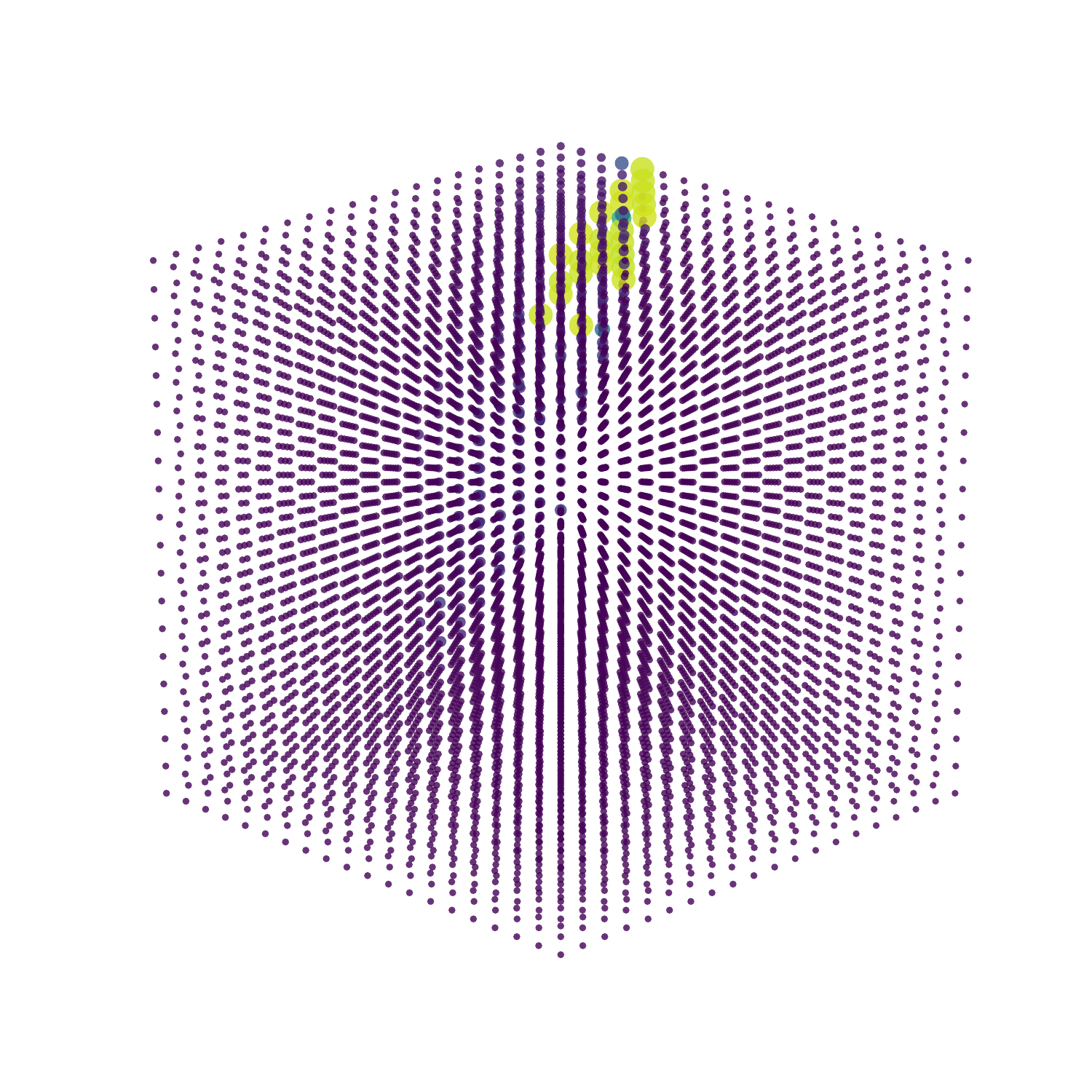} &
        \includegraphics[trim=0 0 0 80, clip, width=\imgwidth, valign=m]{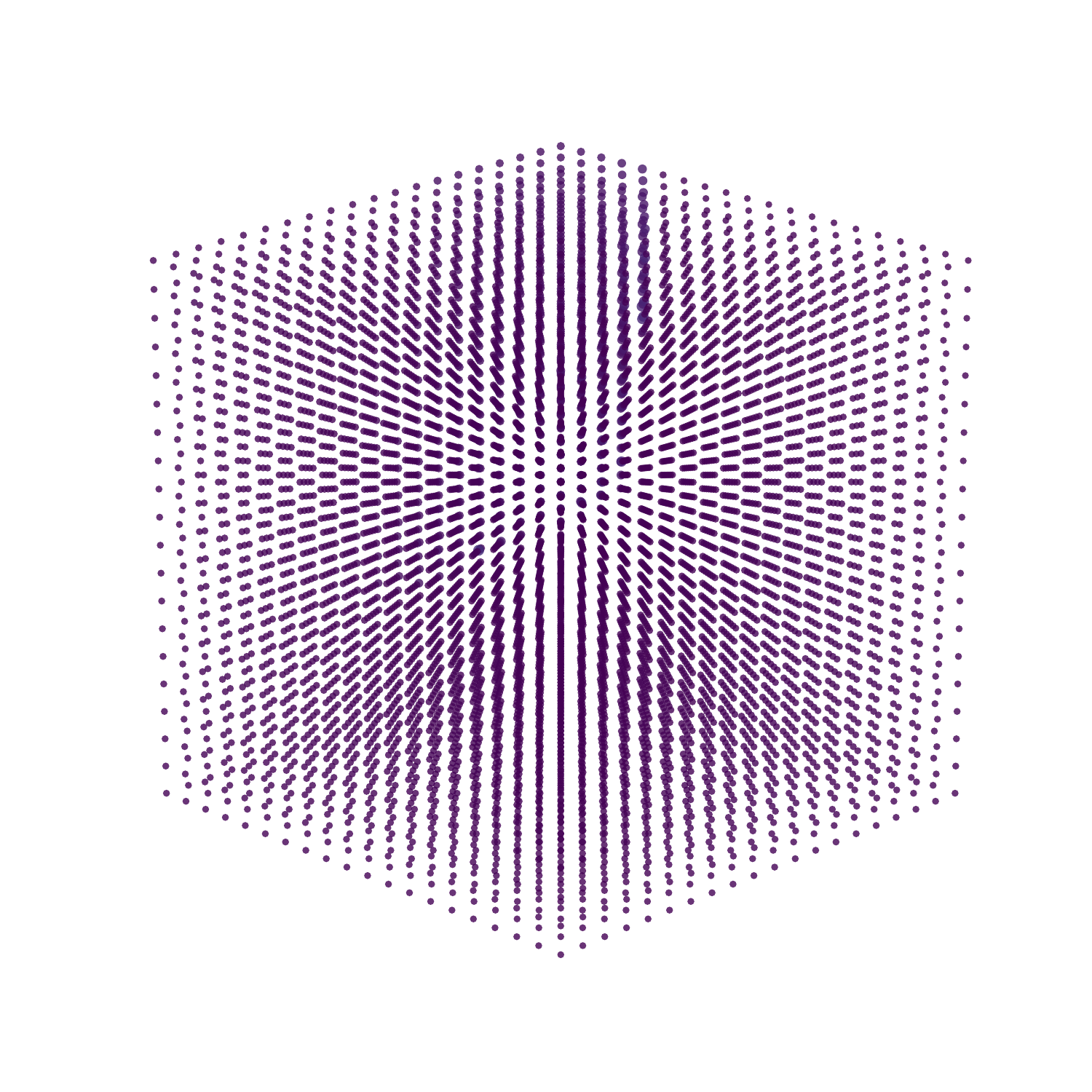} &
        \includegraphics[trim=0 0 0 80, clip, width=\imgwidth, valign=m]{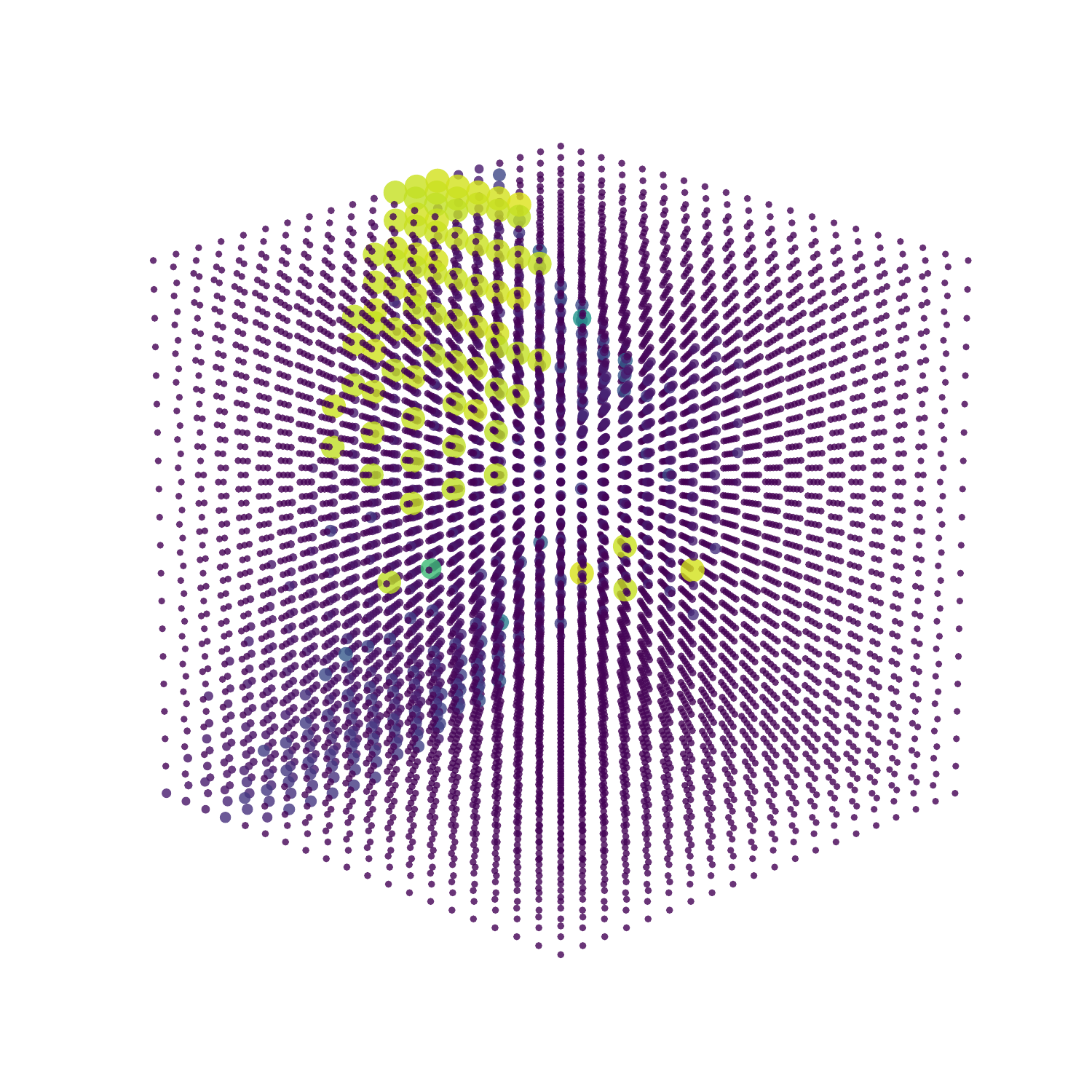} &
        \includegraphics[trim=0 0 0 80, clip, width=\imgwidth, valign=m]{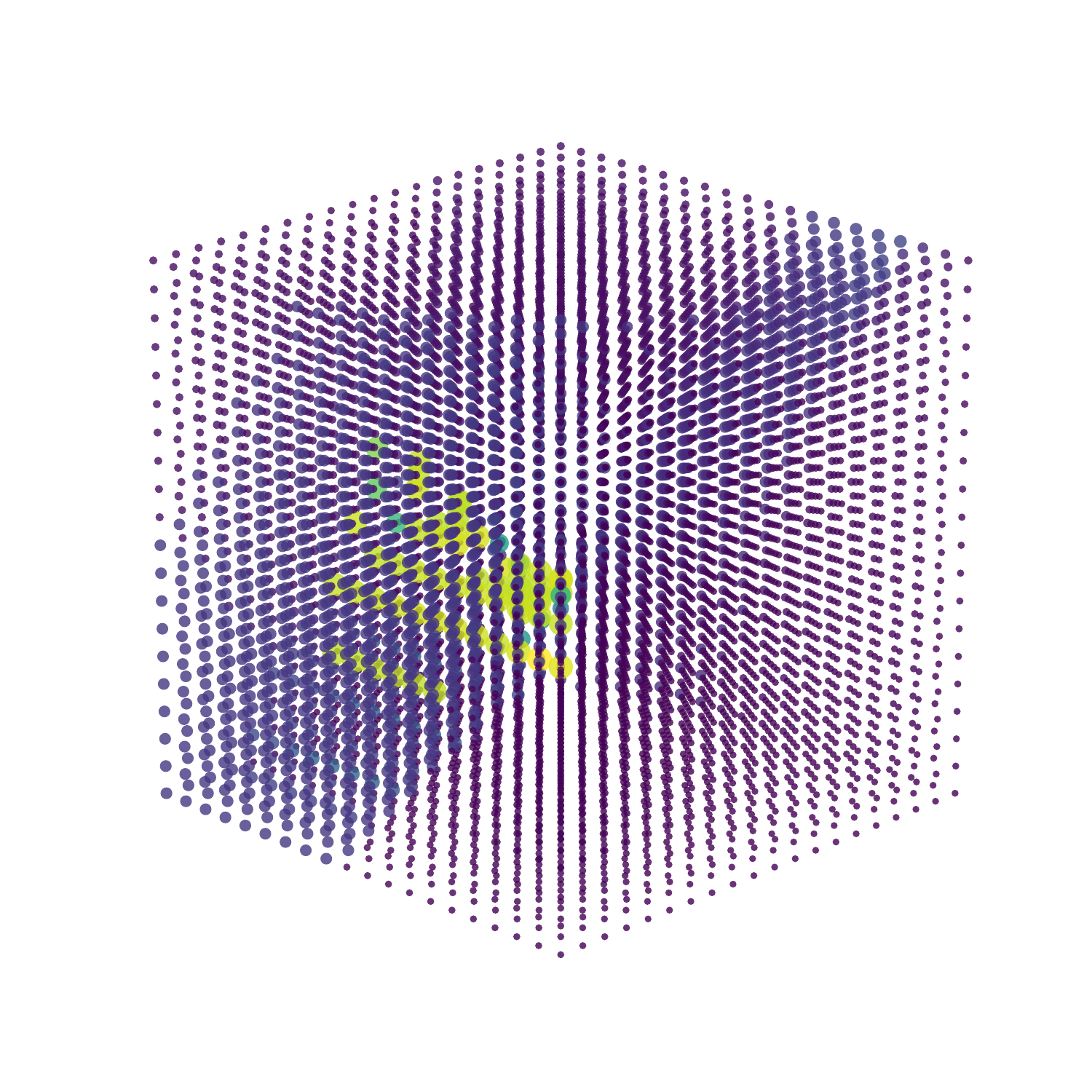} \\[-2.3ex]
        
        \rotatebox{90}{\hspace{0pt}\codeblue{Forward}} &
        \includegraphics[trim=0 0 0 80, clip, width=\imgwidth, valign=m]{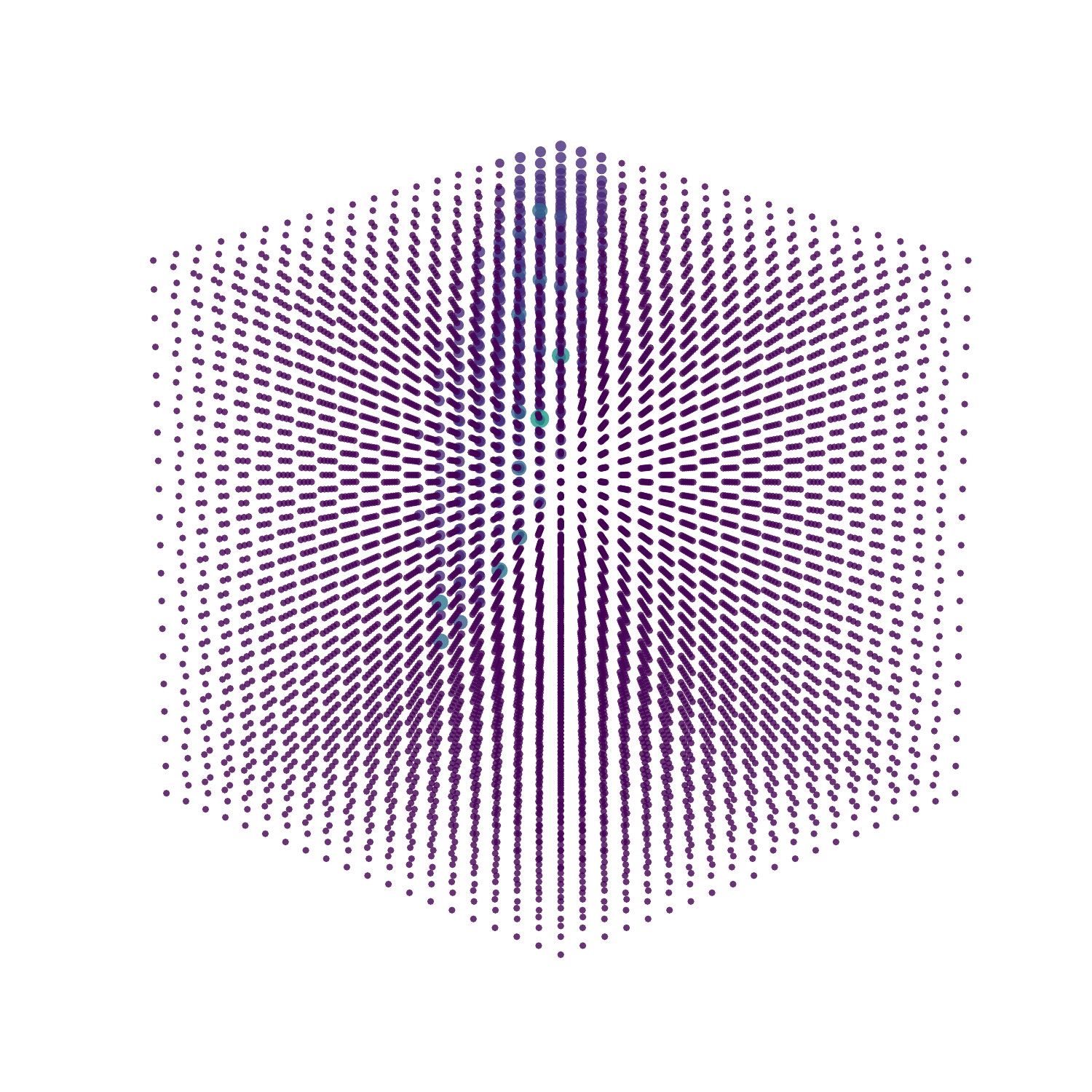} &
        \includegraphics[trim=0 0 0 80, clip, width=\imgwidth, valign=m]{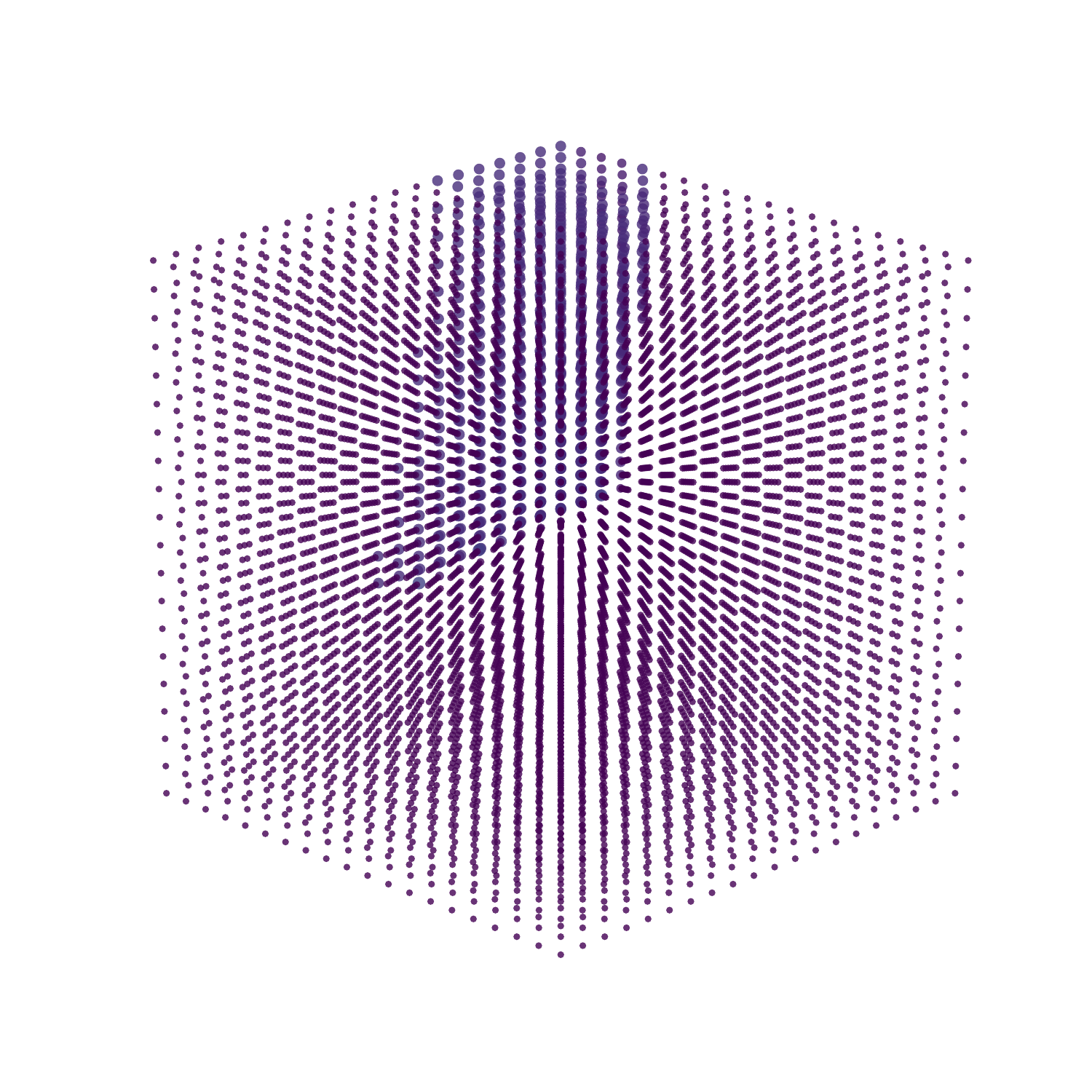} &
        \includegraphics[trim=0 0 0 80, clip, width=\imgwidth, valign=m]{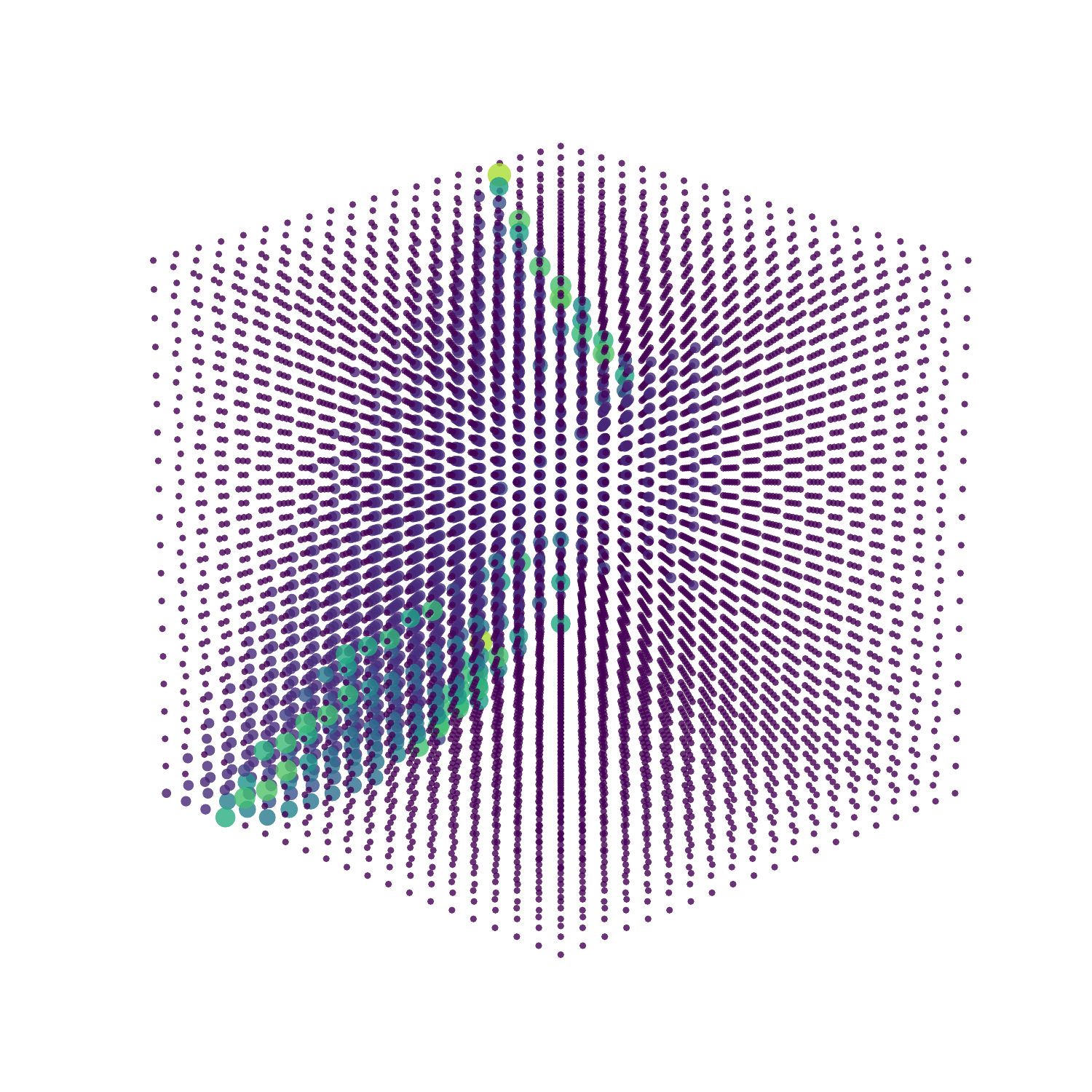} &
        \includegraphics[trim=0 0 0 80, clip, width=\imgwidth, valign=m]{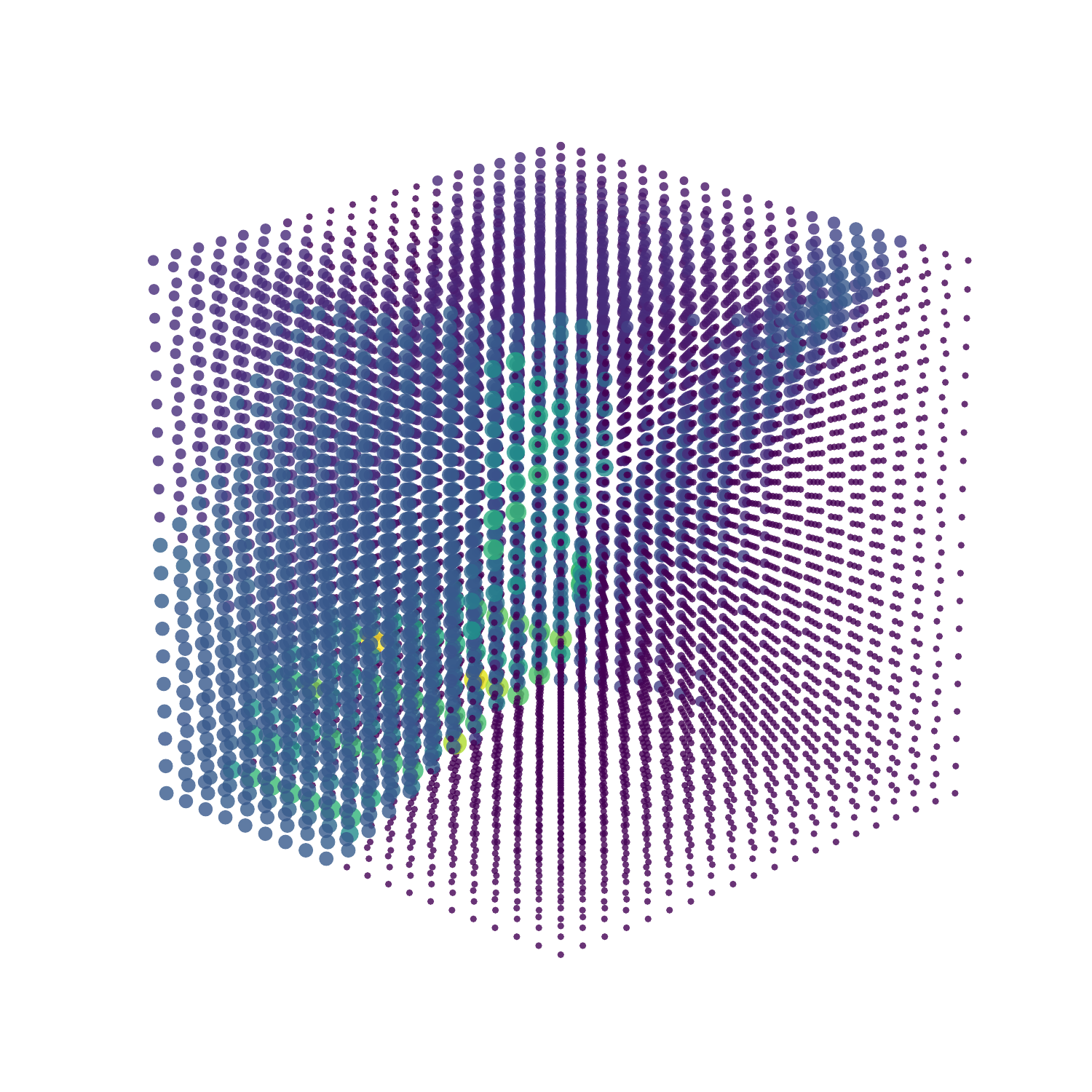} \\[-2.3ex]

        \rotatebox{90}{\hspace{0pt}\codeblue{Backward}} &
        \includegraphics[trim=0 0 0 80, clip, width=\imgwidth, valign=m]{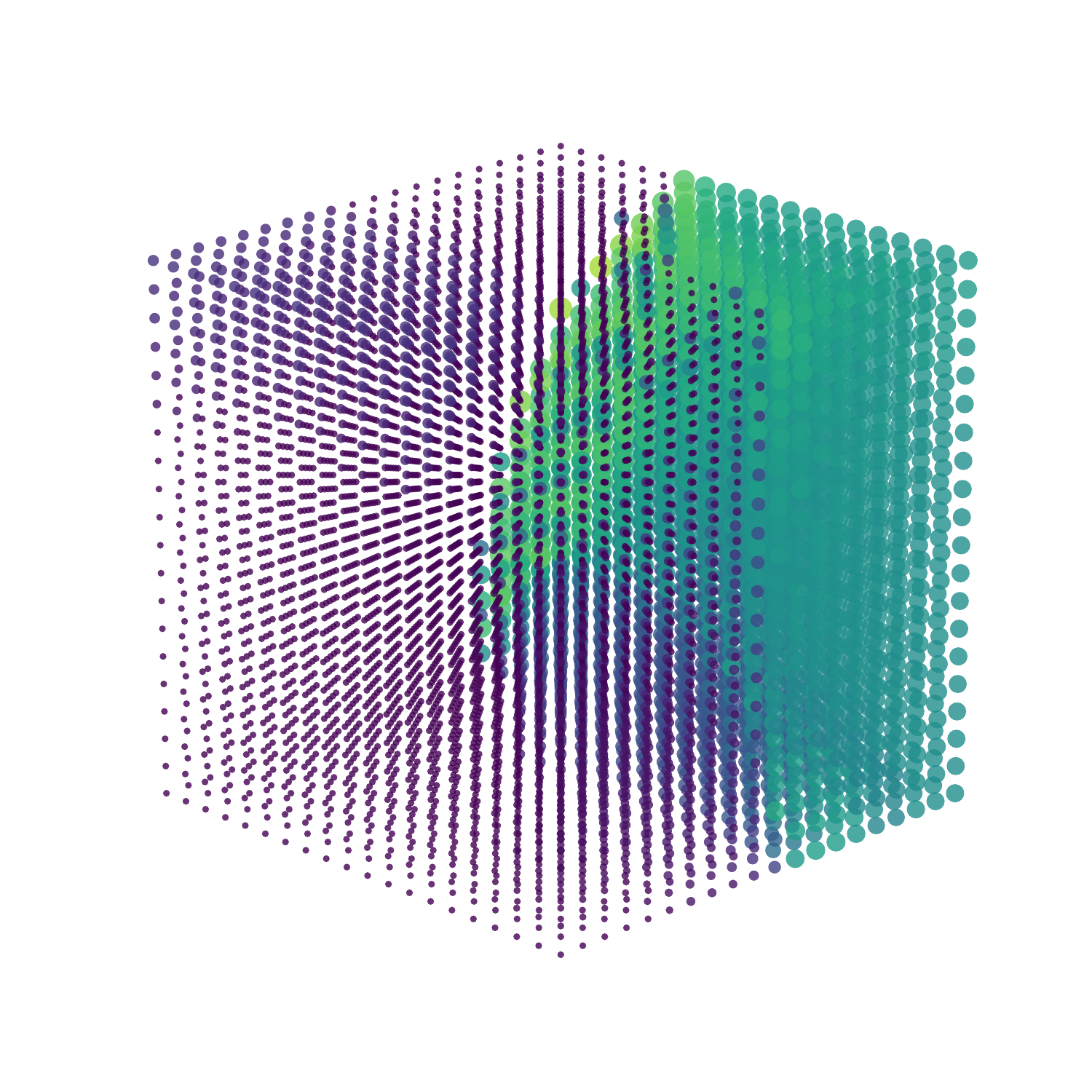} &
        \includegraphics[trim=0 0 0 80, clip, width=\imgwidth, valign=m]{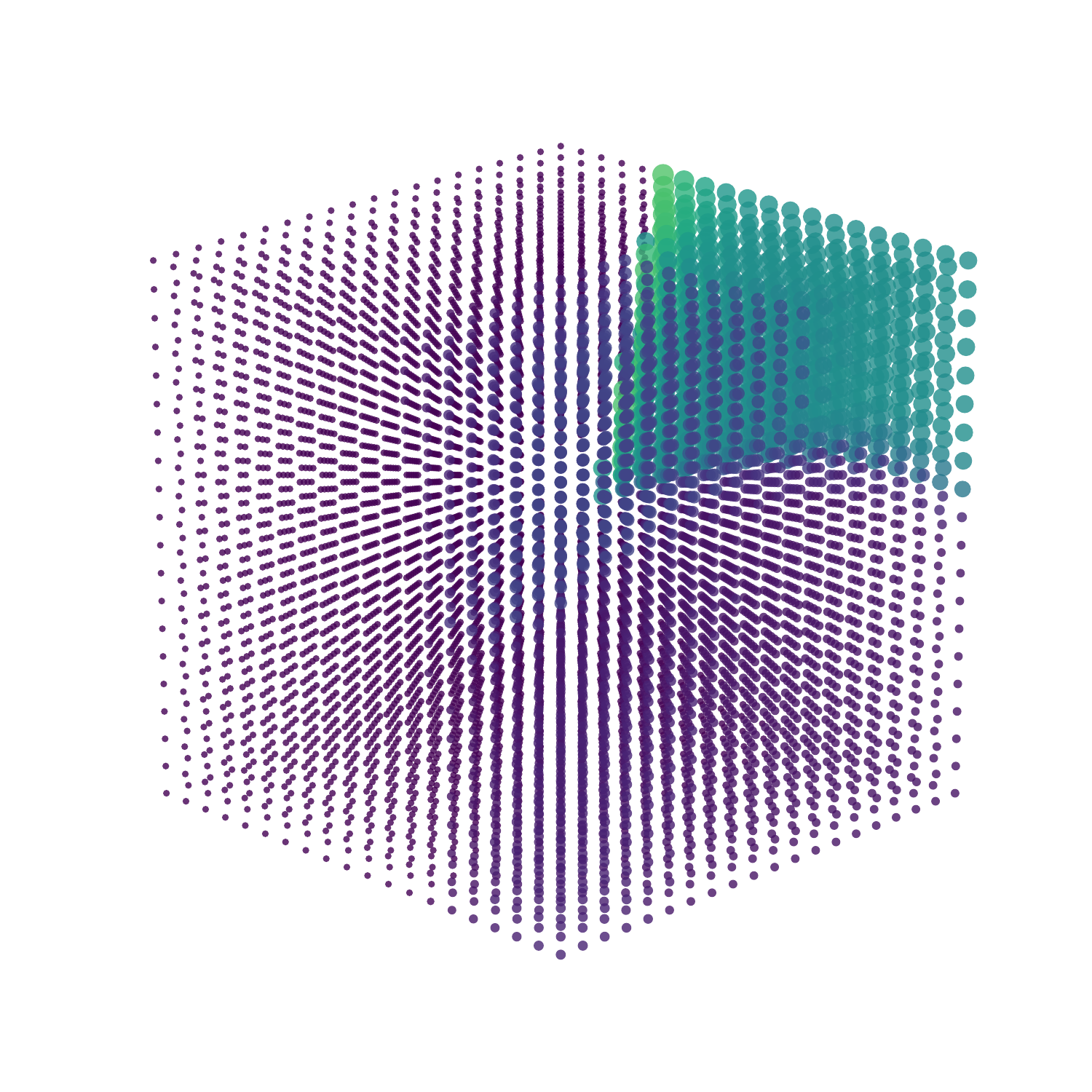} &
        \includegraphics[trim=0 0 0 80, clip, width=\imgwidth, valign=m]{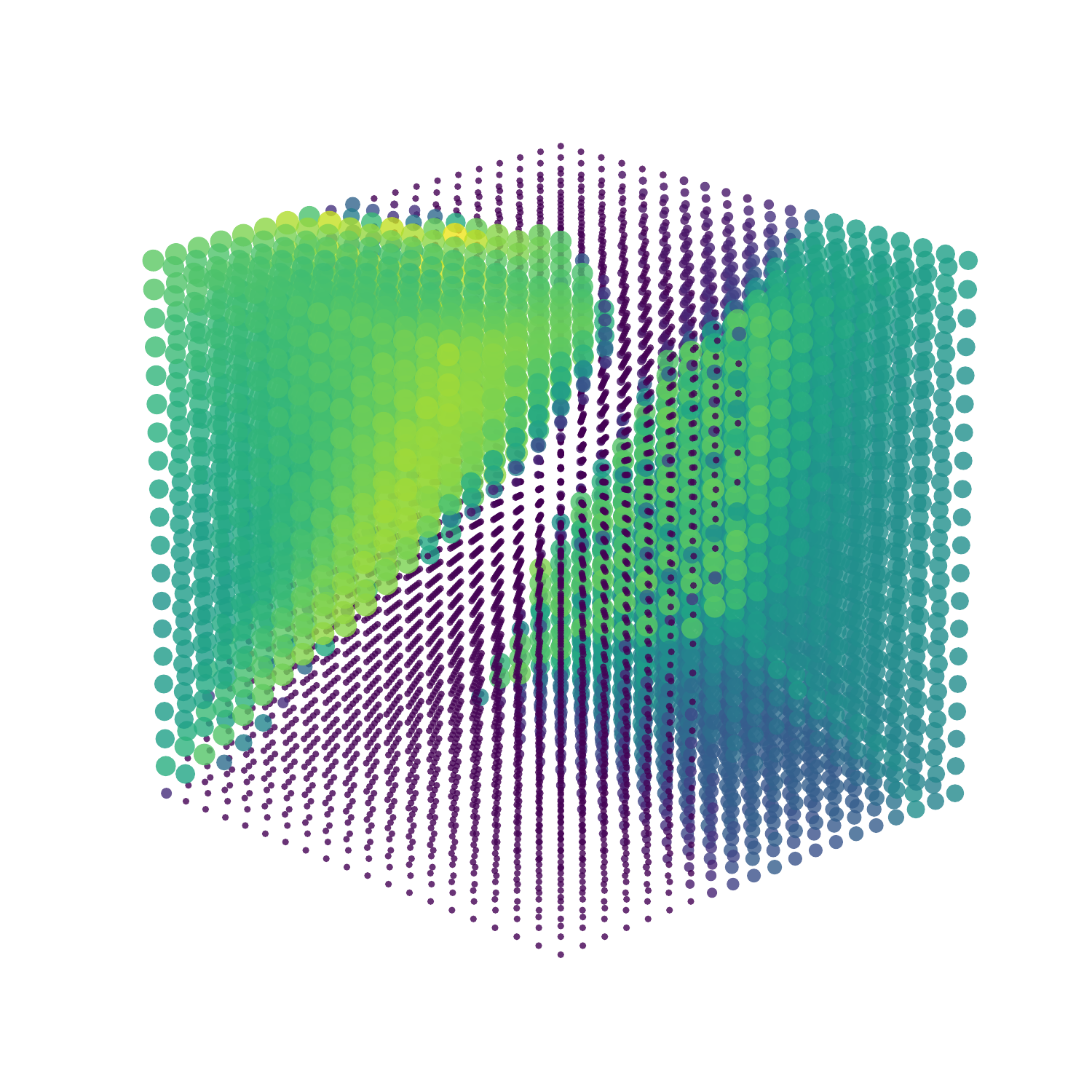} &
        \includegraphics[trim=0 0 0 80, clip, width=\imgwidth, valign=m]{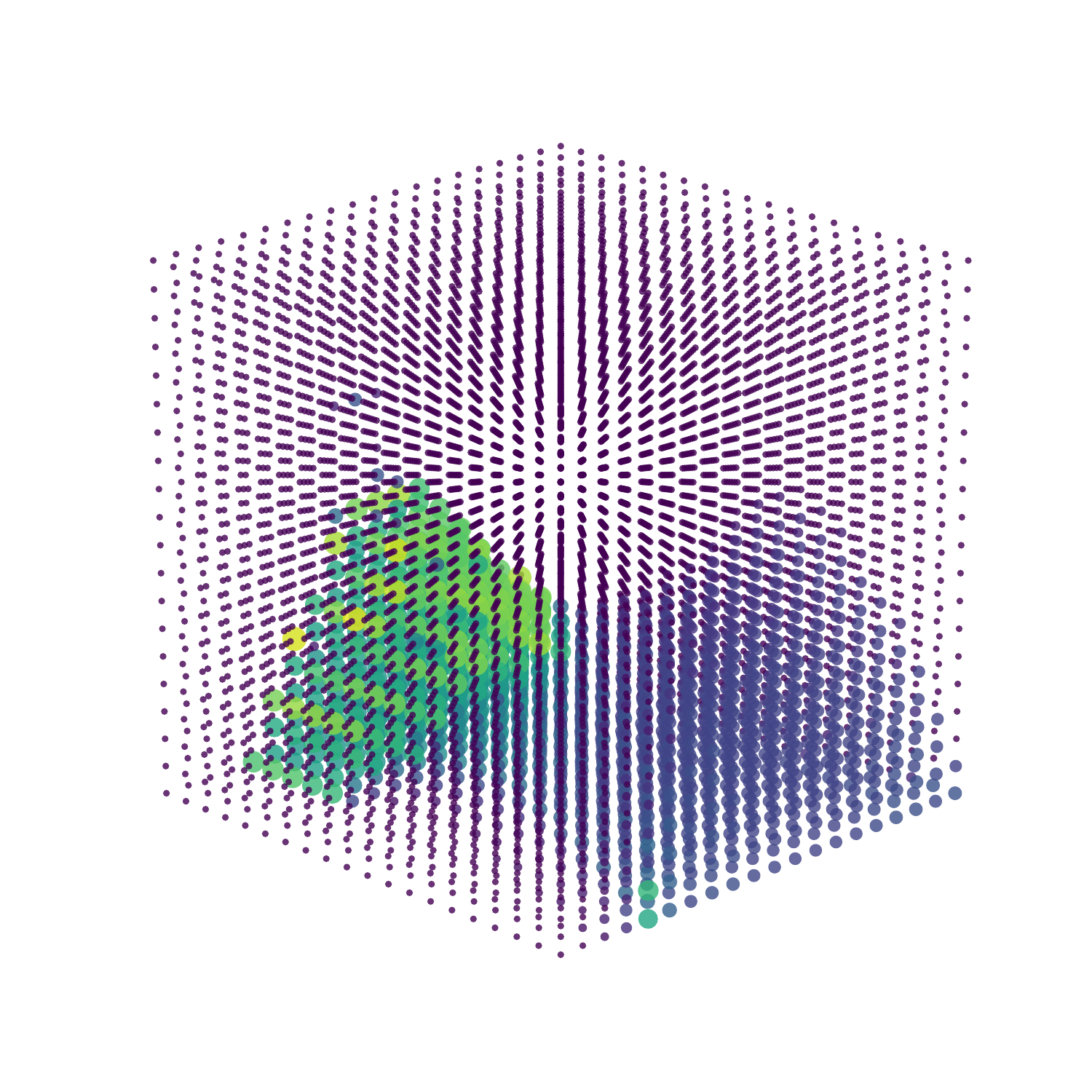} \\[-2.3ex]

        \rotatebox{90}{\hspace{0pt}\codeblue{Standstill}} &
        \includegraphics[trim=0 75 0 75, clip, width=\imgwidth, valign=m]{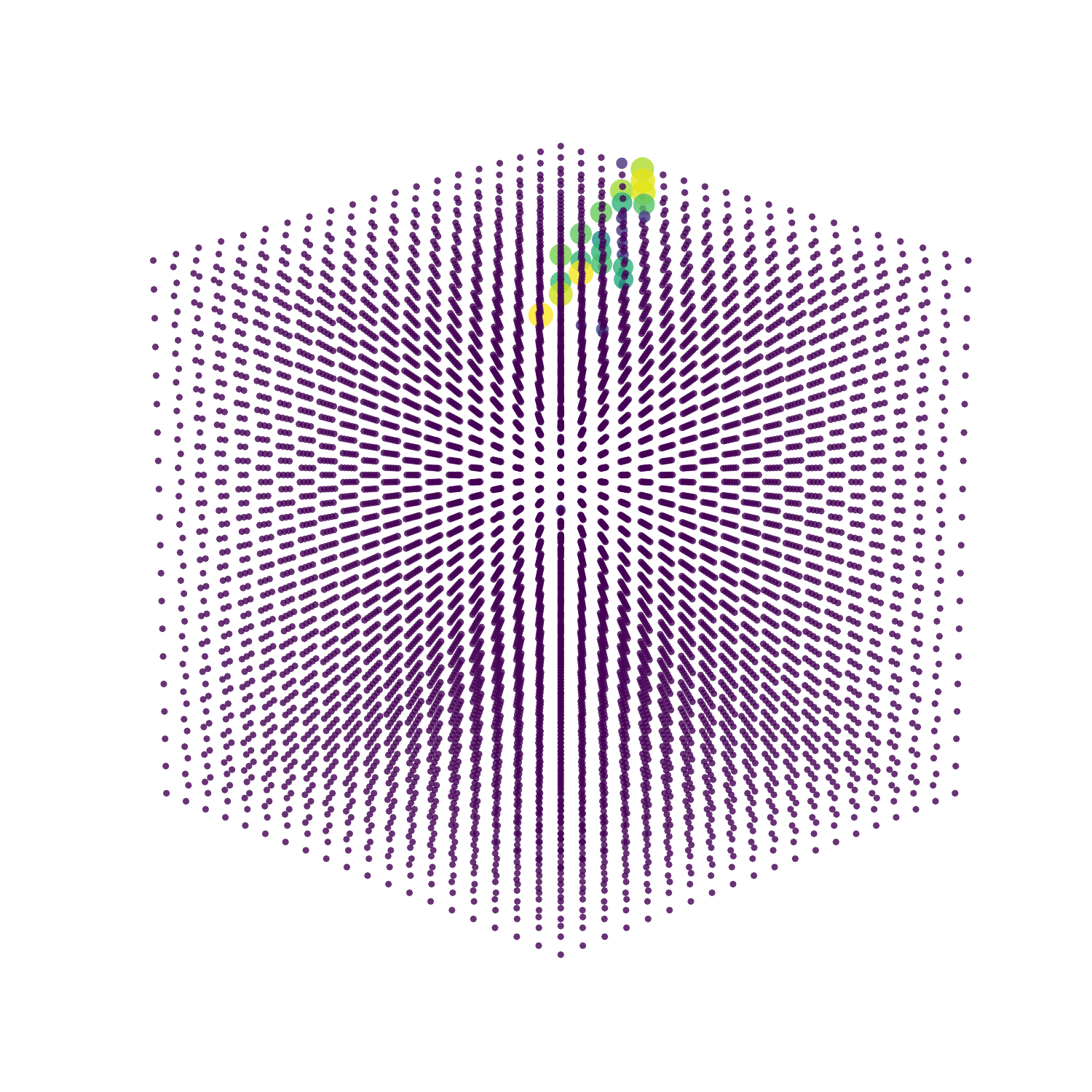} &
        \includegraphics[trim=0 75 0 75, clip, width=\imgwidth, valign=m]{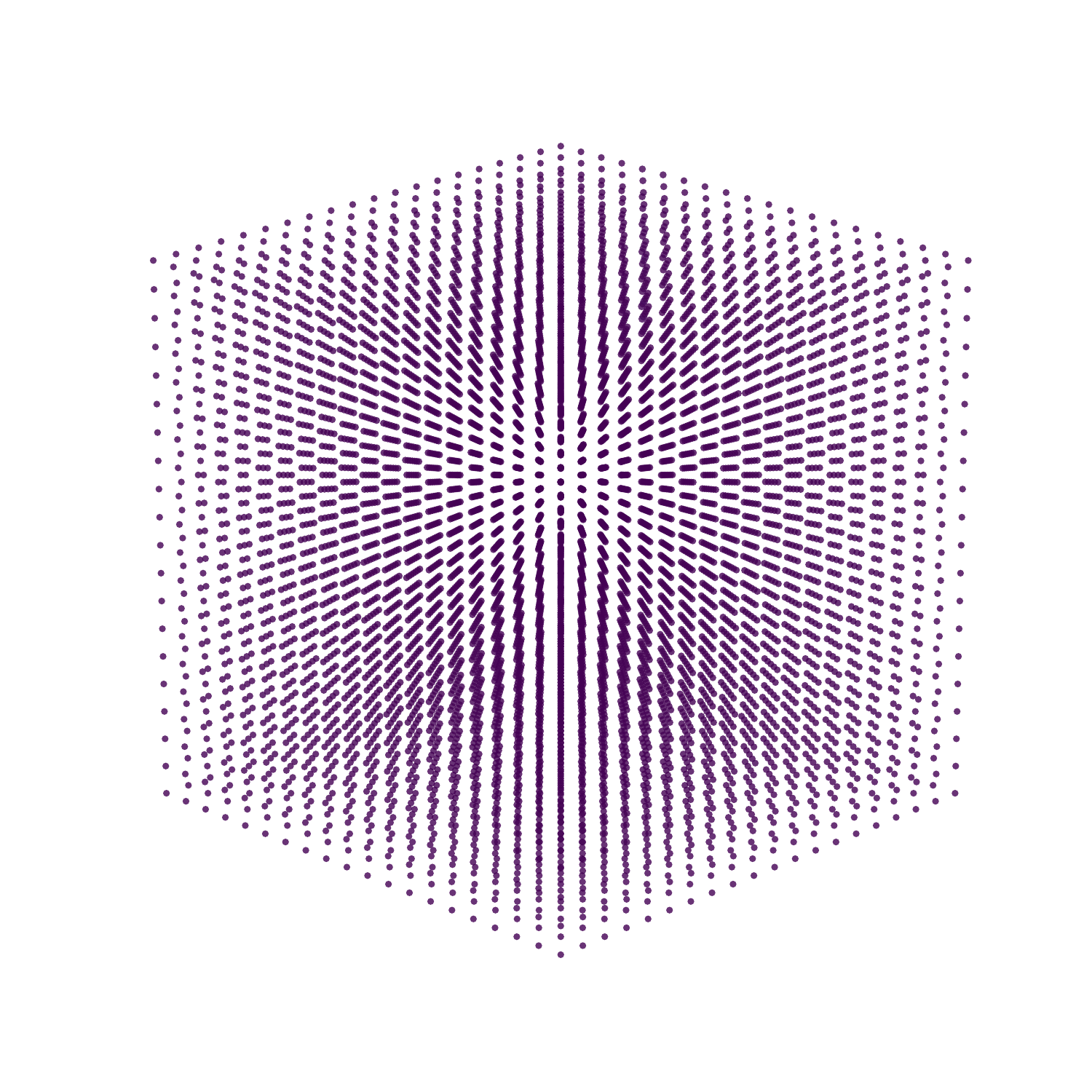} &
        \includegraphics[trim=0 75 0 75, clip, width=\imgwidth, valign=m]{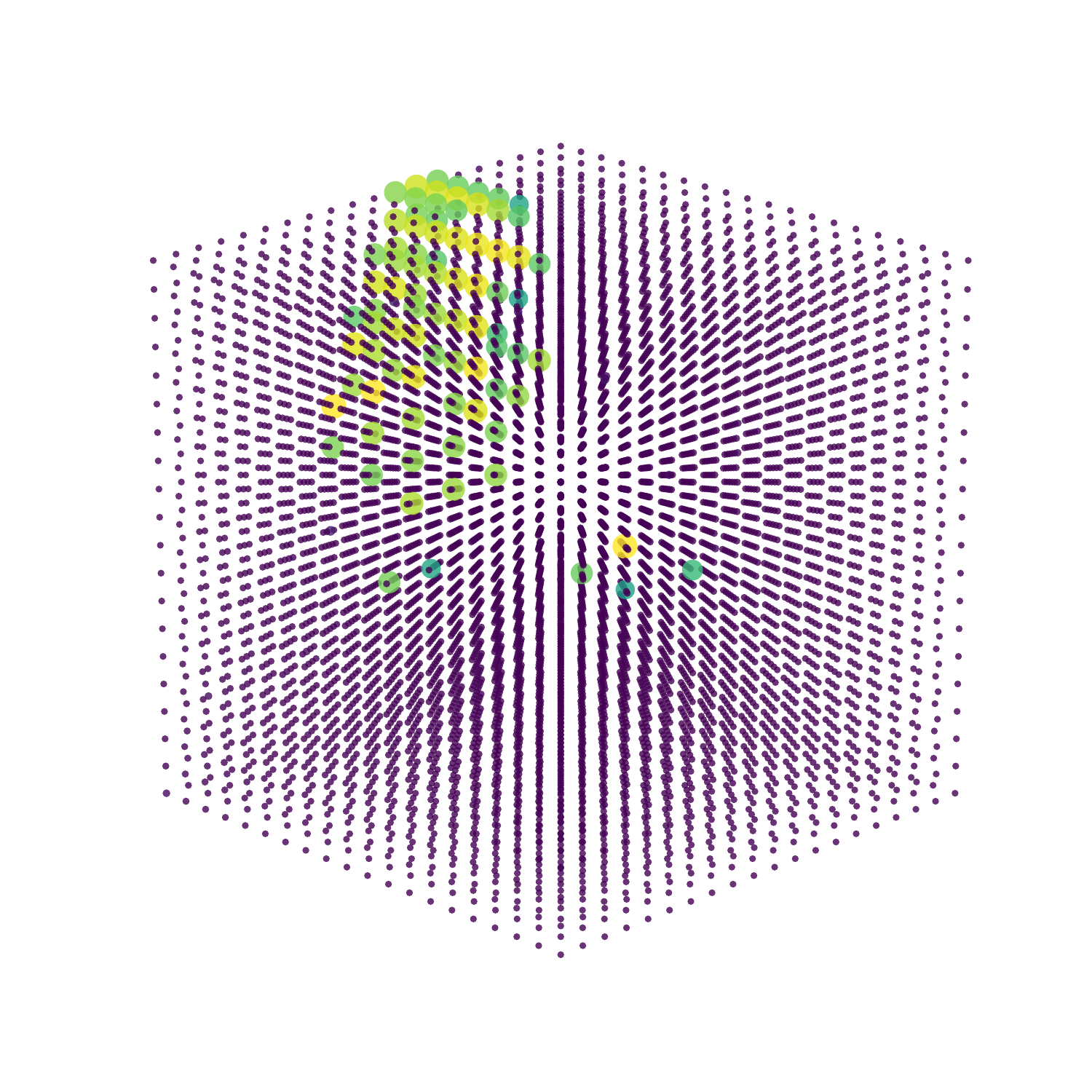} &
        \includegraphics[trim=0 75 0 75, clip, width=\imgwidth, valign=m]{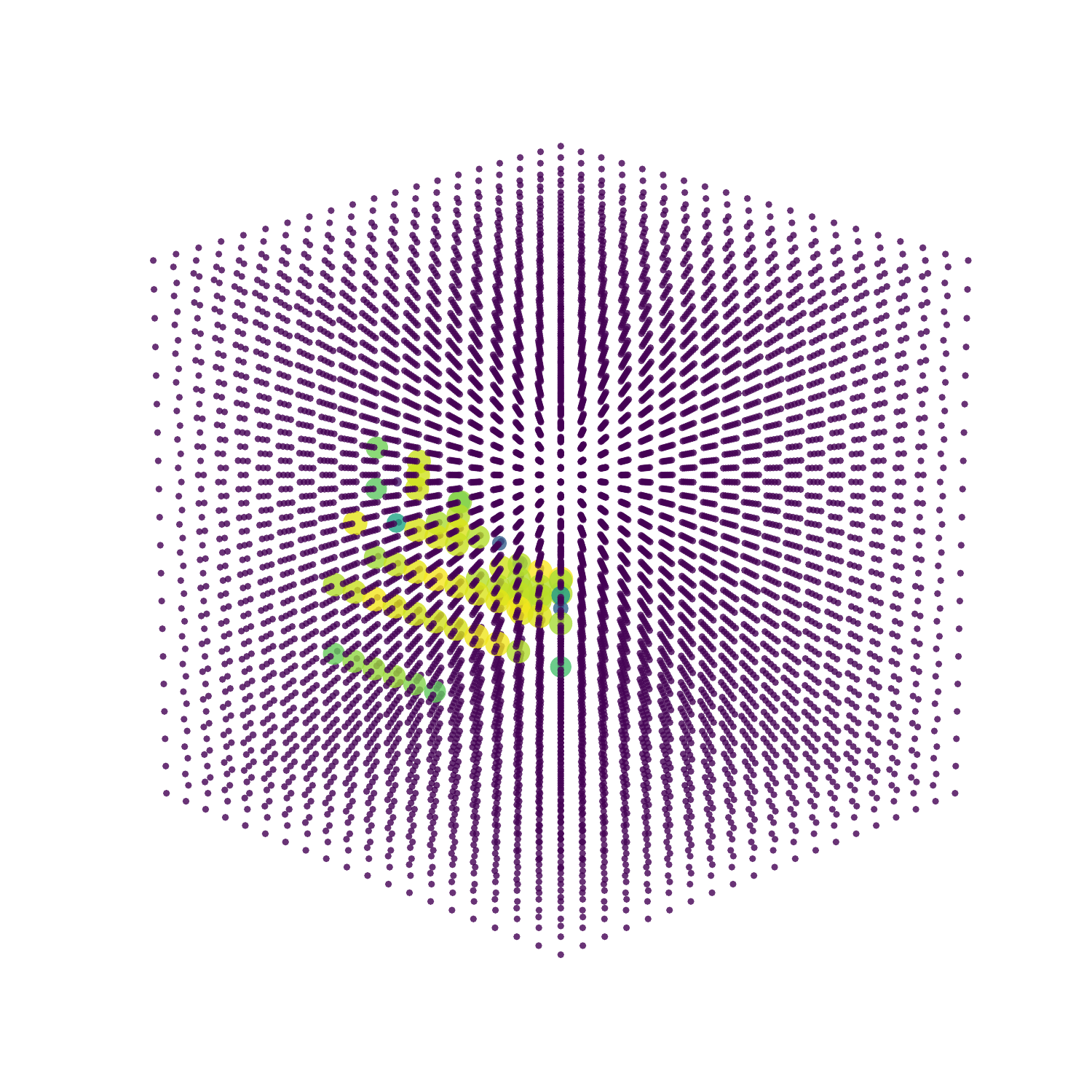} \\
    \end{tabular}
    \caption{HP, seed 999, 3D}
    \label{fig:full_latent_comparison_999_3d}
\end{figure*}

\subsection{Latent Behavior Optimization}
\label{app:add_optimization}
In this experiment, we compare the optimization performances of our pipeline (O-PGPE) against similar methods and other competitive DRL baselines, which include: P-PGPE, i.e., standard PGPE running in the parameter space, SAC, DDPG, and TD3, all utilizing policies of the same shape as the one being compressed by our autoencoder. Additionally, we compare our results with A-PGPE (Latent PGPE on the Action-based Policy Compression manifolds). Both O-PGPE and A-PGPE are tested on the same bottleneck dimensions, namely: $k=1,2,3$ for MC, and $k=3,5,8$ for RC and HP. Finally, we evaluate the performance of WS-O-PGPE (Warm Start O-PGPE) with bottleneck size $k=3$, $5$, and $8$.

\begin{figure}[t]

    \centering 
    \includegraphics[width=0.9\textwidth]{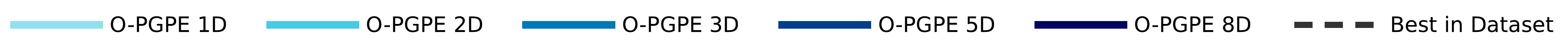}
    \includegraphics[width=0.5\textwidth]{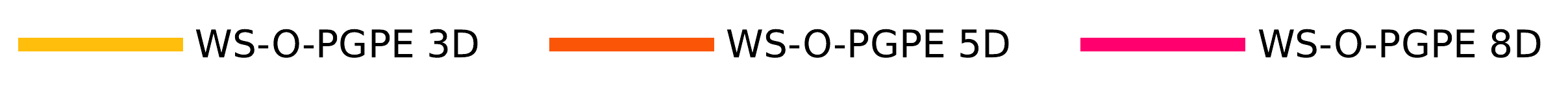}
    \vfill
    \begin{subfigure}[b]{0.24\textwidth}
        \includegraphics[width=\textwidth]{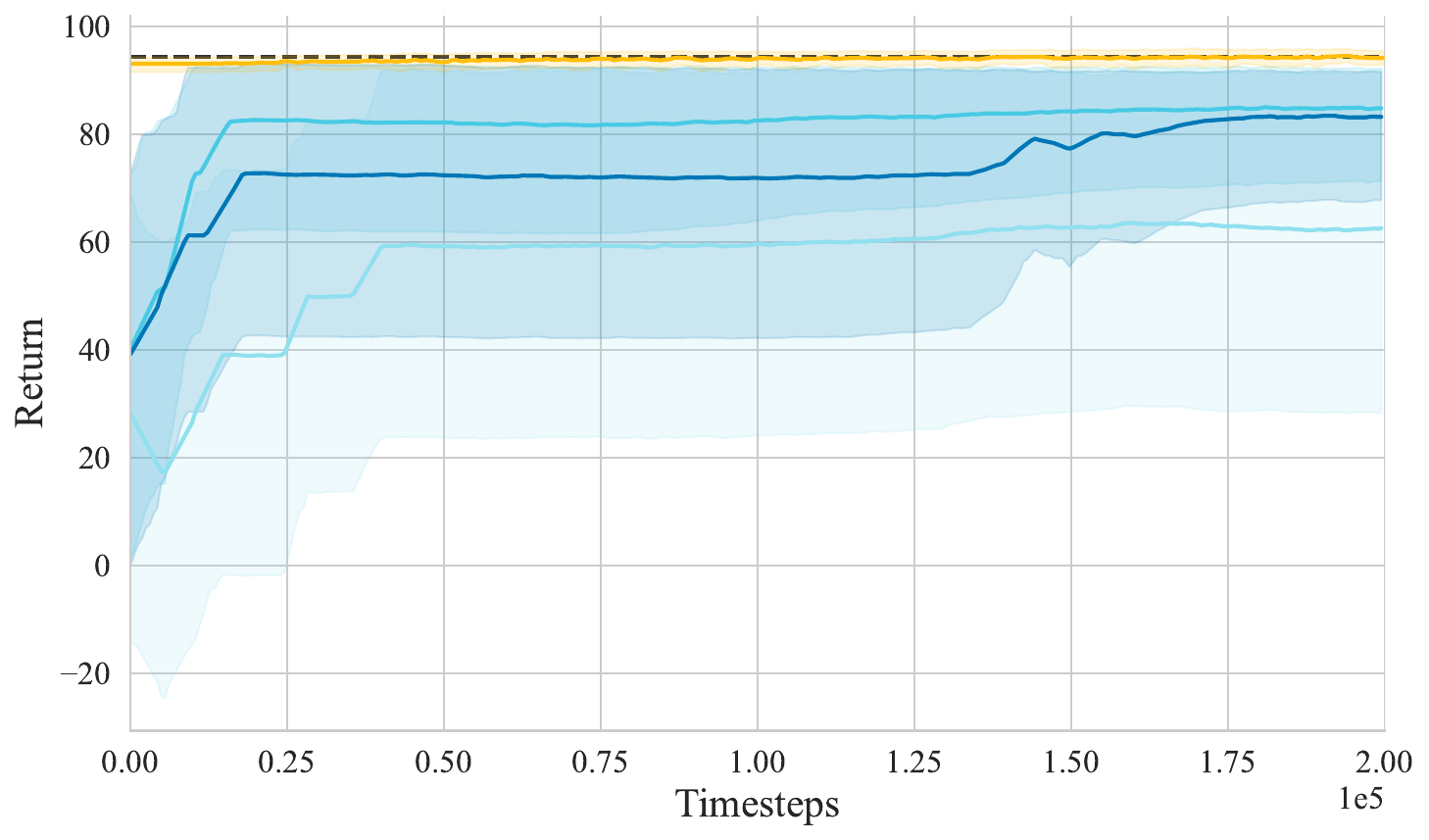}
        \caption{MC, \codeblue{standard}}
        
    \end{subfigure}
    \hfill 
    \begin{subfigure}[b]{0.24\textwidth}
        \includegraphics[width=\textwidth]{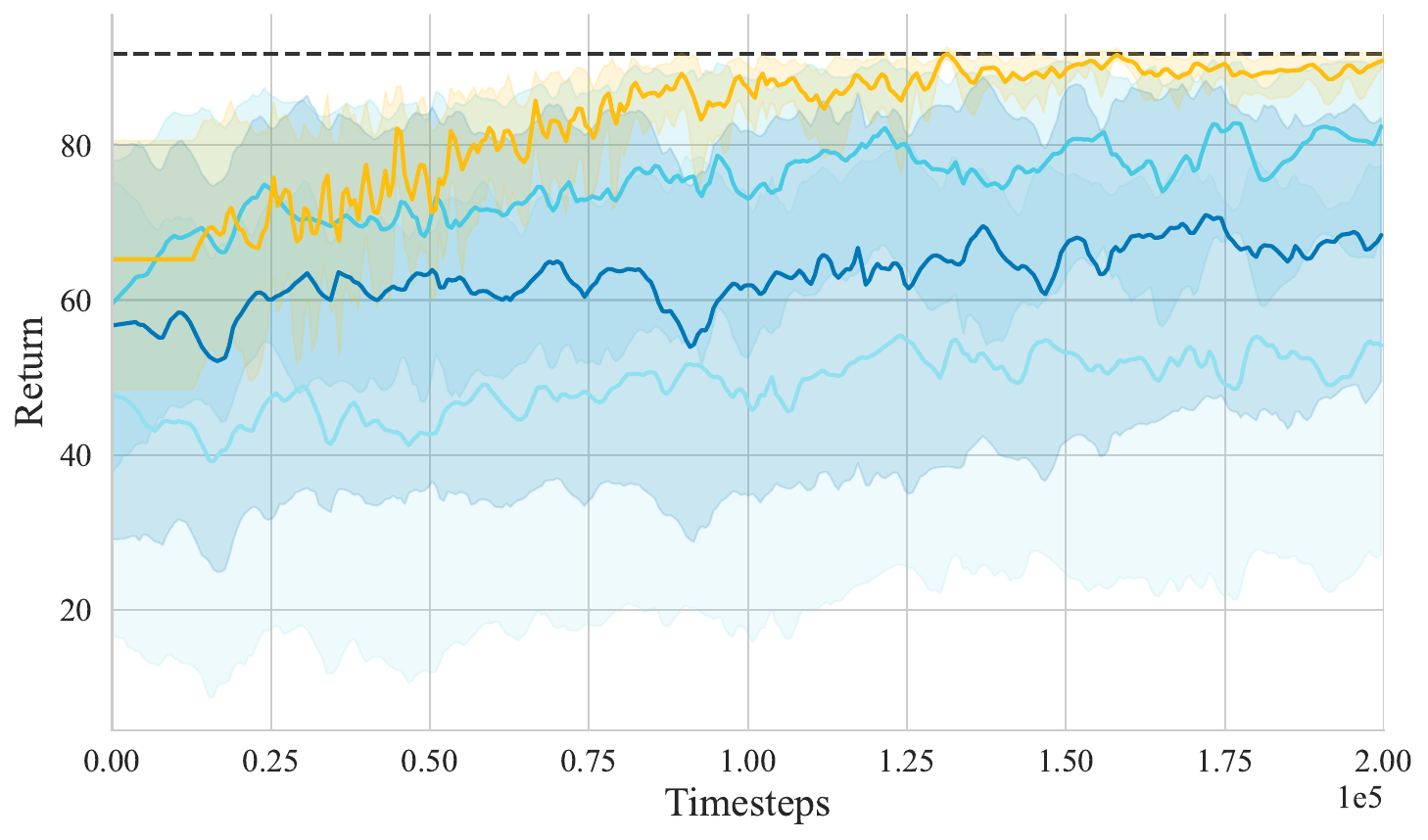}
        \caption{MC, \codeblue{left}}
        
    \end{subfigure}
    \hfill
    \begin{subfigure}[b]{0.24\textwidth}
        \includegraphics[width=\textwidth]{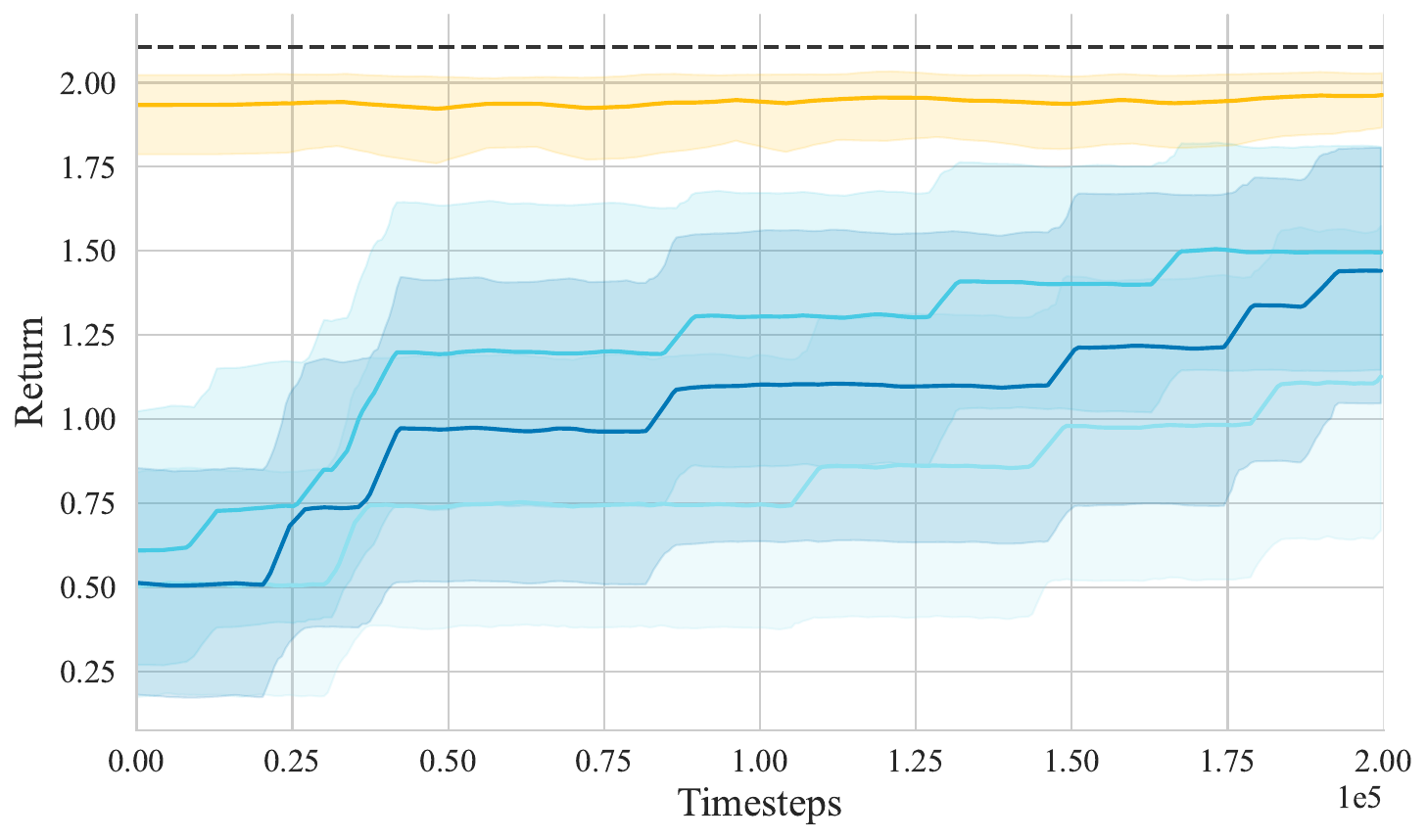}
        \caption{MC, \codeblue{speed}}
        
    \end{subfigure}
    \hfill
    \begin{subfigure}[b]{0.24\textwidth}
        \includegraphics[width=\textwidth]{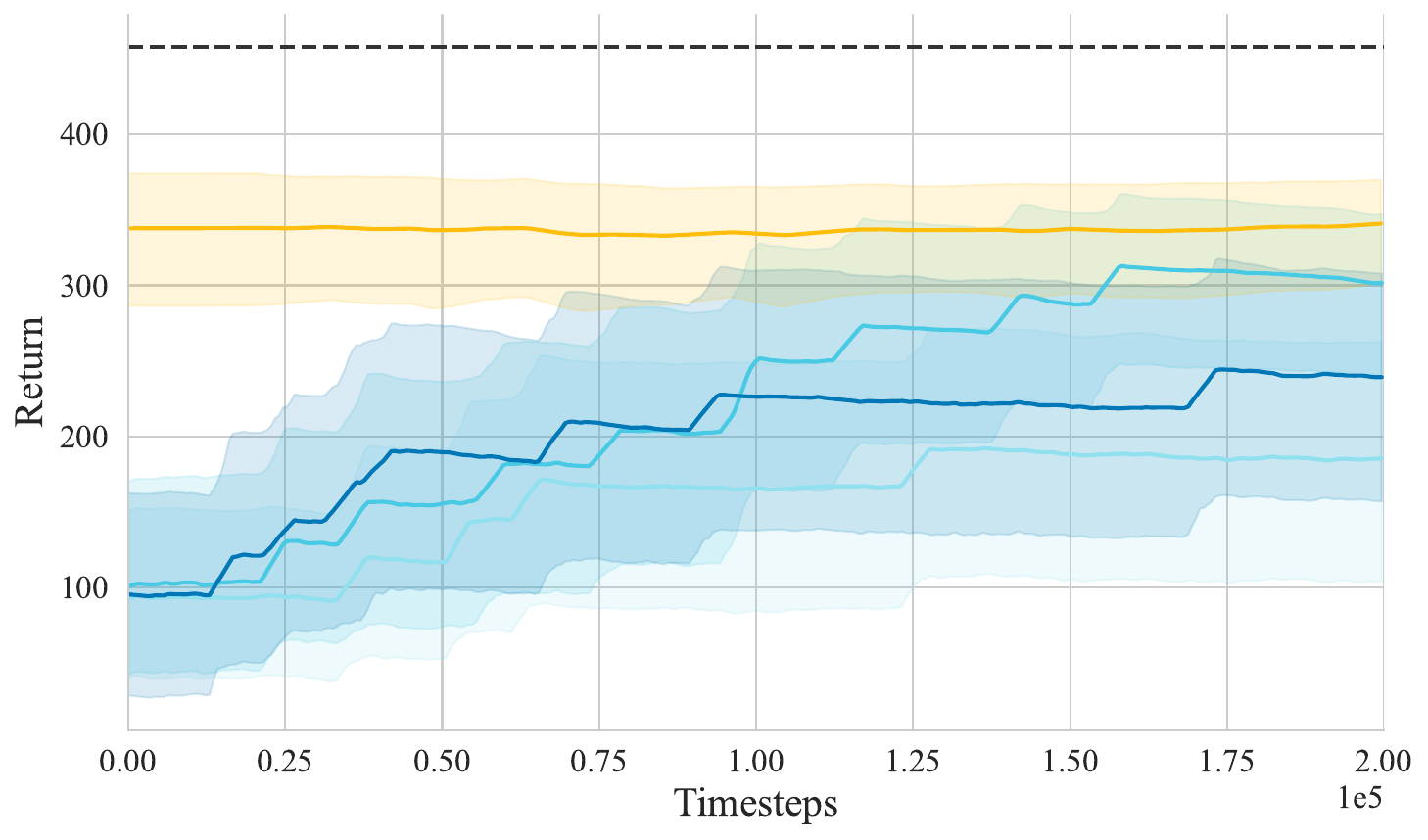}
        \caption{MC, \codeblue{height}}
        
    \end{subfigure}

    \begin{subfigure}[b]{0.24\textwidth}
        \includegraphics[width=\textwidth]{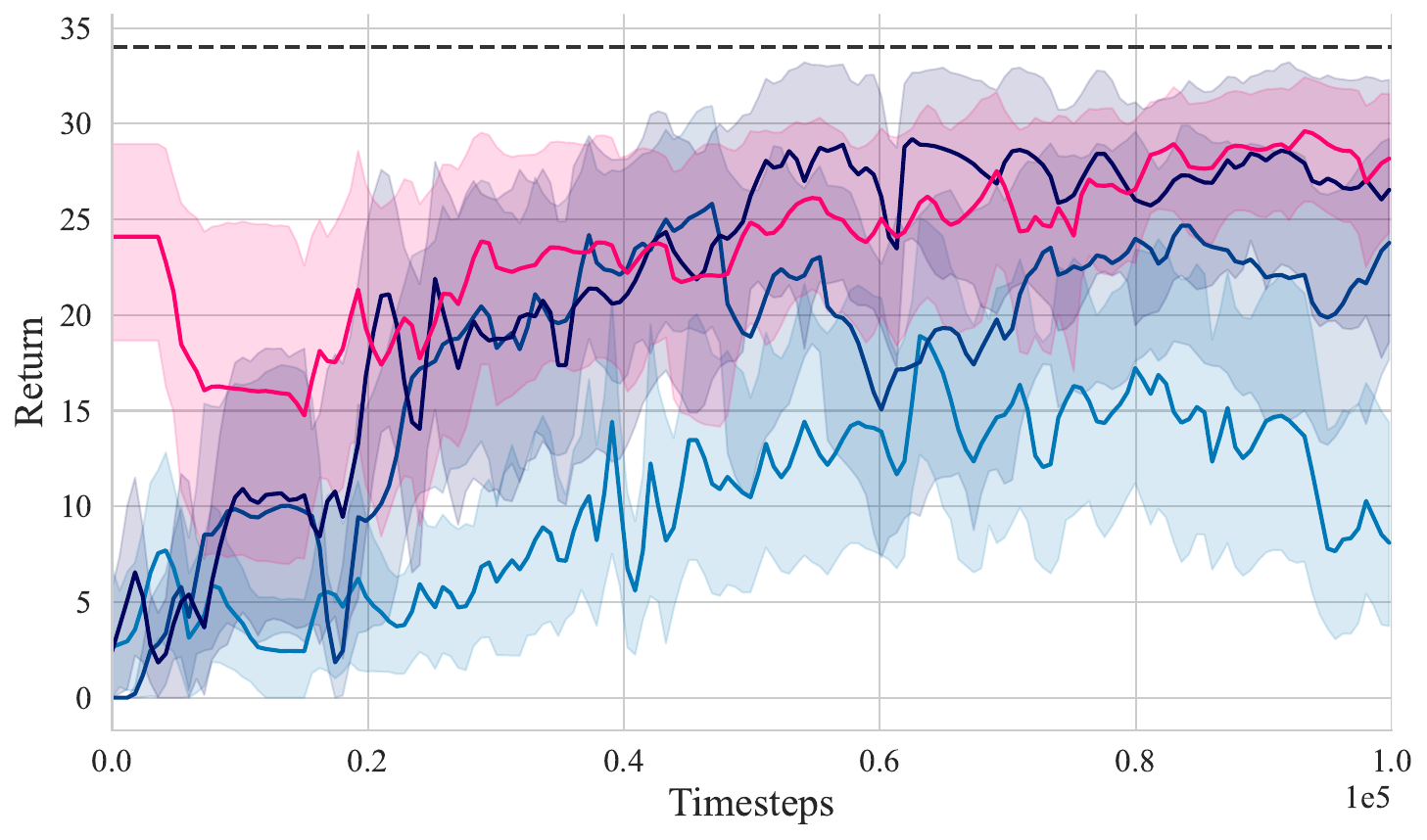}
         \caption{RC, \codeblue{speed}}
        
    \end{subfigure}
    \hfill 
    \begin{subfigure}[b]{0.24\textwidth}
        \includegraphics[width=\textwidth]{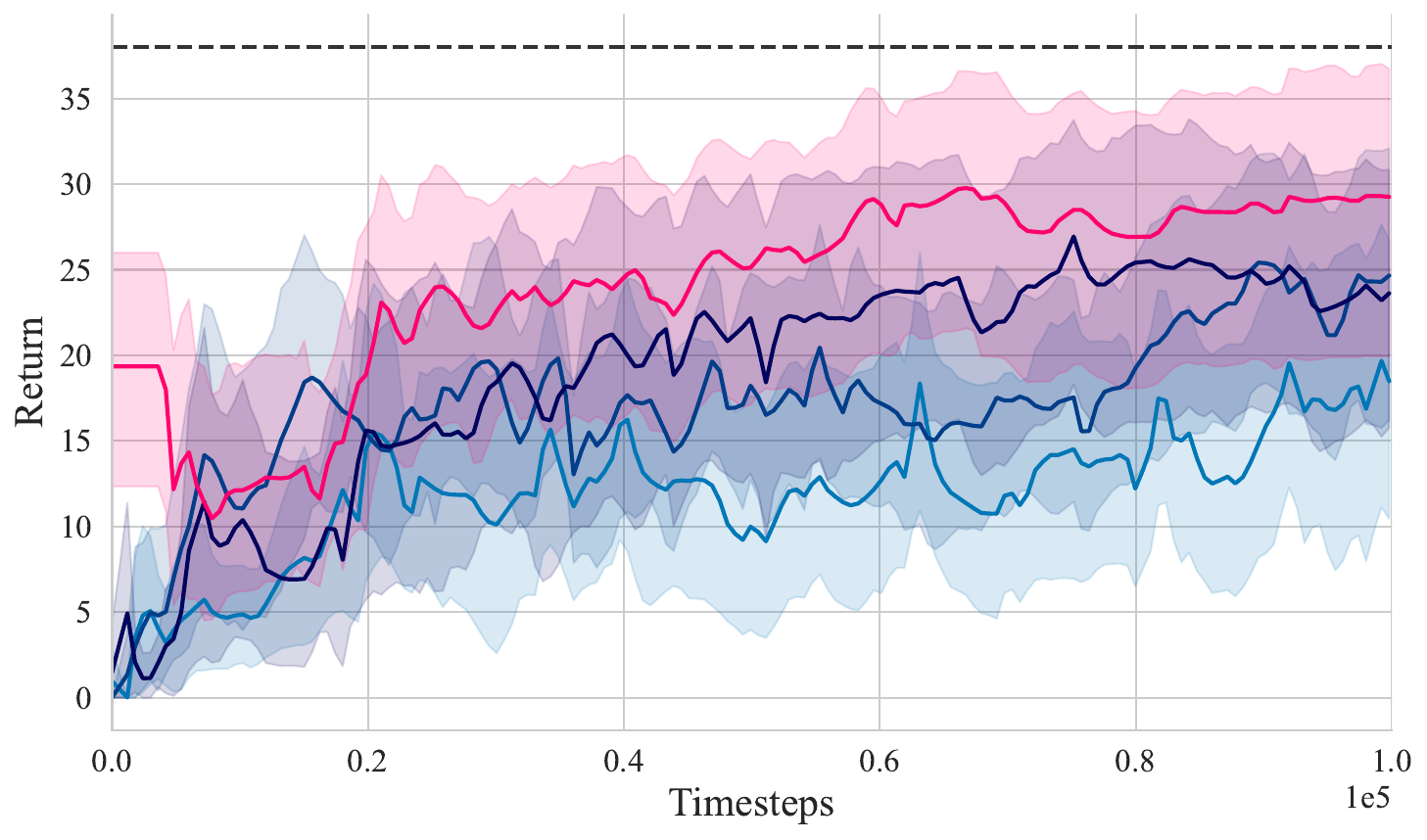}
        \caption{RC, \codeblue{clockwise}}
        
    \end{subfigure}
    \hfill
    \begin{subfigure}[b]{0.24\textwidth}
        \includegraphics[width=\textwidth]{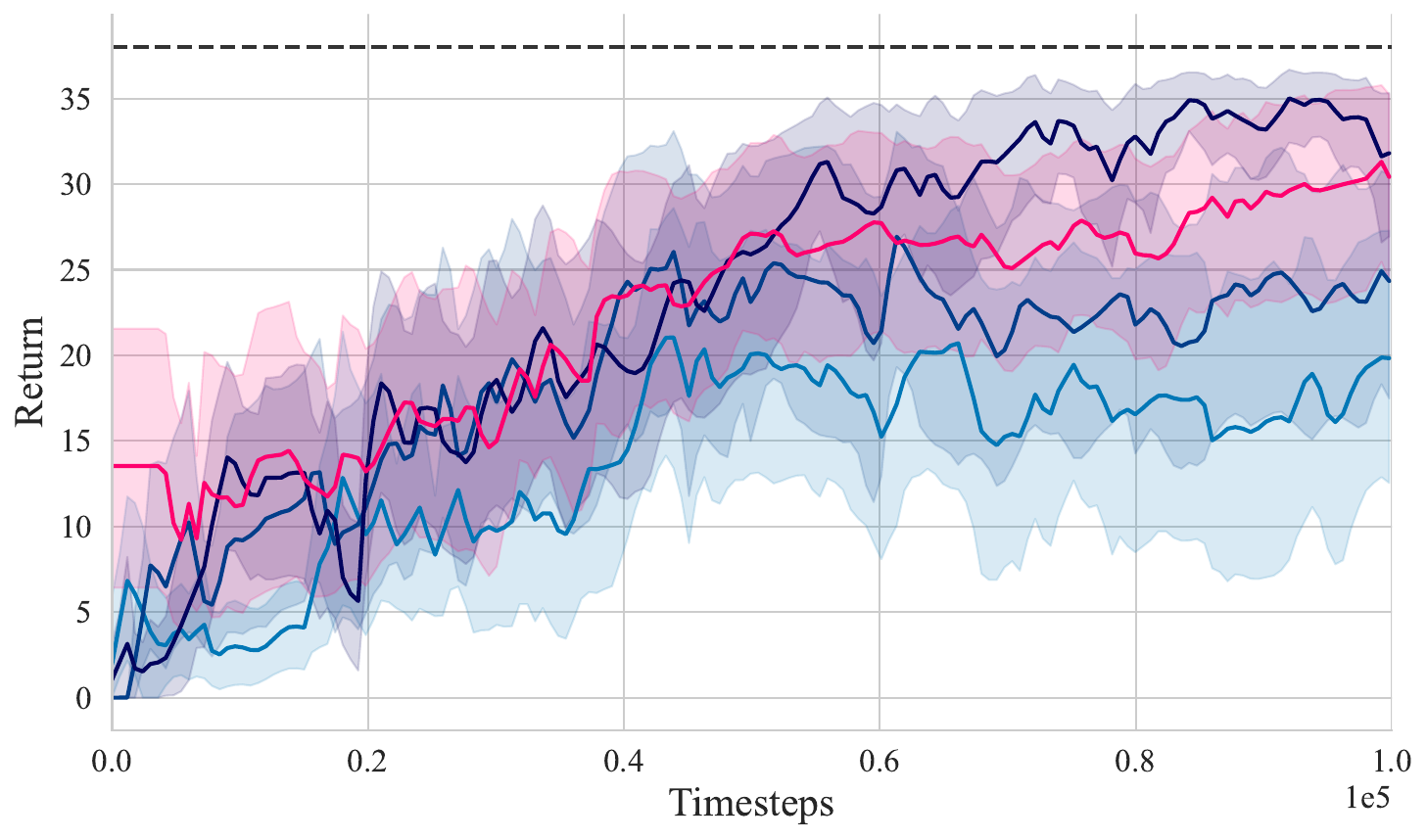}
        \caption{RC, \codeblue{c-clockwise}}
        
    \end{subfigure}
    \hfill
    \begin{subfigure}[b]{0.24\textwidth}
        \includegraphics[width=\textwidth]{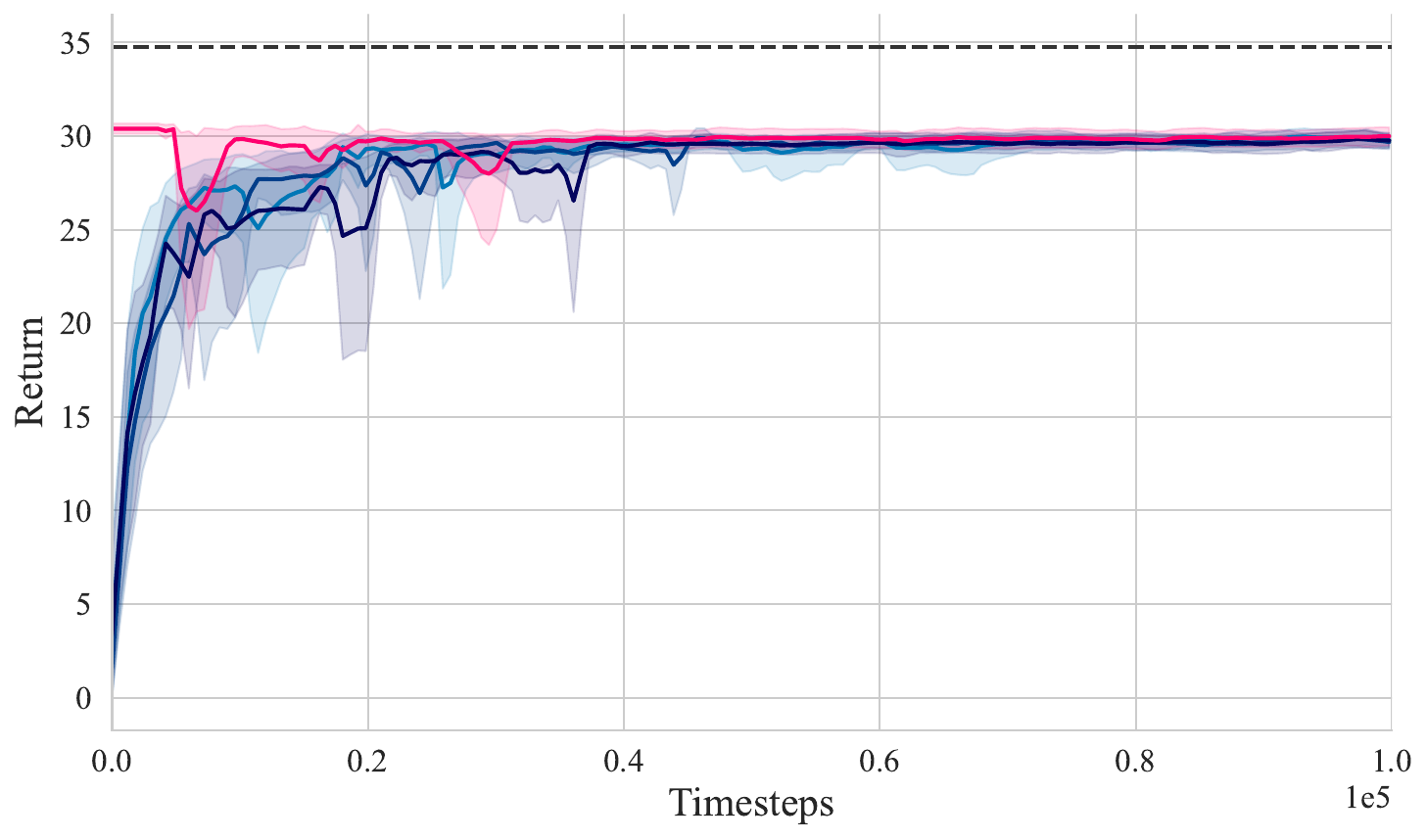}
        \caption{RC, \codeblue{radial}}
        
    \end{subfigure}

    \begin{subfigure}[b]{0.24\textwidth}
        \includegraphics[width=\textwidth]{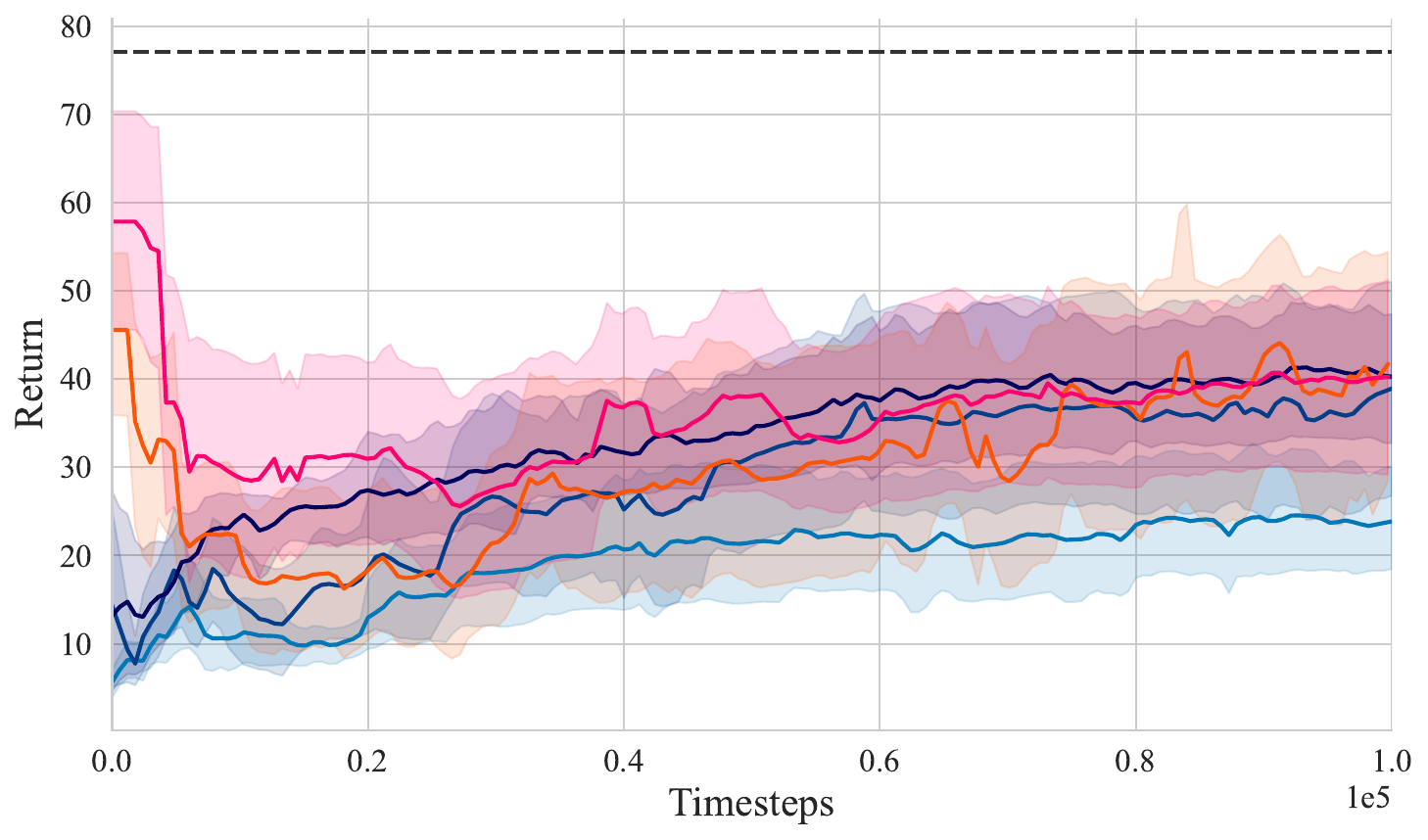}
        \caption{HP, \codeblue{forward}}
        
    \end{subfigure}
    \hfill 
    \begin{subfigure}[b]{0.24\textwidth}
        \includegraphics[width=\textwidth]{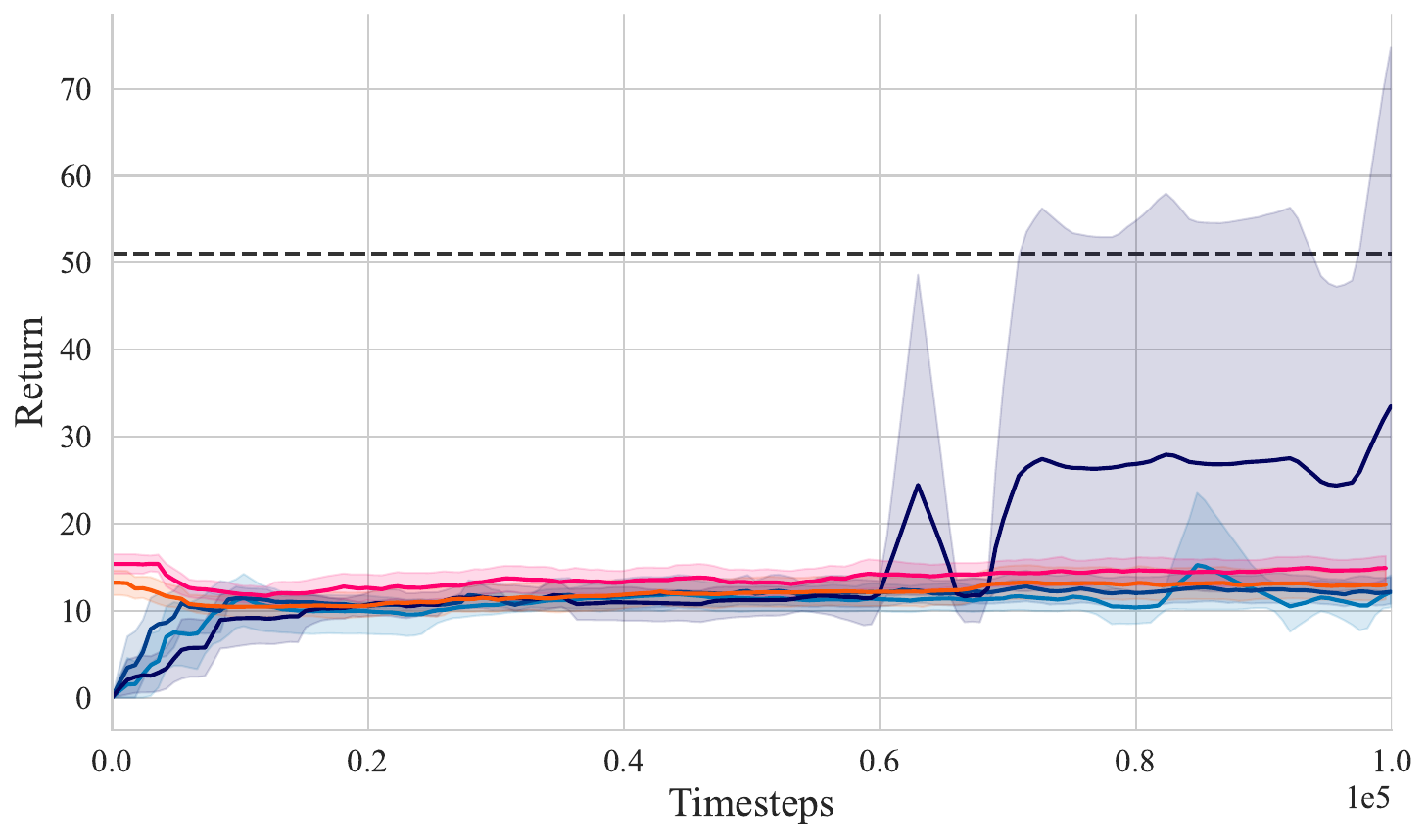}
        \caption{HP, \codeblue{backward}}
        
    \end{subfigure}
    \hfill
    \begin{subfigure}[b]{0.24\textwidth}
        \includegraphics[width=\textwidth]{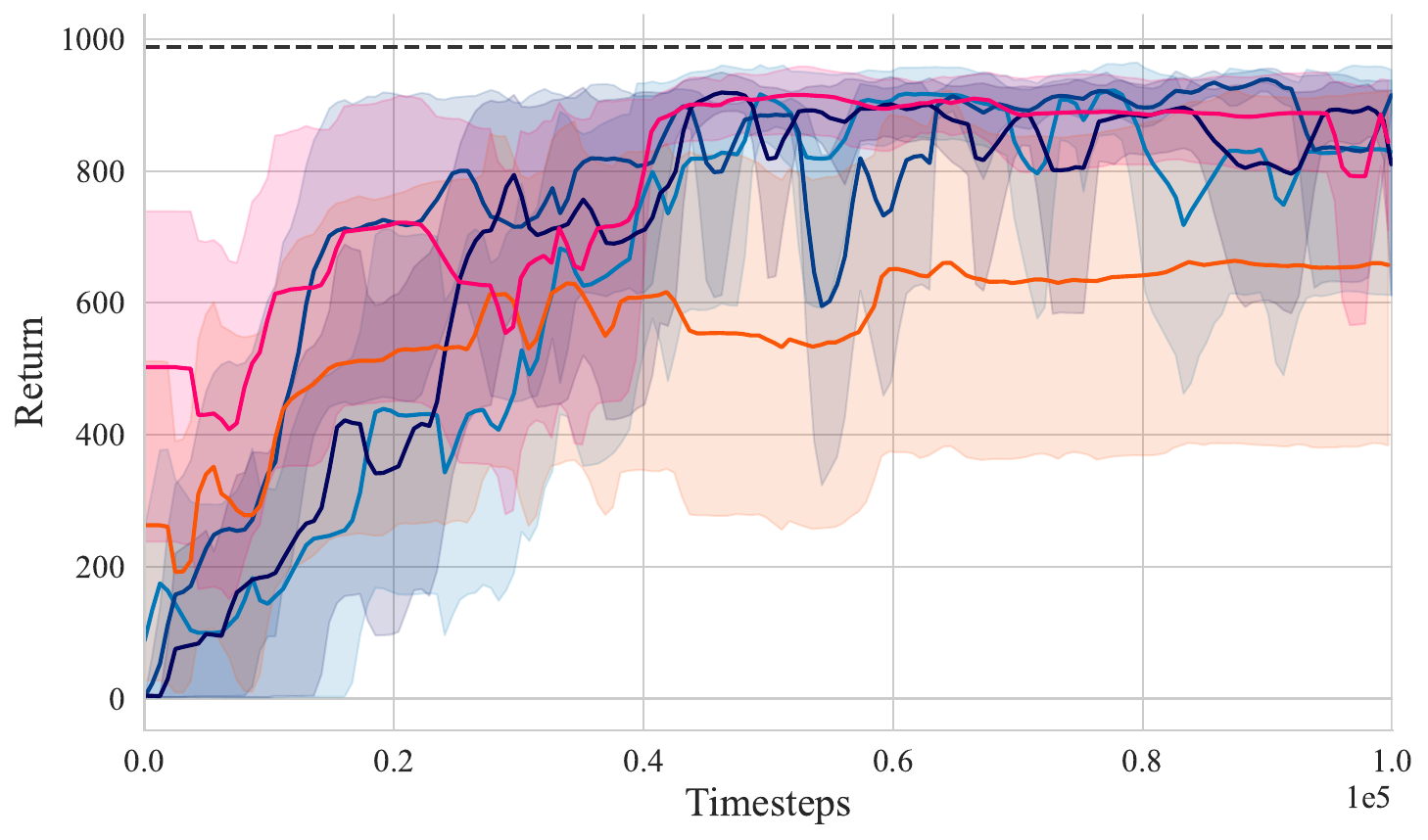}
        \caption{HP, \codeblue{standstill}}
        
    \end{subfigure}
    \hfill
    \begin{subfigure}[b]{0.24\textwidth}
        \includegraphics[width=\textwidth]{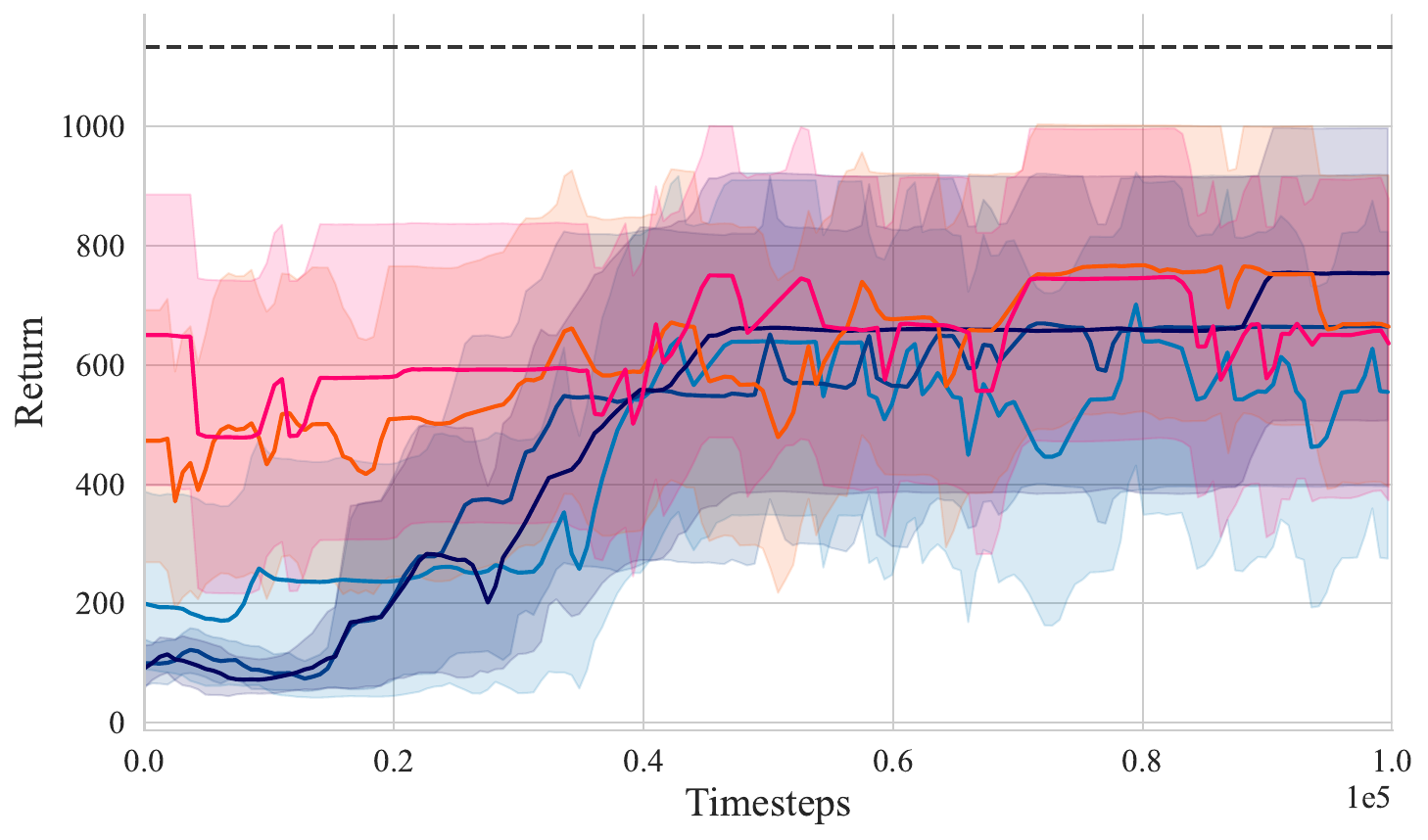}
        \caption{HP, \codeblue{standard}}
        
    \end{subfigure}
    \caption{Performance comparison in MC, RC, and HP for different tasks using \textbf{Occupancy-based Policy Compression}.}
    \label{fig:all_opc_performances}
\end{figure}

\begin{figure}[t]

    \centering 
    \includegraphics[width=0.7\textwidth]{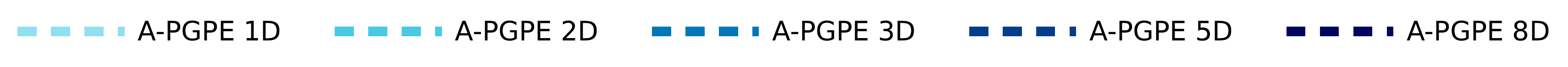}
    \vfill
    \begin{subfigure}[b]{0.24\textwidth}
        \includegraphics[width=\textwidth]{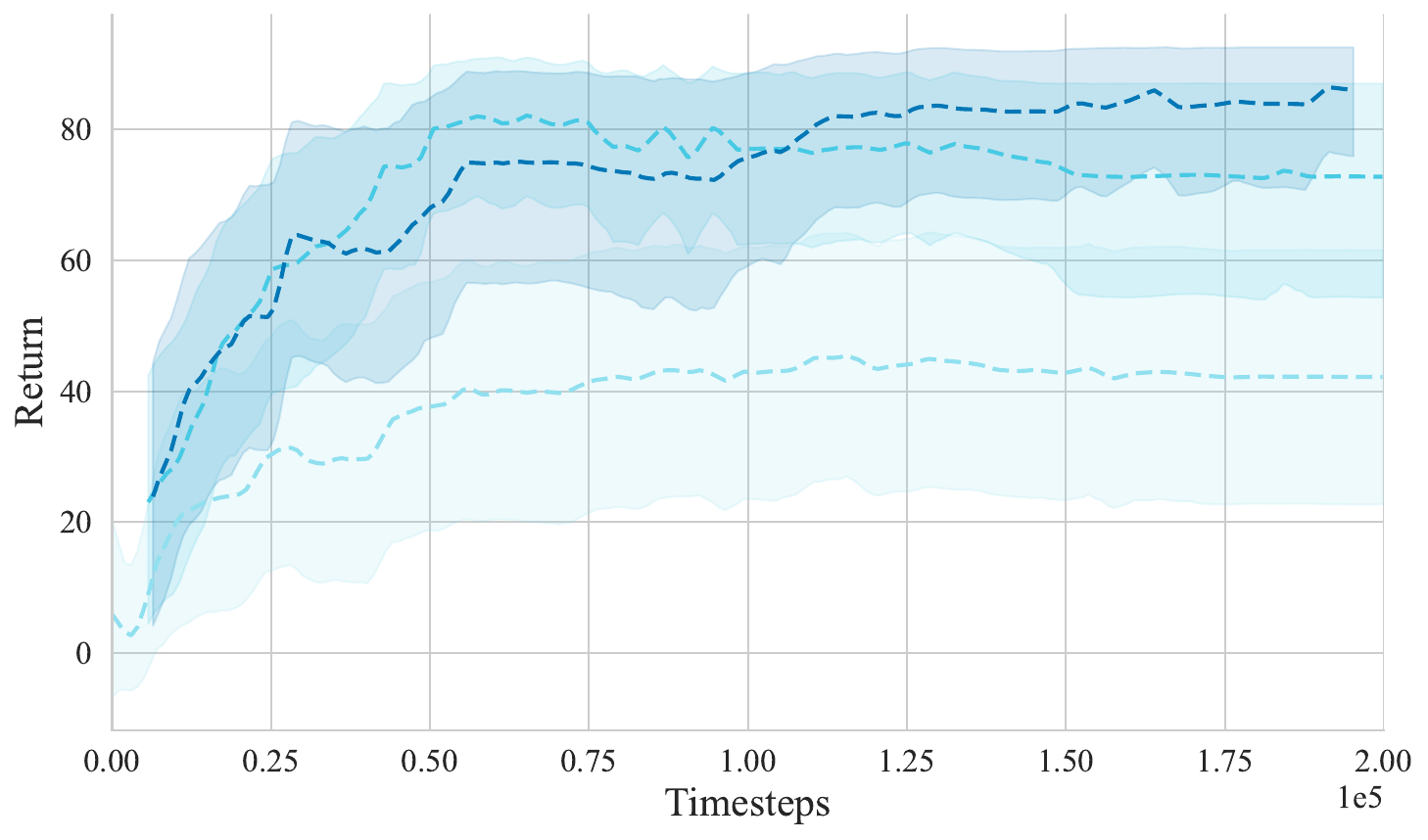}
        \caption{MC, \codeblue{standard}}
        
    \end{subfigure}
    \hfill 
    \begin{subfigure}[b]{0.24\textwidth}
        \includegraphics[width=\textwidth]{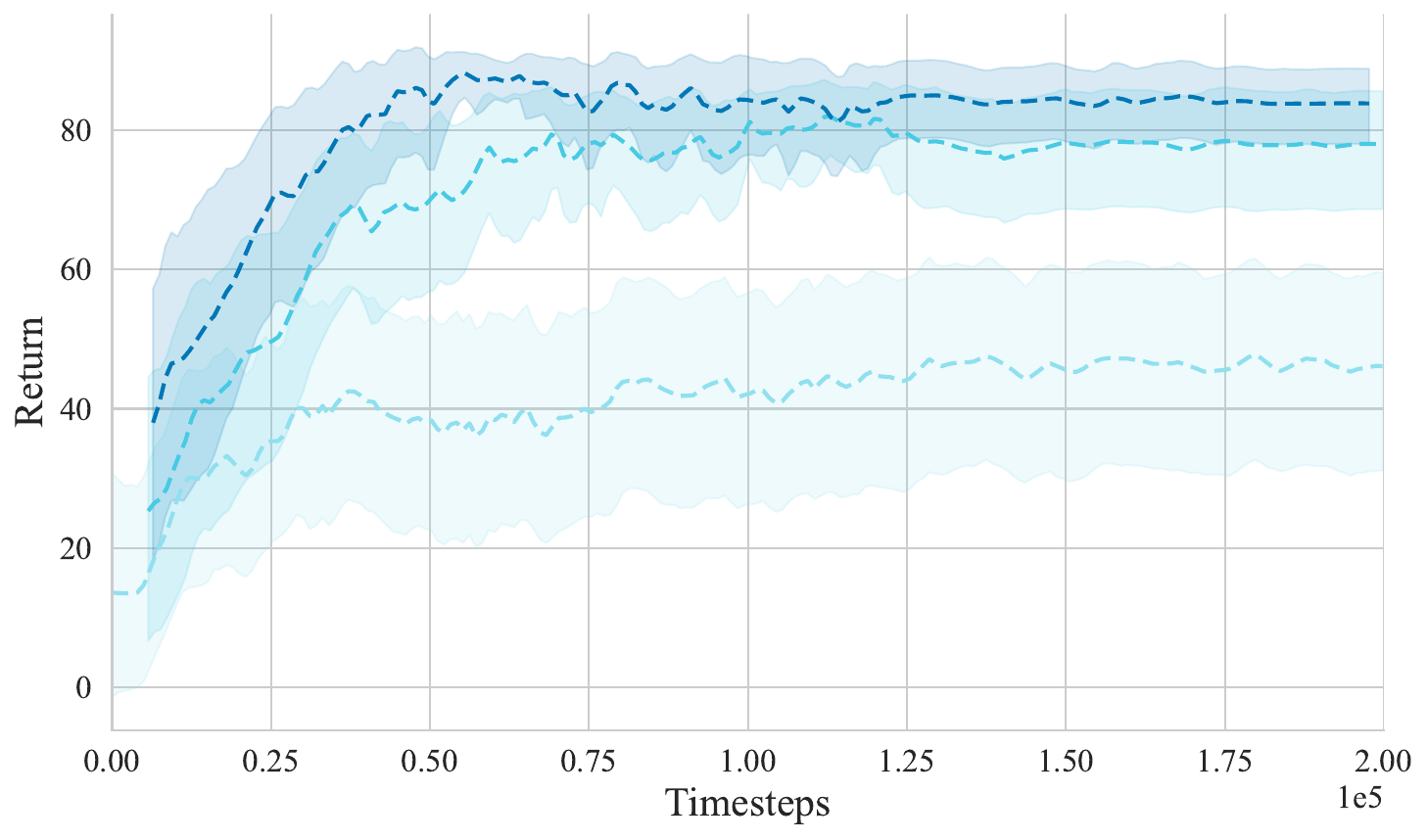}
        \caption{MC, \codeblue{left}}
        
    \end{subfigure}
    \hfill
    \begin{subfigure}[b]{0.24\textwidth}
        \includegraphics[width=\textwidth]{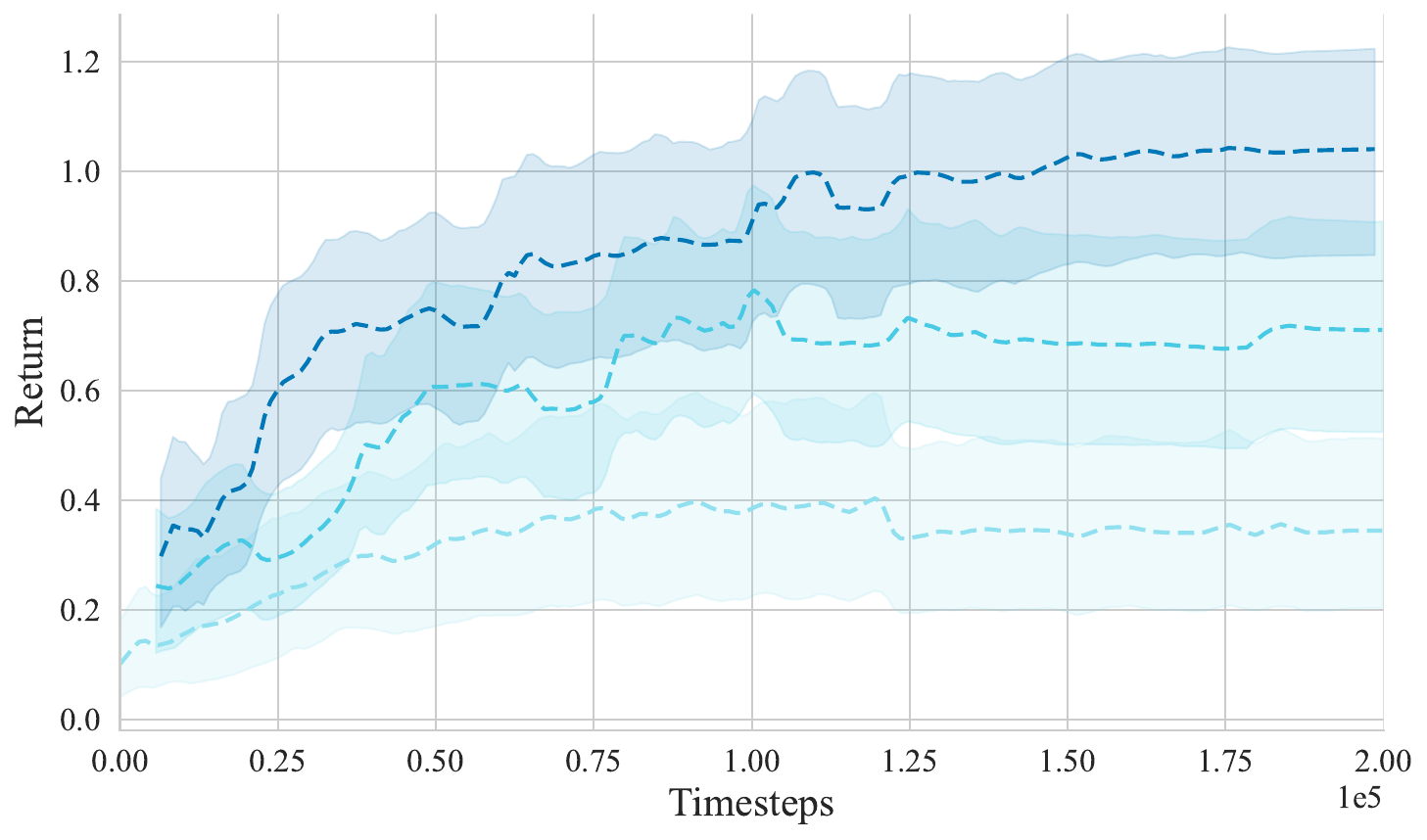}
        \caption{MC, \codeblue{speed}}
        
    \end{subfigure}
    \hfill
    \begin{subfigure}[b]{0.24\textwidth}
        \includegraphics[width=\textwidth]{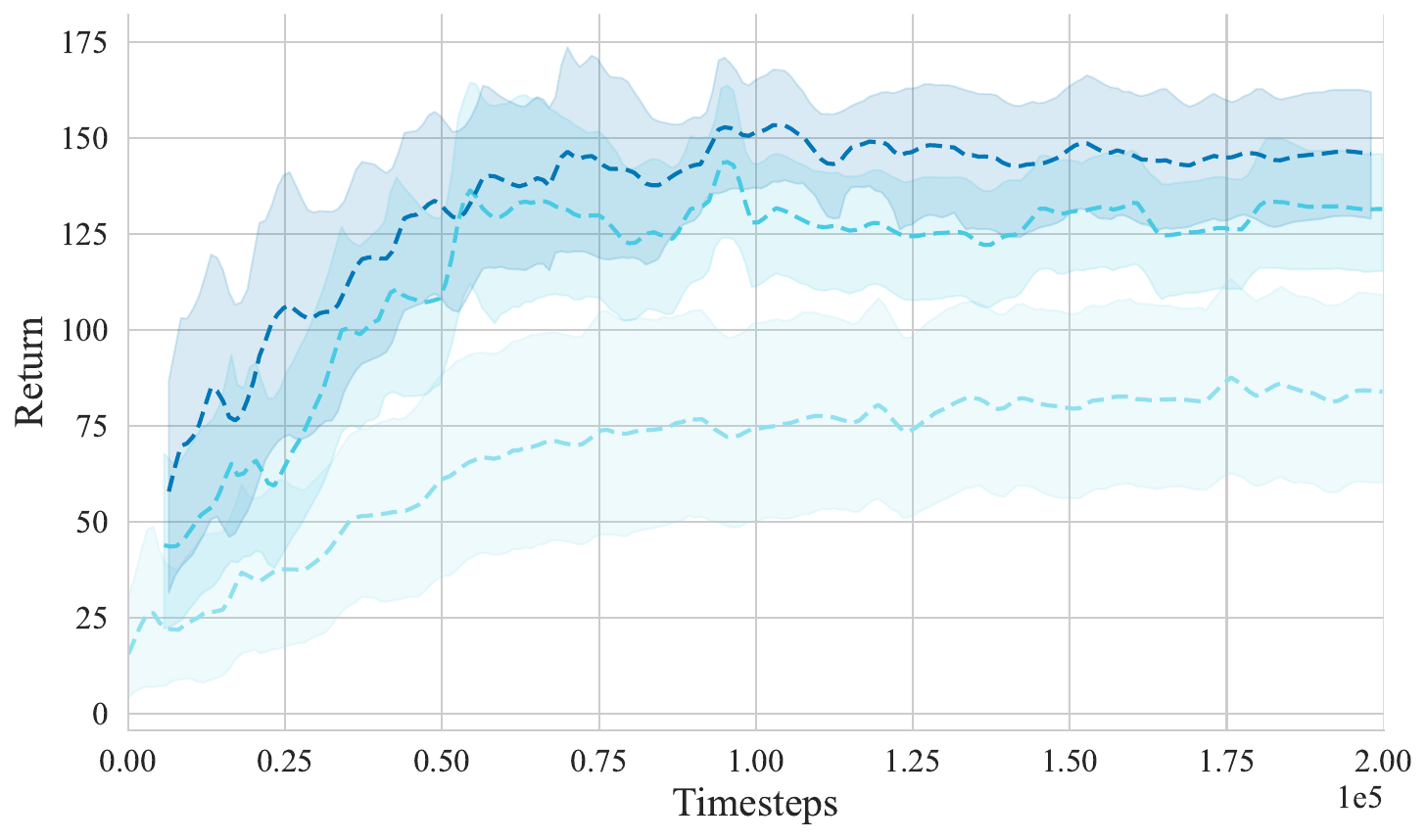}
        \caption{MC, \codeblue{height}}
        
    \end{subfigure}

    \begin{subfigure}[b]{0.24\textwidth}
        \includegraphics[width=\textwidth]{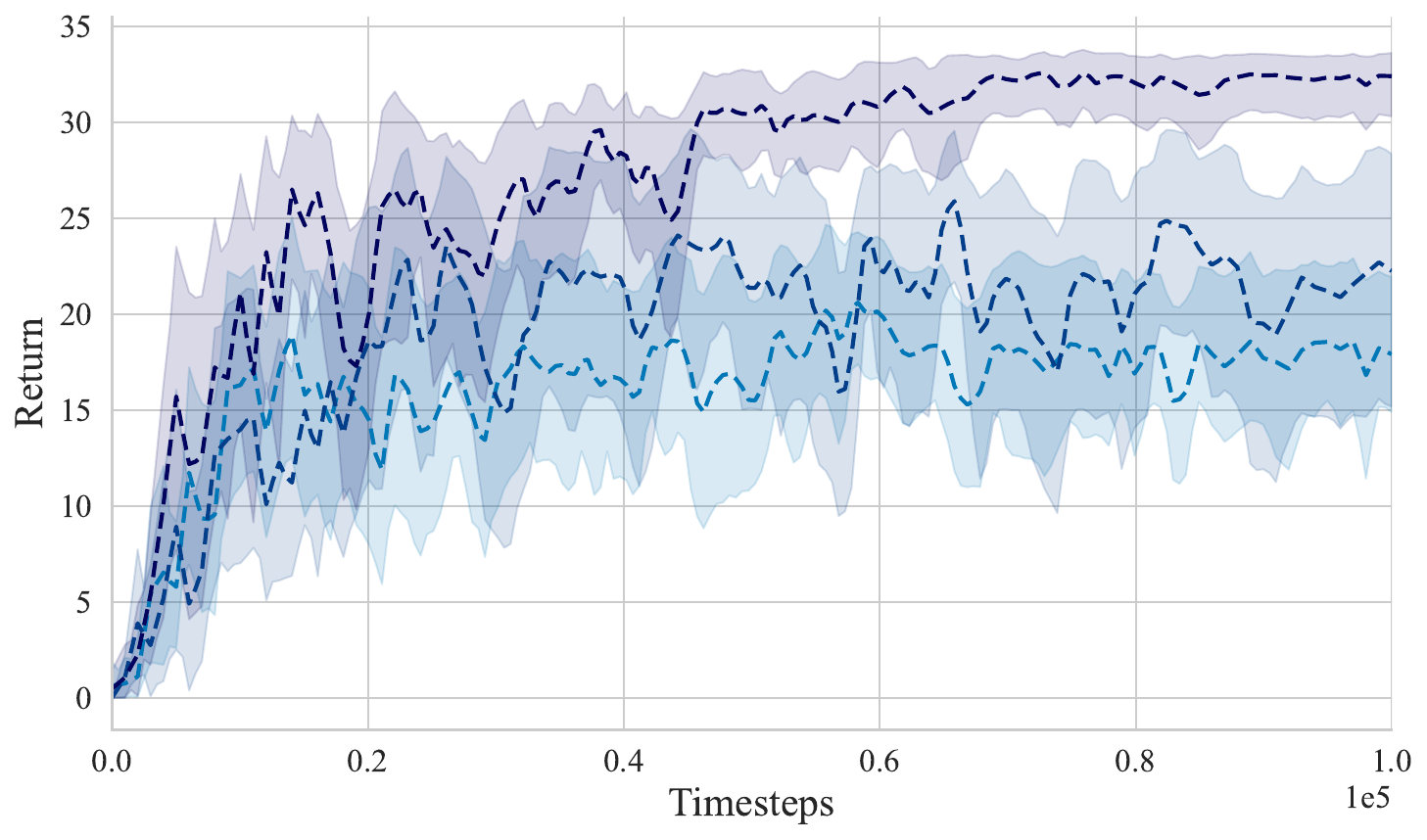}
         \caption{RC, \codeblue{speed}}
        
    \end{subfigure}
    \hfill 
    \begin{subfigure}[b]{0.24\textwidth}
        \includegraphics[width=\textwidth]{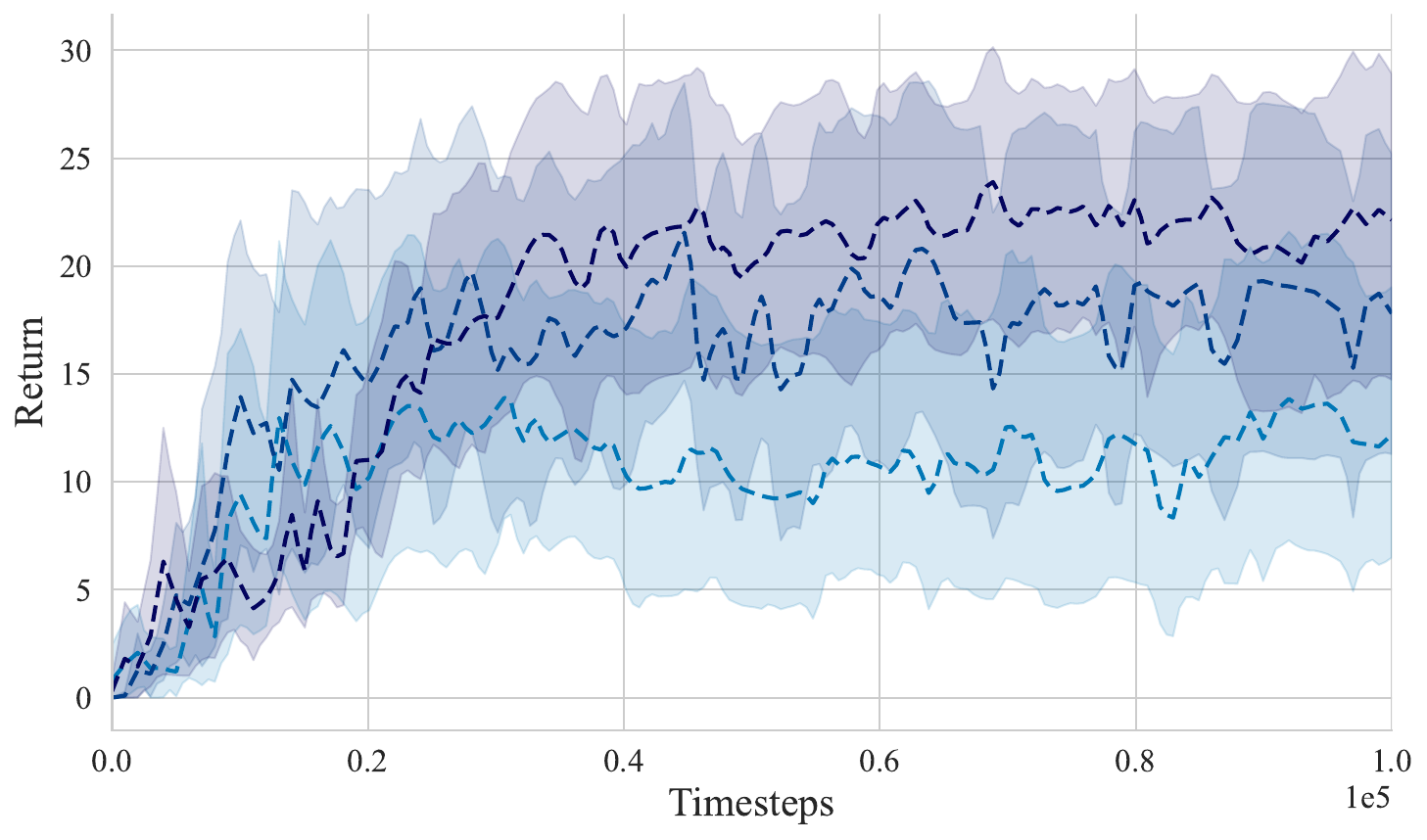}
        \caption{RC, \codeblue{clockwise}}
        
    \end{subfigure}
    \hfill
    \begin{subfigure}[b]{0.24\textwidth}
        \includegraphics[width=\textwidth]{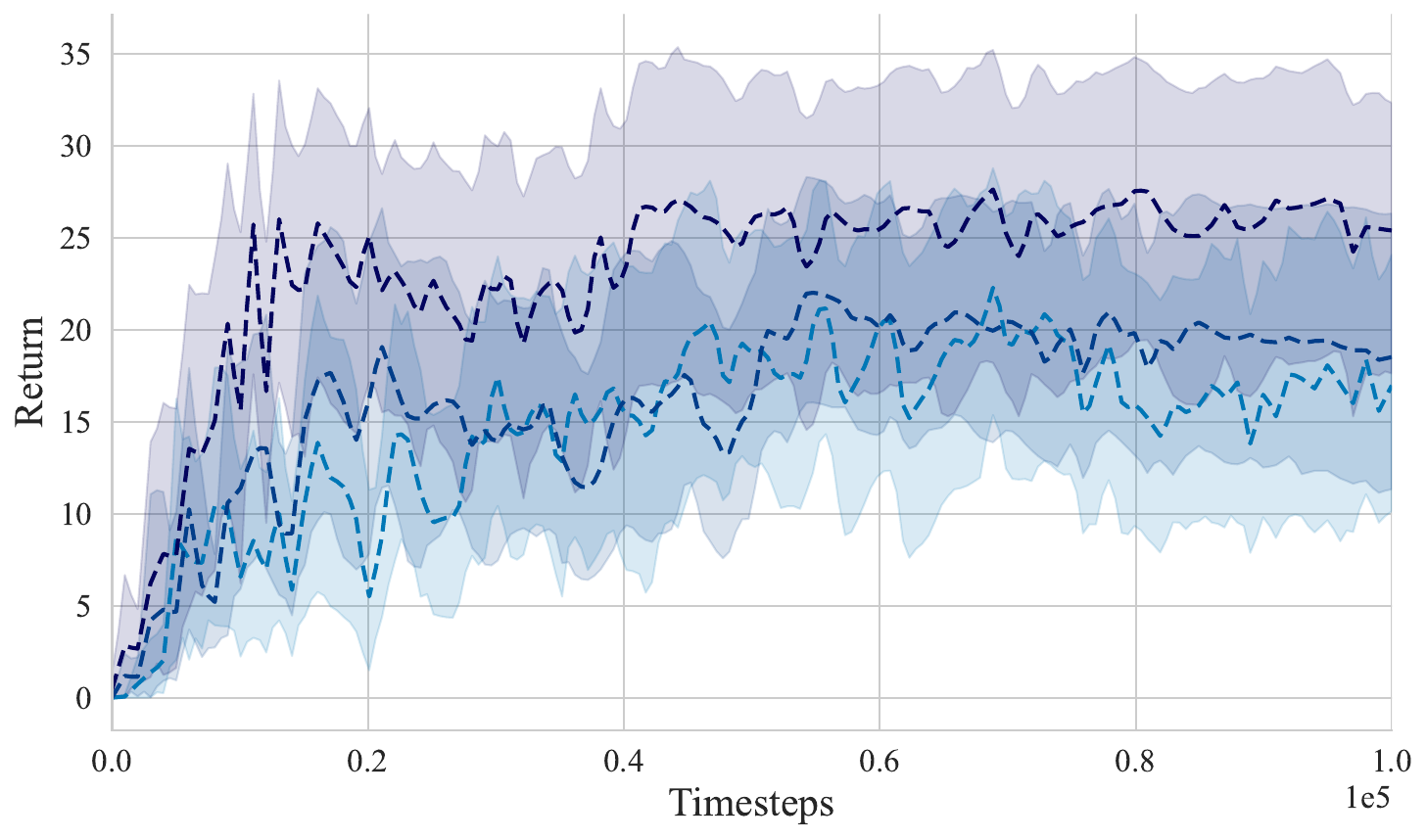}
        \caption{RC, \codeblue{c-clockwise}}
        
    \end{subfigure}
    \hfill
    \begin{subfigure}[b]{0.24\textwidth}
        \includegraphics[width=\textwidth]{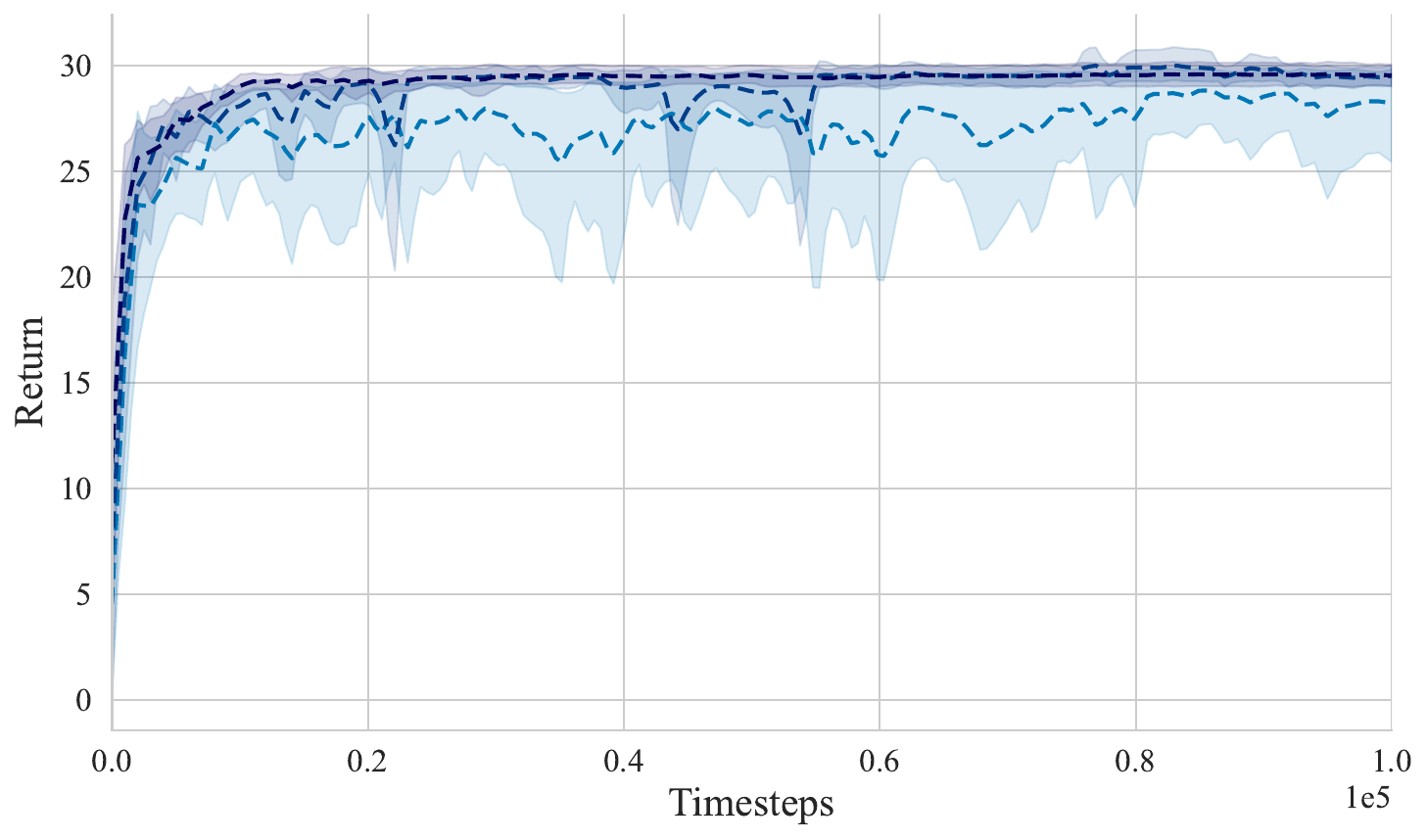}
        \caption{RC, \codeblue{radial}}
        
    \end{subfigure}

    \caption{Performance comparison in MC, RC, and HP for different tasks using \textbf{Action-based Policy Compression}.}

    \label{fig:all_apc_performances}
\end{figure}

\begin{figure}[t]

    \centering 
    \includegraphics[width=0.65\textwidth]{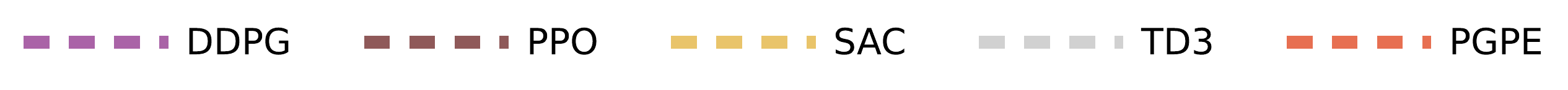}
    \vfill
    \begin{subfigure}[b]{0.24\textwidth}
        \includegraphics[width=\textwidth]{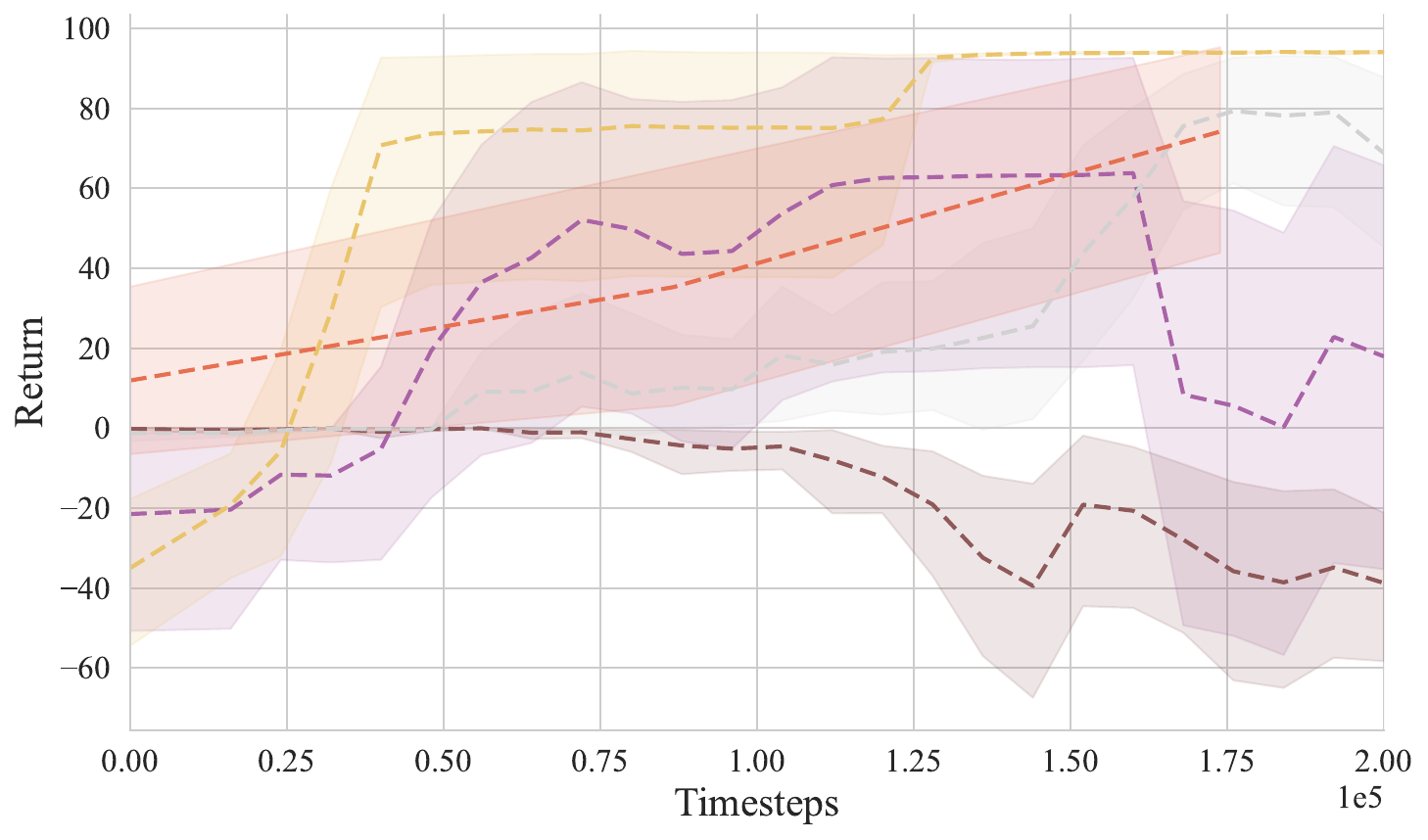}
        \caption{MC, \codeblue{standard}}
        
    \end{subfigure}
    \hfill 
    \begin{subfigure}[b]{0.24\textwidth}
        \includegraphics[width=\textwidth]{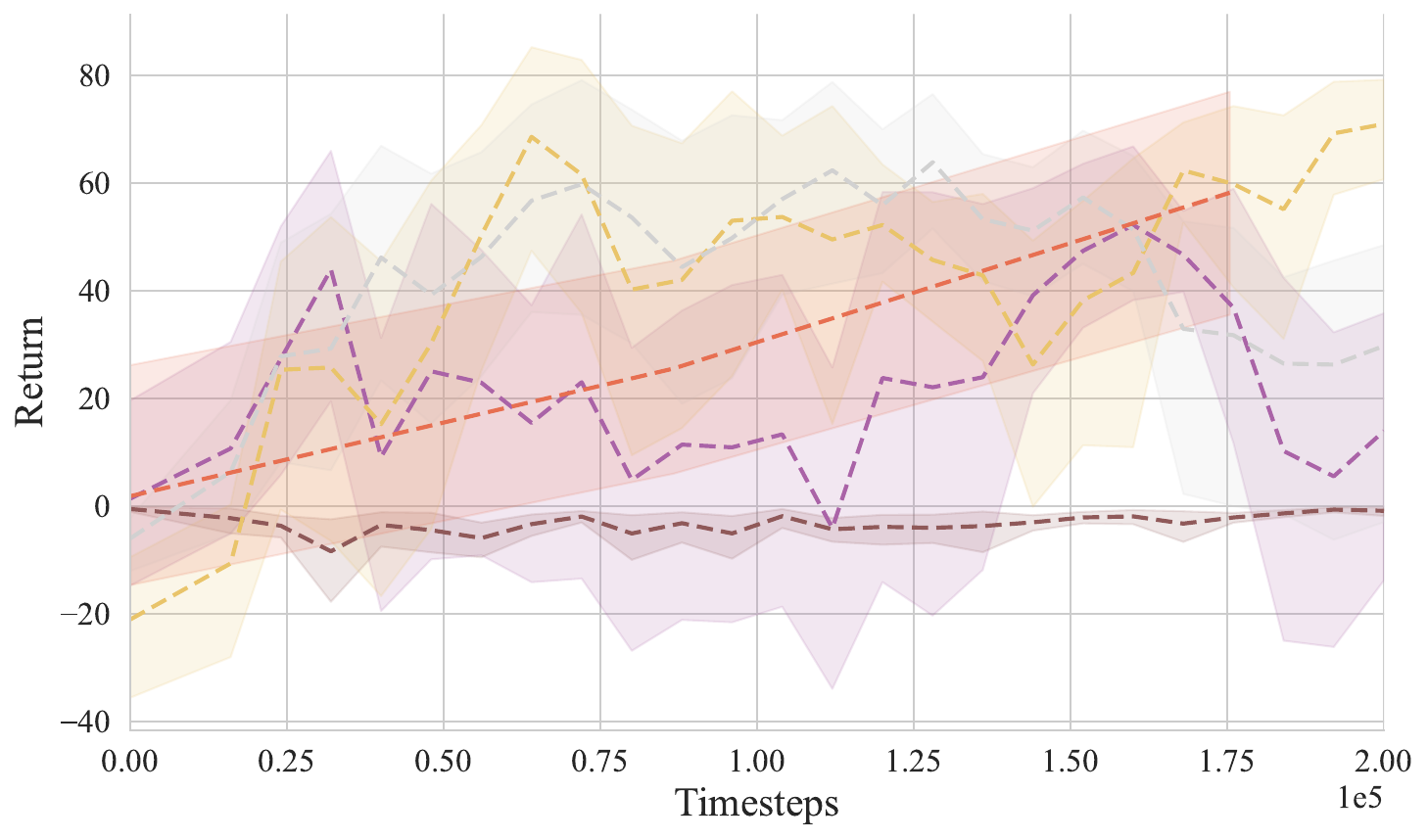}
        \caption{MC, \codeblue{left}}
        
    \end{subfigure}
    \hfill
    \begin{subfigure}[b]{0.24\textwidth}
        \includegraphics[width=\textwidth]{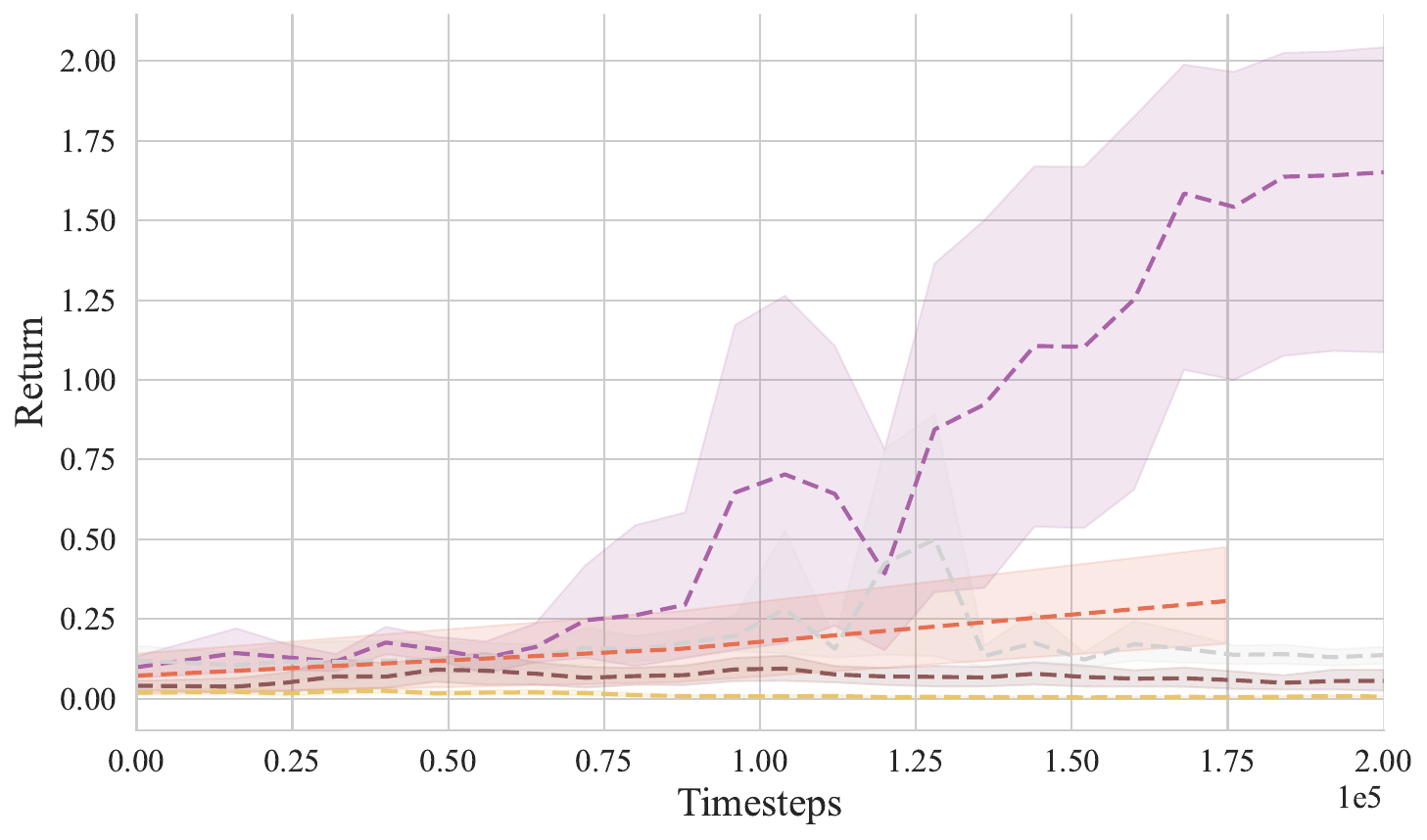}
        \caption{MC, \codeblue{speed}}
        
    \end{subfigure}
    \hfill
    \begin{subfigure}[b]{0.24\textwidth}
        \includegraphics[width=\textwidth]{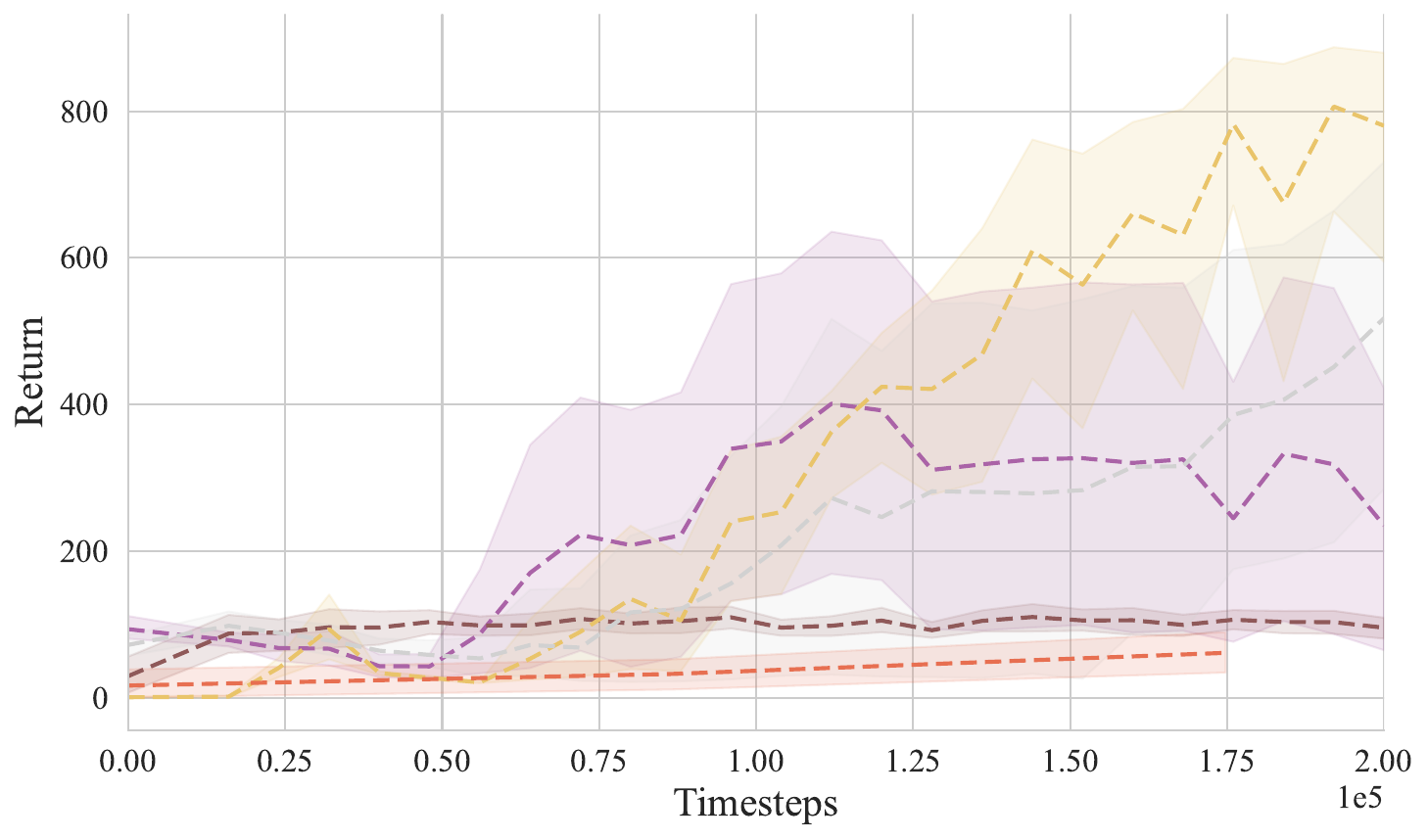}
        \caption{MC, \codeblue{height}}
        
    \end{subfigure}

    \begin{subfigure}[b]{0.24\textwidth}
        \includegraphics[width=\textwidth]{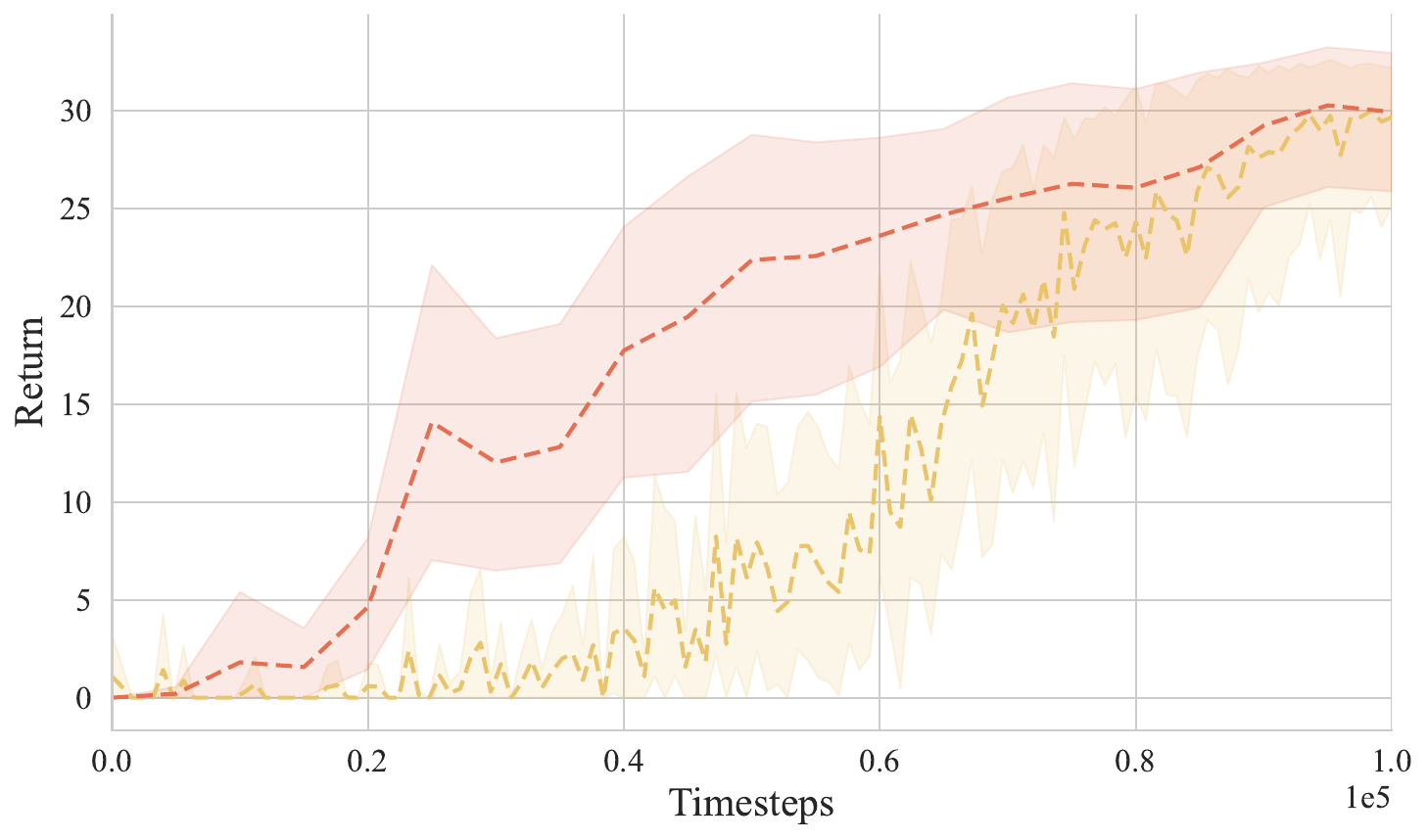}
         \caption{RC, \codeblue{speed}}
        
    \end{subfigure}
    \hfill 
    \begin{subfigure}[b]{0.24\textwidth}
        \includegraphics[width=\textwidth]{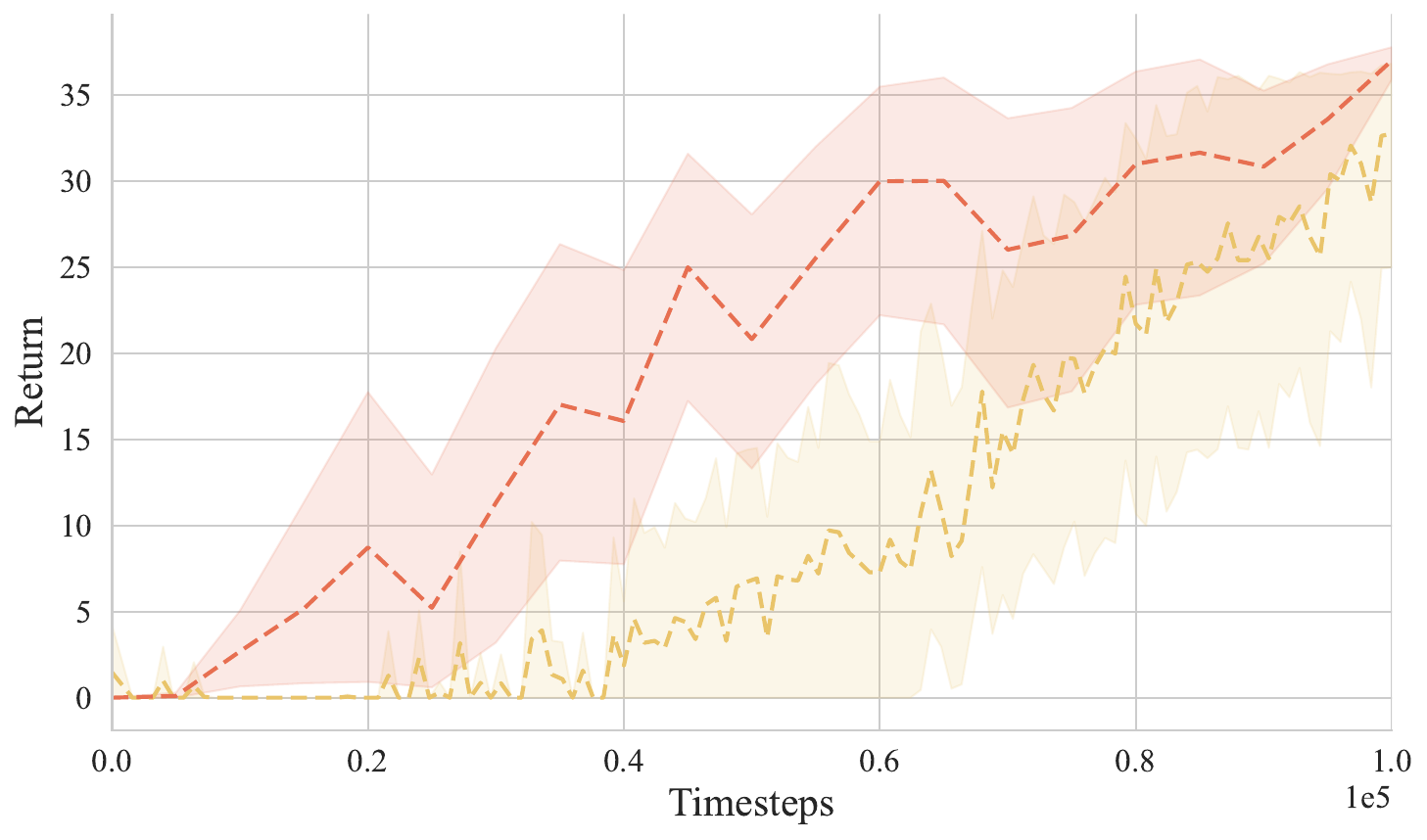}
        \caption{RC, \codeblue{clockwise}}
        
    \end{subfigure}
    \hfill
    \begin{subfigure}[b]{0.24\textwidth}
        \includegraphics[width=\textwidth]{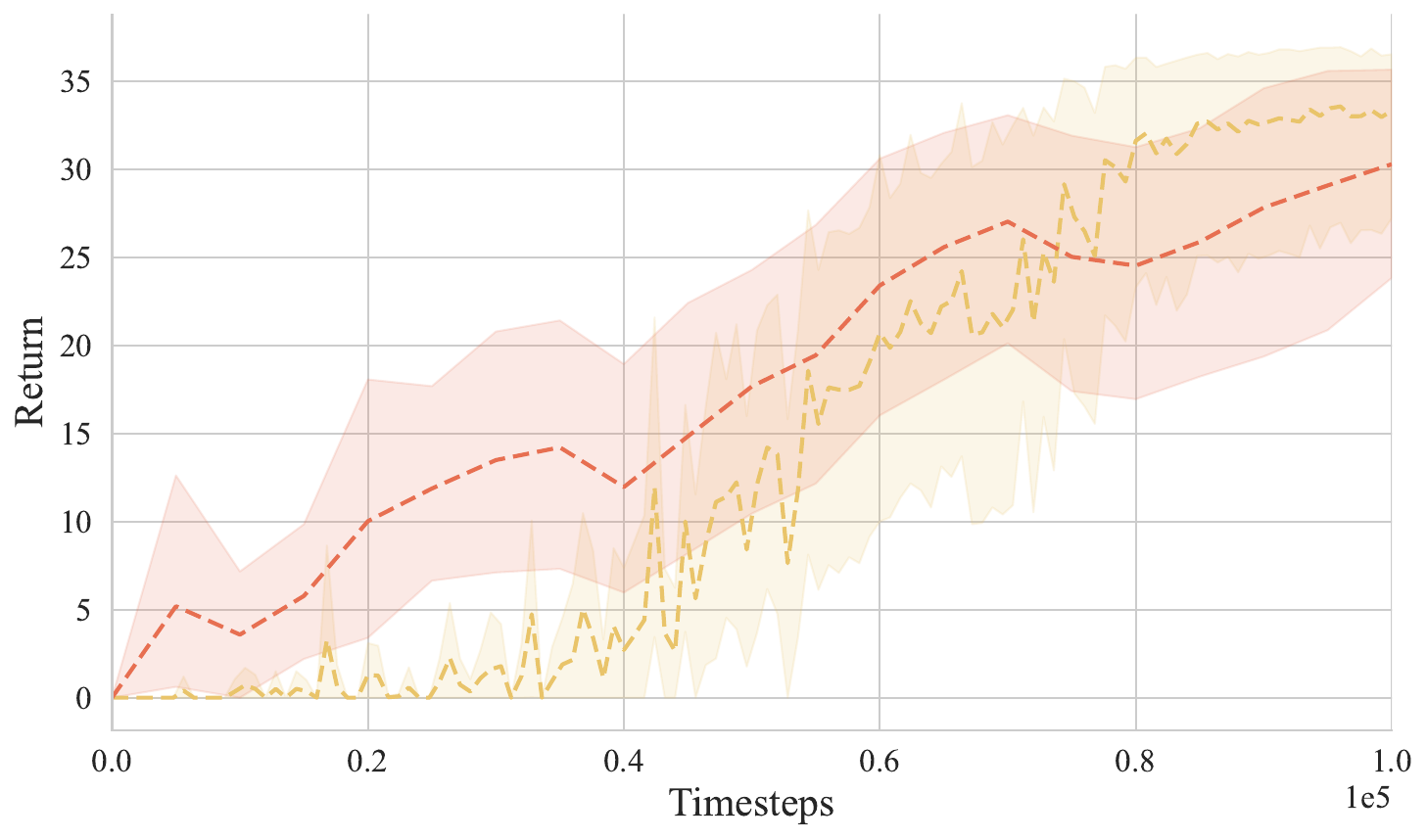}
        \caption{RC, \codeblue{c-clockwise}}
        
    \end{subfigure}
    \hfill
    \begin{subfigure}[b]{0.24\textwidth}
        \includegraphics[width=\textwidth]{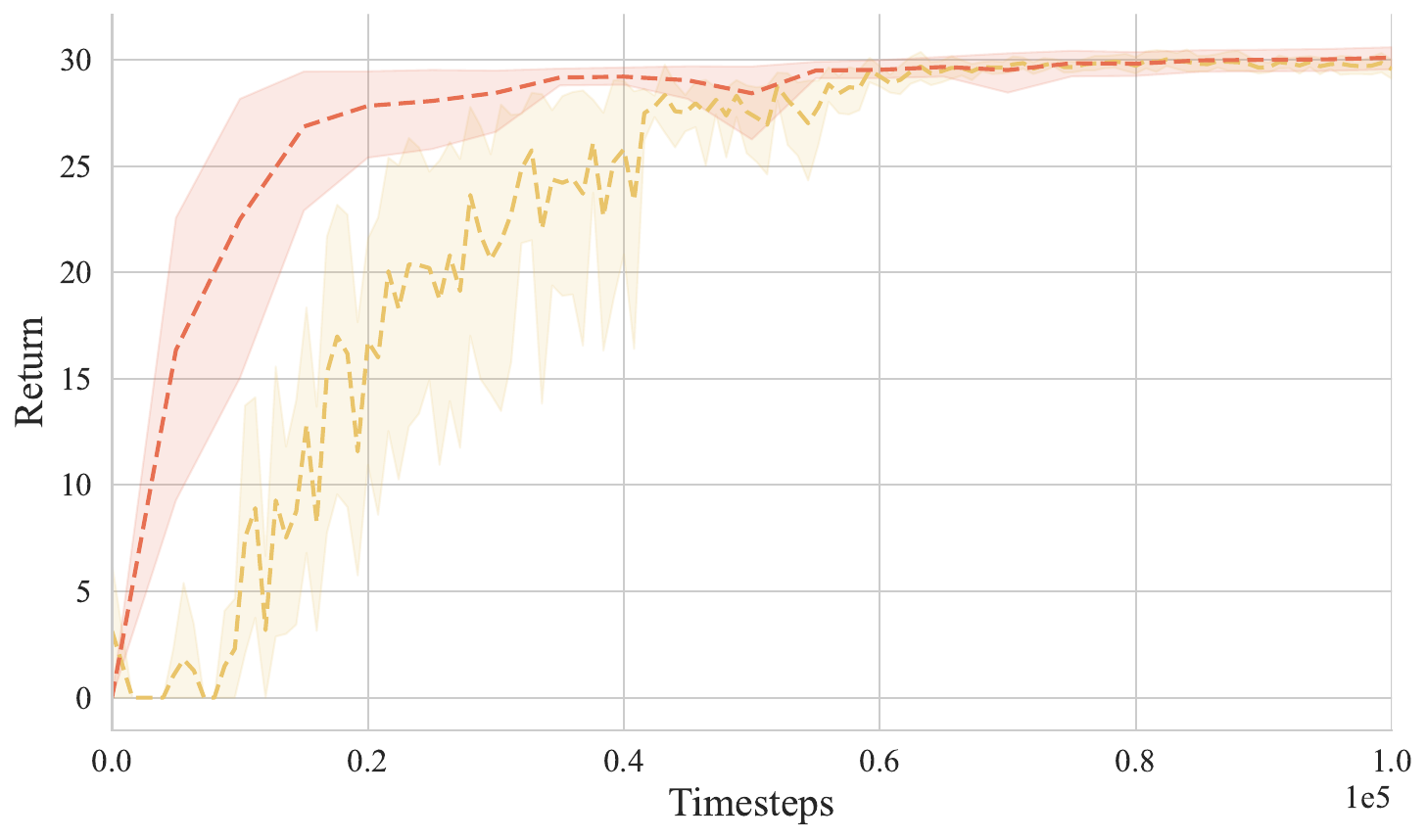}
        \caption{RC, \codeblue{radial}}
        
    \end{subfigure}

    \begin{subfigure}[b]{0.24\textwidth}
        \includegraphics[width=\textwidth]{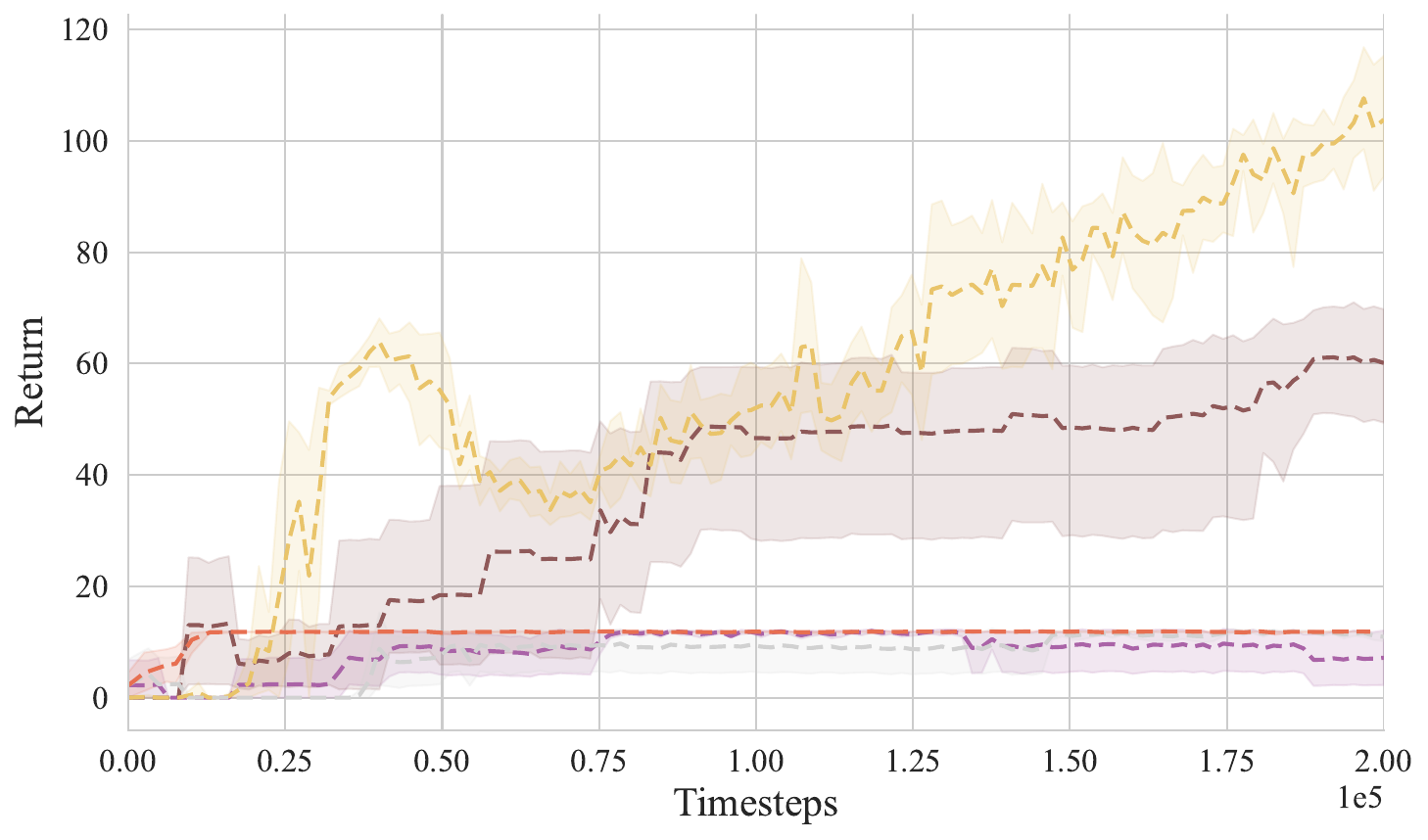}
        \caption{HP, \codeblue{forward}}
        
    \end{subfigure}
    \hfill 
    \begin{subfigure}[b]{0.24\textwidth}
        \includegraphics[width=\textwidth]{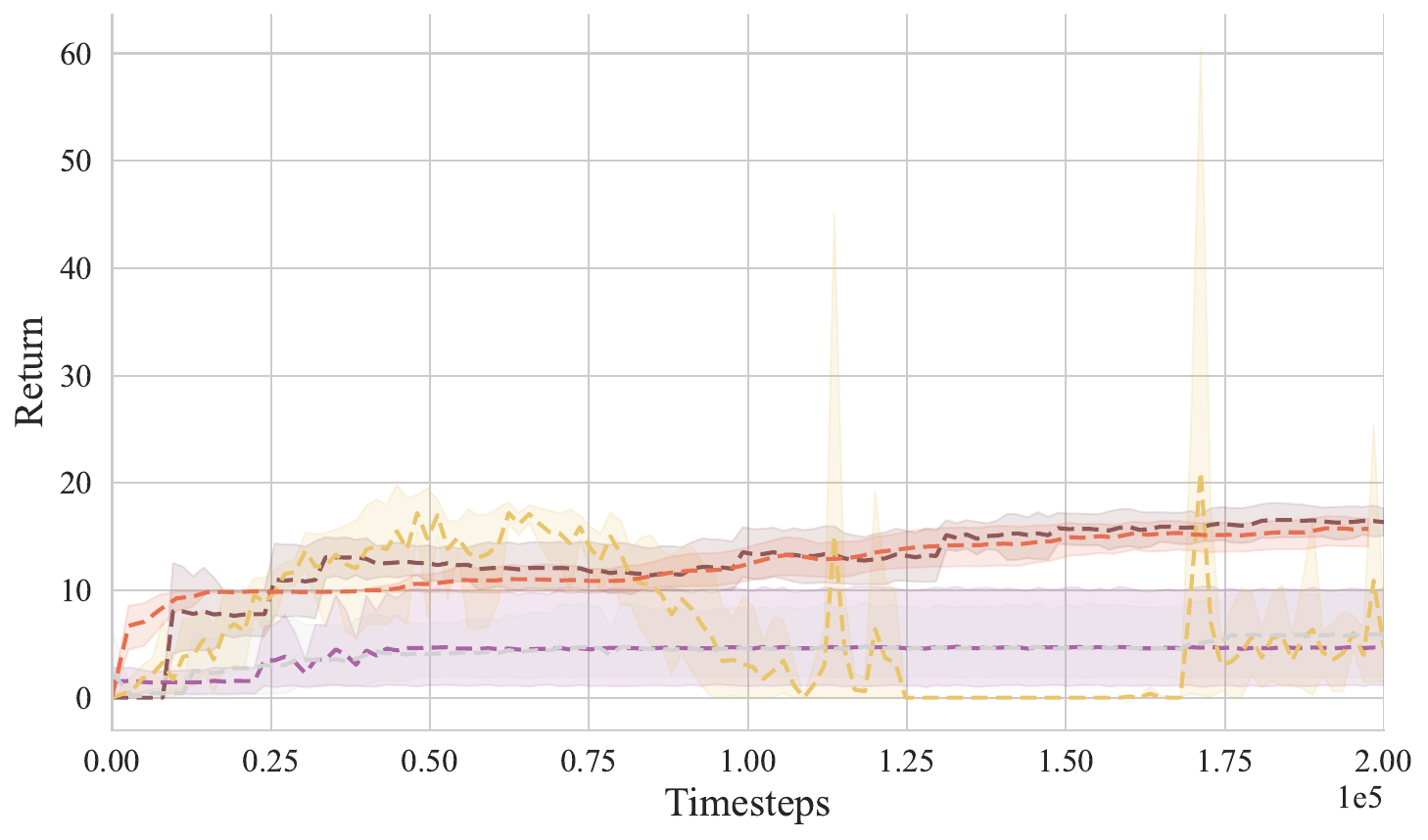}
        \caption{HP, \codeblue{backward}}
        
    \end{subfigure}
    \hfill
    \begin{subfigure}[b]{0.24\textwidth}
        \includegraphics[width=\textwidth]{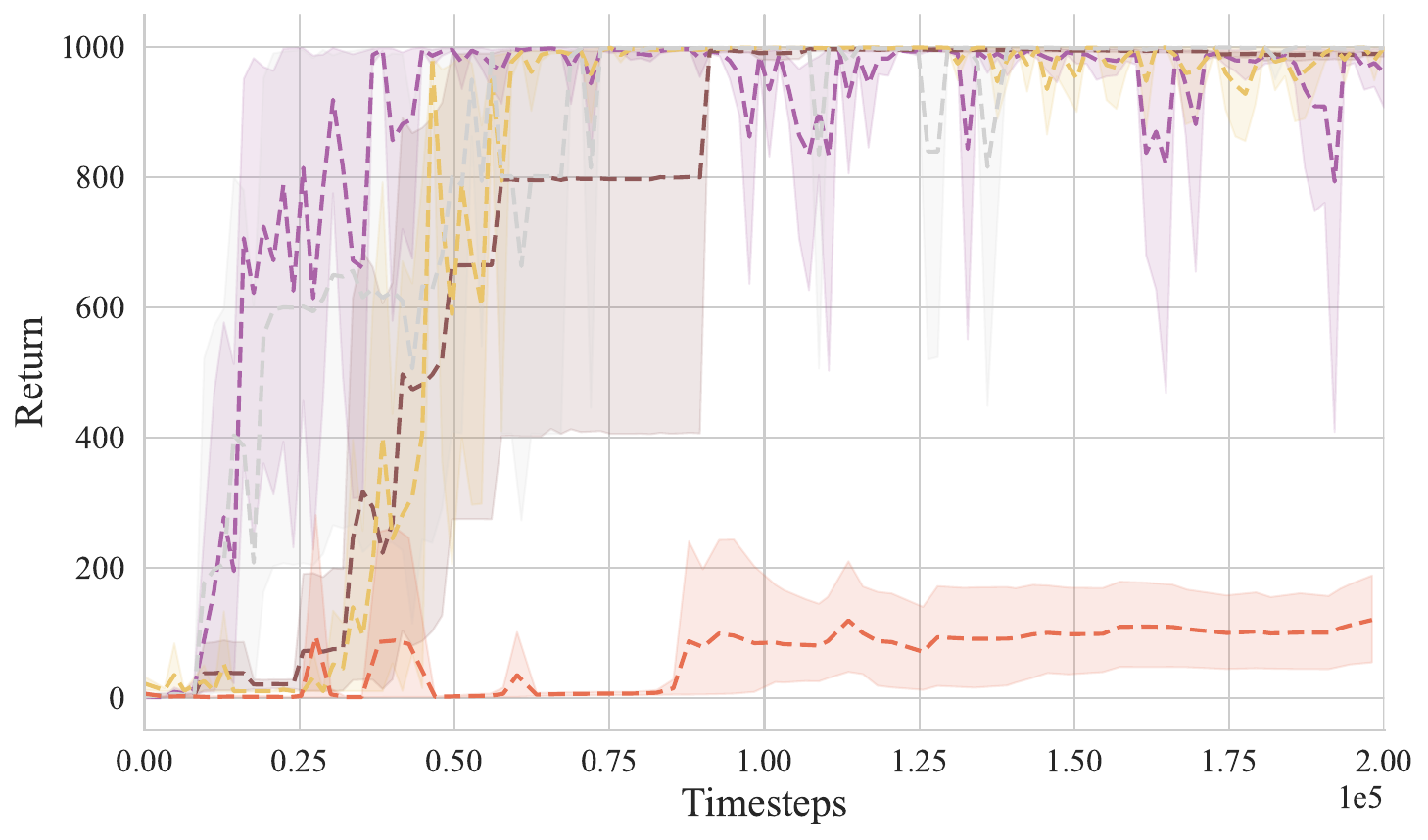}
        \caption{HP, \codeblue{standstill}}
        
    \end{subfigure}
    \hfill
    \begin{subfigure}[b]{0.24\textwidth}
        \includegraphics[width=\textwidth]{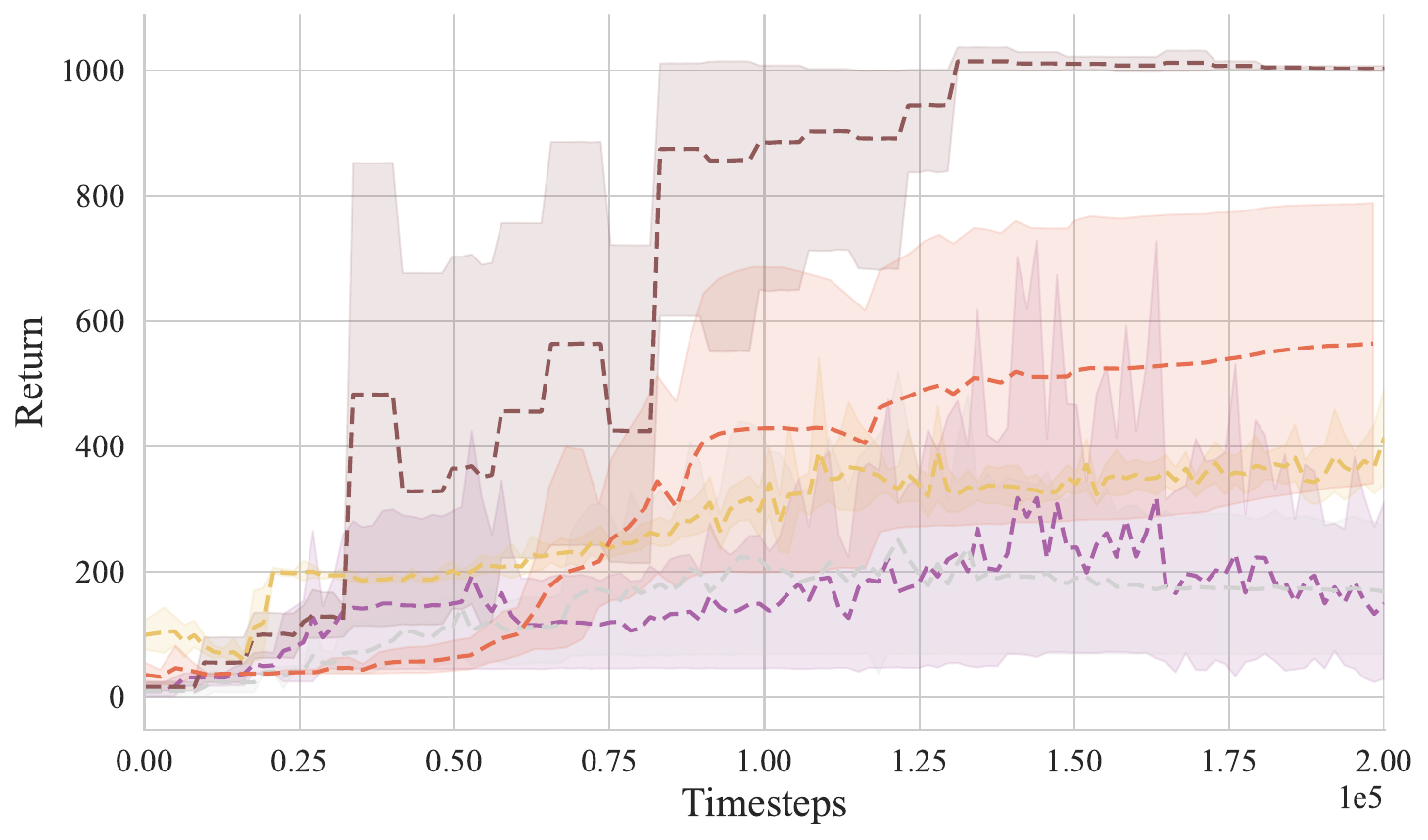}
        \caption{HP, \codeblue{standard}}
        
    \end{subfigure}
    \caption{Performance comparison in MC, RC, and HP for different tasks across all \textbf{baselines}.}

    \label{fig:all_baselines_performances}
\end{figure}

We present the results in categories: Occupancy-based Policy Compression (\Figgref{fig:all_opc_performances}), Action-based Policy Compression (\Figgref{fig:all_apc_performances}), and baselines (\Figgref{fig:all_baselines_performances}).

